\documentclass{article}

\PassOptionsToPackage{numbers, sort&compress}{natbib}

\usepackage[main, final]{neurips_2026}

\usepackage[utf8]{inputenc} 
\usepackage[T1]{fontenc}    
\usepackage{hyperref}       
\usepackage{url}            
\usepackage{booktabs}       
\usepackage{amsfonts}       
\usepackage{nicefrac}       
\usepackage{microtype}      
\usepackage{xcolor}         
\usepackage{algorithm}

\usepackage{microtype}
\usepackage{graphicx}
\usepackage{subcaption}
\usepackage{booktabs} 

\usepackage{mathtools}
\usepackage{paralist}
\usepackage{multicol, multirow}
\usepackage{caption}
\usepackage{makecell}
\usepackage{enumitem}
\usepackage{algorithmic}
\usepackage{float}
\usepackage{wrapfig}

\usepackage{amsmath}
\usepackage{amssymb}
\usepackage{mathtools}
\usepackage{amsthm}

\theoremstyle{plain}
\newtheorem{theorem}{Theorem}[section]

\newtheorem{lemma}[theorem]{Lemma}

\theoremstyle{definition}
\newtheorem{definition}[theorem]{Definition}

\theoremstyle{remark}
\newtheorem{remark}[theorem]{Remark}
\title{Closed-Form Implicit Neural Representations}

\author{
Woojin Cho\thanks{Equal contribution.} \\
TelePIX \\
\texttt{woojin@telepix.net} \\
\And
Junghwan Park\footnotemark[1] \\
TelePIX \\
\texttt{junghwan@telepix.net} \\
}

\begin{document}

\maketitle

\begin{abstract}
Training implicit neural representations (INRs) to capture fine-scale details typically relies on iterative backpropagation and is often hindered by spectral bias when the target exhibits highly non-uniform frequency content. We propose ELM-INR, a backpropagation-free INR that decomposes the domain into overlapping subdomains and fits each local problem using an Extreme Learning Machine (ELM) in closed form, replacing iterative optimization with stable linear least-squares solutions. This design yields fast and numerically robust reconstruction by combining local predictors through a partition of unity. To understand where approximation becomes difficult under fixed local capacity, we analyze the method from a spectral Barron norm perspective, which reveals that global reconstruction error is dominated by regions with high spectral complexity. Building on this insight, we introduce BEAM, an adaptive mesh refinement strategy that balances spectral complexity across subdomains to improve reconstruction quality in capacity-constrained regimes.
\end{abstract}

\section{Introduction}\label{sec:intro}
Implicit neural representations (INRs) provide a simple and flexible way to model data as continuous functions. Instead of storing discrete samples on a fixed grid, an INR learns a coordinate-based mapping that returns the signal value at any queried location~\cite{essakine2024we}. This viewpoint naturally enables resolution-free evaluation, continuous resampling, and compact representations that transfer across discretizations. As a result, INRs have become a broadly useful parameterization for complex signals, ranging from images~\cite{sitzmann2020implicit} and shapes~\cite{park2019deepsdf} to spatiotemporal fields~\cite{pumarola2021d}.

Despite these advantages, most INR pipelines still rely on iterative backpropagation to fit each instance. This makes optimization slow in wall-clock time, especially when the target contains fine-scale structures. The problem is amplified by spectral bias, where standard neural networks tend to learn low-frequency components early and fit high-frequency content much later, which can lead to delayed convergence or underfitting of localized details~\cite{rahaman2019spectral, tancik2020fourier}. In addition, compared to classical numerical analysis pipelines~\cite{reddy1993introduction}, the resulting learned models are often harder to interpret because the solution is embedded implicitly in network parameters produced by a nonconvex training process.

\begin{wrapfigure}{r}{0.49\linewidth}
    \centering
    \vspace{-2pt}
    \includegraphics[width=\linewidth]{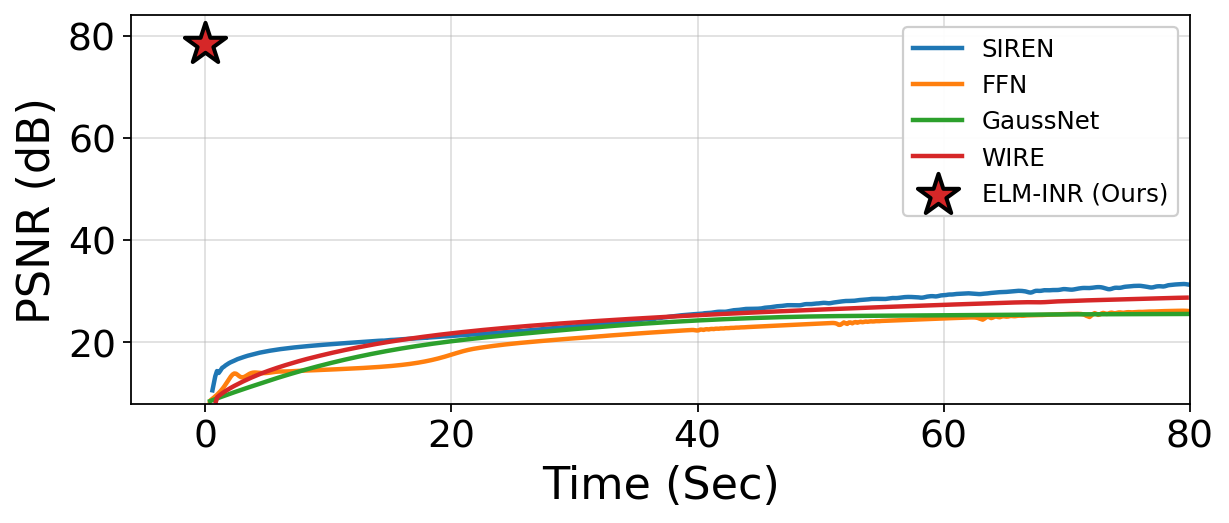}
    \vspace{-15pt}
    \caption{\textbf{PSNR over runtime on the \texttt{kodim05} image.} Unlike iteratively trained INR baselines, ELM-INR uses closed-form solutions across various subdomain sizes and hidden node counts.}
    \label{fig:time_psnr_kodim05}
    \vspace{-10pt}
\end{wrapfigure}

We propose ELM-INR to overcome both spectral bias induced slowdowns and lengthy iterative training. ELM-INR decomposes the domain into overlapping subdomains and fits each subproblem using an Extreme Learning Machine (ELM)~\cite{huang2004extreme, huang2006extreme} whose output weights are obtained by a closed form least squares solve. The local predictors are then blended through a partition of unity~\cite{babuvska1997partition} to produce a smooth global reconstruction. This design removes end to end backpropagation and yields stable and predictable optimization behavior. 
Figure~\ref{fig:time_psnr_kodim05} illustrates this efficiency more concretely by reporting reconstruction quality over wall-clock time on the \texttt{kodim05} image. While conventional INR baselines rely on backpropagation and therefore improve gradually through iterative optimization, ELM-INR is trained in a backpropagation-free manner and achieves high-quality reconstruction in under 20 milliseconds.

We further analyze ELM-INR through the lens of Barron space~\cite{marwah2023neural, chen2021representation}, using spectral complexity to explain why global reconstruction error is dominated by the most difficult regions. This perspective motivates BEAM, a Barron enhanced adaptive mesh refinement strategy that constructs spectrally balanced subproblems so that each local model uses its finite basis effectively. BEAM is especially beneficial in capacity constrained regimes where the number of basis functions is small, providing consistent gains without increasing iterative training cost.

Finally, we validate ELM-INR on diverse modalities beyond conventional images, including multispectral satellite imagery and scientific data such as Navier–Stokes, ERA5, and MRI. Despite using markedly less computation, our method consistently achieves near-perfect reconstruction of the observed signal in our benchmark settings, substantially outperforming existing INR models.

To sum up, our contributions are threefold.
\begin{itemize}
    \item We introduce a backpropagation-free INR framework based on closed form local ELM solves and partition of unity coupling. 
    \item We provide a Barron space based explanation that links approximation difficulty to local spectral complexity and motivates spectral balancing. 
    \item Finally, we propose BEAM, an adaptive partitioning algorithm that improves ELM-INR particularly when the local basis size is limited.
\end{itemize}


\section{Related Work}
\paragraph{Neural representations.}
Implicit neural representations (INRs), also known as coordinate-based neural fields, model a signal $f$ as a continuous function
$\hat{f}:\Omega\subset\mathbb{R}^d\to\mathbb{R}^c$ that maps coordinates to signal values~\cite{sitzmann2020implicit, molaei2023implicit, cho2025neural}.
Beyond 2D imagery, INRs are a standard parameterization in 3D vision where NeRF-style models represent scenes as continuous radiance and density fields and enable high-fidelity novel-view synthesis from multi-view observations~\cite{mildenhall2021nerf, barron2021mip, pumarola2021d}.

A chronic challenge is spectral bias because standard MLPs tend to fit low-frequency components earlier than high-frequency details, which can slow convergence and reduce fidelity on highly textured signals~\cite{rahaman2019spectral, krishnapriyan2021characterizing, yuce2022structured}.
To mitigate this issue, prior work enriches the input with frequency-rich encodings such as Fourier or positional features or learned coordinate mappings~\cite{zhu2024disorder}, or adopts periodic and localized function bases such as sinusoidal~\cite{sitzmann2020implicit}, Gabor-like~\cite{ramasinghe2022beyond}, or wavelet-like parameterizations~\cite{saragadam2023wire}.

Recently, another line of work leverages trained INRs as representations of individual instances in a dataset~\cite{dupont2021coin, cho2025fourier}.
In this view, each signal or field is first fit by an INR and the resulting parameters are then mapped to a vector space for retrieval, clustering, and conditional generation, while curated collections of pretrained INRs facilitate reuse and benchmarking across domains~\cite{dupont2022data, ma2024implicit, jo2025pdefuncta}.
INRs have also been explored for compression by transmitting compact network parameters instead of raw pixels, yet most INR-based codecs still require per-instance iterative optimization at encoding time, which limits deployment under tight compute and power budgets~\cite{luigi2023deep, bauer2023spatial}.

\begin{figure*}[t]
\centering
\includegraphics[width=1.0\columnwidth]{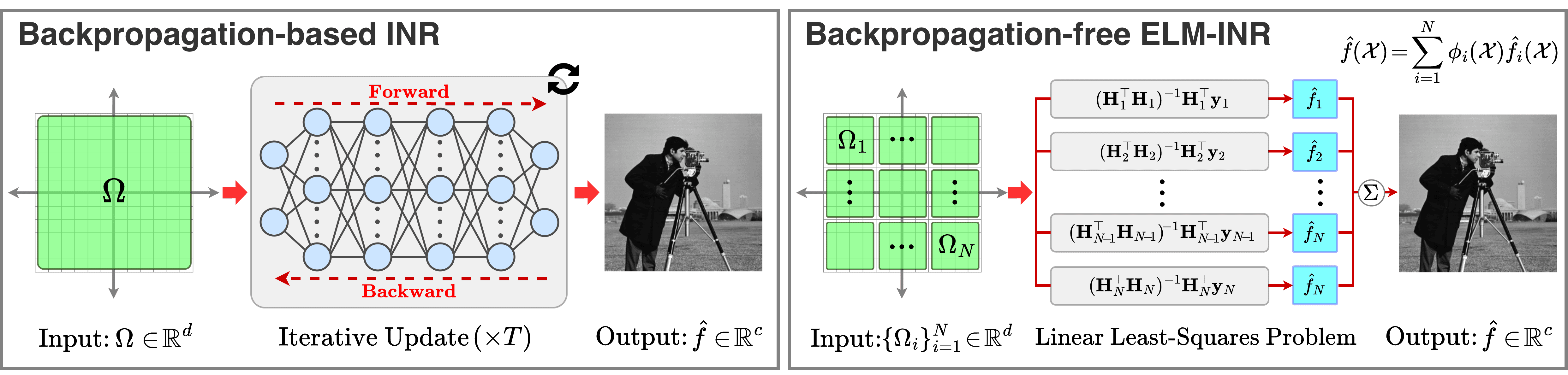}
    \caption{\textbf{Backpropagation-based INR vs. ELM-INR.} Standard INRs require iterative gradient-based training, while ELM-INR fits subdomain ELMs via one-shot least squares and blends them with partition-of-unity for a smooth global reconstruction.}
    \label{fig:main}
\end{figure*}

\paragraph{Backpropagation-free optimization and domain decomposition.}
Backpropagation-free learning is closely related to random feature models, where hidden layer parameters are frozen after initialization, reducing the training process to a linear regression problem solvable via efficient linear algebra routines rather than iterative gradient descent. A canonical instance is the Extreme Learning Machine (ELM)~\cite{huang2004extreme, huang2006extreme, huang2011extreme}, which yields predictable runtimes and avoids local minima by solving for output weights in closed form. To handle multi-scale signals, Partition-of-Unity (PoU) methods~\cite{babuvska1997partition, lee2021partition} decompose the global domain into overlapping subdomains blended via smooth window functions, a concept foundational to methods like Generalized Finite Elements~\cite{melenk1996partition}.

Recent works have extended these ideas to neural fields in scientific machine learning~\cite{dwivedi2020physics}. For instance, FBPINNs~\cite{moseley2023finite} and ELM-FBPINN~\cite{anderson2024elm} demonstrate the potential of combining PINNs~\cite{raissi2019physics, karniadakis2021physics} with domain decomposition to enhance stability. However, standard INRs continue to struggle with spectral bias and the computational cost of iterative backpropagation. Addressing these limitations, we propose ELM-INR, a method that integrates adaptive domain decomposition with the closed-form efficiency of ELMs to bypass iterative optimization and the slow convergence caused by spectral bias, enabling fast and high-fidelity reconstruction. Unlike ELM-based PDE solvers such as ELM-FBPINN, where the solution is unknown before training and the domain is therefore decomposed into predefined subdomains, ELM-INR observes the target signal directly and can adapt the partition itself to local spectral complexity before fitting, as described in Section~\ref{sec:beam}.

\section{Extreme Learning Machines for Implicit Neural Representations}\label{sec:elm_inr}
In this section, we develop the theoretical foundations of ELM-INR, our method that integrates Extreme Learning Machines with implicit neural representations. We first formalize how local function approximation is performed using ELMs, and how these local models are stitched together via a partition-of-unity to form a global approximation. We then analyze the computational complexity advantages of this approach. Finally, we introduce the Barron space framework to quantify the spectral complexity of target functions, which provides a rigorous motivation for our adaptive, spectral-based domain decomposition strategy leading into Section~\ref{sec:beam}.

\subsection{Local Approximation via ELMs}
An ELM constitutes a single-layer feedforward network that avoids iterative optimization by fixing random hidden weights. Consequently, training reduces to solving a linear least squares problem in closed form. In our framework, we employ ELMs for local function approximation, which can be mathematically expressed as  Eq.~\eqref{eq:elm}

\begin{definition}[Local Extreme Learning Machine]
An ELM is a single-hidden-layer feedforward network where the input weights and biases are randomly initialized and frozen, while only the output weights are trainable. On a specific subdomain $\Omega_i \subset \mathbb{R}^d$, the local ELM $\hat{f}_i(\mathcal{X})$ is defined as follows:
\begin{equation}
    \hat{f}_i(\mathcal{X}) = \sum_{j=1}^{m} \alpha_{i,j}\,\sigma(w_{i,j}^\top \mathcal{X} + b_{i,j}), \quad \forall \mathcal{X} \in \Omega_i\label{eq:elm}
\end{equation}
where $\mathcal{X} \in \Omega_i$ is a continuous input coordinate, $m$ is the hidden width, $\sigma(\cdot)$ is the activation function, and $({w}_{i,j}, b_{i,j})$ are fixed random parameters. The coefficients $\boldsymbol{\alpha}_i = [\alpha_{i,1}, \dots, \alpha_{i,m}]^\top$ are the only trainable parameters for $\hat{f}_i(\mathcal{X})$.
\end{definition}

Unlike conventional INRs that require iterative backpropagation, the fixed nature of the hidden parameters simplifies the learning process significantly. Since the mapping from the hidden layer to the output is linear, finding the optimal weights becomes a convex optimization problem.

\begin{lemma}[ELM as Linear Least-Squares]
Training the local ELM $\hat{f}_i$ on $\Omega_i$ reduces to a linear least-squares problem. Let $\mathbf{H}_i \in \mathbb{R}^{S_i \times m}$ be the hidden layer activation matrix for $S_i$ sample points in $\Omega_i$, defined as $(\mathbf{H}_i)_{kj} = \sigma(w_{i,j}^\top x_k + b_{i,j})$. The optimal output weights $\boldsymbol{\alpha}_i^*$ that minimize the squared error 
$\|\mathbf{H}_i\boldsymbol{\alpha}_i-\mathbf{y}_i\|_2^2$
can be obtained in closed form:
\begin{equation}
    \boldsymbol{\alpha}_i^* = (\mathbf{H}_i^\top \mathbf{H}_i)^{-1} \mathbf{H}_i^\top \mathbf{y}_i
    \label{eq:ls}
\end{equation}
where $\mathbf{y}_i \in \mathbb{R}^{S_i}$ is the target vector sampled from $f$ at points inside $\Omega_i$. This closed-form solution can be computed efficiently with standard linear solvers, enabling rapid training while avoiding sensitivity to local minima common in gradient-based optimization.
\end{lemma}

\subsection{Global Reconstruction via Partition of Unity}
To construct a continuous global representation $f(\mathcal{X})$ from these independent local models, we employ an overlapping domain decomposition strategy. We define a collection of smooth window functions $\{\phi_i(\mathcal{X})\}_{i=1}^{N}$ associated with the overlapping subdomains $\{\Omega_i\}_{i=1}^{N}$.
\begin{definition}[Partition of Unity]
The window functions form a Partition of Unity (PoU) over the domain $\Omega$, and the global ELM-INR approximation is the weighted sum of local solutions:
\begin{equation}
    \sum_{i=1}^N \phi_i(\mathcal{X}) = 1, \;\; \forall{\mathcal{X}} \in \Omega, \;\; \text{supp}(\phi_i) \subset \Omega_i,
    \qquad
    \hat{f}(\mathcal{X}) = \sum_{i=1}^N \phi_i(\mathcal{X}) \hat{f}_i(\mathcal{X}).
    \label{eq:pou}
\end{equation}
This formulation ensures that the global approximation $\hat{f}(\mathcal{X})$ remains smooth and continuous across subdomain boundaries. Furthermore, because the global output is a linear combination of local outputs, the training remains decoupled and each $\boldsymbol{\alpha}_i$ can be solved independently. Since the local linear systems are independent, they can be batched and solved in parallel on a GPU, providing substantial wall-clock speedups in practice (See Figure~\ref{fig:time_psnr_kodim05}).

\end{definition}



\subsection{Barron Space and Spectral Complexity}
This section establishes the theoretical foundation for our proposed methods. It utilizes the Barron space framework to quantify the spectral cost of approximating a function, proving that the global error of a neural representation is bottlenecked by regions with high spectral complexity.

\begin{definition}[Barron Space $\mathcal{B}$ and Function Class $\Gamma$~\cite{barron2002universal}]
The Barron space $\mathcal{B}$ is a function space consisting of functions that can be efficiently approximated by shallow (single-hidden-layer) neural networks with dimension-independent error rates. Unlike classical Sobolev spaces~\cite{adams2003sobolev, maz2013sobolev}, where approximation error suffers from the curse of dimensionality $\mathcal{O}(m^{-1/d})$, functions in the Barron space admit an approximation rate of $\mathcal{O}(1/\sqrt{m})$, where $m$ is the network width~\cite{ma2022barron}.

To make this computationally tractable, the Barron space is concretely realized through a Fourier-based function class $\Gamma$, defined by the integrability of the function's Fourier spectrum and its first moment:
\begin{equation}
\Gamma\!\coloneqq\!\!\Bigl\{f\!:\! \int_{\mathbb{R}^d}\!\! |\mathcal{F}({\xi})|\, d{\xi}\!<\!\infty,
\int_{\mathbb{R}^d}\!\! \|{\xi}\|_{1}\,|\mathcal{F}({\xi})|\, d{\xi}\! <\!\infty
\Bigr\},
\end{equation}
where $\mathcal{F}$ denotes the Fourier transform of $f$ and $\xi \in \mathbb{R}^d$ represents the continuous frequency variable.
\end{definition}

\begin{definition}[Spectral Barron Norm~\cite{chen2021representation, liao2025spectral}]
For a function $f \in \Gamma$, the difficulty of approximation is governed by the Spectral Barron Norm. This norm acts as a proxy for the function's spectral complexity, weighing the frequency components by their magnitude:
\begin{equation}
    \|f\|_{\mathcal{B}_S} \coloneqq \int_{\mathbb{R}^d} \|\xi\|_1 |\mathcal{F}(\xi)| d\xi.
\end{equation}
Intuitively, a larger $\|f\|_{\mathcal{B}_{S}}$ indicates that the function possesses significant high-frequency energy or rapid variations, making it inherently harder for a neural network to fit.
\end{definition}



This norm serves as a critical quantity in controlling the approximation error, as formalized by the following bound.

\begin{theorem}[Approximation Error Bound]
\label{thm:barron_bound}
For a function $f$ with a finite spectral Barron norm $\|f\|_{\mathcal{B}_S}$, a neural network with $m$ hidden units can approximate $f$ on a subdomain $\Omega_i$ with an $L^2$ error bounded by:
\begin{equation}
    \|f - \hat{f}_i\|_{L^2(\Omega_i)} \le \frac{C \|f\|_{\mathcal{B}_S}}{\sqrt{m}},
\end{equation}
where $C$ is a constant independent of $m$.
\end{theorem}

This theorem highlights a critical limitation of uniform partitioning. Real-world signals often exhibit non-uniform spectral complexity. If we employ a fixed width $m$ across all subdomains, the global accuracy will be bottlenecked by the most complex region.

\begin{remark}[Error Domination by Spectral Complexity]\label{lemma:error_domination}
Consider a partition of $\Omega$ where each subdomain is approximated by an ELM of width $m$. Let $\beta_i \coloneqq \|f|_{\Omega_i}\|_{\mathcal{B}_S}$ be the local spectral Barron norm. From Theorem \ref{thm:barron_bound}, the worst-case local error is bounded by:
\begin{equation}
    \max_{i} \|f - \hat{f}_i\|_{L^2(\Omega_i)} \le \frac{C}{\sqrt{m}} \left( \max_{i} \beta_i \right).
\label{eq:barron_bound}
\end{equation}
\textbf{Implication:} The global reconstruction quality is limited by the region with the highest spectral complexity ($\max_i \beta_i$). This theoretical insight directly motivates our proposed method, Barron-Enhanced Adaptive Mesh refinement (BEAM), which adaptively partitions $\Omega$ to equalize $\beta_i$ across all subdomains, ensuring uniform high-fidelity reconstruction.
\end{remark}

\begin{figure}[t]
\centering
\begin{minipage}[c]{0.50\columnwidth}
\centering
\includegraphics[width=\linewidth]{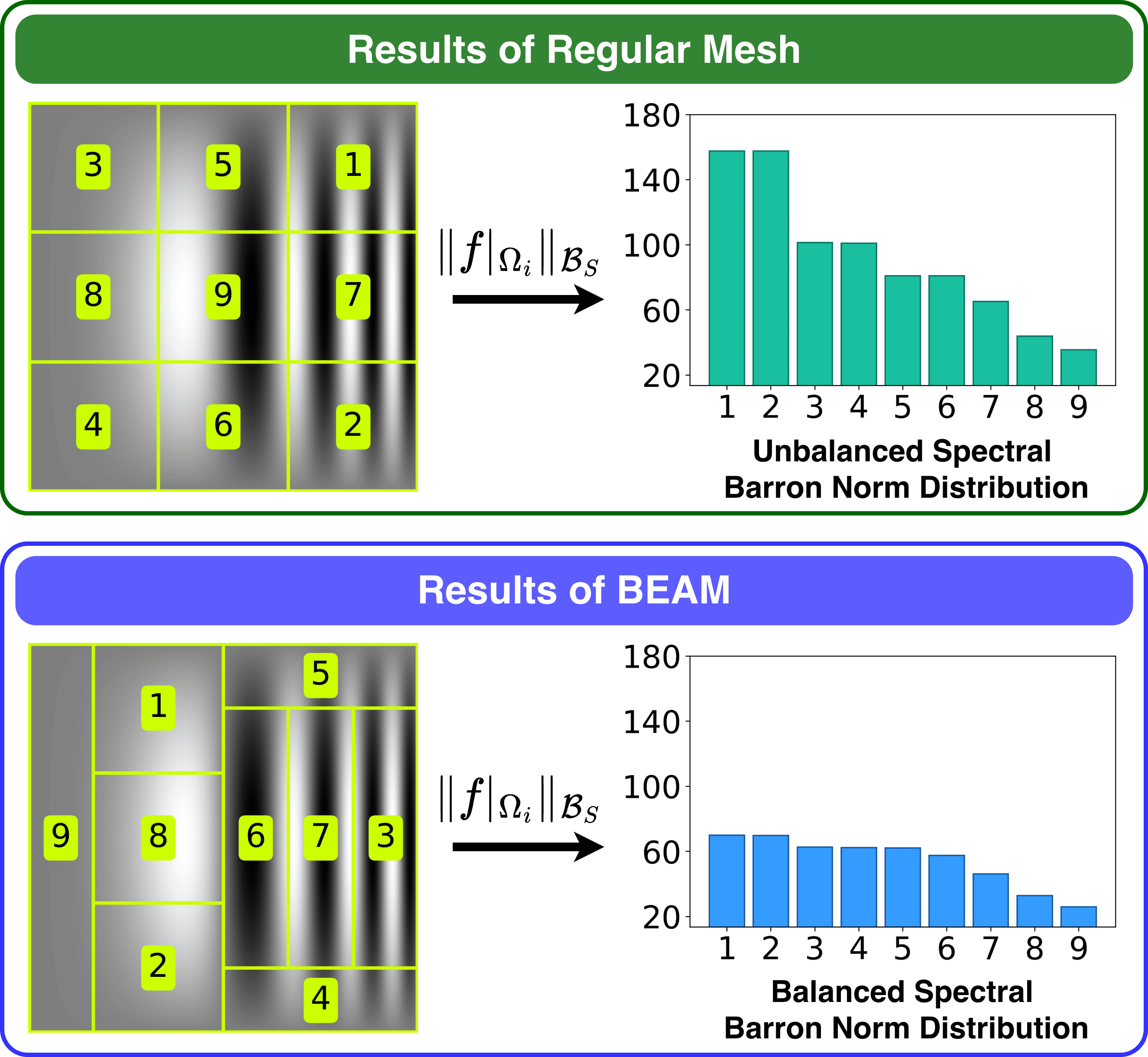}
\caption{\textbf{Regular Mesh vs. BEAM.} Unlike uniform partitioning (top), BEAM (bottom) adaptively merges simple regions and refines complex regions, balancing spectral complexity across subdomains.}
\label{fig:beam_concept}
\end{minipage}
\hfill
\begin{minipage}[c]{0.47\columnwidth}
\centering
\vspace{-10pt}
\begin{algorithm}[H]
   \caption{Barron-Enhanced Adaptive Mesh Refinement}
   \label{alg:beam}
\begin{algorithmic}
   \STATE {\bfseries Input:} Image $f$, Threshold $\tau$, Atomic size $s$
   \STATE {\bfseries Output:} Adaptive Partition $\mathcal{P}$
   \STATE Initialize $\mathcal{P}$ with $s \times s$ atomic cells
   \STATE Compute local spectral energy $\|f|_{\Omega_i}\|_{\mathcal{B}_S}$ for each $\Omega_i \in \mathcal{P}$
   \REPEAT
       \STATE Sort $\Omega_i \in \mathcal{P}$ by energy (ascending)
       \FOR{each $\Omega_i$ where $\|f|_{\Omega_i}\|_{\mathcal{B}_{S}} < \tau$}
           \STATE Find neighbor $\Omega_j$ that minimizes \\$E_{new} = \|f|_{\Omega_i \cup \Omega_j}\|_{\mathcal{B}_{S}}$
           \IF{$E_{new} \le \tau$}
               \STATE Merge $\Omega_i, \Omega_j \rightarrow \Omega_{new}$ in $\mathcal{P}$
               \STATE Update energy for $\Omega_{new}$ and remove $\Omega_i, \Omega_j$
           \ENDIF
       \ENDFOR
   \UNTIL{no merges occur}
   \STATE \textbf{return} $\mathcal{P}$
\end{algorithmic}
\end{algorithm}
\end{minipage}
\end{figure}

\section{Barron-Enhanced Adaptive Mesh}\label{sec:beam}
Building on the theoretical motivation established in Section~\ref{sec:elm_inr}, we propose BEAM (Barron-Enhanced Adaptive Mesh refinement).
This domain decomposition algorithm is explicitly designed to equalize the spectral complexity across all subdomains.
By strictly enforcing a spectral capacity constraint, BEAM transforms a computationally hard global fitting problem into a collection of spectrally balanced, manageable subproblems.

This design is directly motivated by the error bound in Eq.~\eqref{eq:barron_bound}, which shows that the worst-case approximation error is governed by the largest local spectral Barron norm, scaled by the inverse square root of the model capacity $m$.
While increasing $m$ can in principle suppress the error even in spectrally complex regions, doing so is often impractical in real-world INR settings due to computational, memory, or energy constraints.
In such capacity-limited regimes, the dominant source of error arises from spectral imbalance across subdomains, and simply enlarging the model width becomes an inefficient remedy.
BEAM addresses this issue by adaptively partitioning the domain to equalize local spectral Barron norms, thereby reducing the worst-case error under a fixed and moderate model capacity.

\subsection{Practical Estimation of Spectral Barron Norm}

Since the continuous spectral Barron norm is generally intractable to compute
for real signals, we rely on a discrete Fourier-based spectral proxy~\cite{marwah2023neural} that is suitable for finite data. For a region $\Omega_i$, we compute spectral Barron norm as follows:
\begin{equation}
\|f|_{\Omega_i}\|_{\mathcal{B}_{S}} \approx \sum_{k} \|k\|_{1} |\mathcal{F}_{i}(k)|,
\end{equation}
where $k$ indexes discrete frequency components in the DFT domain.
This measure directly reflects frequency-weighted spectral energy and is sufficient for BEAM, as the algorithm relies only on relative complexity comparisons under a threshold $\tau$.

\subsection{Principle of Spectral Balancing}
Standard uniform partitioning fails to account for the spatial non-uniformity of signal complexity. To illustrate this, we consider a toy example derived from a Poisson equation with the target field $f(x_1, x_2) = \sin(2\pi (4x_1^3))\sin(\pi x_2)$, which exhibits spatially varying frequencies due to the cubic term. Figure~\ref{fig:beam_concept} demonstrates that applying a regular mesh to this field results in a highly skewed distribution of spectral complexity: subdomains covering high-frequency oscillations exhibit excessive spectral Barron norms, creating error bottlenecks, while smooth regions remain underutilized.

BEAM addresses this imbalance by constructing a partition $\mathcal{P} = \{\Omega_i\}_{i=1}^N$ where every subdomain satisfies the spectral constraint $\|f|_{\Omega_i}\|_{\mathcal{B}_{S}} \le \tau$. As visualized in the bottom panel of Figure~\ref{fig:beam_concept}, our algorithm adaptively merges simpler regions while isolating complex features, equilibrating the spectral Barron norms across the domain. This ensures that the difficulty of each subproblem matches the capacity of the local ELM, preventing any single network from being overwhelmed.

\subsection{Bottom-Up Spectral Merging}
Unlike traditional top-down subdivision, BEAM employs a bottom-up merging strategy to allow flexible subdomain shapes, as summarized in Algorithm~\ref{alg:beam}. The procedure initializes the domain as a fine grid of atomic cells and then iteratively identifies regions with low spectral complexity to merge them with spatial neighbors. To refine the partition, we use a greedy rule that evaluates all valid neighboring candidates and chooses the merge that minimizes the combined spectral energy while ensuring that it remains below the threshold $\tau$. Repeating this step gradually aggregates background areas into larger subdomains while naturally isolating high-frequency details into smaller patches.
Since the subdomains can be irregular, we compute the spectral Barron norm on the bounding box of each region, masking pixels outside the region to zero to evaluate the local signal content.

\section{Experiments}\label{sec:experiments}
\paragraph{Experimental Setup.}
We first describe the common experimental settings used throughout this section, including implementation details, baseline configurations, and the default setup of ELM-INR.
All experiments are implemented in PyTorch~\cite{paszke2019pytorch} and conducted on a workstation equipped with an NVIDIA A6000 GPU and an AMD EPYC 9224 24-Core processor.
For baseline INR methods (SIREN~\cite{sitzmann2020implicit}, FFN~\cite{tancik2020fourier}, GaussNet~\cite{ramasinghe2022beyond}, and WIRE~\cite{saragadam2023wire}), we use the Adam optimizer~\cite{kingma2014adam} and tune model-specific hyperparameters; detailed configurations are provided in the Appendix~\ref{app:detailed_experimental_setups}.

ELM-INR uses a single-hidden-layer Extreme Learning Machine with ReLU activations for all local models. Other nonlinear activations perform comparably, as shown in Appendix~\ref{app:activation}. Unless explicitly stated otherwise, we use square subdomains with $S_i = 32^2$ and fix the number of hidden nodes to $m = 1024$, shared across all subdomains; results reported without further specification correspond to ELM-INR without BEAM.
Within each subdomain, pixel coordinates are normalized to the range $[-1,1]$ and arranged as
$\mathcal{X} \in \mathbb{R}^d$.
We encode input coordinates using Random Fourier Features (RFF)~\cite{rahimi2007random, tancik2020fourier, he2024random, rudi2017generalization}, following the standard positional encoding configuration of~\cite{tancik2020fourier} (see Appendix~\ref{app:detailed_experimental_setups} for hyperparameter details). The per-subdomain least-squares problems are fully decoupled and are solved in parallel on the GPU, as detailed in Appendix~\ref{app:detailed_experimental_setups}.

\subsection{ELM-INR: Qualitative Reconstruction Results}
\begin{figure}[ht!]
    \centering
    \setlength{\tabcolsep}{2pt}
    \renewcommand{\arraystretch}{1.0}
    \begin{tabular}{cccccc}
        {\footnotesize SIREN}{\scriptsize (30.5 dB)} &
        {\footnotesize FFN}{\scriptsize (25.5 dB)} &
        {\footnotesize GaussNet}{\scriptsize (23.6 dB)} &
        {\footnotesize WIRE}{\scriptsize (28.9 dB)} &
        {\footnotesize \textbf{ELM-INR}}{\scriptsize \textbf{(78.4 dB)}} &
        {\footnotesize GT} \\
        \includegraphics[width=0.155\linewidth]{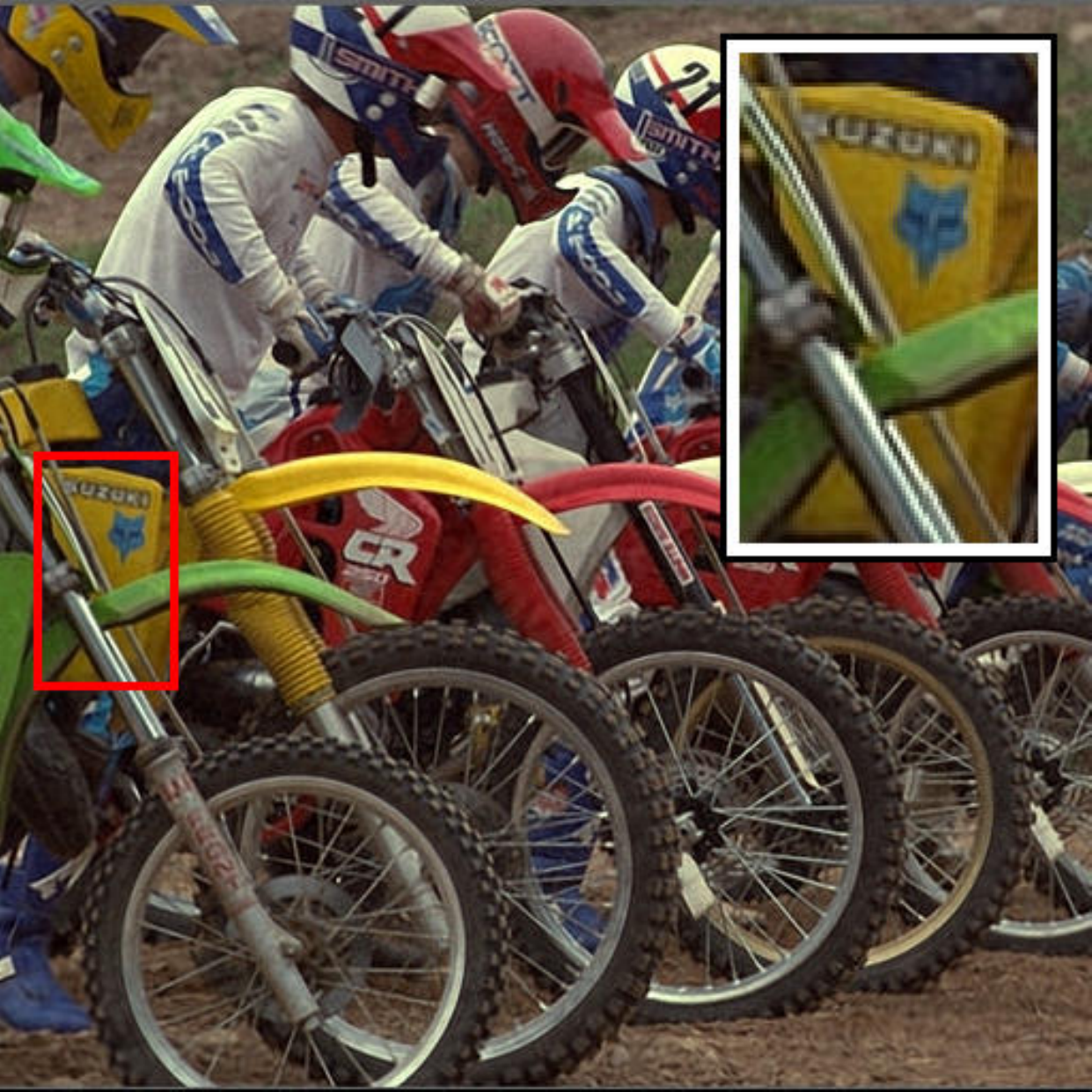} &
        \includegraphics[width=0.155\linewidth]{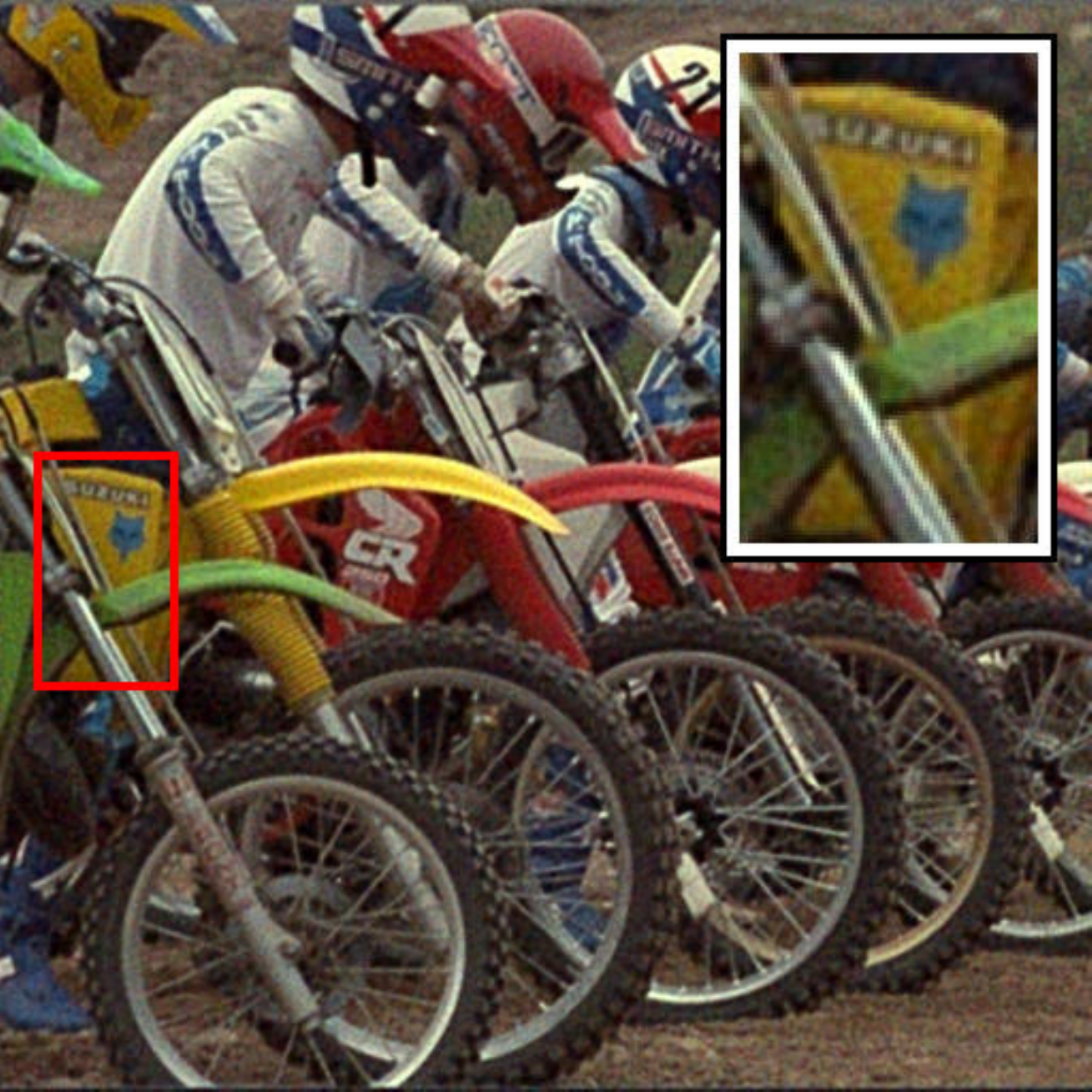} &
        \includegraphics[width=0.155\linewidth]{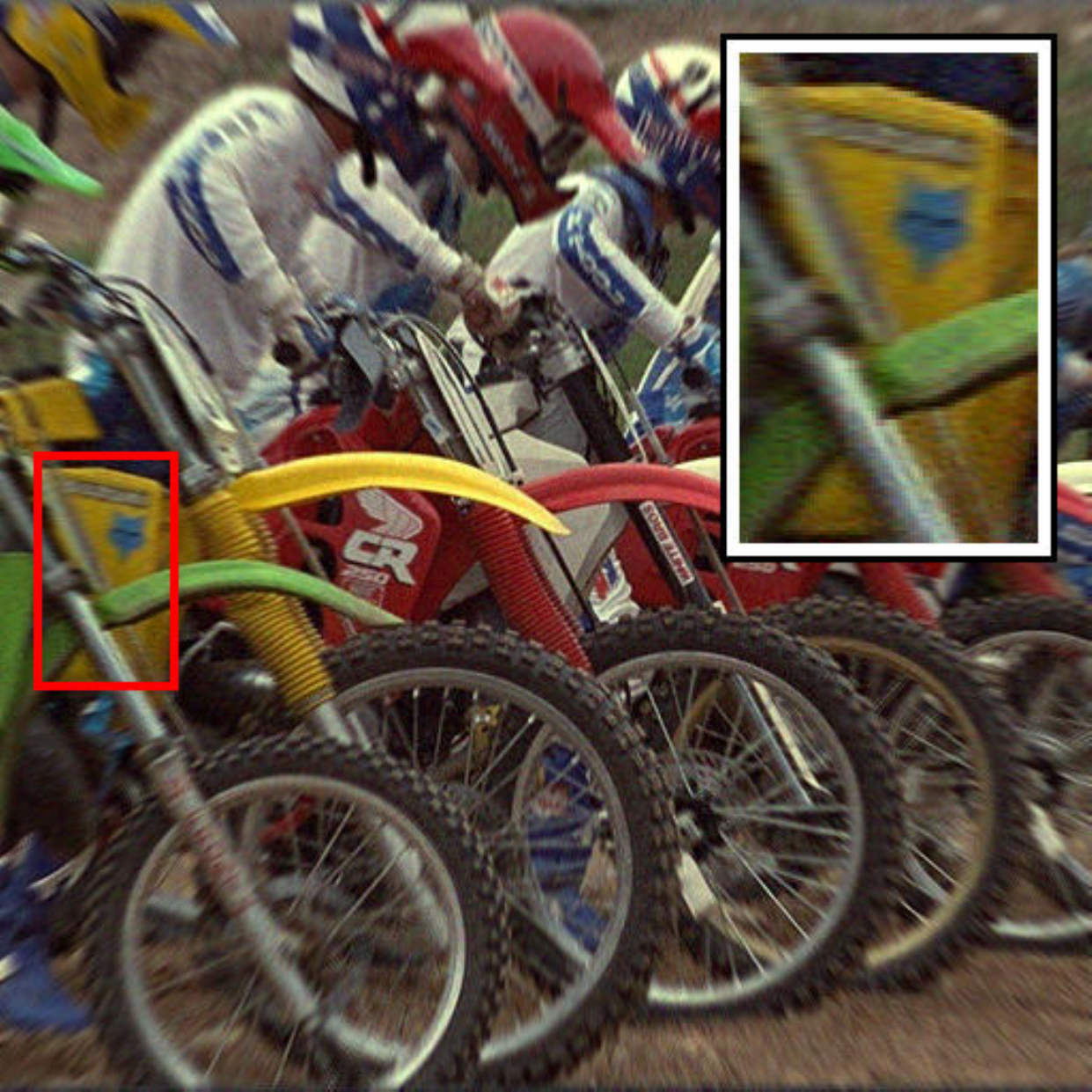} &
        \includegraphics[width=0.155\linewidth]{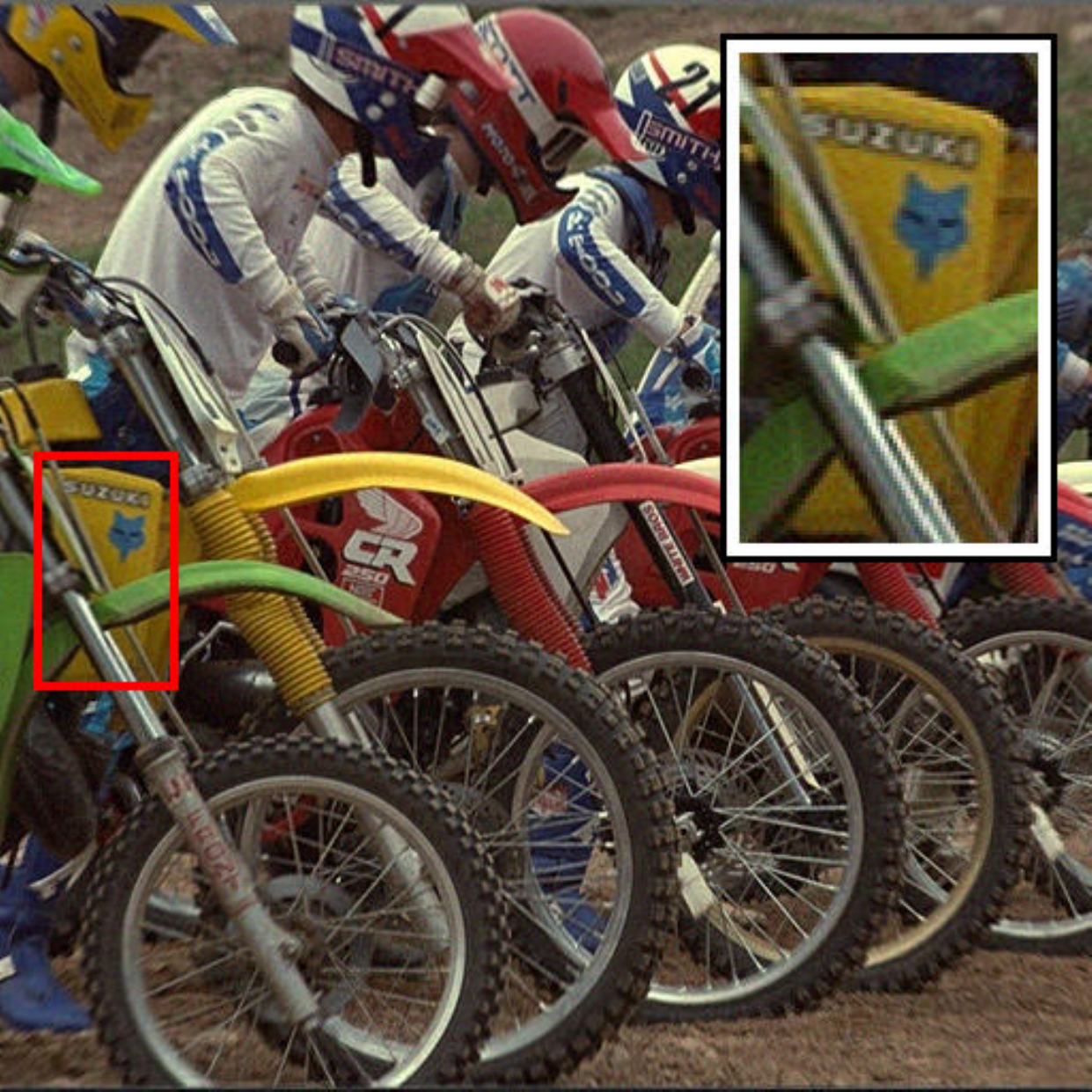} &
        \includegraphics[width=0.155\linewidth]{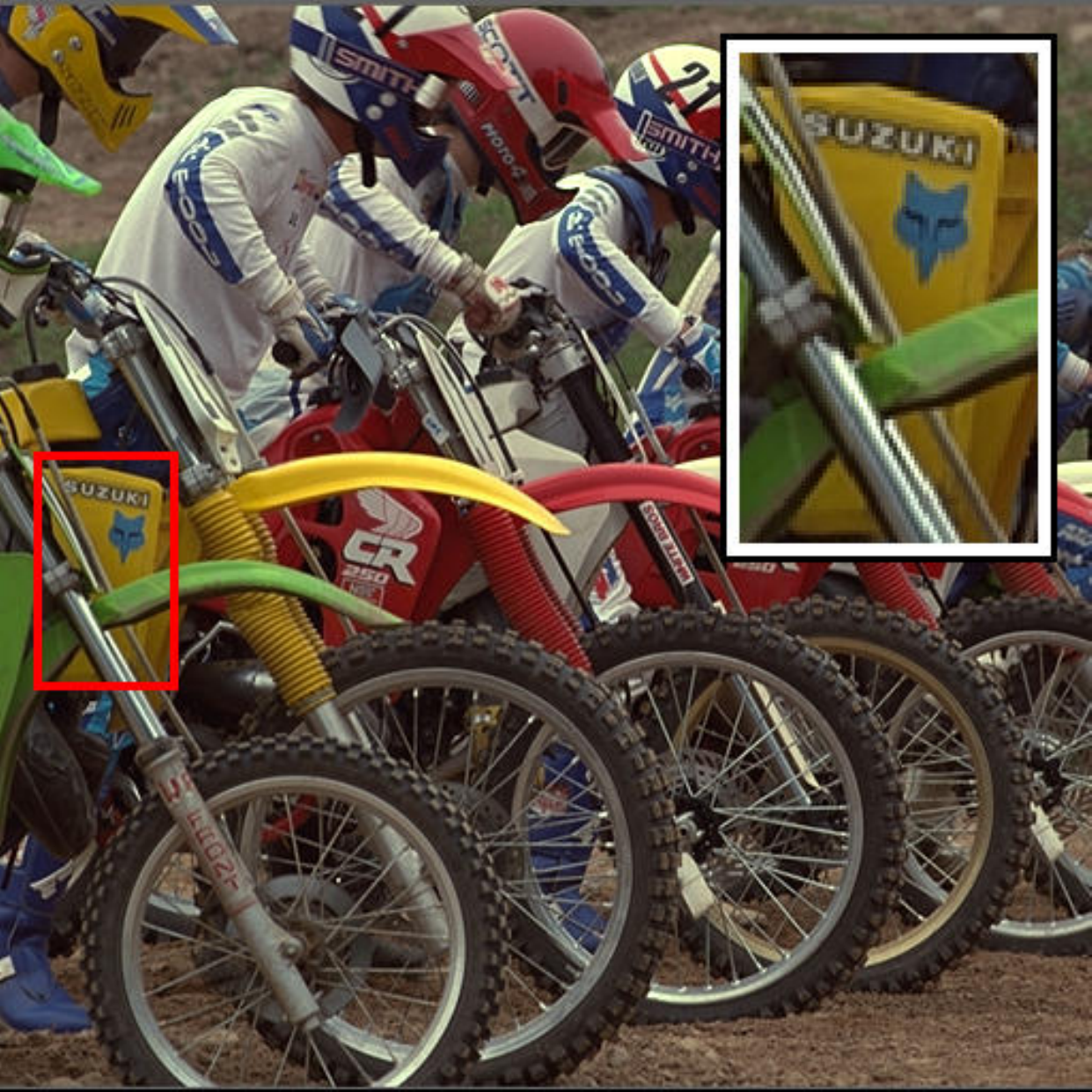} &
        \includegraphics[width=0.155\linewidth]{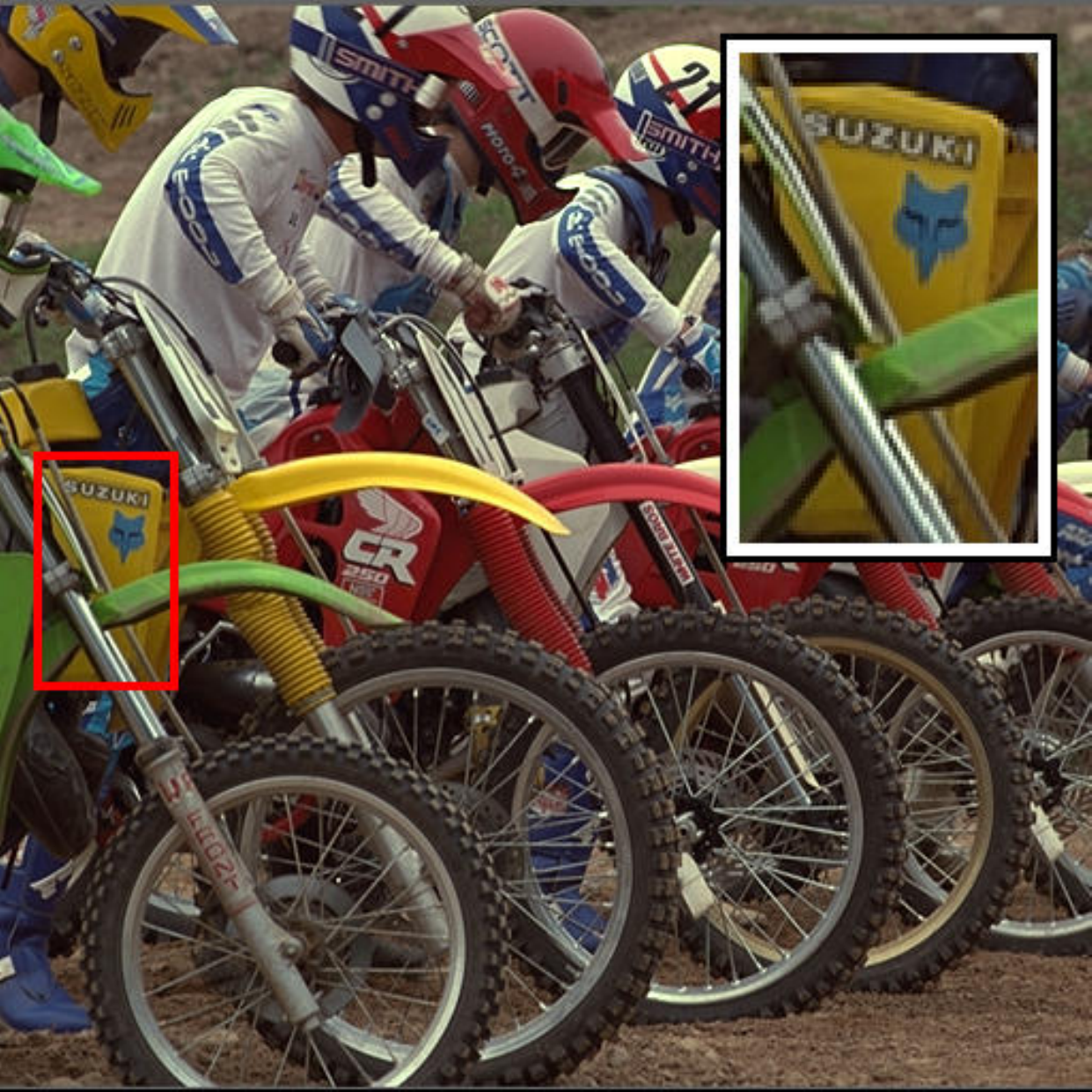}
    \end{tabular}
    \vspace{-0.5pt}
    \caption{
    Qualitative comparison on the \texttt{kodim05} image.
    All baseline models are trained for the same number of epochs.
    }
    \label{fig:kodim05_final_qualitative}
    \vspace{-4pt}
\end{figure}

\paragraph{Qualitative Reconstruction on Standard Image.}
Figure~\ref{fig:kodim05_final_qualitative} shows qualitative reconstruction results on the \texttt{kodim05} image under a standard training protocol.
All models are trained for 1000 epochs to reconstruct a $512 \times 512$ RGB image and are evaluated at convergence. While backpropagation-based INR baselines exhibit residual blurring and fail to recover fine-scale structures, ELM-INR achieves near-perfect reconstruction, faithfully capturing both low-frequency content and high-frequency details, as reflected by its substantially higher PSNR.

\begin{figure}[ht!]
    \centering
    \vspace{-5pt}
    \setlength{\tabcolsep}{2pt}
    \renewcommand{\arraystretch}{1.0}
    \begin{tabular}{cccccc}
    {\footnotesize SIREN}{\scriptsize (33.06 dB)} &
    {\footnotesize FFN}{\scriptsize (38.60 dB)} &
    {\footnotesize GaussNet}{\scriptsize (30.61 dB)} &
    {\footnotesize WIRE}{\scriptsize (30.89 dB)} &
    {\footnotesize \textbf{ELM-INR}}{\scriptsize \textbf{(79.19 dB)}} &
    {\footnotesize GT} \\
    \includegraphics[width=0.155\linewidth]{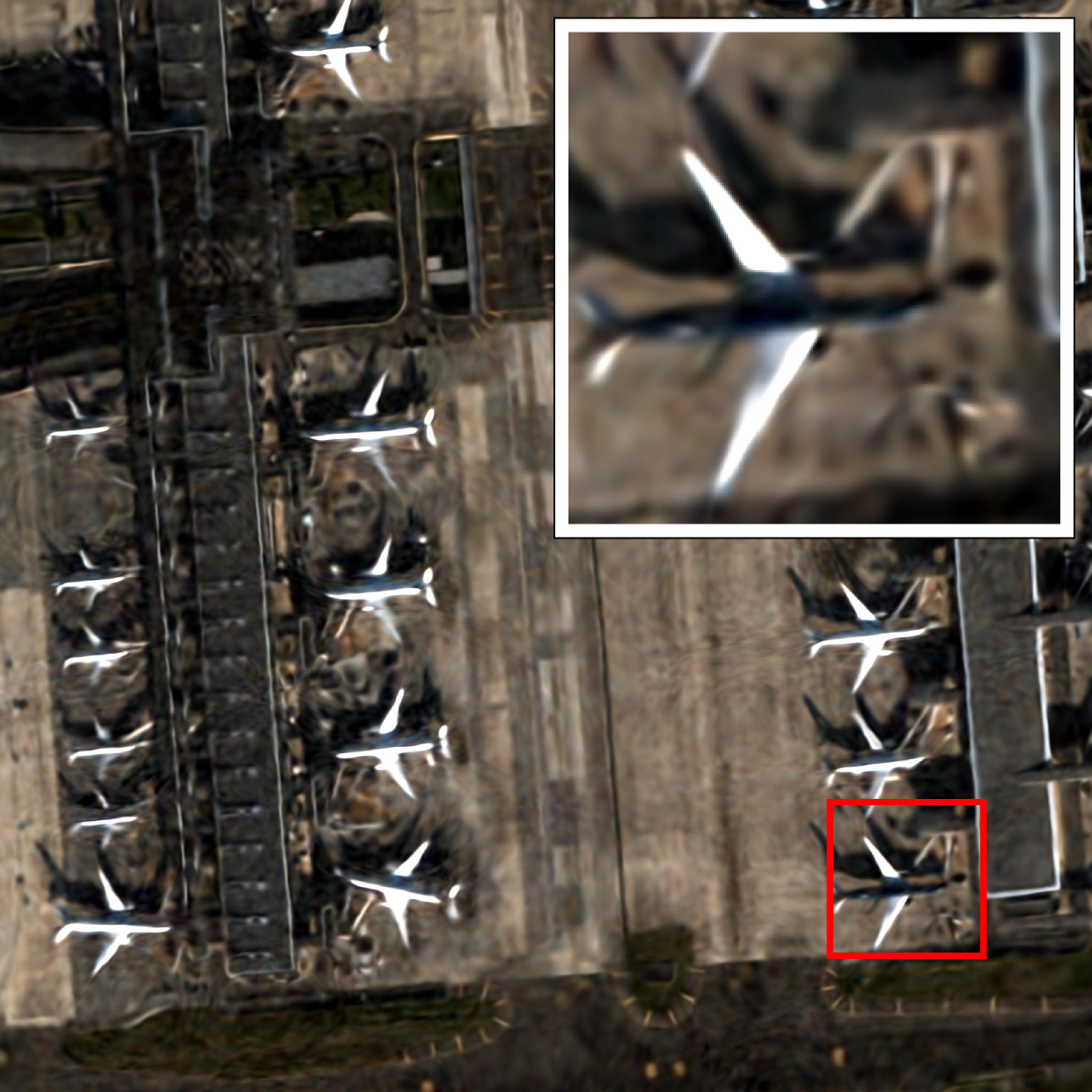} &
    \includegraphics[width=0.155\linewidth]{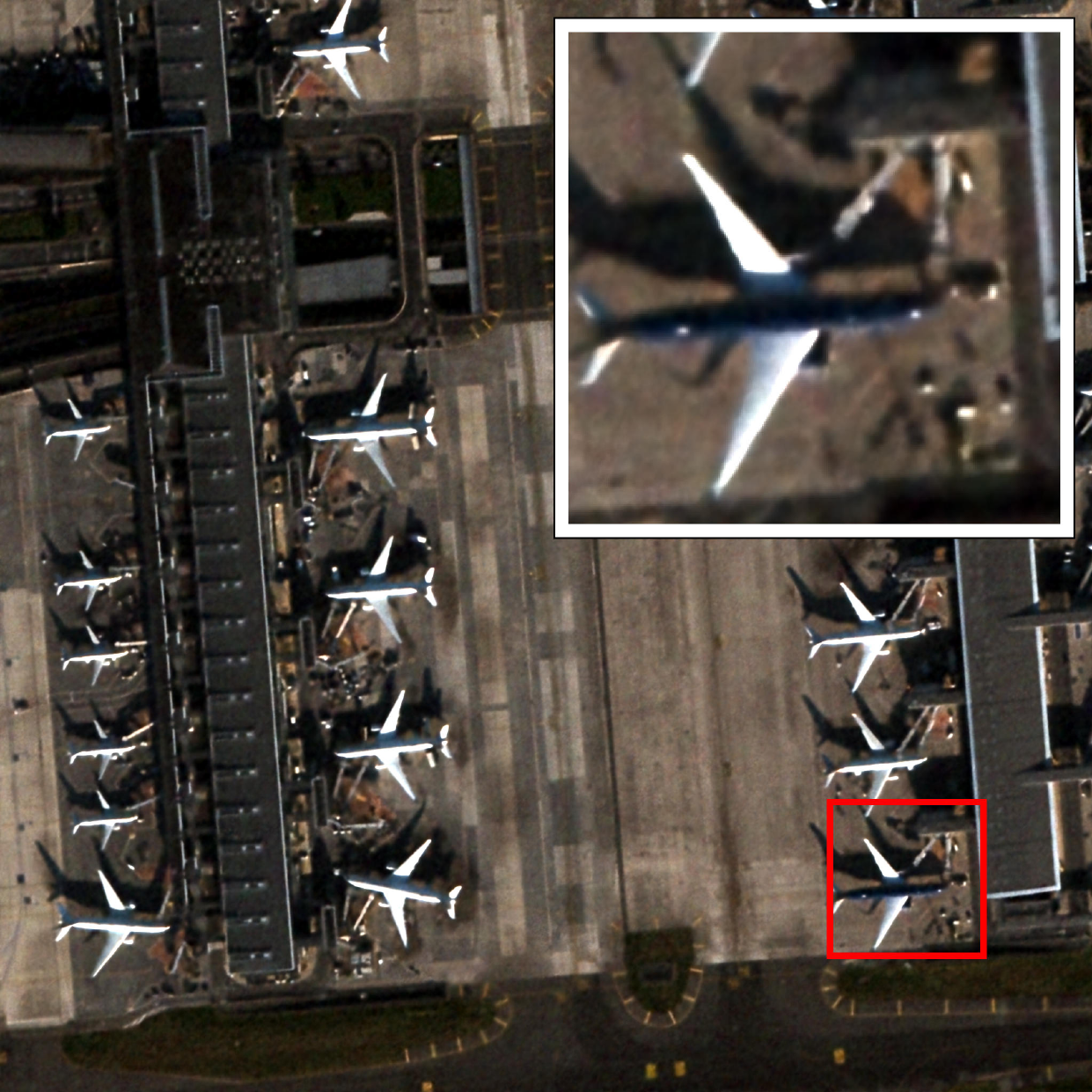} &
    \includegraphics[width=0.155\linewidth]{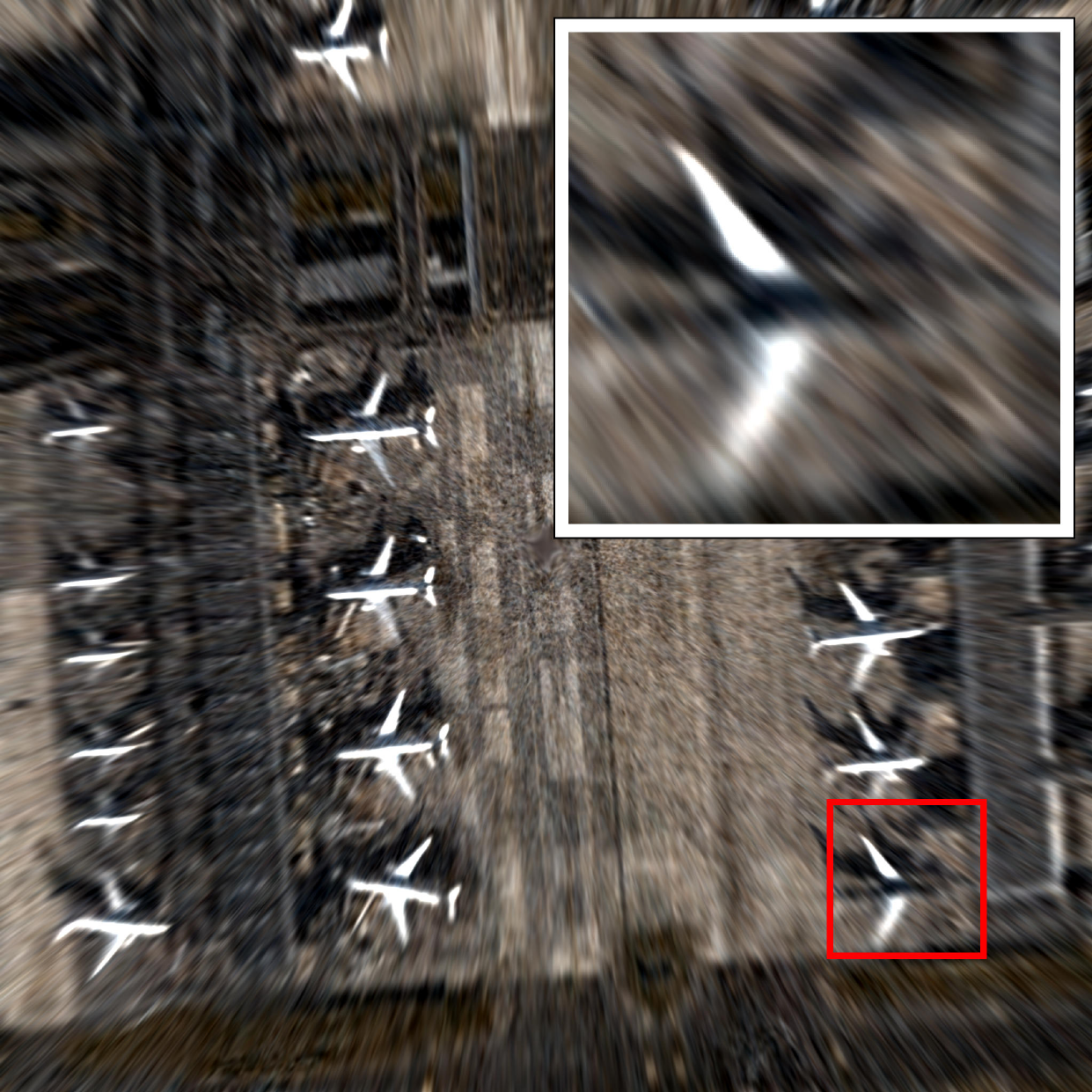} &
    \includegraphics[width=0.155\linewidth]{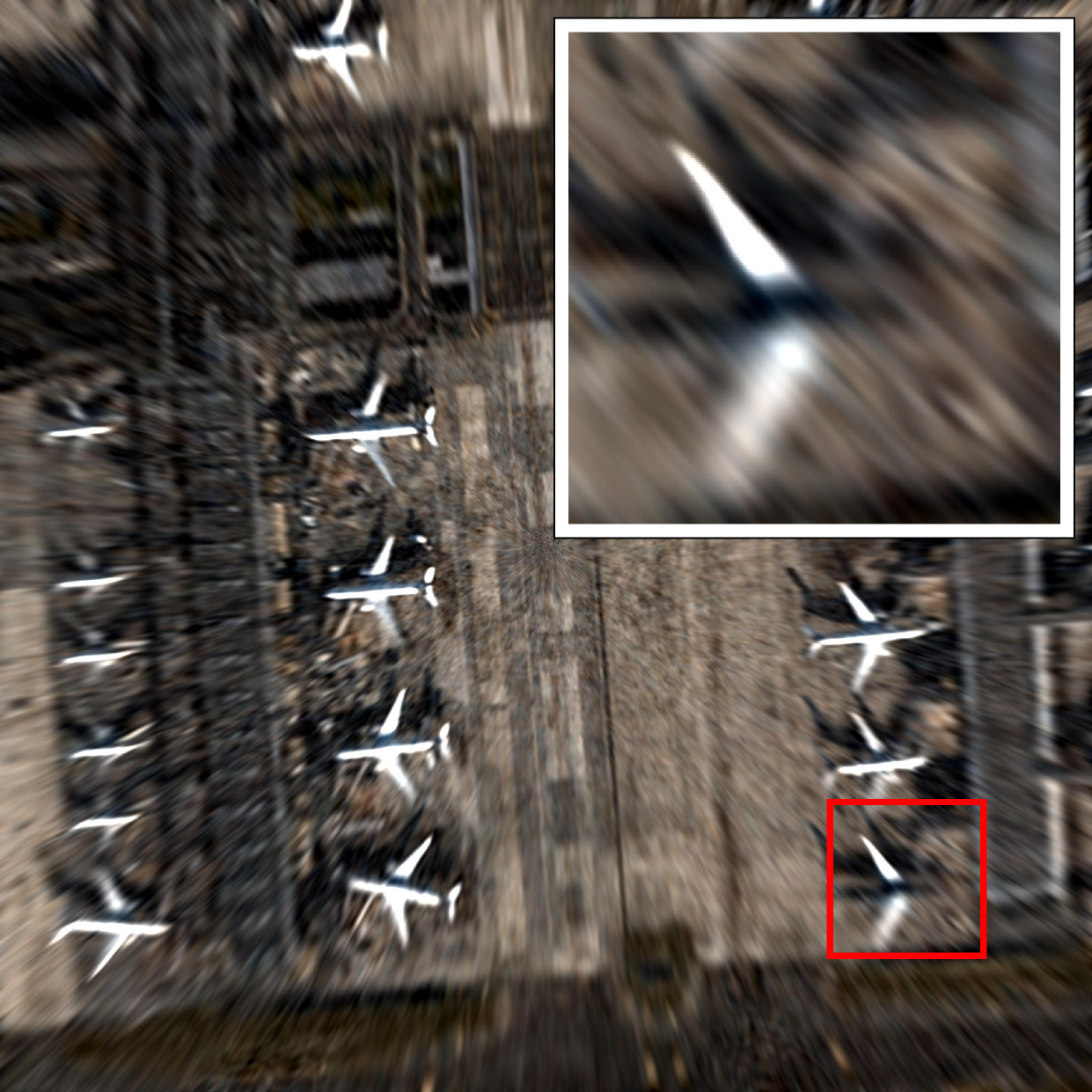} &
    \includegraphics[width=0.155\linewidth]{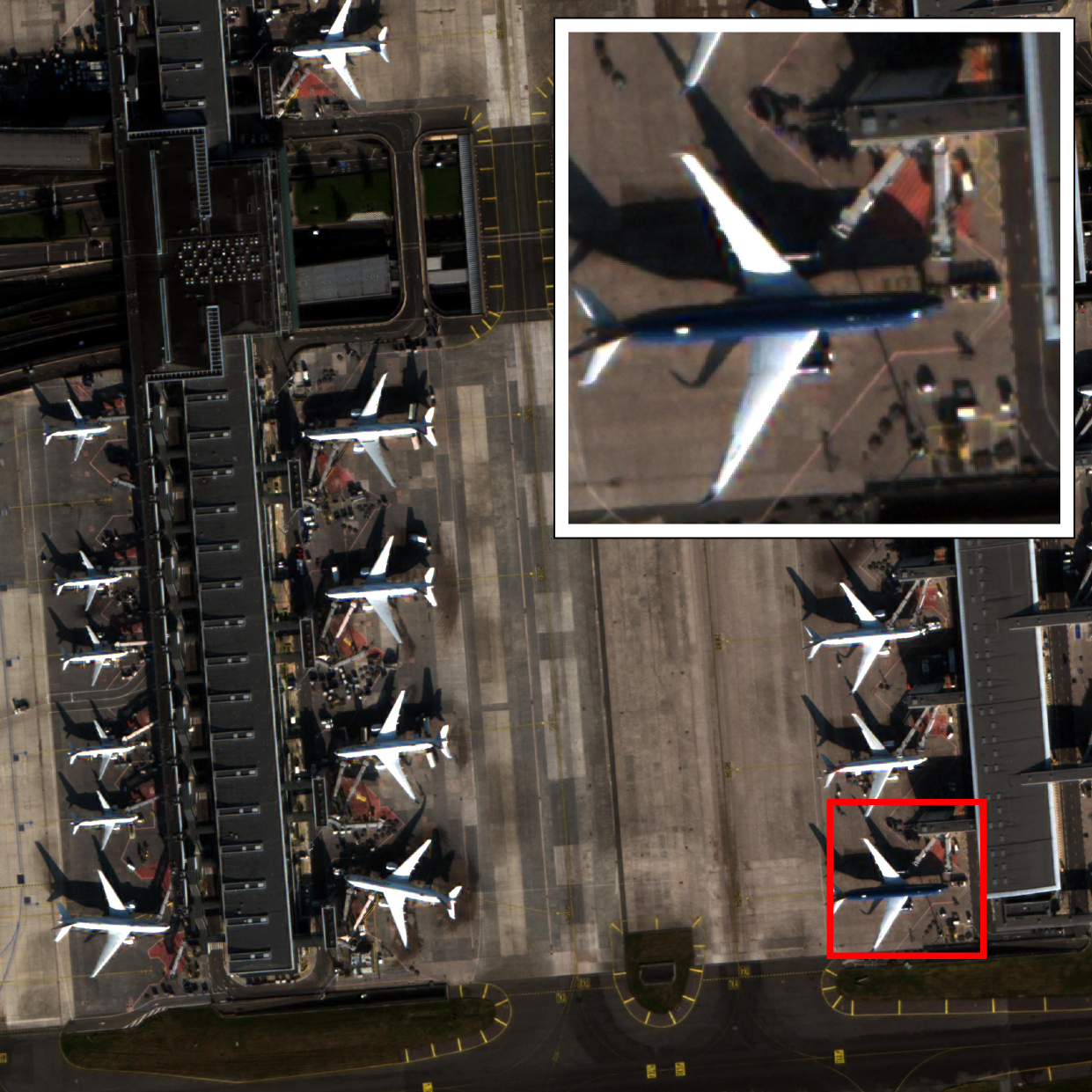} &
    \includegraphics[width=0.155\linewidth]{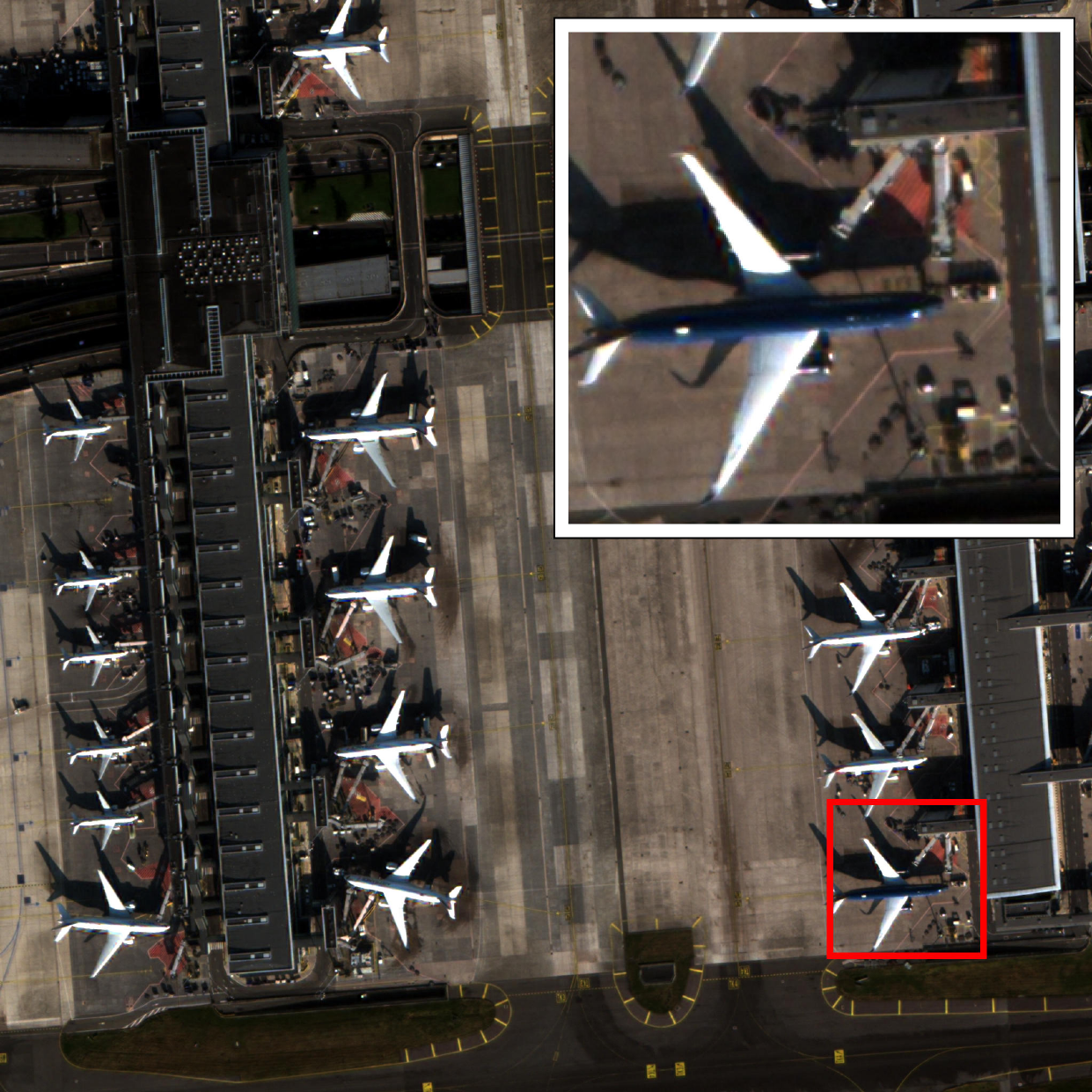}
    \end{tabular}
    \vspace{-0.5pt}
    \caption{
    Qualitative comparison on a high-frequency region from a WorldView-3 multispectral image.
    The image consists of 8 spectral bands with a spatial resolution of $2048 \times 2048$ pixels, while the visualization shows only the RGB channels for clarity.}
    \vspace{-6pt}
    \label{fig:wv3_fr}
\end{figure}

\paragraph{Multispectral Image Reconstruction.}
To further evaluate the robustness of INR models beyond standard single- or three-channel standard images, we extend our experiments to high-resolution multispectral satellite imagery. Figure \ref{fig:wv3_fr} presents a qualitative comparison on a challenging high-frequency crop from a WorldView-3 scene over a Paris Charles de Gaulle Airport (CDG), consisting of 8 spectral bands at a spatial resolution of $2048 \times 2048$. All models are trained to regress the full spectral vector at each spatial coordinate, while only the RGB composite is visualized for clarity. For ELM-INR, we use $S_i = 64^2$ samples per subdomain and an ELM with $m = 4096$ hidden units.

This setting is substantially more demanding than conventional image reconstruction, as the model must simultaneously capture fine-grained spatial details and complex cross-band correlations. As shown in Figure \ref{fig:wv3_fr}, backpropagation-based INR baselines (SIREN, FFN, GaussNet, and WIRE) suffer from pronounced blurring, texture loss, and distortion in high-frequency regions such as edges and man-made structures. These artifacts are indicative of both spectral bias and optimization difficulties when scaling INR training to high-resolution, multi-channel signals.

In contrast, ELM-INR achieves near-perfect reconstruction, faithfully preserving sharp edges and fine structural details across the scene. The substantial PSNR gap highlights the effectiveness of the proposed closed-form, backpropagation-free formulation in handling spectrally complex, multi-band data. These results demonstrate that the advantages of ELM-INR are not limited to standard RGB imagery, but extend naturally to realistic multispectral remote sensing scenarios where iterative INR training becomes increasingly brittle and expensive.

\begin{figure}[t]
    \centering
    \setlength{\tabcolsep}{2pt}
    \renewcommand{\arraystretch}{1.0}
    \begin{tabular}{cccccc}
        {\footnotesize SIREN} {\scriptsize(4.22e-3)} &
        {\footnotesize FFN} {\scriptsize(7.02e-2)} &
        {\footnotesize GaussNet} {\scriptsize(2.36e-2)} &
        {\footnotesize WIRE} {\scriptsize(2.09e-2)} &
        {\footnotesize ELM-INR} {\scriptsize(5.68e-5)} &
        {\footnotesize GT} \\
        
        \includegraphics[width=0.155\linewidth]{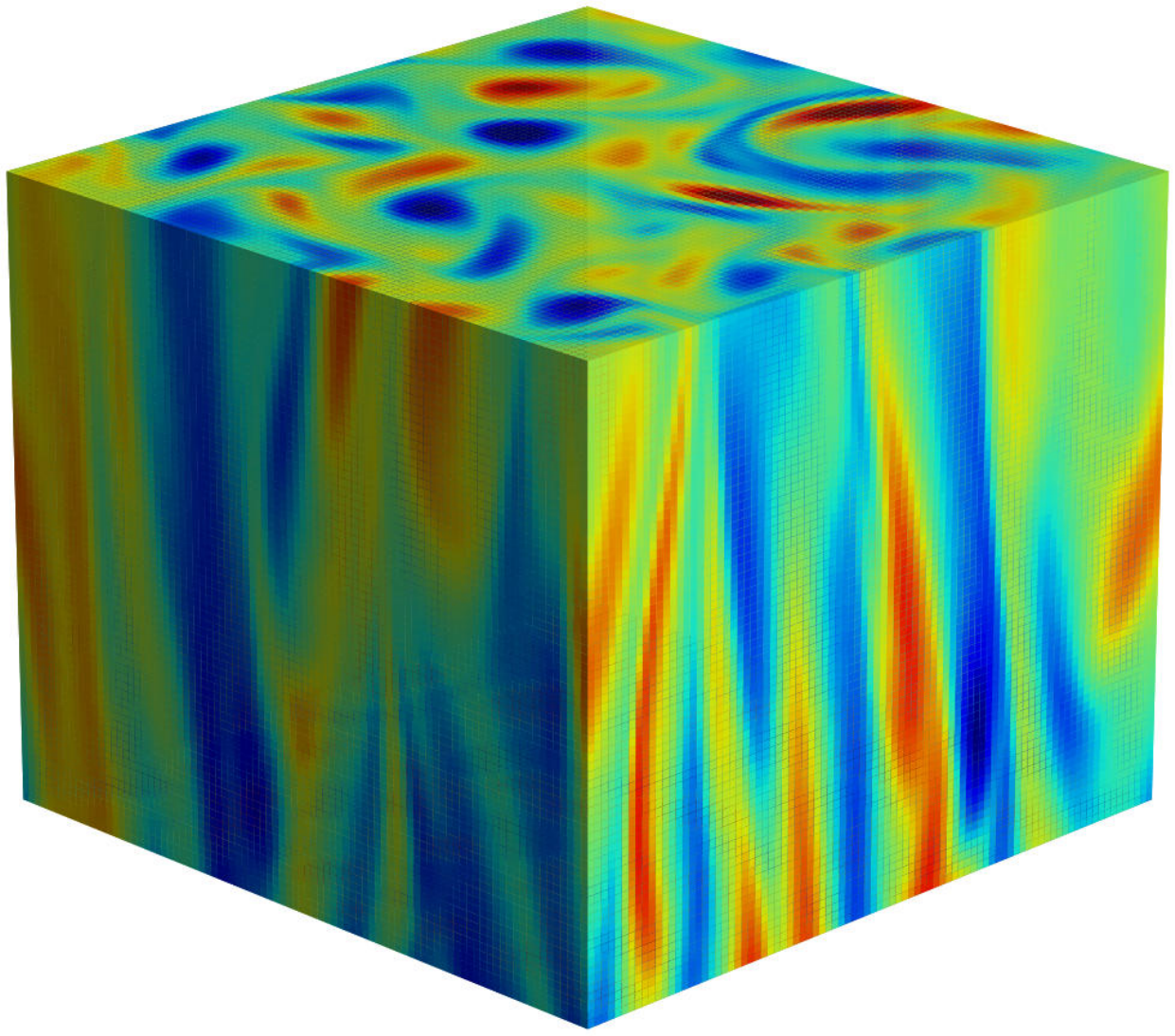} &
        \includegraphics[width=0.155\linewidth]{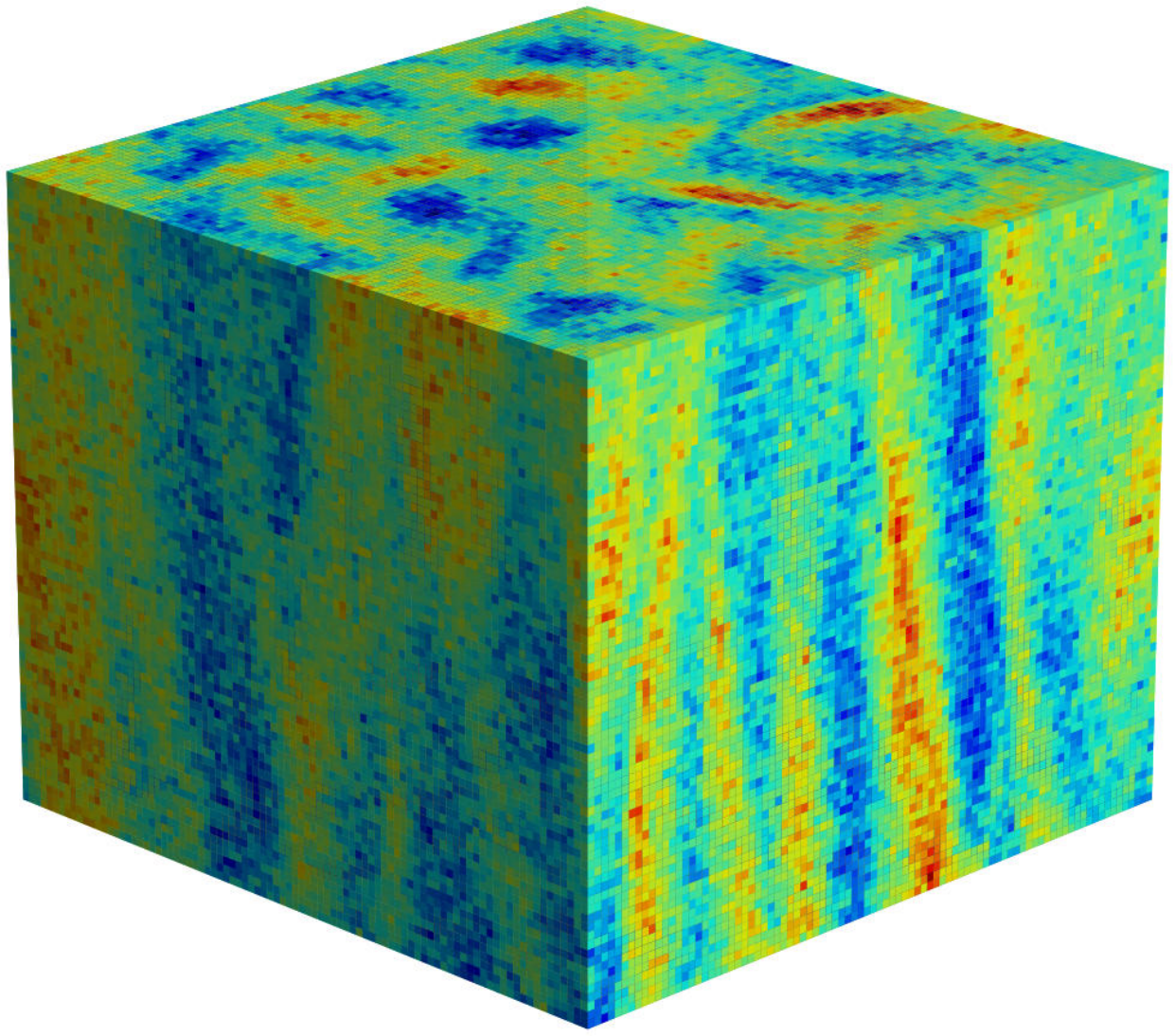} &
        \includegraphics[width=0.155\linewidth]{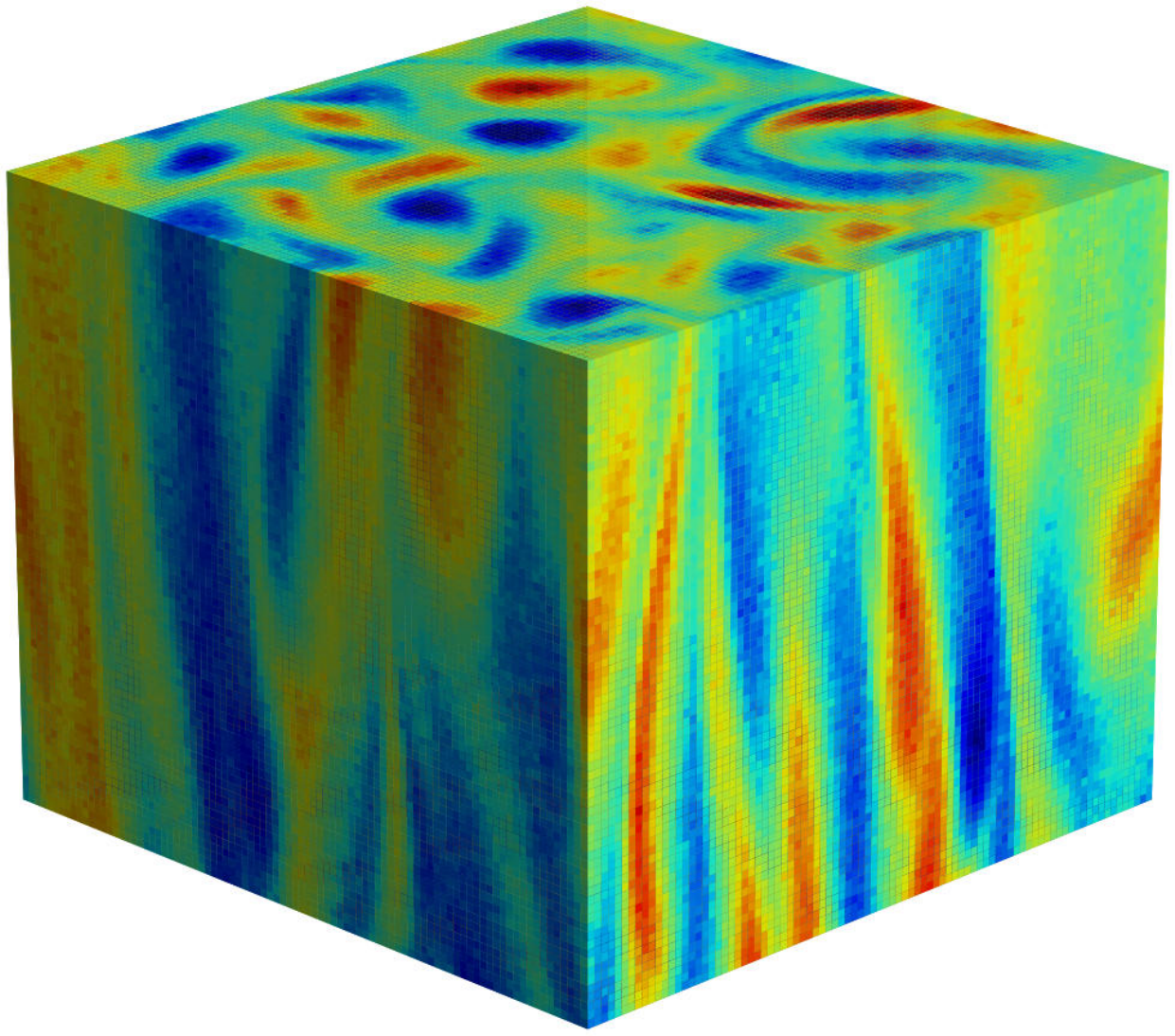} &
        \includegraphics[width=0.155\linewidth]{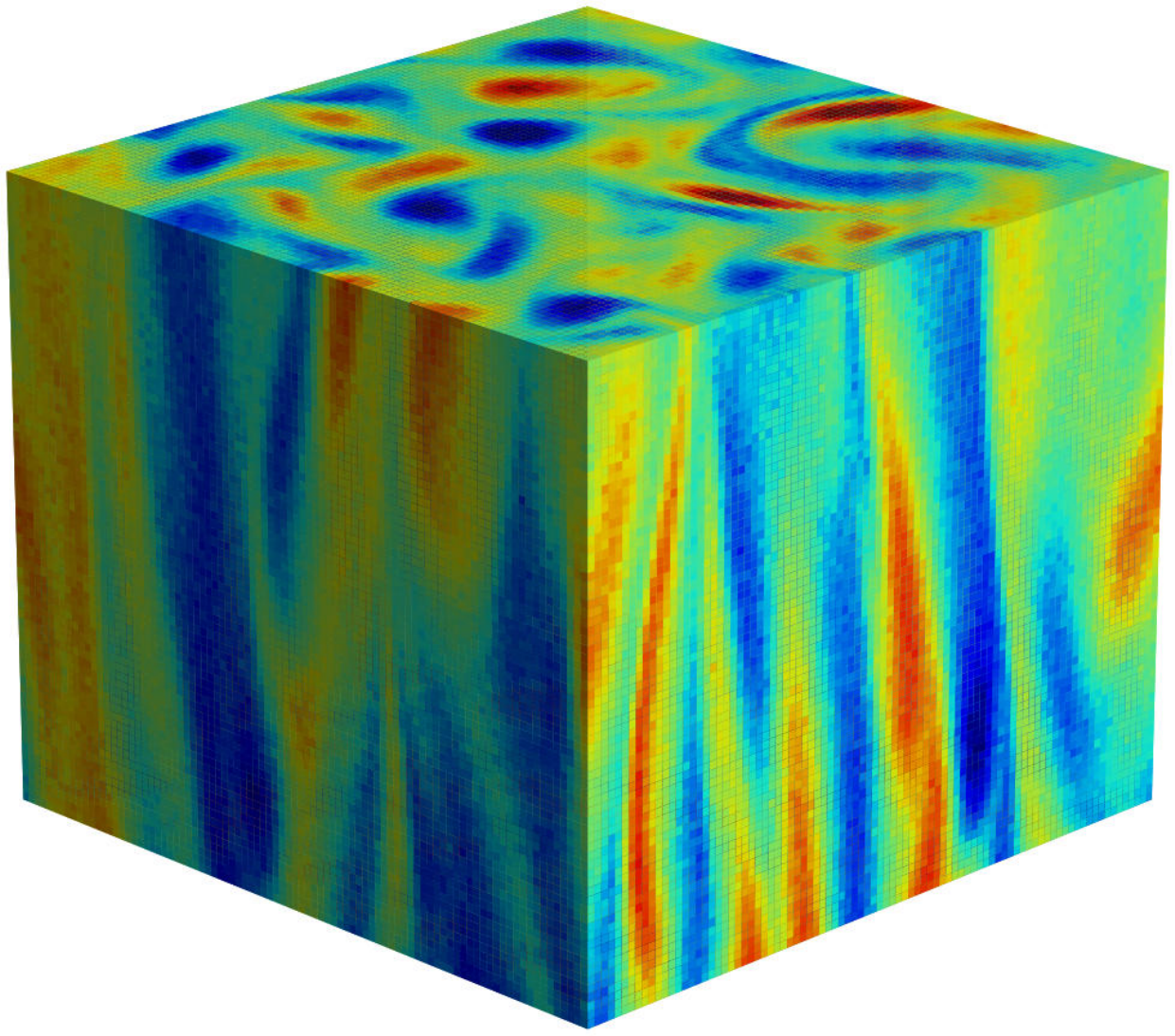} &
        \includegraphics[width=0.155\linewidth]{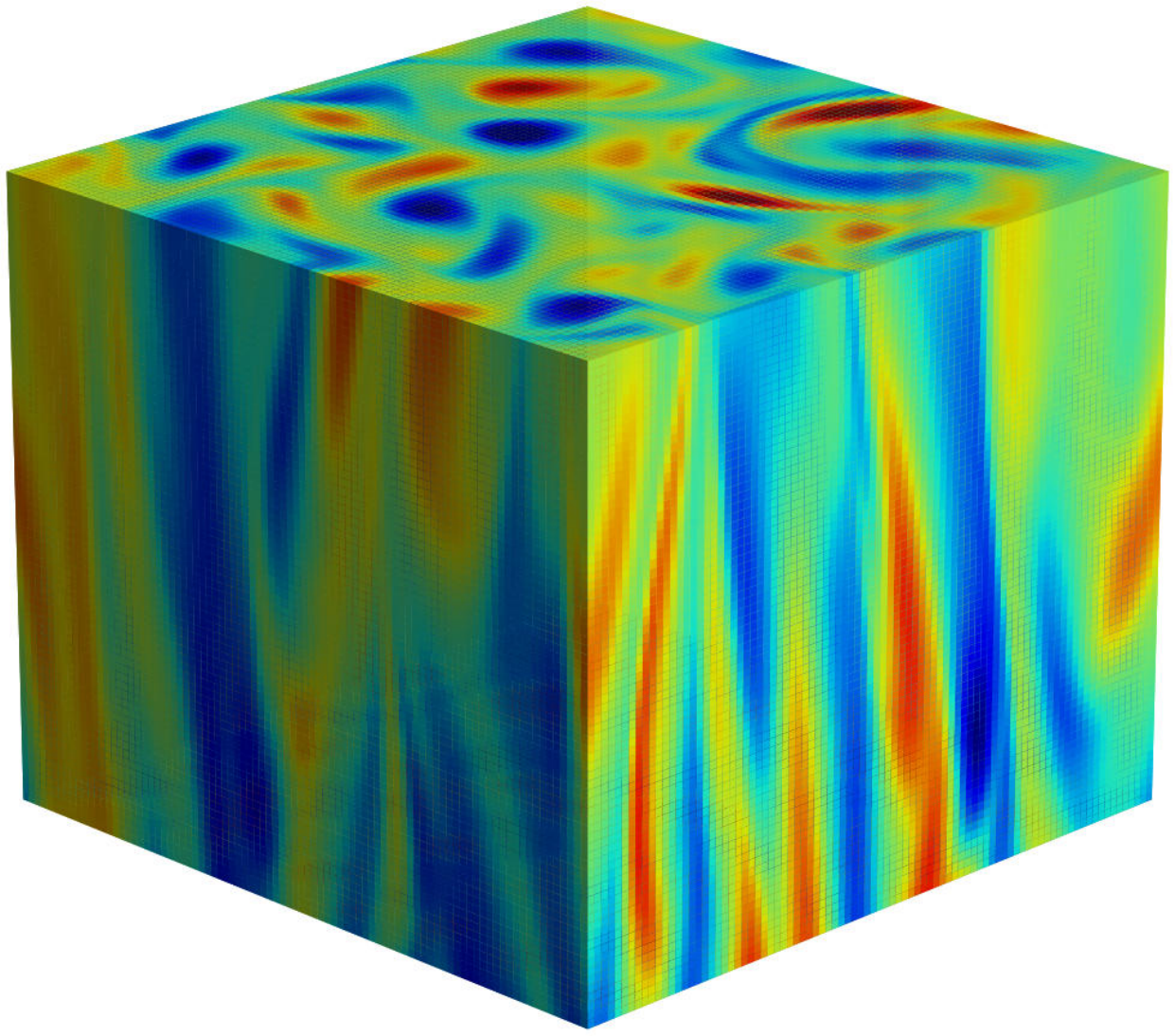} &
        \includegraphics[width=0.155\linewidth]{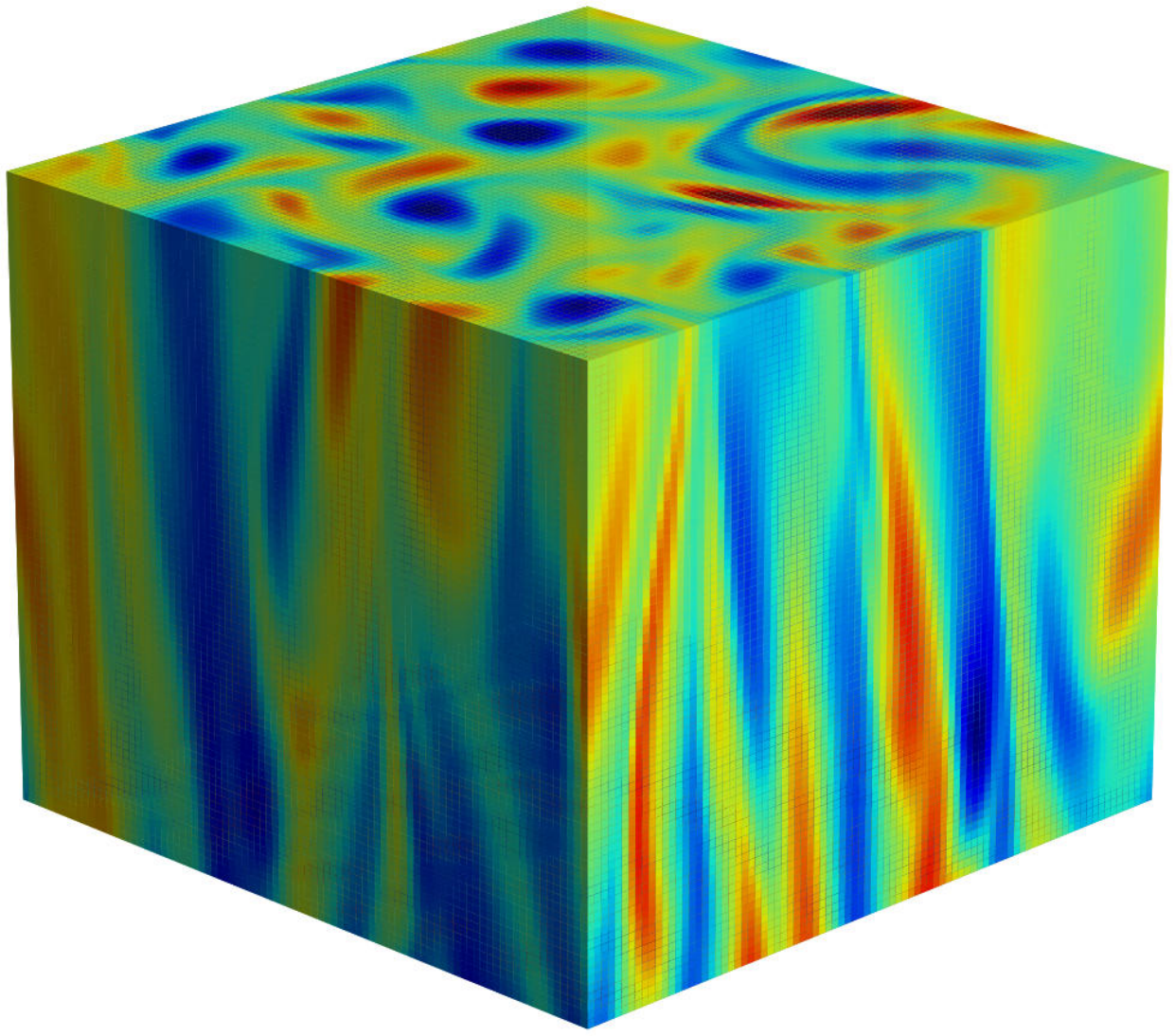}
    \end{tabular}
    \caption{\textbf{Navier--Stokes equation volume reconstruction results.} The horizontal axes are spatial, and the height corresponds to time. Numbers in parentheses denote MAE.}
    \label{fig:pde_3d}
\end{figure}

\paragraph{Spatiotemporal PDE Reconstruction.}
We further evaluate the expressiveness of INR models on a three-dimensional spatiotemporal signal derived from the Navier--Stokes equations. Figure~\ref{fig:pde_3d} compares volume reconstruction results, where the two horizontal axes correspond to spatial dimensions and the vertical axis represents time. This task is particularly challenging due to the presence of multi-scale structures and rapidly varying dynamics across both space and time.

For this experiment, ELM-INR is configured differently from the image-based settings to explicitly account for temporal structure. Since the temporal dimension contains 100 time steps, the domain is partitioned not only spatially but also along the time axis, with temporal patches of size 10.
Each spatiotemporal subdomain uses $S_i = 64^2$ samples and an ELM with $m = 4096$ hidden units, allowing the local models to capture fine-grained temporal variations while maintaining numerical stability.

As shown in Figure~\ref{fig:pde_3d}, backpropagation-based INR baselines fail to recover fine-scale temporal structures. In contrast, ELM-INR achieves a near-exact reconstruction of the full volume, yielding orders-of-magnitude lower error. These results indicate that the proposed closed-form, partitioned representation naturally extends to complex spatiotemporal fields governed by physical dynamics.

\paragraph{Training Efficiency.}
We analyze reconstruction quality as a function of wall-clock time to compare optimization efficiency across methods.
Figure~\ref{fig:time_psnr_kodim05} reports PSNR versus runtime on the same image.
Backpropagation-based INR baselines improve gradually due to iterative optimization and converge to moderate reconstruction quality.
In contrast, ELM-INR reaches very high PSNR within a short runtime, as the solution is obtained via closed-form linear solves rather than iterative gradient updates.

Across different configurations of ELM-INR, we observe consistently fast convergence and high reconstruction quality.
Both settings achieve high PSNR within a short runtime, indicating that ELM-INR remains efficient even when varying the number of subdomains and model capacity.
These results further emphasize that the proposed method is not only substantially faster than existing INRs, but also robust to reasonable choices of its internal configuration.

\subsection{Ablation Study}
\begin{wrapfigure}{r}{0.6\textwidth}
    \centering
    \vspace{-12pt}
    \begin{subfigure}[t]{0.47\linewidth}
        \centering
        \includegraphics[width=\linewidth]{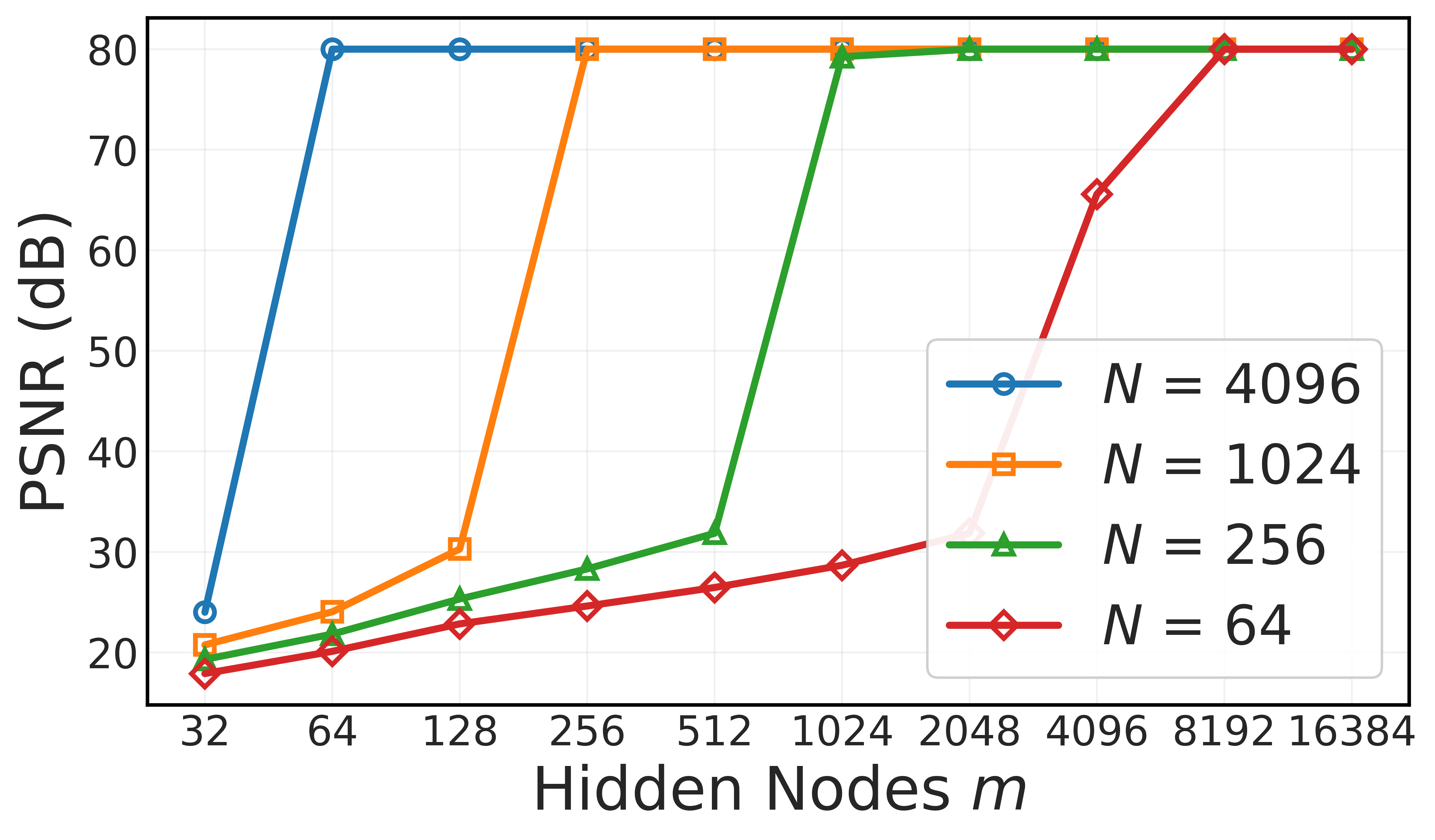}
        \caption{RFF (w)}
        \label{fig:ablation_hidden_nodes_subdomains_ff}
    \end{subfigure}
    \hfill
    \begin{subfigure}[t]{0.47\linewidth}
        \centering
        \includegraphics[width=\linewidth]{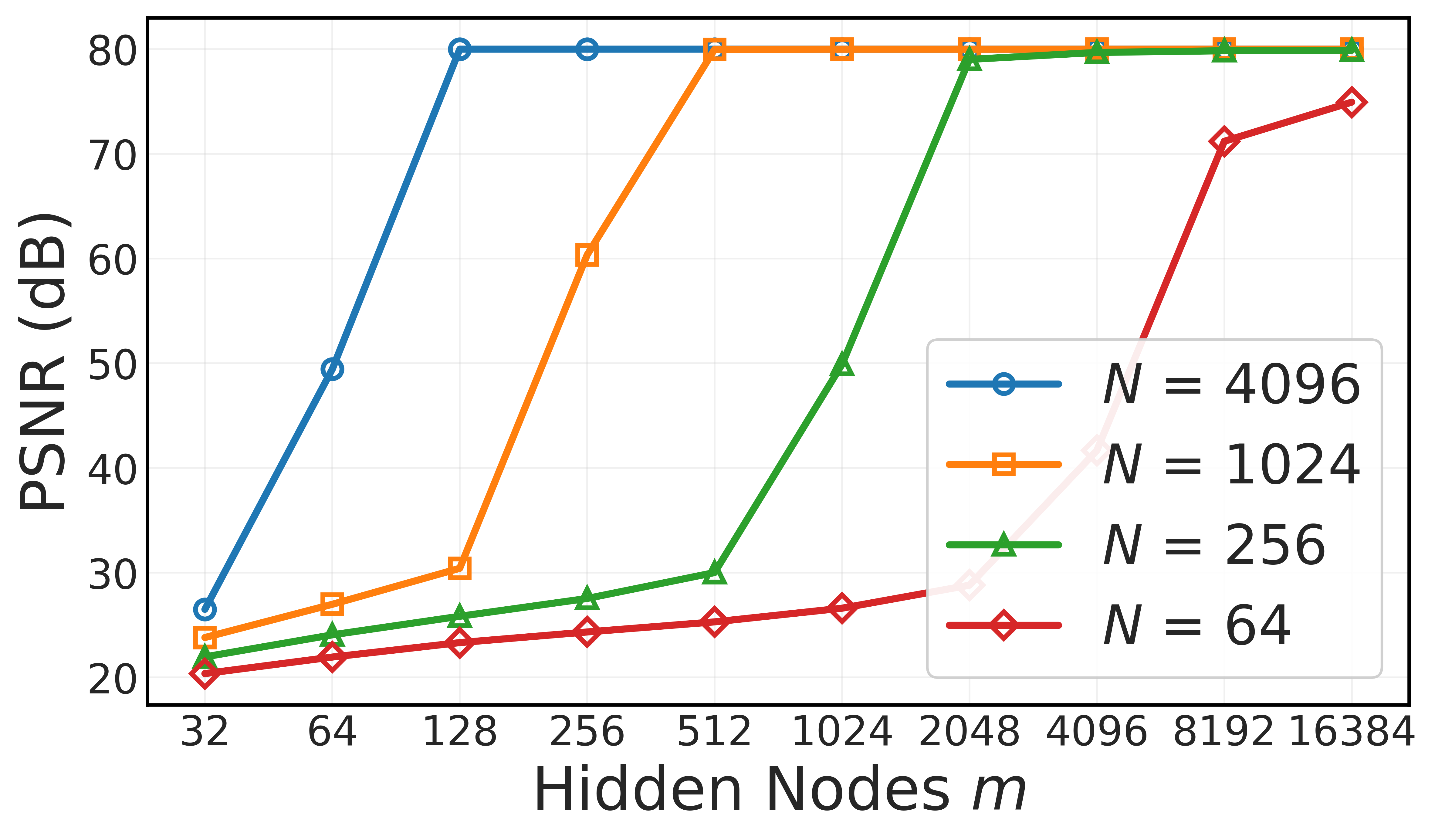}
        \caption{RFF (w/o)}
        \label{fig:ablation_hidden_nodes_subdomains_noff}
    \end{subfigure}
    \vspace{-4pt}
    \caption{
    PSNR versus the number of hidden nodes $m$ for different numbers of subdomains $N$.
    }
    \vspace{-10pt}
    \label{fig:ablation_hidden_nodes_subdomains}
\end{wrapfigure}

We conduct an ablation study on the number of basis functions, implemented as the number of hidden nodes $m$, under different numbers of subdomains $N$, for ELM-INR using the \texttt{cameraman} image at a resolution of $512 \times 512 \times 1$. Figure~\ref{fig:ablation_hidden_nodes_subdomains} reports the reconstruction performance in terms of PSNR as $m$ increases for each choice of $N$ and input representation.

Across all settings, increasing $m$ consistently improves reconstruction quality, but the rate at which performance saturates depends strongly on both the input encoding and the degree of domain decomposition.
When Fourier feature mapping is employed (Figure~\ref{fig:ablation_hidden_nodes_subdomains_ff}), PSNR saturates rapidly as $m$ increases, indicating that relatively few basis functions are sufficient to capture high-frequency content.
In contrast, when raw coordinates are used without Fourier features (Figure~\ref{fig:ablation_hidden_nodes_subdomains_noff}), PSNR improves more gradually and requires substantially larger hidden widths to reach saturation.

The effect of domain decomposition is particularly evident when the number of subdomains is small, where each local model must approximate a more complex signal.
In this regime, the absence of Fourier features leads to a shift of the saturation point toward larger values of $m$, reflecting the increased representational burden.
Overall, Fourier feature mapping significantly reduces the required model capacity, while domain decomposition further simplifies local approximation tasks.

Notably, saturation is typically reached at $m \approx |\Omega_i|$ across all configurations, providing a simple practical guideline for choosing model capacity and reducing the need for hyperparameter tuning.

\subsection{Effect of BEAM on ELM-INR}

\begin{figure}[t]
\centering
\begin{minipage}[t]{0.48\linewidth}
    \centering
    \captionsetup[subfigure]{font=scriptsize}
    \begin{subfigure}{0.31\linewidth}
        \centering
        \includegraphics[width=\linewidth]{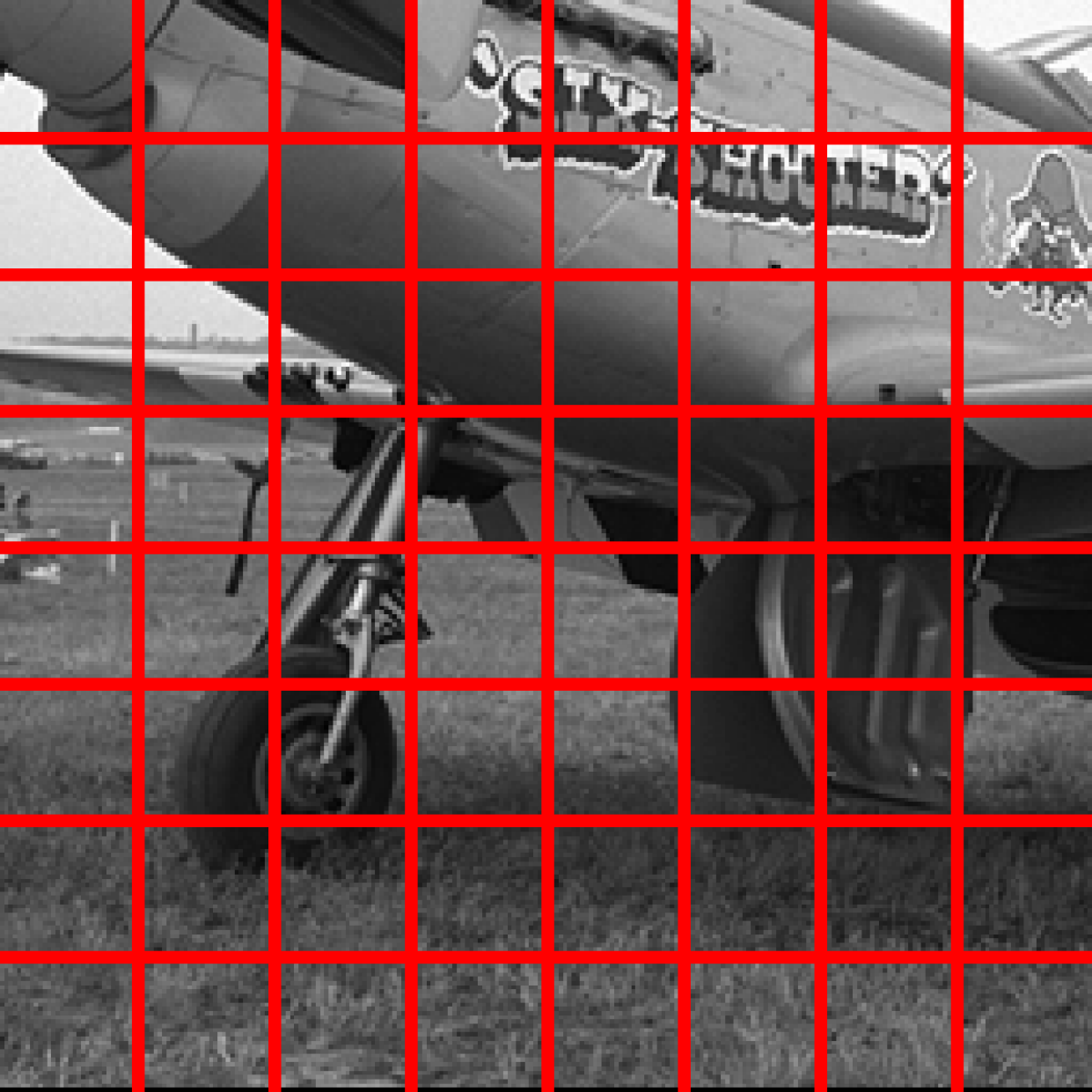}
        \caption{Regular Mesh}
        \label{fig:kodim20_beam:a}
    \end{subfigure}
    \hfill
    \begin{subfigure}{0.31\linewidth}
        \centering
        \includegraphics[width=\linewidth]{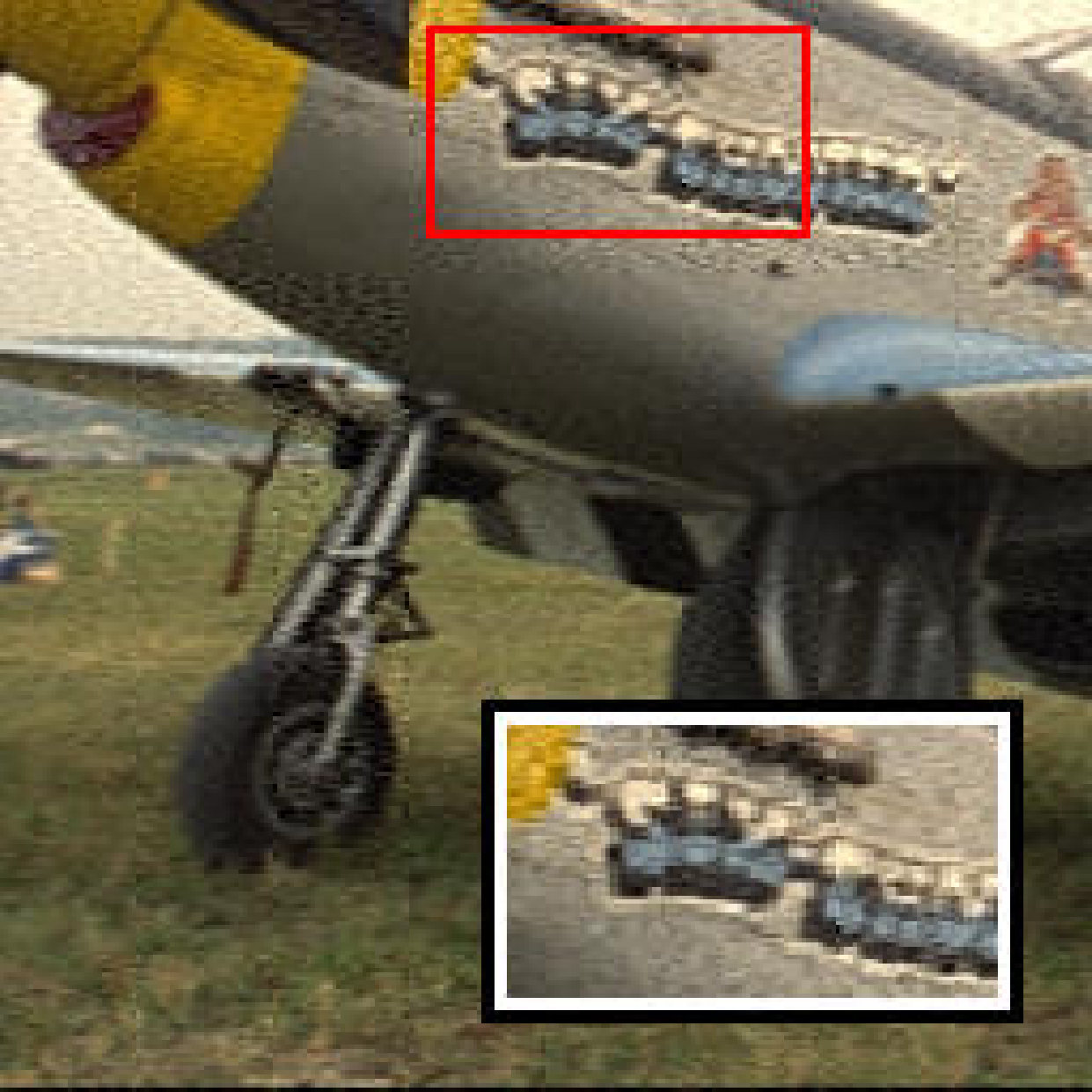}
        \caption{$m$=256 (26.0 dB)}
        \label{fig:kodim20_beam:b}
    \end{subfigure}
    \hfill
    \begin{subfigure}{0.31\linewidth}
        \centering
        \includegraphics[width=\linewidth]{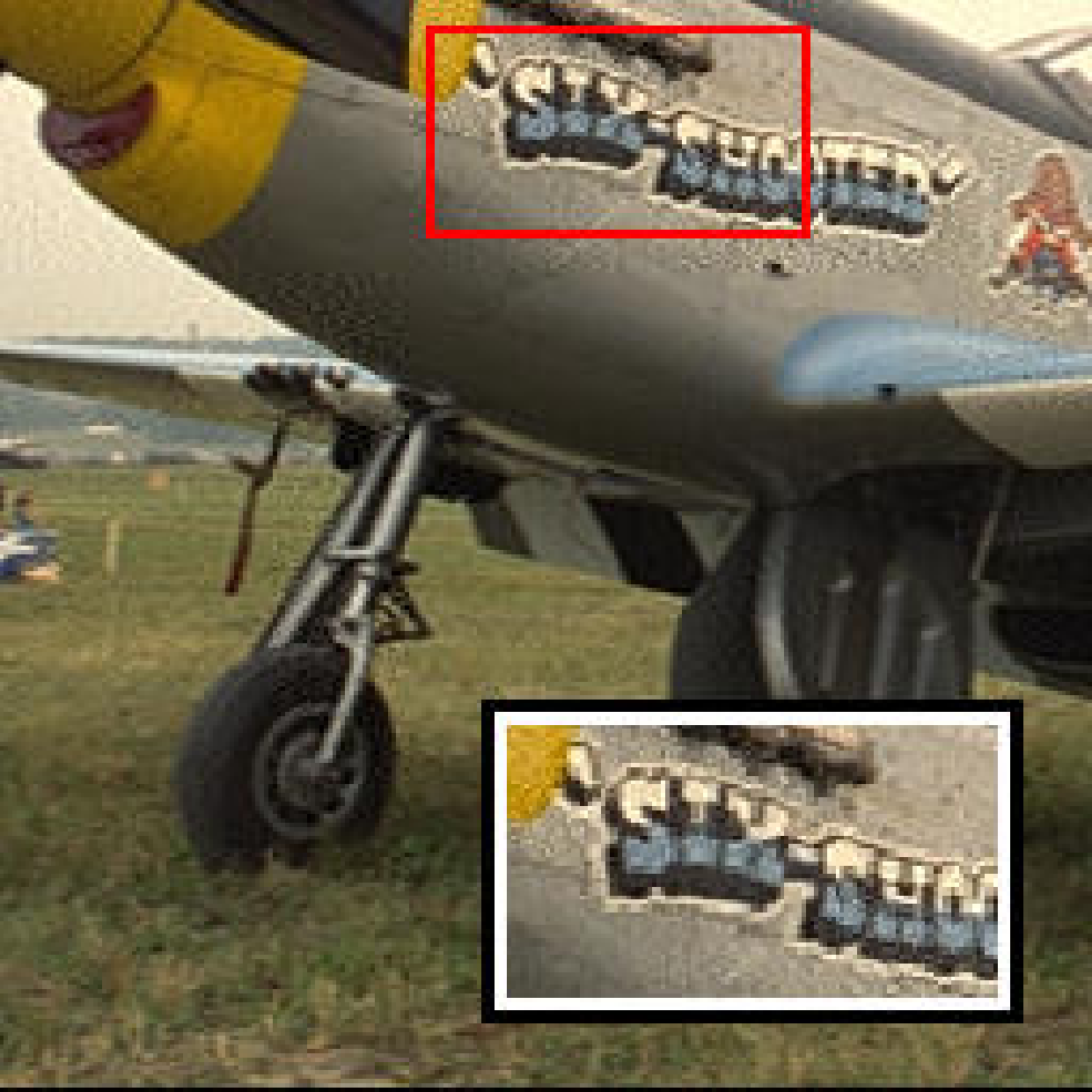}
        \caption{$m$=512 (29.6 dB)}
        \label{fig:kodim20_beam:c}
    \end{subfigure}
    
    \begin{subfigure}{0.31\linewidth}
        \centering
        \includegraphics[width=\linewidth]{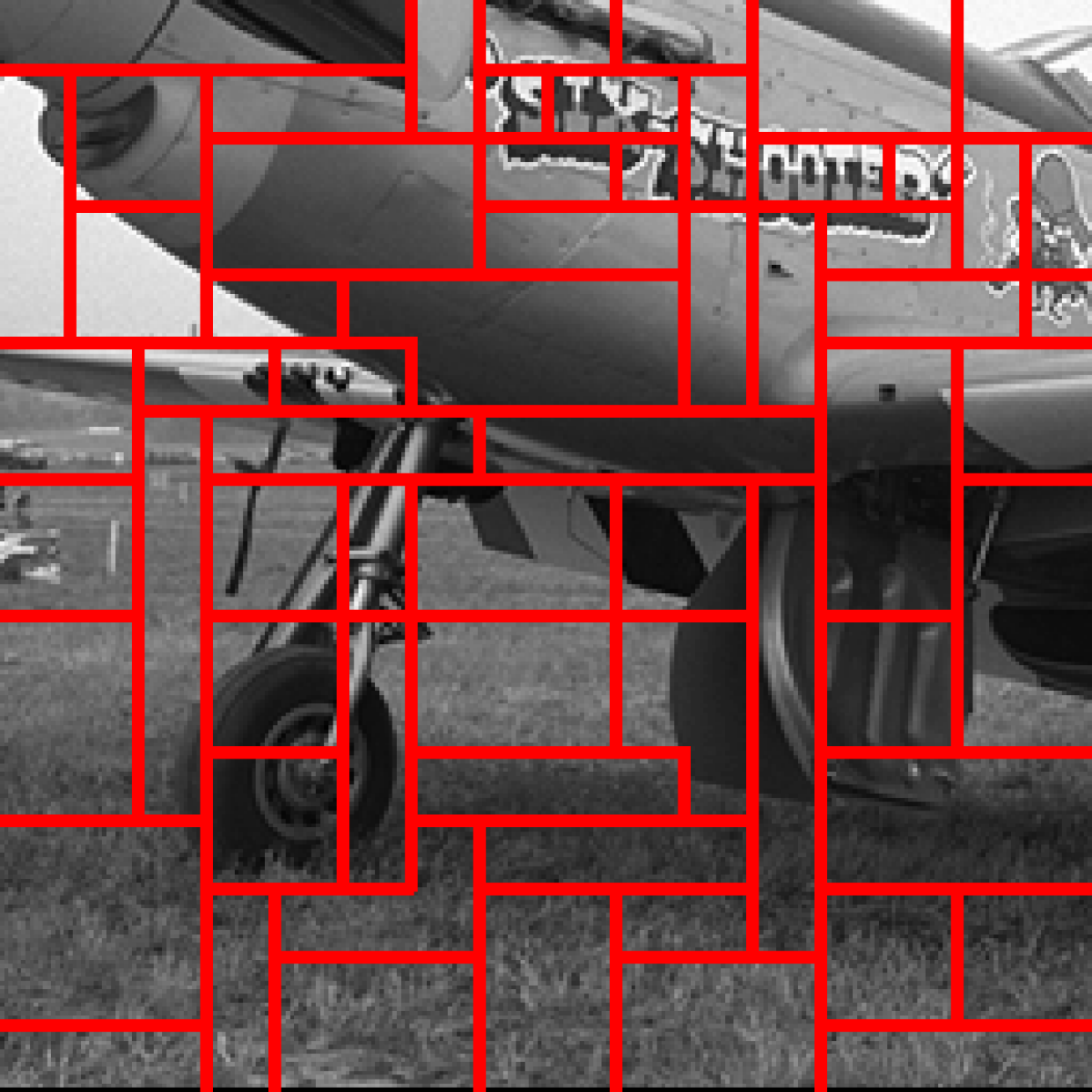}
        \caption{BEAM}
        \label{fig:kodim20_beam:d}
    \end{subfigure}
    \hfill
    \begin{subfigure}{0.31\linewidth}
        \centering
        \includegraphics[width=\linewidth]{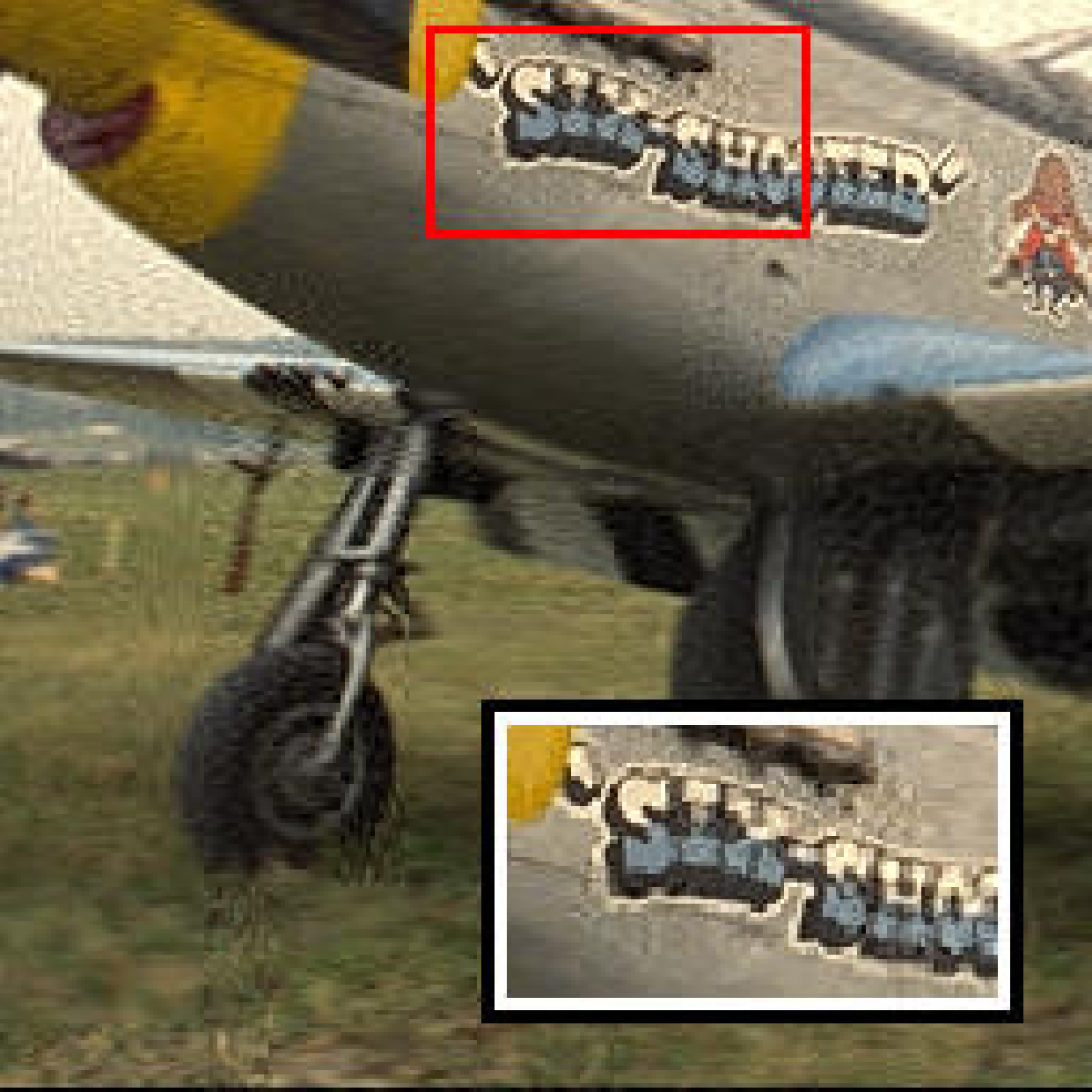}
        \caption{$m$=256 (27.5 dB)}
        \label{fig:kodim20_beam:e}
    \end{subfigure}
    \hfill
    \begin{subfigure}{0.31\linewidth}
        \centering
        \includegraphics[width=\linewidth]{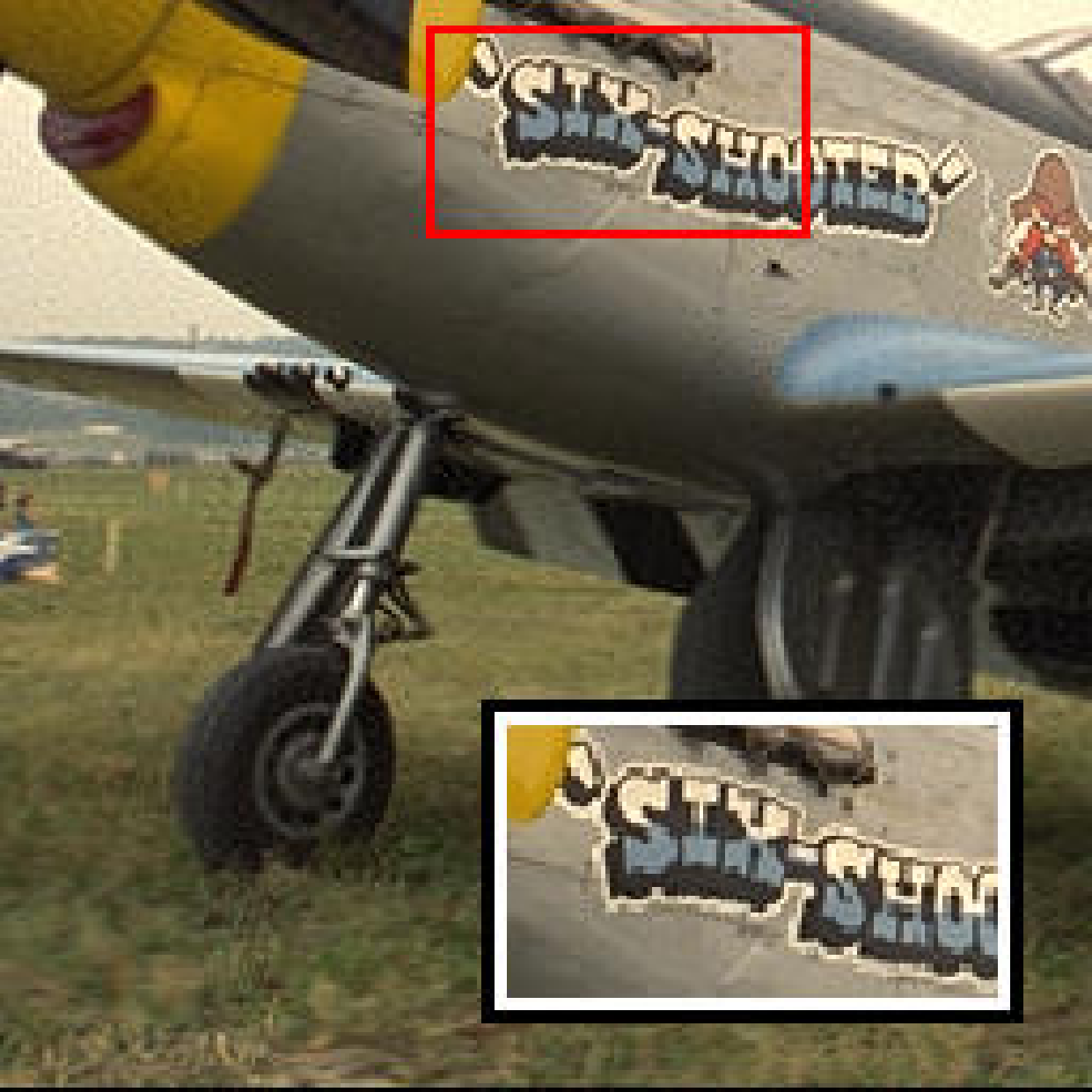}
        \caption{$m$=512 (32.3 dB)}
        \label{fig:kodim20_beam:f}
    \end{subfigure}
    \caption{Comparison between ELM-INR without BEAM (top row) and ELM-INR with BEAM (bottom row) on the \texttt{kodim20} image (256$\times$256).}
    \label{fig:kodim20_beam}
\end{minipage}
\hfill
\begin{minipage}[t]{0.48\linewidth}
    \centering
    \setlength{\tabcolsep}{1.5pt}
    \renewcommand{\arraystretch}{1.0}
    \begin{tabular}{ccc}
        \includegraphics[width=0.31\linewidth]{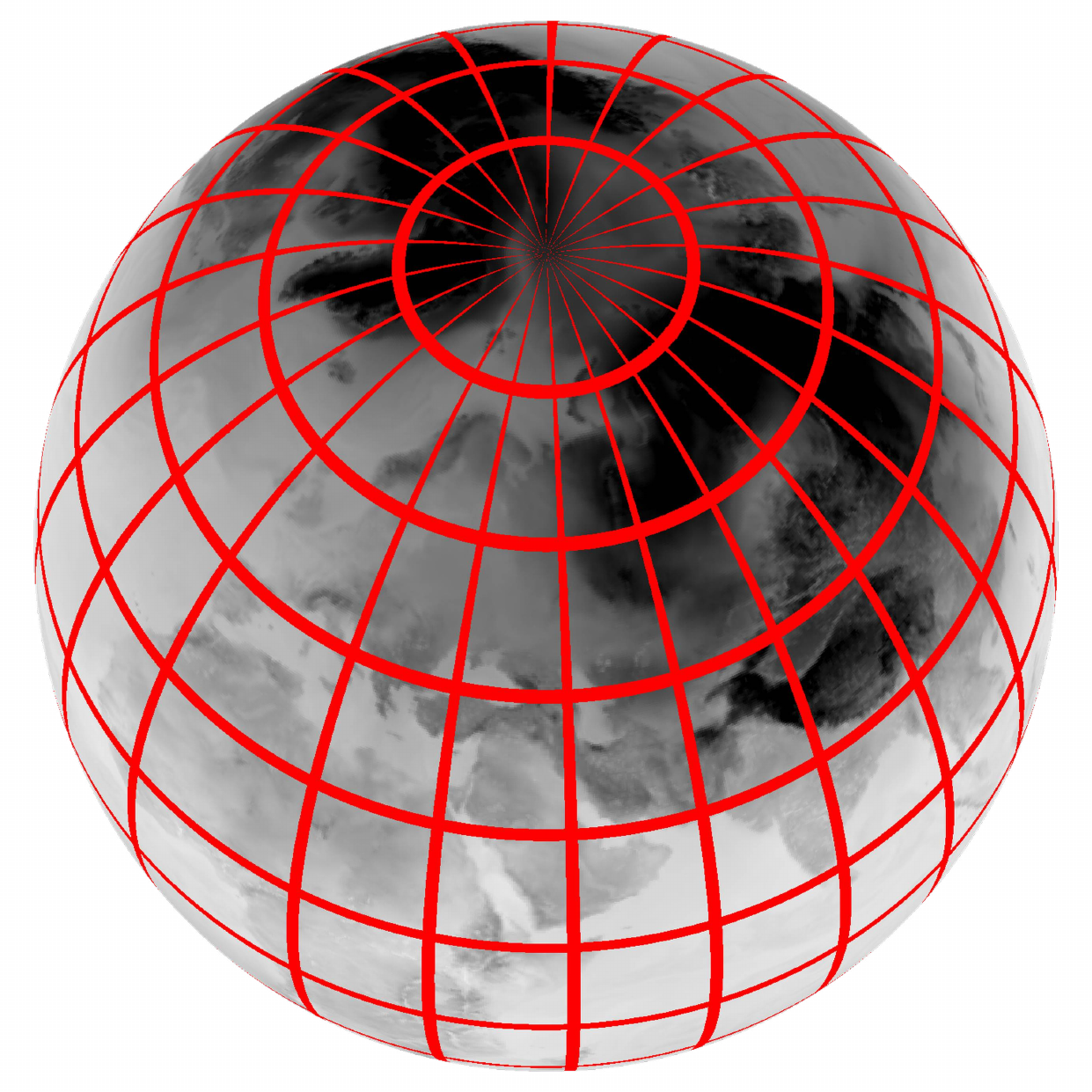} &
        \includegraphics[width=0.31\linewidth]{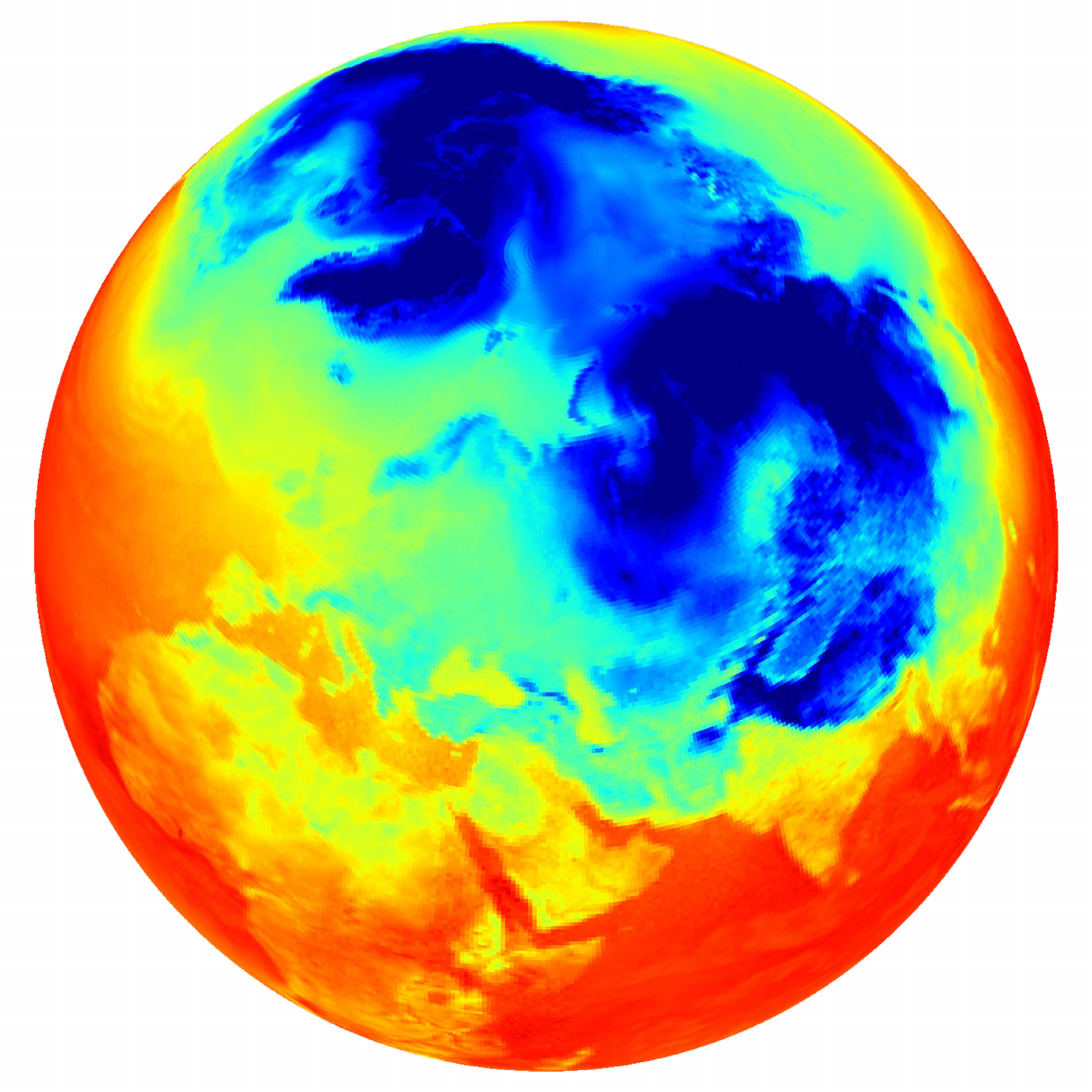} &
        \includegraphics[width=0.31\linewidth]{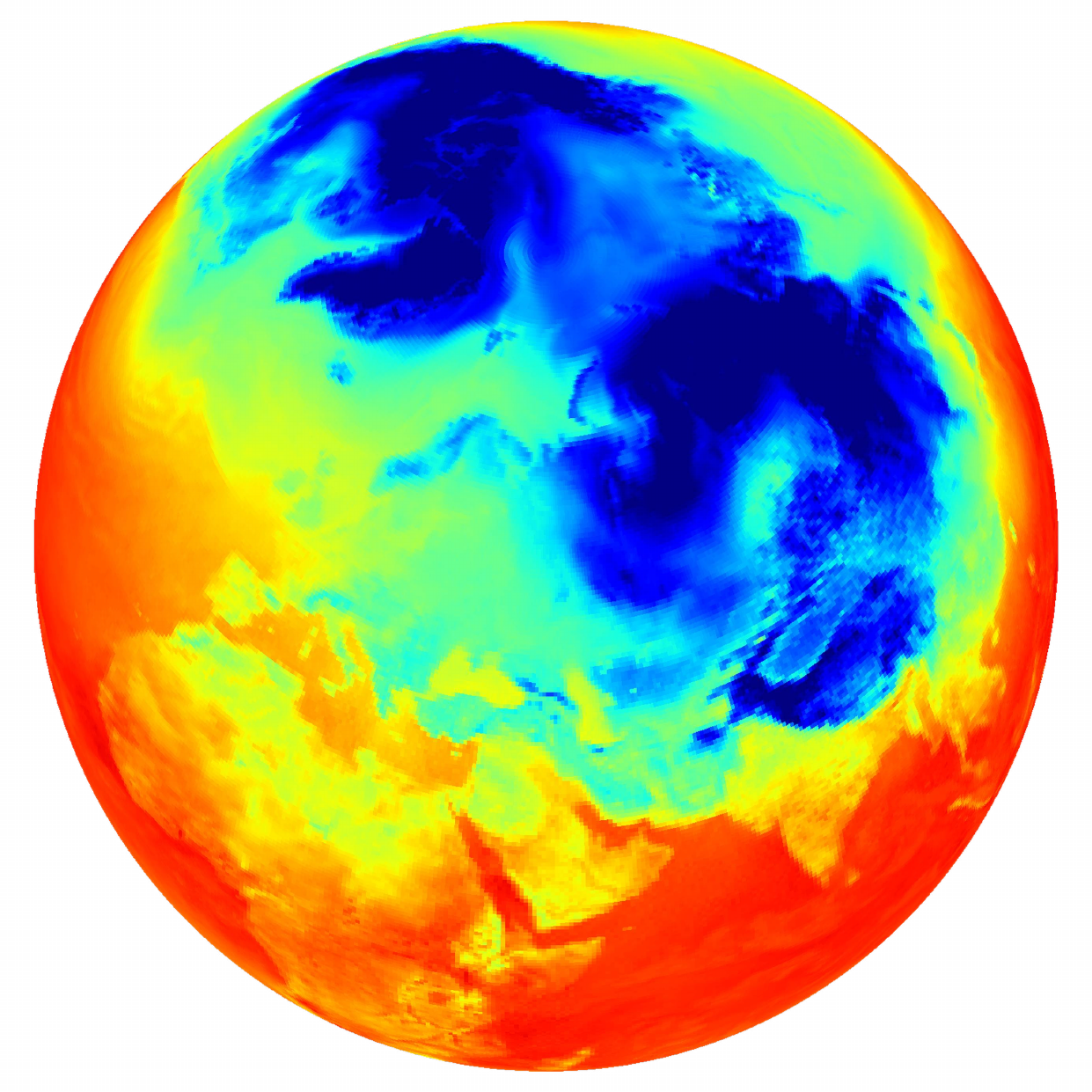} \\
        \scriptsize (a) Regular Mesh &
        \scriptsize (b) $m$=1024 (0.398) &
        \scriptsize (c) $m$=2048 (0.317) \\[6pt]
        \includegraphics[width=0.31\linewidth]{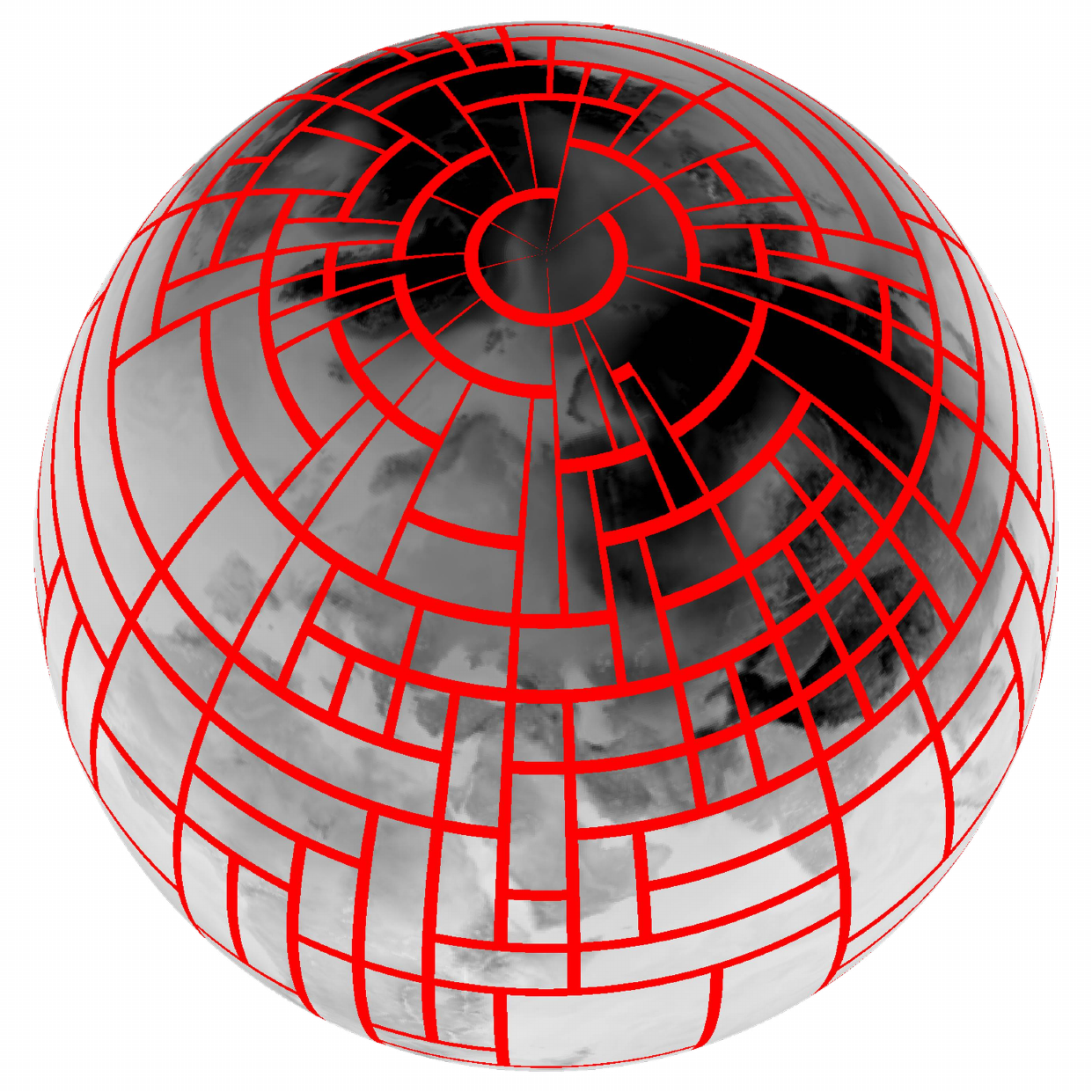} &
        \includegraphics[width=0.31\linewidth]{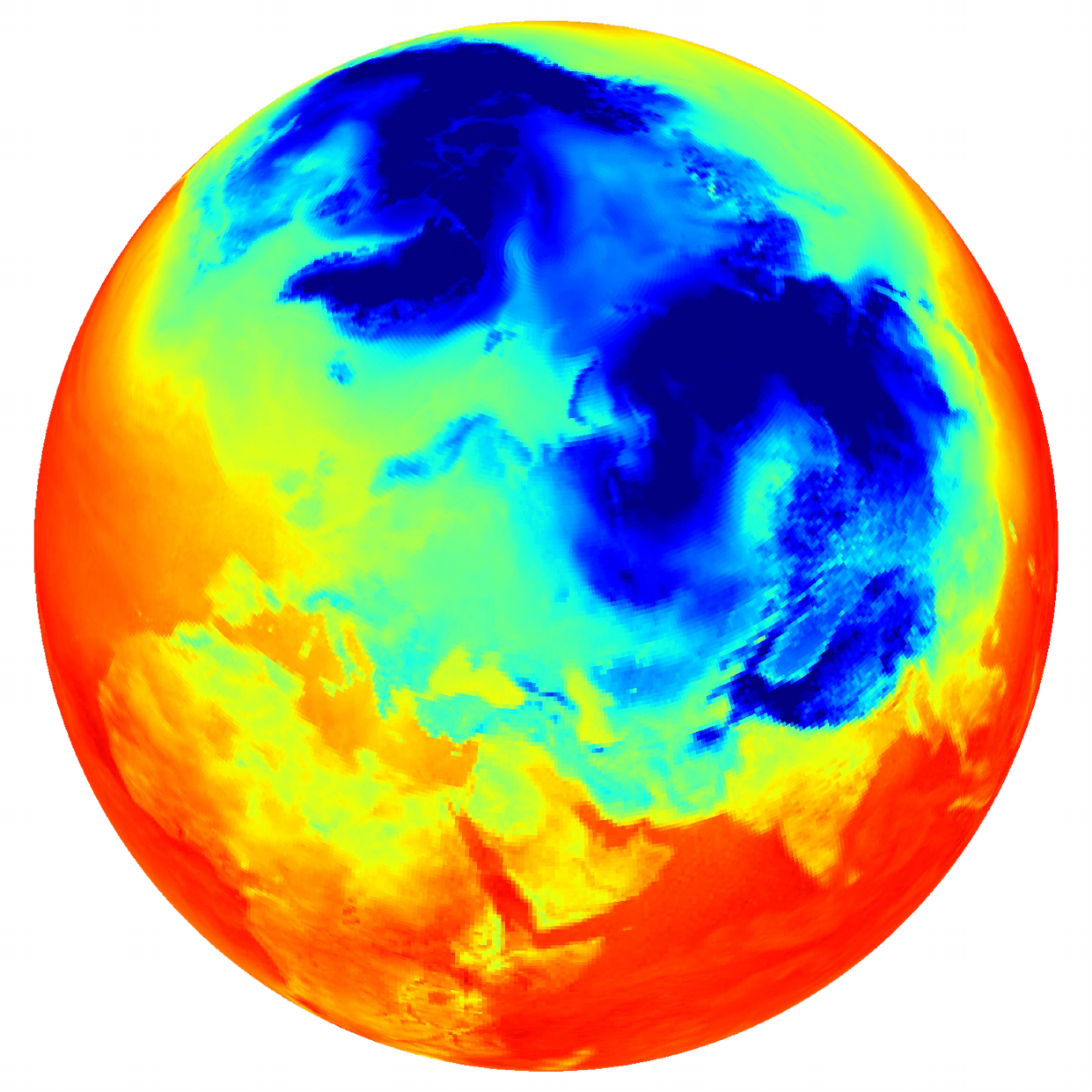} &
        \includegraphics[width=0.31\linewidth]{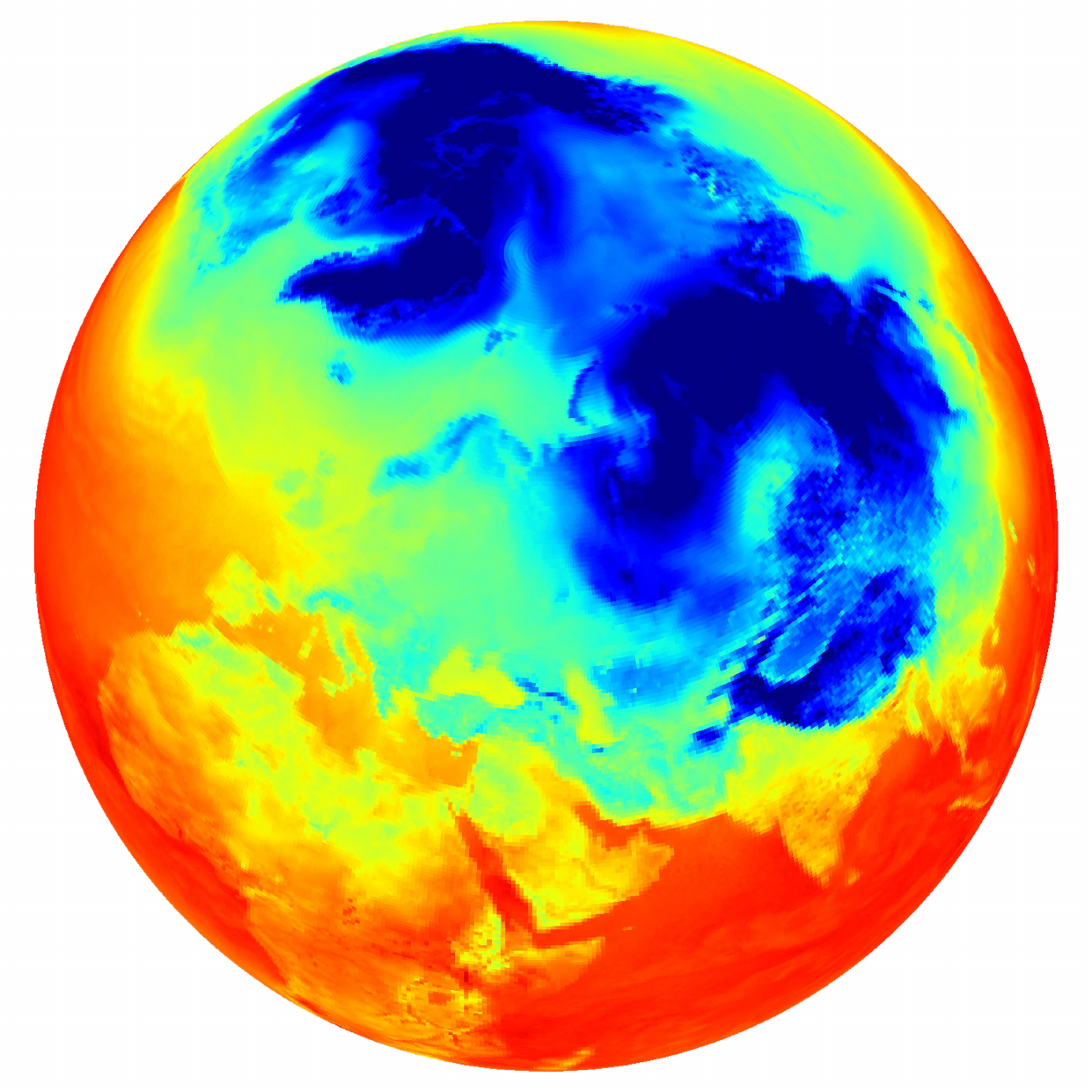} \\
        \scriptsize (d) BEAM &
        \scriptsize (e) $m$=1024 (0.363) &
        \scriptsize (f) $m$=2048 (0.288)
    \end{tabular}
    \caption{Comparison between ELM-INR without BEAM (top row) and ELM-INR with BEAM (bottom row) on full-resolution \texttt{ERA5} data.}
    \label{fig:era5}
\end{minipage}
\end{figure}

\paragraph{Standard Image Reconstruction.}
Figure~\ref{fig:kodim20_beam} demonstrates the effect of BEAM on ELM-INR using the \texttt{kodim20} image. Compared to uniform partitioning, BEAM adaptively refines subdomains by allocating smaller patches to spectrally complex regions and larger patches to smoother areas, resulting in a more balanced partition, as shown in Figure~\ref{fig:kodim20_beam} (\subref{fig:kodim20_beam:a} and \subref{fig:kodim20_beam:d}). For BEAM, the adaptive partition is constructed with an atomic patch size of $s=32$, while both Regular Mesh and BEAM use the same total number of subdomains, fixed to $N=64$, ensuring a fair comparison.

This adaptive partitioning leads to consistent reconstruction improvements. For $m=256$, BEAM improves PSNR from 26.0~dB to 27.5~dB, and for $m=512$, from 29.6~dB to 32.3~dB, as illustrated in Figure~\ref{fig:kodim20_beam} (\subref{fig:kodim20_beam:b}, \subref{fig:kodim20_beam:c}, \subref{fig:kodim20_beam:e}, and \subref{fig:kodim20_beam:f}). The runtime overhead of constructing the BEAM partition in this experiment is reported in Appendix~\ref{app:beam_runtime}. Visual comparisons further show sharper edges and reduced artifacts with BEAM, particularly in regions with complex structures. These results align with the error bound in Eq.~\eqref{eq:barron_bound}, which links reconstruction accuracy to both model capacity and local spectral complexity. While increasing the number of hidden nodes generally improves performance, BEAM provides clear gains at fixed and moderate model capacities by more effectively utilizing available capacity.

\paragraph{Geophysical Field Reconstruction (ERA5).}
In addition to standard image benchmarks, we further evaluate the effect of BEAM on ELM-INR using full-resolution ERA5 climate data~\cite{hersbach2020era5, rasp2023weatherbench}. Figure~\ref{fig:era5} presents reconstruction results on global temperature fields, where each seasonal snapshot has a spatial resolution of latitude $=721$ and longitude $=1440$. The reconstruction error is measured using mean absolute error (MAE), which is suitable for continuous-valued geophysical fields.

For ERA5 experiments, ELM-INR uses square spatial patches with $S_i = 64^2$ samples per subdomain. When constructing adaptive partitions with BEAM, the atomic patch size is set to $s=32$. Importantly, to ensure a fair comparison, the total number of subdomains is fixed to be identical for both Regular Mesh and BEAM, with $N=276$ in all cases. This setting isolates the effect of adaptive partitioning from that of increased model capacity or additional subdomains.

As shown in Figure~\ref{fig:era5}, BEAM consistently reduces reconstruction error compared to uniform partitioning at the same model capacity. For example, in the winter season, BEAM improves MAE from $0.398$ to $0.363$ for $m=1024$, and from $0.317$ to $0.288$ for $m=2048$. Despite using the same number of subdomains, BEAM allocates smaller patches to spectrally complex regions and larger patches to smoother areas, leading to a more effective use of the fixed representational budget. This further corroborates the role of spectral balancing in improving reconstruction quality under capacity-constrained regimes.

\section{Conclusion}
In this work, we proposed ELM-INR, a backpropagation-free framework that dramatically accelerates INR training by utilizing Extreme Learning Machines with closed-form solutions. We provided a theoretical analysis through the lens of Barron space, demonstrating that approximation error is fundamentally bounded by the local spectral complexity of the target signal. Guided by this insight, we introduced BEAM (Barron-Enhanced Adaptive Mesh refinement), an algorithm that adaptively partitions the domain to equalize spectral energy across subproblems. BEAM proves particularly effective in capacity-constrained regimes, ensuring high-fidelity reconstruction even when the number of hidden neurons ($m$) is small, by transforming a hard global fit into a set of spectrally balanced, manageable local fits across diverse modalities.

\section{Limitations and Broader Impact}\label{app:limitation}
ELM-INR provides fast and stable reconstruction for signals with direct coordinate--value supervision, but extending it to 3D novel view synthesis with neural radiance fields requires additional consideration. ELM-INR obtains its output weights by solving, on each subdomain $\Omega_i$, a linear least-squares problem $\boldsymbol{\alpha}_i^{*} = (\mathbf{H}_i^{\top}\mathbf{H}_i)^{-1}\mathbf{H}_i^{\top}\mathbf{y}_i$, which requires every input coordinate to be paired with a directly observed target value. In NeRF-style settings, supervision is indirect: pixel colors arise from volumetric rendering along camera rays, where a nonlinear transmittance term couples the radiance and density fields. No single 3D coordinate carries an observed value, and each supervised pixel constrains an entire ray that typically traverses many subdomains at once. This breaks the two properties that make ELM-INR efficient---linearity in the output weights and decoupling across subdomains---and prevents a direct reduction to independent local least-squares solves. Possible remedies include linearizing the rendering operator around a coarse initialization or alternating schemes that combine ray-wise consistency with subdomain-local ELM solves, which we leave to future work.

\paragraph{Broader Impact}
By replacing iterative training with a closed-form solve, ELM-INR substantially reduces the compute and energy cost of fitting INRs, lowering the barrier to deploying neural representations in resource-constrained scientific pipelines such as medical imaging, Earth observation, and climate data analysis. As with any signal reconstruction method, careful consideration is needed when it is applied to sensitive or security-critical data.

\begin{ack}
We thank the anonymous reviewers for their constructive comments. This work was supported by TelePIX.
\end{ack}

\bibliographystyle{unsrt}
\bibliography{neurips_2026}

\clearpage
\appendix

\section{Symbol Definitions}
Table~\ref{tab:symbol_definitions} provides a consolidated summary of the mathematical symbols used throughout the paper.
\begin{table}[ht!]
\centering
\renewcommand{\arraystretch}{1.0}
\caption{Summary of symbols used throughout the paper.}
\label{tab:symbol_definitions}
{\setlength{\tabcolsep}{20pt}
\begin{tabular}{l @{\hspace{3cm}} l}
\specialrule{1pt}{2pt}{2pt}
\textbf{Symbol} & \textbf{Description} \\
\specialrule{1pt}{2pt}{2pt}

\multicolumn{2}{l}{\textbf{\emph{Global function and domain notation}}} \\
$\mathcal{X} \in \mathbb{R}^d$ & Continuous input coordinate in $d$-dimensional space \\
$f(\mathcal{X})$ & Ground-truth target function \\
$\hat{f}(\mathcal{X})$ & Global ELM-INR approximation of $f$ \\
$d$ & Input dimensionality \\
$\Omega \subset \mathbb{R}^d$ & Global spatial domain \\

\midrule
\multicolumn{2}{l}{\textbf{\emph{Domain decomposition and partition-of-unity (PoU)}}} \\
$\Omega_i$ & $i$-th local subdomain of $\Omega$ \\
$N$ & Number of subdomains in the partition \\
$\phi_i(\mathcal{X})$ & Partition-of-unity window function associated with $\Omega_i$ \\
$\mathrm{supp}(\phi_i)$ & Support of the PoU function $\phi_i$ \\

\midrule
\multicolumn{2}{l}{\textbf{\emph{Local Extreme Learning Machine (ELM) representation}}} \\
$\hat{f}_i(\mathcal{X})$ & Local ELM approximation on subdomain $\Omega_i$ \\
$m$ & Number of hidden neurons (ELM width) \\
$\sigma(\cdot)$ & Activation function \\
$w_{i,j}$ & Fixed random input weight of neuron $j$ in $\Omega_i$ \\
$b_{i,j}$ & Fixed random bias of neuron $j$ in $\Omega_i$ \\
$\alpha_{i,j}$ & Trainable output weight of neuron $j$ in $\Omega_i$ \\
$\boldsymbol{\alpha}_i$ & Output weight vector for the $i$-th local ELM \\

\midrule
\multicolumn{2}{l}{\textbf{\emph{ELM training and least-squares formulation}}} \\
$S_i$ & Number of training samples in subdomain $\Omega_i$ \\
$\mathbf{H}_i \in \mathbb{R}^{S_i \times m}$ & Hidden-layer activation matrix for $\Omega_i$ \\
$\mathbf{y}_i \in \mathbb{R}^{S_i}$ & Target values sampled from $f$ on $\Omega_i$ \\

\midrule
\multicolumn{2}{l}{\textbf{\emph{Barron space and spectral complexity measures}}} \\
$\mathcal{F}(\xi)$ & Fourier transform of function $f$ \\
$\xi \in \mathbb{R}^d$ & Continuous frequency variable \\
$\Gamma$ & Fourier-based Barron function class \\
$\mathcal{B}$ & Barron space \\
$\|f\|_{\mathcal{B}}$ & Barron norm of function $f$ \\
$\|f\|_{\mathcal{B}_S}$ & Spectral Barron norm of function $f$ \\
$\beta_i$ & Local spectral Barron norm on subdomain $\Omega_i$ \\

\midrule
\multicolumn{2}{l}{\textbf{\emph{Approximation error and theoretical bounds}}} \\
$C$ & Constant independent of network width $m$ \\
$\|\cdot\|_{L^2(\Omega)}$ & $L^2$ norm over domain $\Omega$ \\

\midrule
\multicolumn{2}{l}{\textbf{\emph{BEAM algorithm and adaptive partitioning}}} \\
$\mathcal{P}$ & Adaptive partition of the domain produced by BEAM \\
$\tau$ & Spectral Barron norm threshold used in BEAM \\
$s$ & Atomic cell size for BEAM initialization \\
$E_{new}$ & Spectral energy after merging two subdomains \\
$k$ & Discrete frequency index in the DFT domain \\
$\mathcal{F}_i(k)$ & Discrete Fourier coefficient on subdomain $\Omega_i$ \\

\specialrule{1pt}{2pt}{2pt}
\end{tabular}}
\end{table}

\clearpage
\section{Comprehensive Quantitative Comparison with INR Baselines}
\label{app:comprehensive_comparison}

This appendix consolidates the quantitative results comparing ELM-INR with four
backpropagation-based coordinate INR baselines (SIREN~\cite{sitzmann2020implicit},
FFN~\cite{tancik2020fourier}, GaussNet~\cite{ramasinghe2022beyond}, and
WIRE~\cite{saragadam2023wire}) across the image and remote-sensing benchmarks
reported in the paper. For each benchmark, we follow the experimental
protocol described in Appendix~\ref{app:detailed_experimental_setups}: baselines are
trained with the Adam optimizer using their best configuration found via
a lightweight hyperparameter search, while ELM-INR uses fixed
per-modality settings ($S_i\!=\!32^2$, $m\!=\!1024$ for standard images;
$S_i\!=\!64^2$, $m\!=\!4096$ for multispectral data). For Kodak images,
we report the mean and standard deviation of PSNR over 5 random seeds.
Best results in each row are shown in \textbf{bold}.

\begin{table}[ht]
\centering
\small
\caption{All benchmarks are reported as mean\,$\pm$\,std over 5 random seeds. PSNR is averaged over all spectral bands for the multispectral benchmarks.}
\label{tab:app_psnr}
\begin{tabular}{lccccc}
\specialrule{1pt}{2pt}{2pt}
Dataset & SIREN & FFN & GaussNet & WIRE & ELM-INR \\
\midrule
kodim05 & 31.12{\scriptsize\,$\pm$0.22} & 25.51{\scriptsize\,$\pm$0.25}
        & 24.32{\scriptsize\,$\pm$0.12} & 30.68{\scriptsize\,$\pm$0.27}
        & \textbf{79.03}{\scriptsize\,$\pm$1.59} \\
kodim08 & 27.63{\scriptsize\,$\pm$0.35} & 22.88{\scriptsize\,$\pm$0.15}
        & 21.97{\scriptsize\,$\pm$0.08} & 30.43{\scriptsize\,$\pm$0.16}
        & \textbf{78.82}{\scriptsize\,$\pm$1.73} \\
kodim20 & 40.11{\scriptsize\,$\pm$0.64} & 36.02{\scriptsize\,$\pm$0.12}
        & 32.45{\scriptsize\,$\pm$0.05} & 39.85{\scriptsize\,$\pm$0.94}
        & \textbf{78.42}{\scriptsize\,$\pm$1.20} \\
kodim24 & 28.03{\scriptsize\,$\pm$0.40} & 25.97{\scriptsize\,$\pm$0.26}
        & 24.50{\scriptsize\,$\pm$0.06} & 31.10{\scriptsize\,$\pm$0.05}
        & \textbf{79.11}{\scriptsize\,$\pm$1.49} \\
WorldView-3, CDG
        & 33.06{\scriptsize\,$\pm$0.42} & 38.60{\scriptsize\,$\pm$0.55}
        & 30.61{\scriptsize\,$\pm$0.38} & 30.89{\scriptsize\,$\pm$0.61}
        & \textbf{79.19}{\scriptsize\,$\pm$0.42} \\
WorldView-3, LAX
        & 30.56{\scriptsize\,$\pm$0.47} & 35.08{\scriptsize\,$\pm$0.33}
        & 28.87{\scriptsize\,$\pm$0.52} & 29.20{\scriptsize\,$\pm$0.68}
        & \textbf{78.49}{\scriptsize\,$\pm$0.71} \\
Sentinel-2, Gal\'{a}pagos
        & 44.20{\scriptsize\,$\pm$0.36} & 42.54{\scriptsize\,$\pm$0.49}
        & 38.41{\scriptsize\,$\pm$0.62} & 37.40{\scriptsize\,$\pm$0.44}
        & \textbf{79.51}{\scriptsize\,$\pm$0.58} \\
MRI     & 33.70{\scriptsize\,$\pm$0.51} & 33.40{\scriptsize\,$\pm$0.39}
        & 29.20{\scriptsize\,$\pm$0.74} & 34.00{\scriptsize\,$\pm$0.45}
        & \textbf{79.80}{\scriptsize\,$\pm$0.66} \\
\specialrule{1pt}{2pt}{2pt}
\end{tabular}
\end{table}
\paragraph{Summary.}
As summarized in Table~\ref{tab:app_psnr}, across all image and
remote-sensing benchmarks, namely standard 2D Kodak images, WorldView-3 and Sentinel-2 multispectral satellite imagery, and MRI, ELM-INR consistently outperforms every coordinate-based INR baseline by a substantial margin, with PSNR gaps that exceed roughly $30\,\text{dB}$ on every dataset. The standard deviation across 5 random seeds remains small, typically $1$--$2\,\text{dB}$ on Kodak images, confirming that the closed-form solve produces stable reconstructions across random initializations of the hidden weights. These results corroborate the central claim of the paper: closed-form local fitting combined with partition-of-unity coupling provides an efficient and robust alternative to iterative INR training.

\section{Detailed Experimental Setups}\label{app:detailed_experimental_setups}
\paragraph{Software and hardware environments.}
All experiments were conducted on a workstation equipped with a single NVIDIA A6000 GPU (48\,GB VRAM),
an AMD EPYC 9224 24-core CPU (96 logical cores), and 377\,GB of system memory.
The software environment is based on Ubuntu 20.04.6 LTS with Python 3.12.7.
We use PyTorch 2.7.0 compiled with CUDA 12.6 for all experiments.

\paragraph{Hyperparameter settings.}
All baseline INRs are optimized using the Adam optimizer with default momentum parameters and no weight decay.
All models are trained for a fixed number of optimization steps depending on the task:
1000 steps for standard image datasets,
3000 steps for Navier--Stokes experiments,
and 10{,}000 steps for remote-sensing datasets.

\medskip
Rather than fixing a single hyperparameter configuration per model, we perform a lightweight hyperparameter search over a common set of architectural and optimization parameters.
Specifically, the number of hidden nodes is selected from $\{256, 512, 1024\}$,
the number of hidden layers from $\{3,4,5\}$, and the learning rate from $\{5\times10^{-5},\,1\times10^{-4}\}$. For each baseline, we report results using the best-performing configuration.

All remaining model-specific hyperparameters—such as activation functions,
frequency or bandwidth parameters, and feature encodings—are set according to the
original configurations recommended in the respective papers.

\clearpage
\paragraph{ELM-INR.}
ELM-INR uses fixed hyperparameters within each data modality.
For standard image experiments, we set the subdomain size to $S_i = 32^2$
and the number of hidden nodes to $m = 1024$.
For multispectral remote-sensing images and Navier--Stokes experiments,
we use a larger configuration with $S_i = 64^2$ and $m = 4096$. Leveraging the inherently decoupled nature of partition-based formulation,
we solve the closed-form least-squares problem for all subdomains in parallel via a single batched GPU triangular solve against the random-feature design matrix, which is shared across regular subdomains of equal size. This further reduces the wall-clock training time by an order of magnitude compared to a sequential implementation.

\paragraph{Random Fourier Features.}
For coordinate encoding, we use the default RFF configuration of~\cite{tancik2020fourier} across all subdomains and signal modalities. Given a coordinate $\mathcal{X} \in \mathbb{R}^d$, the encoding takes the form
\begin{equation}
    \gamma(\mathcal{X}) = \left[\cos(2\pi \mathbf{B} \mathcal{X}),\ \sin(2\pi \mathbf{B} \mathcal{X})\right],
\end{equation}
where the frequency matrix $\mathbf{B} \in \mathbb{R}^{F \times d}$ is sampled from $\mathcal{N}(0, \sigma_{\mathrm{RFF}}^2)$. We set $F = 10$ and $\sigma_{\mathrm{RFF}} = 1$.

\paragraph{PSNR computation.}
Reconstruction quality is evaluated using the peak signal-to-noise ratio (PSNR).
Following our implementation, predictions and ground-truth signals are first normalized
by the maximum value of the ground truth.
The PSNR is then computed as
\[
\mathrm{PSNR} = -10 \log_{10}(\mathrm{MSE} + \varepsilon),
\]
where $\mathrm{MSE}$ denotes the mean squared error over all evaluated samples
and $\varepsilon$ is a small constant for numerical stability.
In all experiments, we set $\varepsilon = 10^{-8}$.
For multispectral images, PSNR is computed independently for each channel
and the final score is reported as the average across channels.

\section{Dataset Description: Navier--Stokes Equation}\label{app:navier}
We use the same $(2+1)$D incompressible Navier--Stokes dataset as in
SPINN~\cite{cho2023separable}. The dataset consists of a single
spatio-temporal scalar field sampled on a uniform grid of size
$n_x \times n_y \times n_t = 128 \times 128 \times 100$, where the first
two axes correspond to the spatial coordinates $(x_1,x_2)$ and the third
axis corresponds to time $t$.

\paragraph{Governing equation and setting.}
Let $\mathcal{X}=(x_1,x_2,t)\in [0,2\pi]^2 \times [0,1]$ denote the
space--time coordinate. Following SPINN~\cite{cho2023separable}, we
consider a decaying-turbulence setting (no external forcing) for the
$(2+1)$D incompressible NS dynamics in the vorticity formulation:
\begin{equation}
\partial_t \omega(\mathcal{X}) + u(\mathcal{X})\cdot \nabla \omega(\mathcal{X})
\;=\; \nu \Delta \omega(\mathcal{X}),
\qquad
\nabla \cdot u(\mathcal{X})=0,\;\;
\omega(\mathcal{X})=\partial_{x_1}u_2(\mathcal{X})-\partial_{x_2}u_1(\mathcal{X}),
\label{eq:ns_vorticity}
\end{equation}
where $u(\mathcal{X})=(u_1(\mathcal{X}),u_2(\mathcal{X}))$ is the 2D
velocity field, and we fix the viscosity to $\nu=0.01$.

\paragraph{Numerical generation and learning pairs.}
A reference solution is generated using a pseudo-spectral CFD solver
(JAX-CFD) with periodic boundary conditions in space. The initial
condition is sampled from a Gaussian random field (with maximum velocity
$5$ in the SPINN setup), and the resulting field is recorded on the
uniform grid over $[0,2\pi]^2\times[0,1]$. For training coordinate-based
models, each grid point induces a coordinate--value pair
$(\mathcal{X}_{ijk}, \mathbf{y}_{ijk})$, where $\mathbf{y}_{ijk}$
denotes the sampled scalar value at $\mathcal{X}_{ijk}$. We linearly
normalize each coordinate axis to $[-1,1]$ before feeding it to the
network.

\paragraph{Reconstruction results.}
Table~\ref{tab:app_navier_stokes} reports the reconstruction error and
training time of all baselines on this dataset, complementing the
qualitative volume reconstructions shown in Figure~\ref{fig:pde_3d}.
For ELM-INR, the spatio-temporal domain is partitioned both spatially
and temporally, using $S_i\!=\!64^2$ samples per subdomain, temporal
patches of size~10, and $m\!=\!4096$ hidden units; backpropagation-based
baselines are trained for 3{,}000 iterations under their best
hyperparameter configurations. As shown in
Figure~\ref{fig:pde_3d}, backpropagation-based INR baselines fail to
recover fine-scale temporal structure and exhibit pronounced blurring
across the time axis, whereas ELM-INR produces a near-exact
reconstruction of the full spatio-temporal volume. This visual
observation is mirrored quantitatively in
Table~\ref{tab:app_navier_stokes}: ELM-INR achieves an MAE of
$5.68\!\times\!10^{-5}$, more than two orders of magnitude lower than
the strongest backpropagation-based baseline (SIREN at
$4.22\!\times\!10^{-3}$), while training in only $0.015$ seconds---much faster than the iterative baselines, which require
between $1{,}924$ and $2{,}838$ seconds. These results indicate that the
proposed closed-form, partitioned representation extends naturally to
complex spatio-temporal fields governed by physical dynamics, delivering
both substantially higher fidelity and dramatically shorter training
time.

\begin{table}[ht]
\centering
\caption{Reconstruction error and training time on the $(2{+}1)$D
incompressible Navier--Stokes dataset. ELM-INR achieves more than two
orders of magnitude lower MAE while training in a small fraction of
the wall-clock time required by backpropagation-based baselines.}
\label{tab:app_navier_stokes}
\begin{tabular}{lccccc}
\specialrule{1pt}{2pt}{2pt}
Metric & SIREN & FFN & GaussNet & WIRE & ELM-INR \\
\midrule
MAE
    & $4.22\!\times\!10^{-3}$ & $7.02\!\times\!10^{-2}$
    & $2.36\!\times\!10^{-2}$ & $2.09\!\times\!10^{-2}$
    & $\mathbf{5.68\!\times\!10^{-5}}$ \\
Time (sec)
    & 1{,}960 & 1{,}924 & 2{,}838 & 2{,}402 & \textbf{0.015} \\
\specialrule{1pt}{2pt}{2pt}
\end{tabular}
\end{table}

\section{Computational Complexity Analysis}
We analyze the computational cost of ELM-INR in contrast to standard backpropagation-based INRs such as SIREN, FFN, and WIRE. 

To formalize this comparison, let $S$ denote the total number of coordinate samples (e.g., pixels) in the domain, and let $\{\Omega_i\}_{i=1}^{N}$ be a partition of the domain, where each subdomain $\Omega_i$ contains $S_i$ samples such that $\sum_{i=1}^{N} S_i = S$. We assume each local ELM utilizes $m$ hidden units and outputs $C$ channels. For the comparative analysis of backpropagation-based INRs, we denote the number of hidden layers by $L$ and the total number of training iterations by $T$.

\paragraph{ELM-INR (Closed-Form):} 
For each subdomain $\Omega_i$, ELM-INR constructs a hidden activation matrix $\mathbf{H}_i \in \mathbb{R}^{S_i \times m}$ and solves a linear least-squares problem for the output weights. The dominant computational cost arises from forming the normal matrix $\mathbf{H}_i^\top \mathbf{H}_i$ and solving the resulting linear system. The cost per subdomain is therefore $\mathcal{O}(S_i m^2 + m^3)$, where the $S_i m^2$ term dominates in the typical regime where $S_i \gg m$. Summing over all subdomains yields the total cost:
\begin{equation}
\mathcal{O}\left( \sum_{i=1}^{N} (S_i m^2 + m^3) \right)= \mathcal{O}(S m^2 + N m^3).    
\end{equation}
In standard INR settings, where $S \gg m$ and the number of partitions $N \ll S$, the overall cost is effectively linear in the number of samples $S$, up to a quadratic dependence on the hidden width $m$. Importantly, this cost is incurred only once, as ELM-INR does not rely on iterative optimization.

\paragraph{Backpropagation-Based INR:} 
For a standard INR trained via backpropagation, each training iteration consists of a forward pass, a backward pass, and a parameter update. For a network with $L$ hidden layers of width $m$, the computational cost per iteration scales as $\mathcal{O}(L S m^2)$, where the quadratic dependence on $m$ arises from the dense matrix multiplications between hidden layers. Over $T$ training iterations, the total computational cost becomes $\mathcal{O}(T L S m^2)$.

\paragraph{Comparison:} Comparing the two costs, ELM-INR replaces the iterative multiplicative factor $T \times L$ with a linear-algebraic solve. In practice, since the number of iterations $T$ is typically on the order of thousands and the depth $L \ge 2$ for expressive INRs, this difference results in orders-of-magnitude computational savings.

\clearpage
\section{Additional Experimental Results}

\begin{wrapfigure}{r}{0.45\textwidth}
\centering
\vspace{-12pt}
\includegraphics[width=\linewidth]{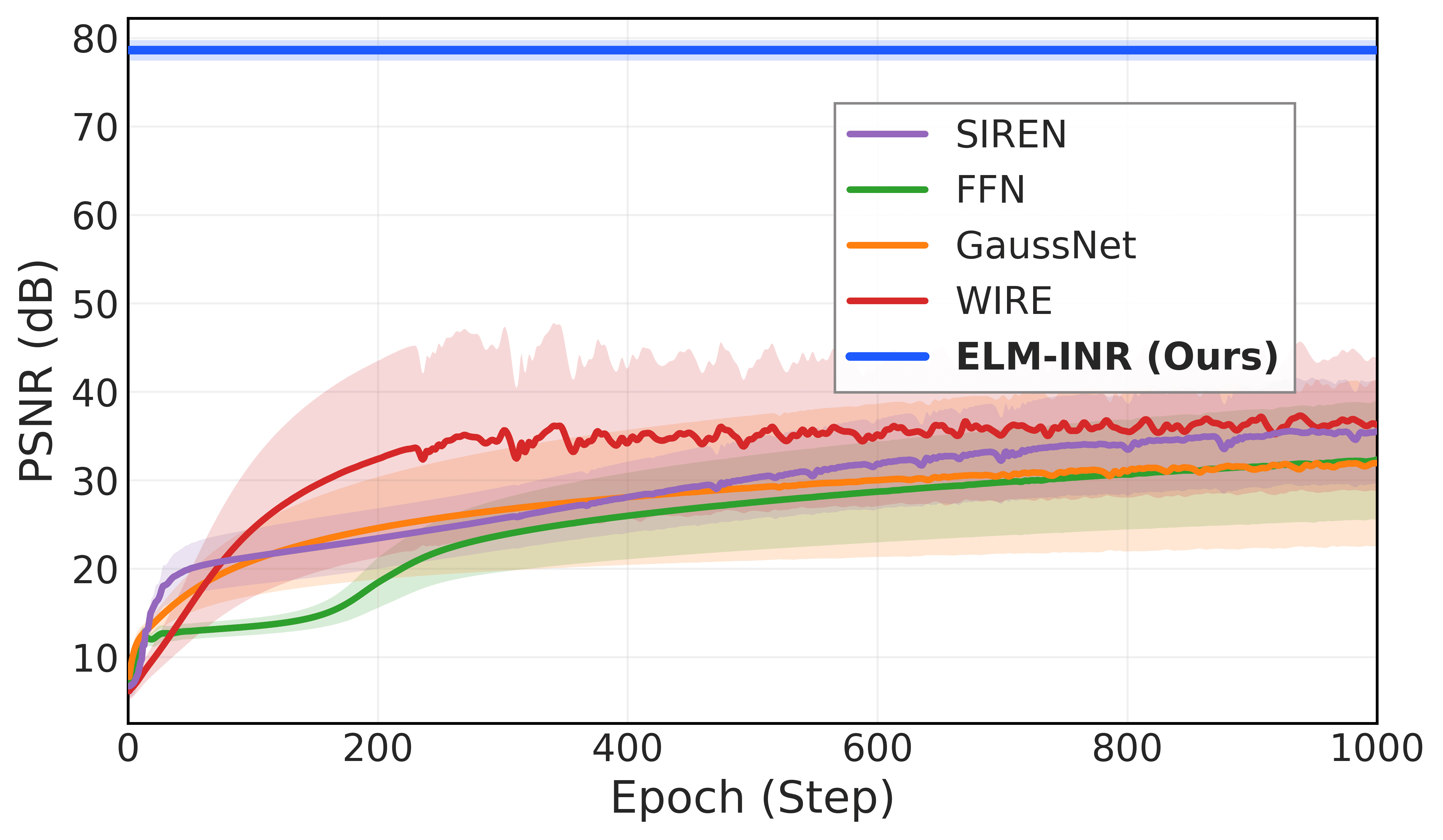}
\caption{Training dynamics of different INR architectures.}
\label{fig:training_dynamics}
\end{wrapfigure}

\paragraph{Interpretation of training dynamics.}
Figure~\ref{fig:training_dynamics} illustrates the evolution of PSNR across
training steps for SIREN, FFN, GaussNet, WIRE, and ELM-INR on \texttt{kodim08}, \texttt{kodim20}, \texttt{kodim24}, and \texttt{cameraman}. Solid lines denote mean performance and shaded regions indicate variability. ELM-INR achieves stable high-quality reconstruction without iterative backpropagation.
Unlike backpropagation-based INRs, ELM-INR does not rely on iterative optimization. Its parameters are obtained via a single closed-form least-squares solve, and therefore it does not admit intermediate training states. For visualization purposes, the ELM-INR curve is replicated across the step axis to highlight the absence of training dynamics.
\subsection{General Dataset in Computer Vision}

\begin{figure}[ht!]
    \centering
    \setlength{\tabcolsep}{2pt}
    \renewcommand{\arraystretch}{1.0}
    \begin{tabular}{cccccc}
        {\footnotesize SIREN}{\scriptsize (27.3 dB)} &
        {\footnotesize FFN}{\scriptsize (23.3 dB)} &
        {\footnotesize GaussNet}{\scriptsize (20.4 dB)} &
        {\footnotesize WIRE}{\scriptsize (25.7 dB)} &
        {\footnotesize \textbf{ELM-INR}}{\scriptsize \textbf{(77.1 dB)}} &
        GT \\
        \includegraphics[width=0.155\linewidth]{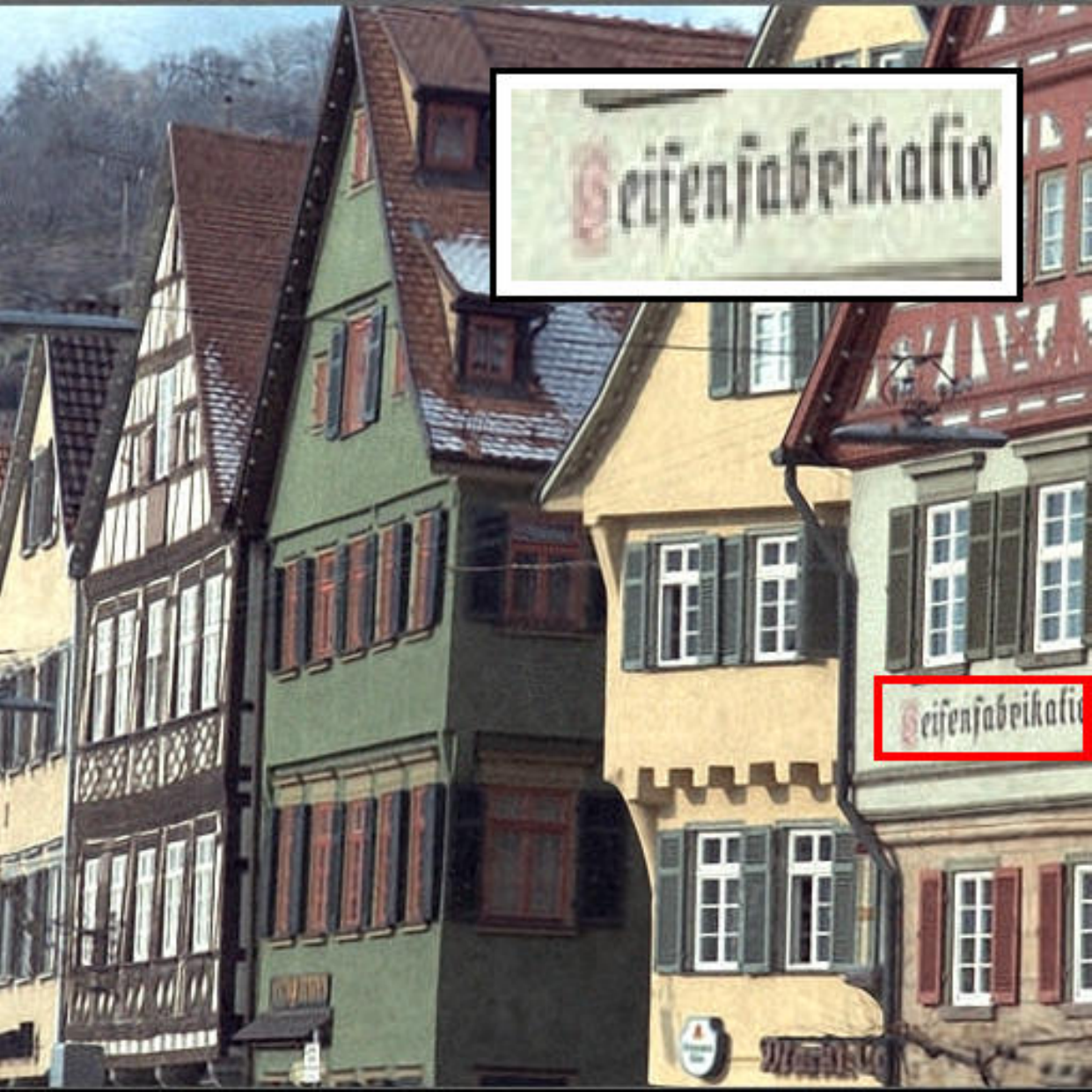} &
        \includegraphics[width=0.155\linewidth]{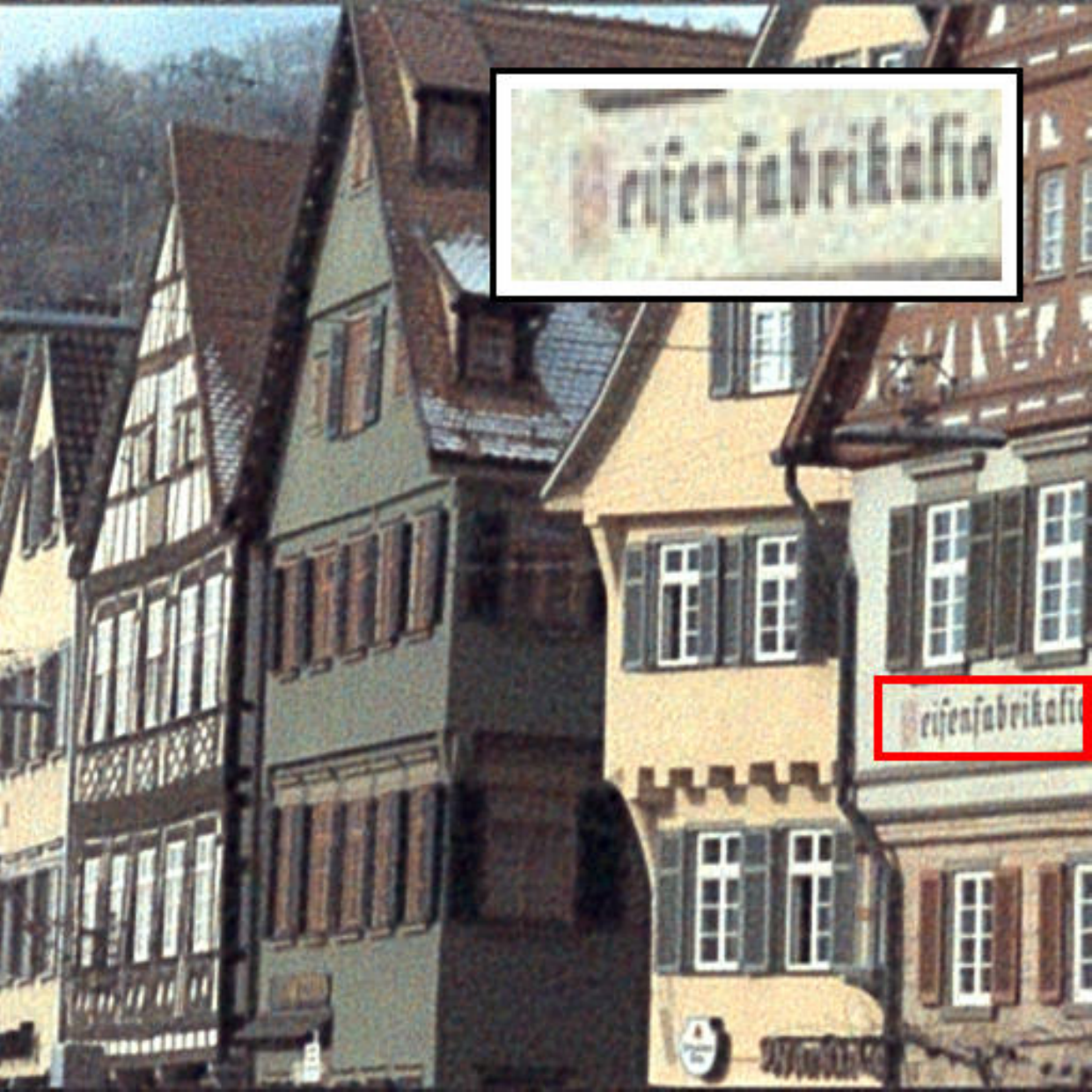} &
        \includegraphics[width=0.155\linewidth]{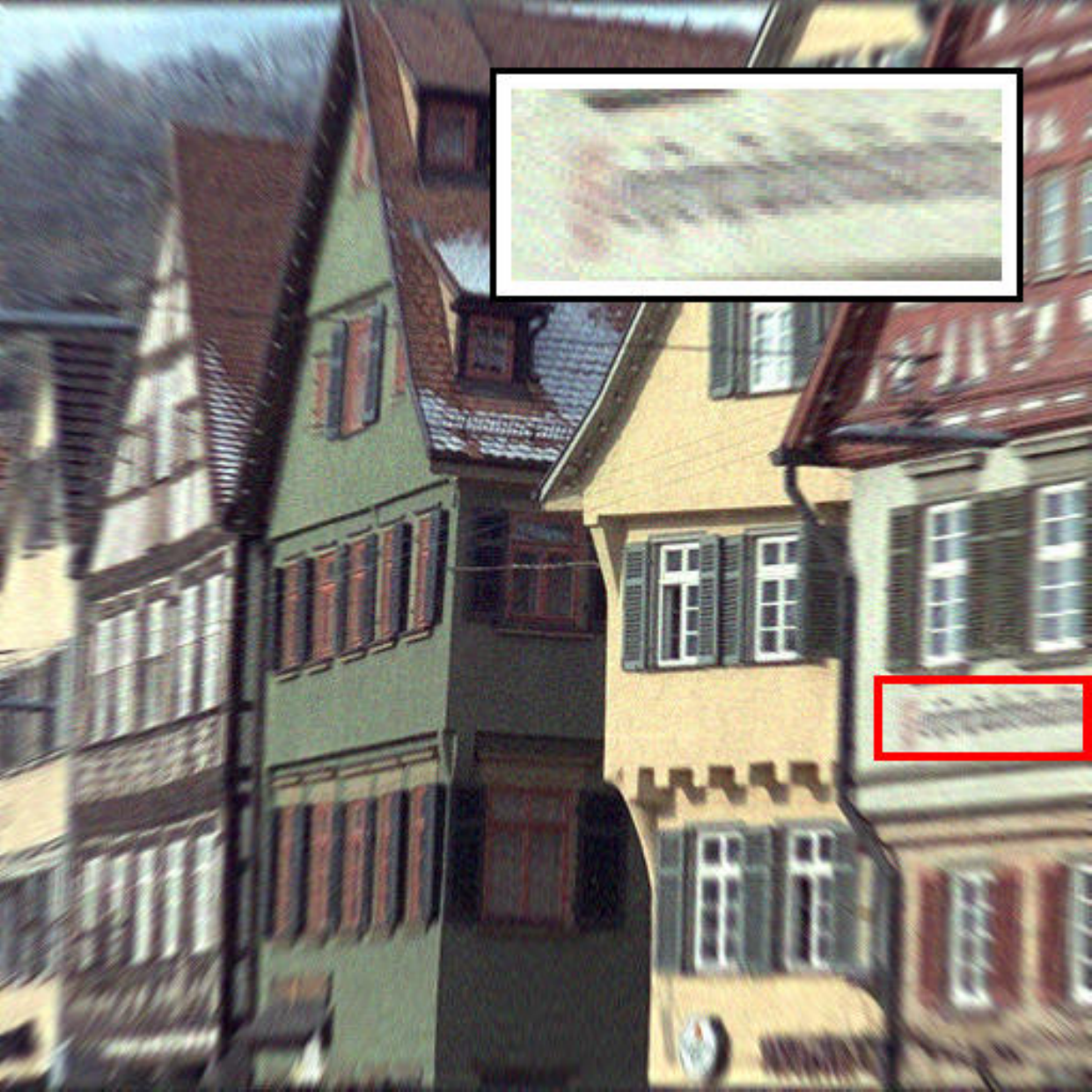} &
        \includegraphics[width=0.155\linewidth]{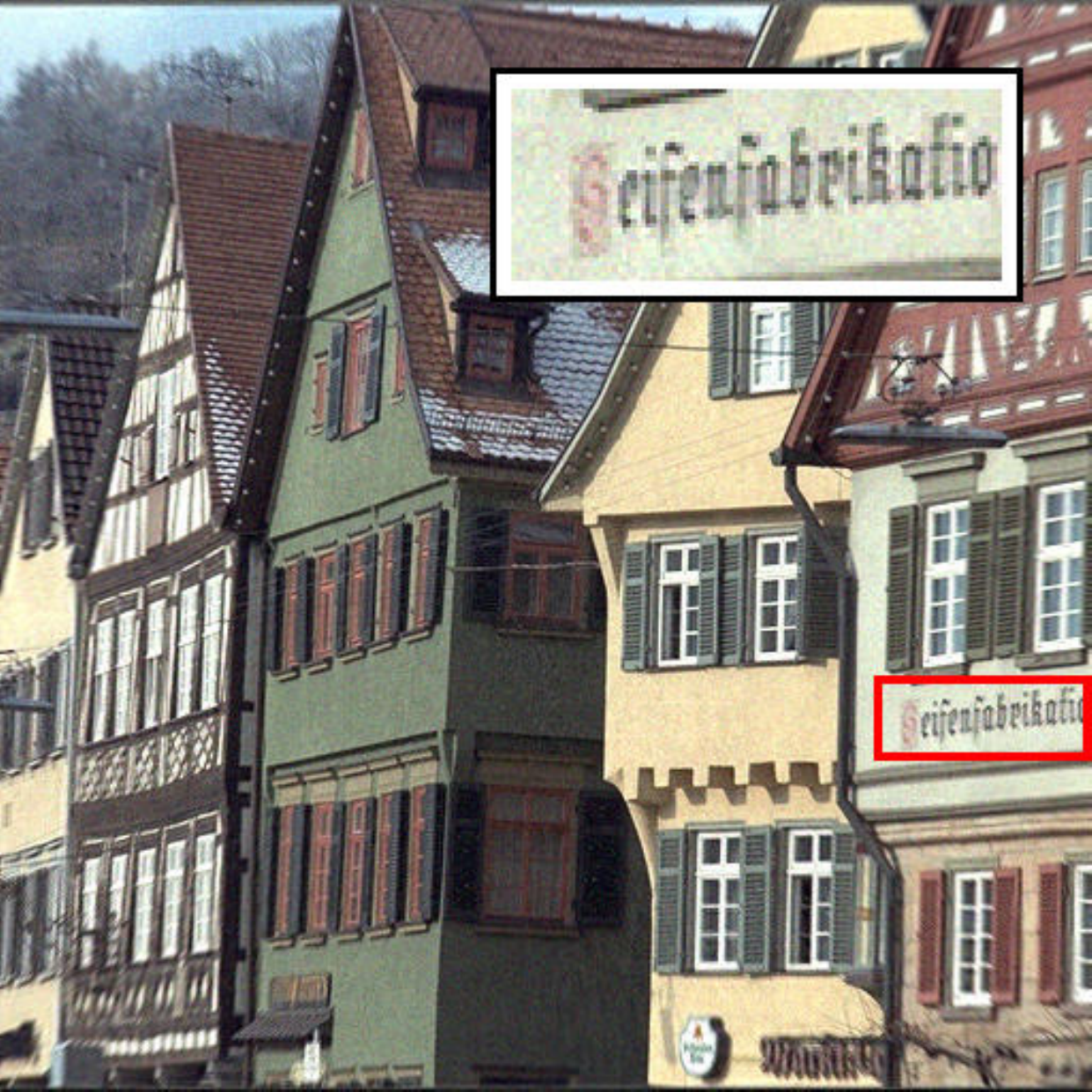} &
        \includegraphics[width=0.155\linewidth]{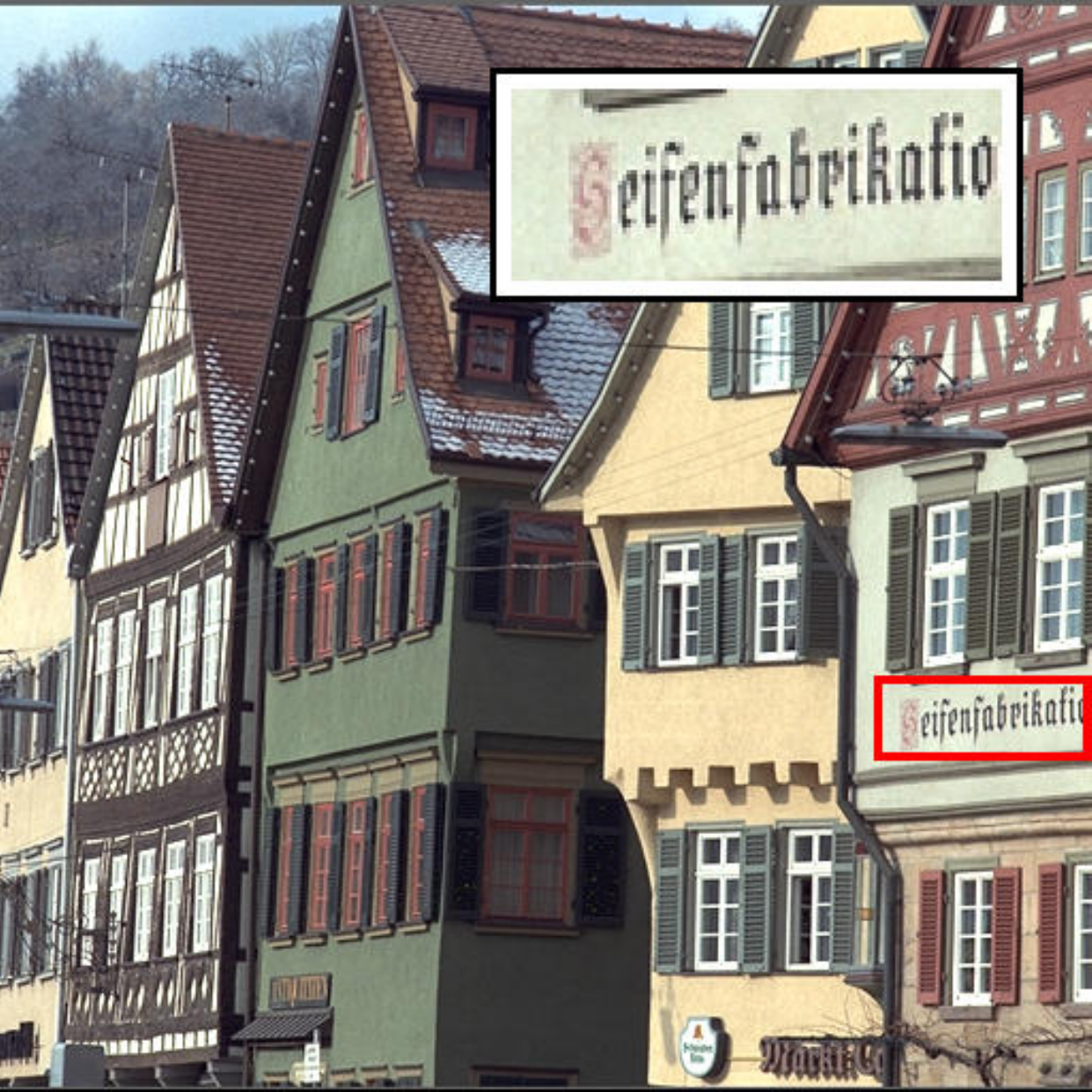} &
        \includegraphics[width=0.155\linewidth]{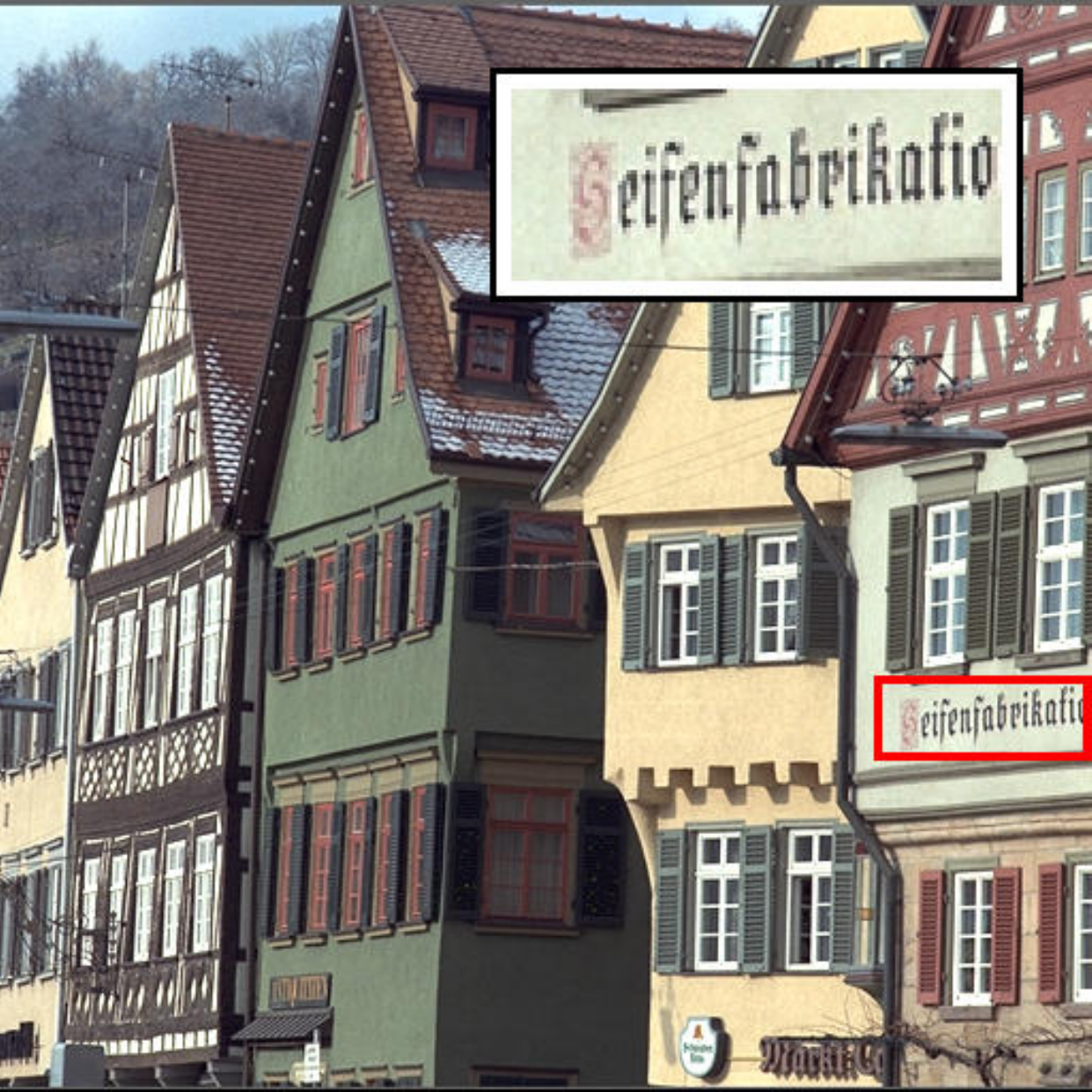}
    \end{tabular}
    \caption{
    Additional qualitative comparison on the \texttt{kodim08} image. ($512 \times 512$)
    }
    \label{fig:kodim08_appendix_qualitative}
\end{figure}

\begin{figure*}[ht!]
    \centering
    \setlength{\tabcolsep}{2pt}
    \renewcommand{\arraystretch}{1.0}
    \begin{tabular}{cccccc}
        {\footnotesize SIREN}{\scriptsize (39.4 dB)} &
        {\footnotesize FFN}{\scriptsize (34.2 dB)} &
        {\footnotesize GaussNet}{\scriptsize (32.1 dB)} &
        {\footnotesize WIRE}{\scriptsize (41.0 dB)} &
        {\footnotesize \textbf{ELM-INR}}{\scriptsize \textbf{(78.4 dB)}} &
        {\footnotesize GT} \\

        \includegraphics[width=0.155\linewidth]{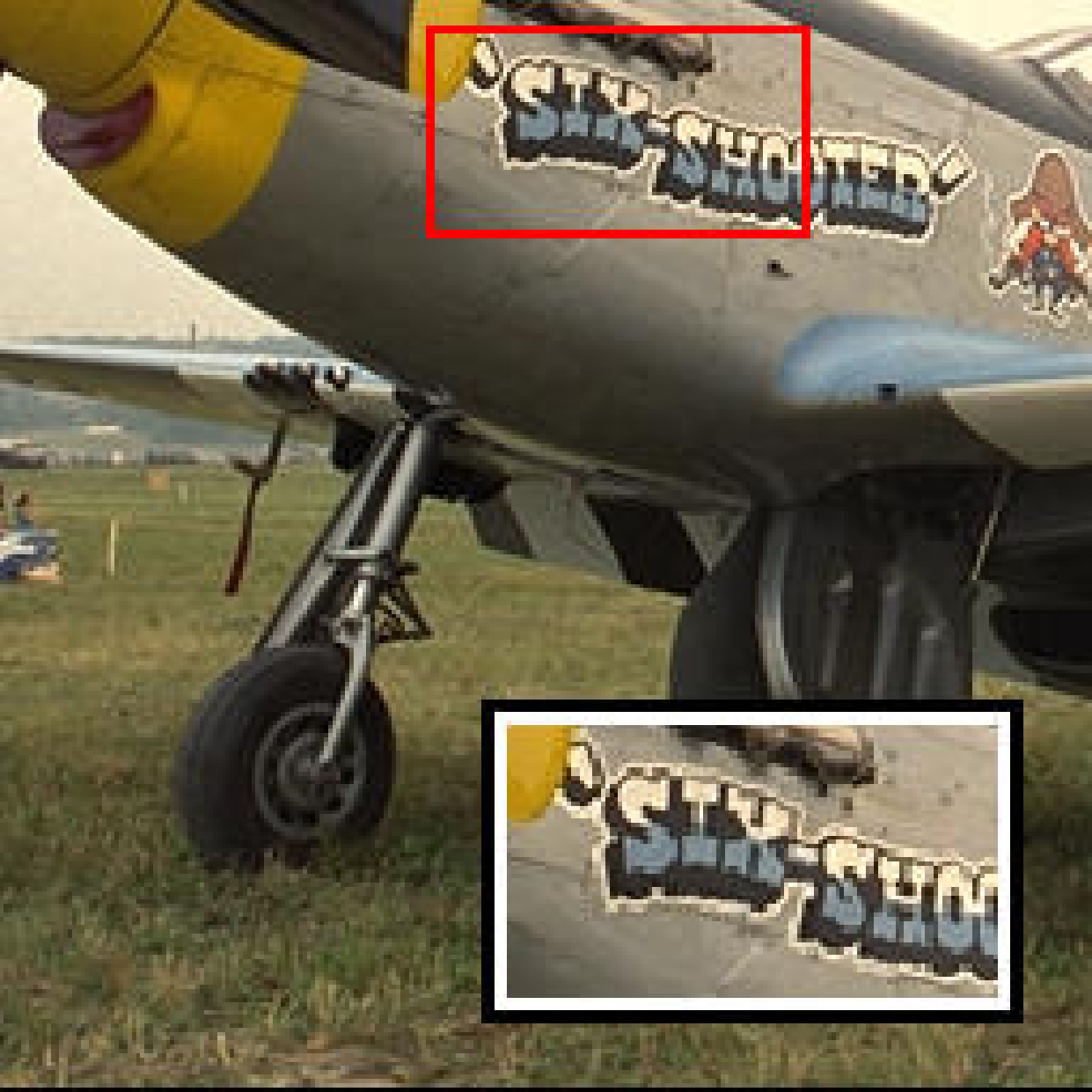} &
        \includegraphics[width=0.155\linewidth]{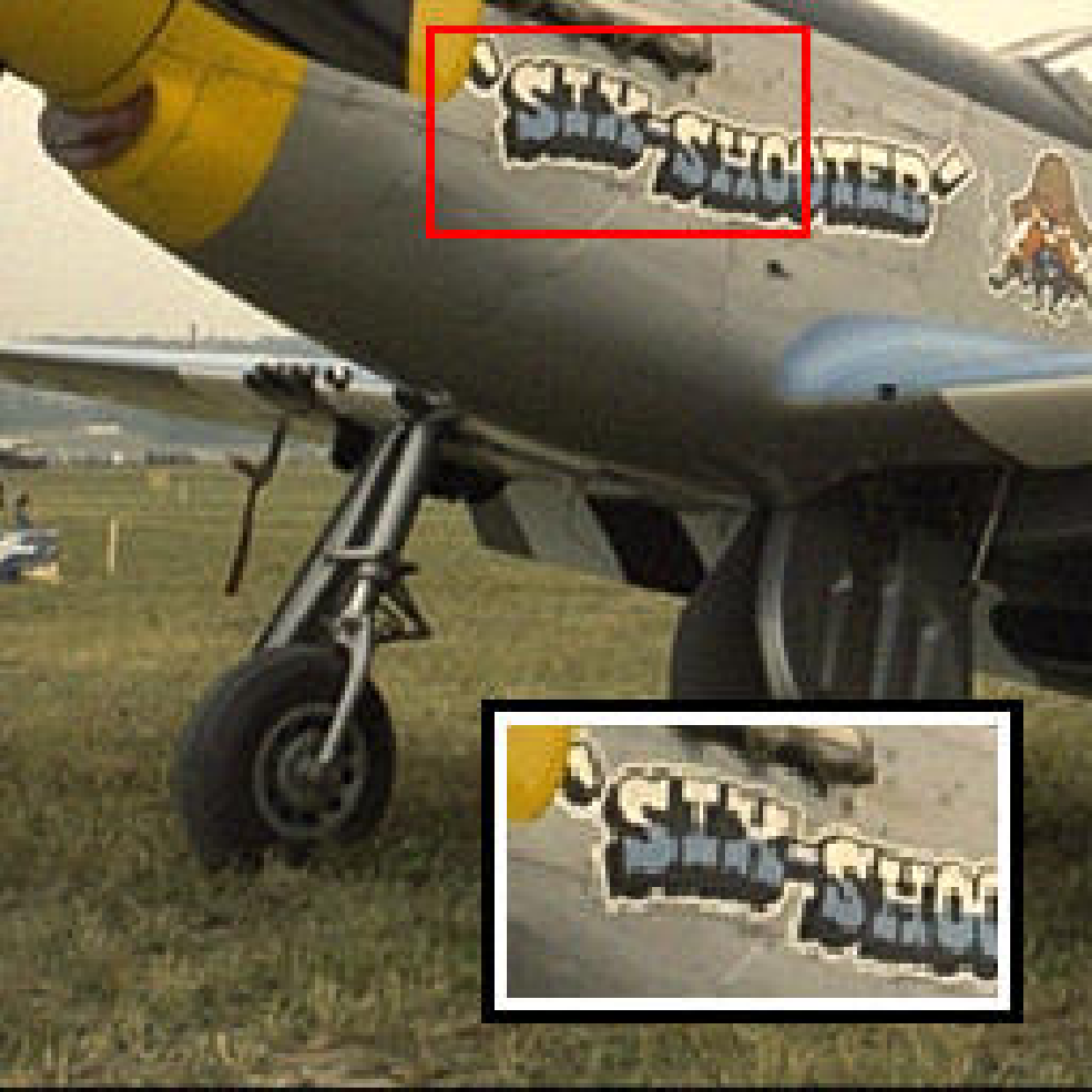} &
        \includegraphics[width=0.155\linewidth]{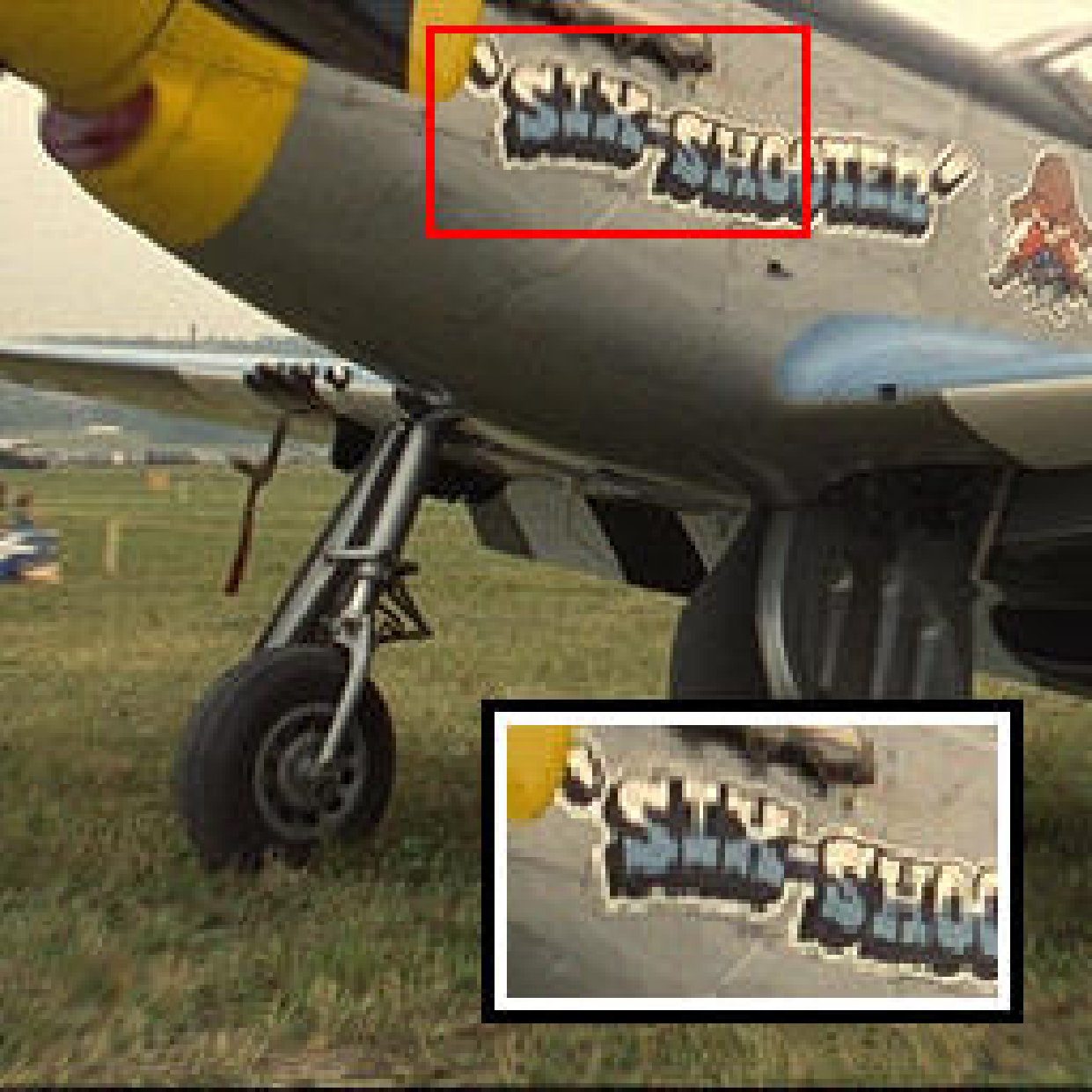} &
        \includegraphics[width=0.155\linewidth]{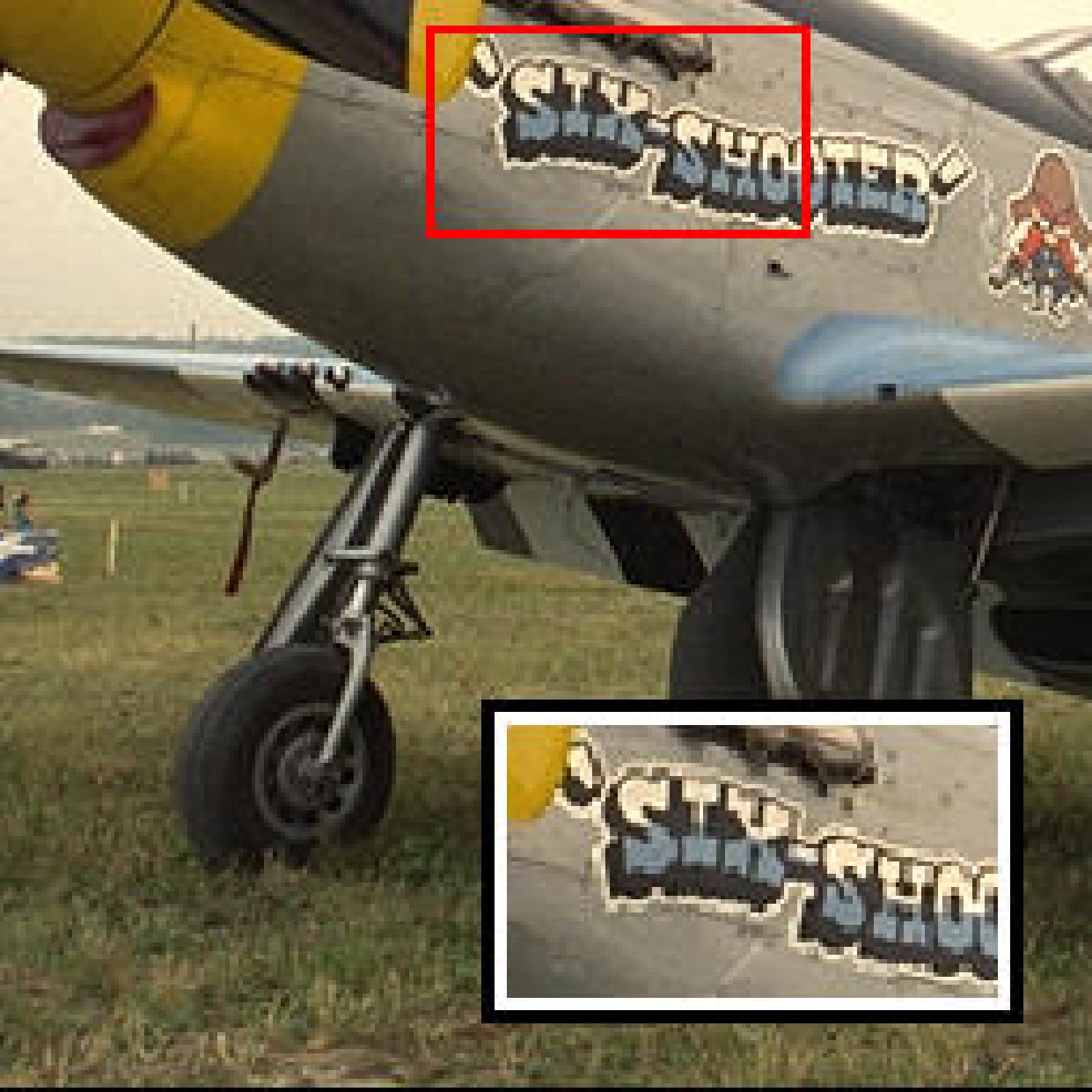} &
        \includegraphics[width=0.155\linewidth]{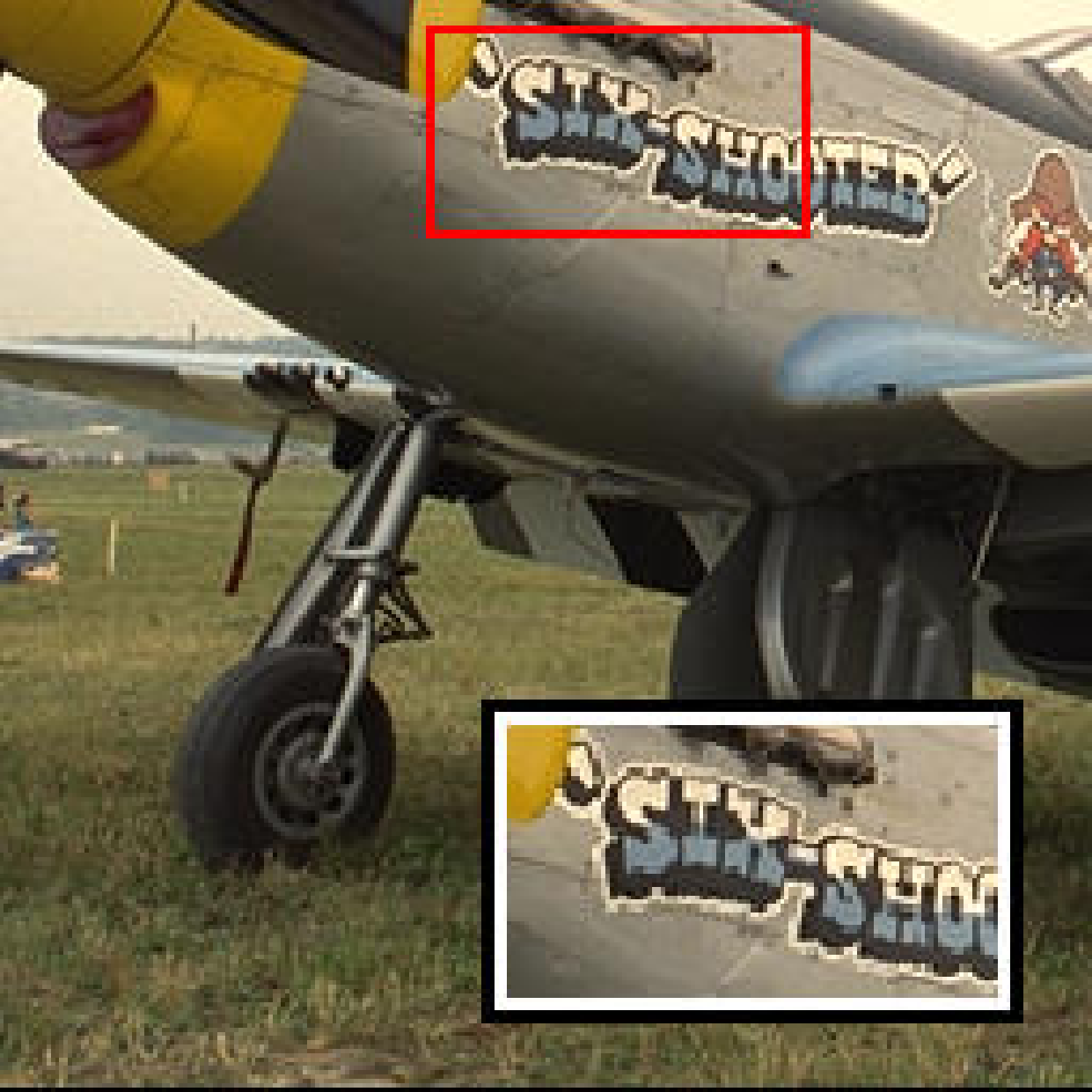} &
        \includegraphics[width=0.155\linewidth]{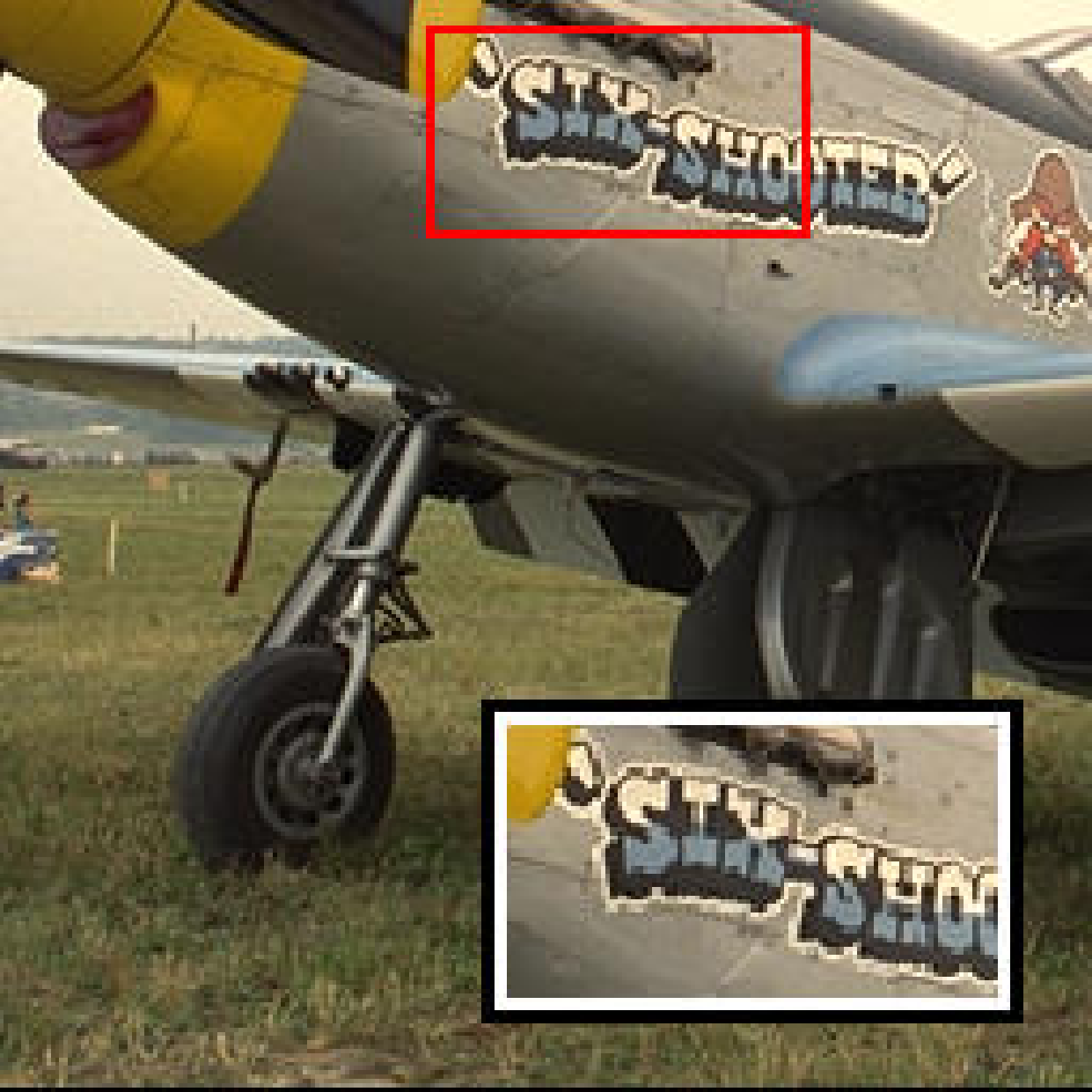}
    \end{tabular}

    \caption{
    Additional qualitative comparison on the \texttt{kodim20} image. ($256 \times 256$)
    }
    \label{fig:kodim20_appendix_qualitative}
\end{figure*}

\begin{figure*}[ht!]
    \centering
    \setlength{\tabcolsep}{2pt}
    \renewcommand{\arraystretch}{1.0}
    \begin{tabular}{cccccc}
        {\footnotesize SIREN}{\scriptsize (27.9 dB)} &
        {\footnotesize FFN}{\scriptsize (26.3 dB)} &
        {\footnotesize GaussNet}{\scriptsize (24.2 dB)} &
        {\footnotesize WIRE}{\scriptsize (28.4 dB)} &
        {\footnotesize \textbf{ELM-INR}}{\scriptsize \textbf{(79.0 dB)}} &
        GT \\
        \includegraphics[width=0.155\linewidth]{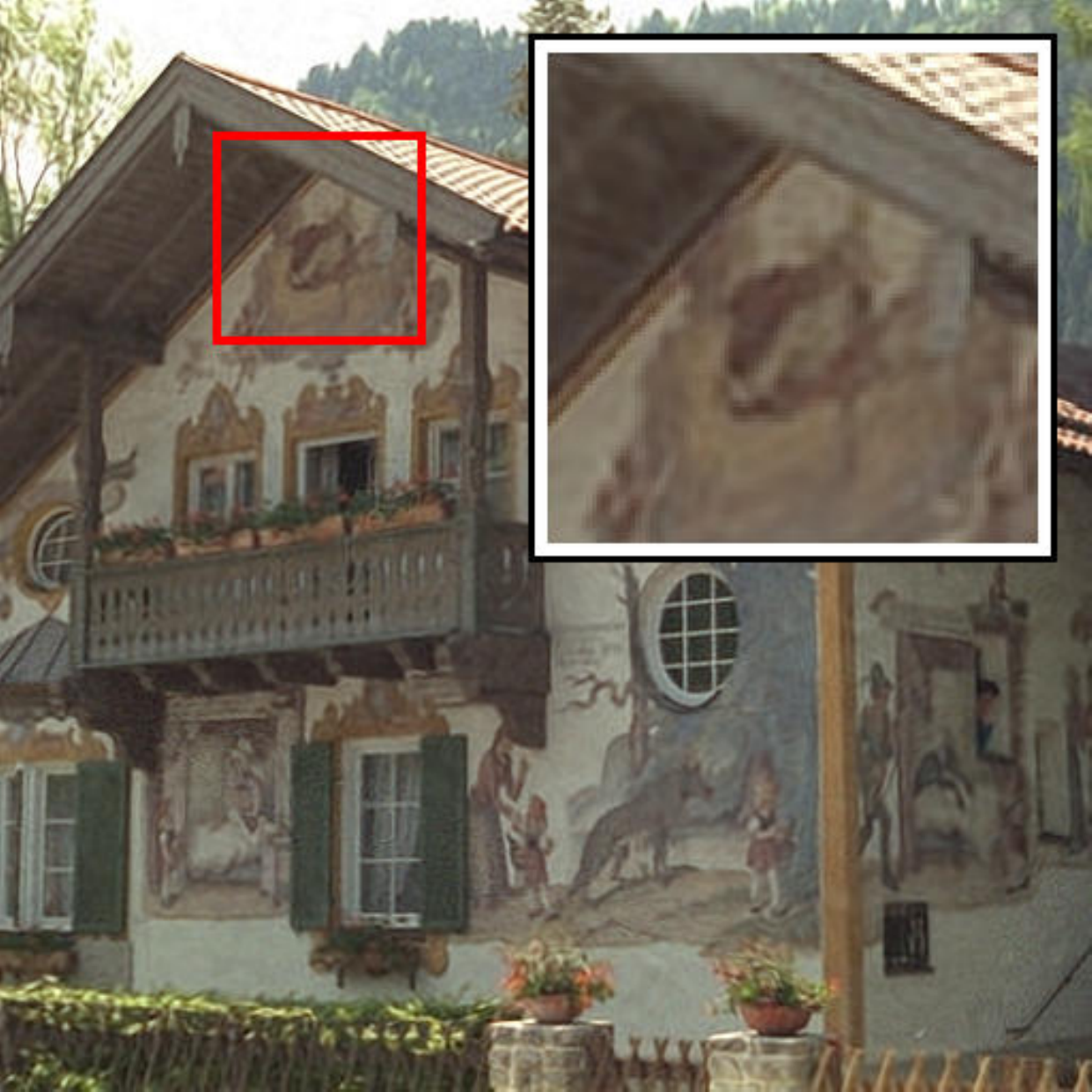} &
        \includegraphics[width=0.155\linewidth]{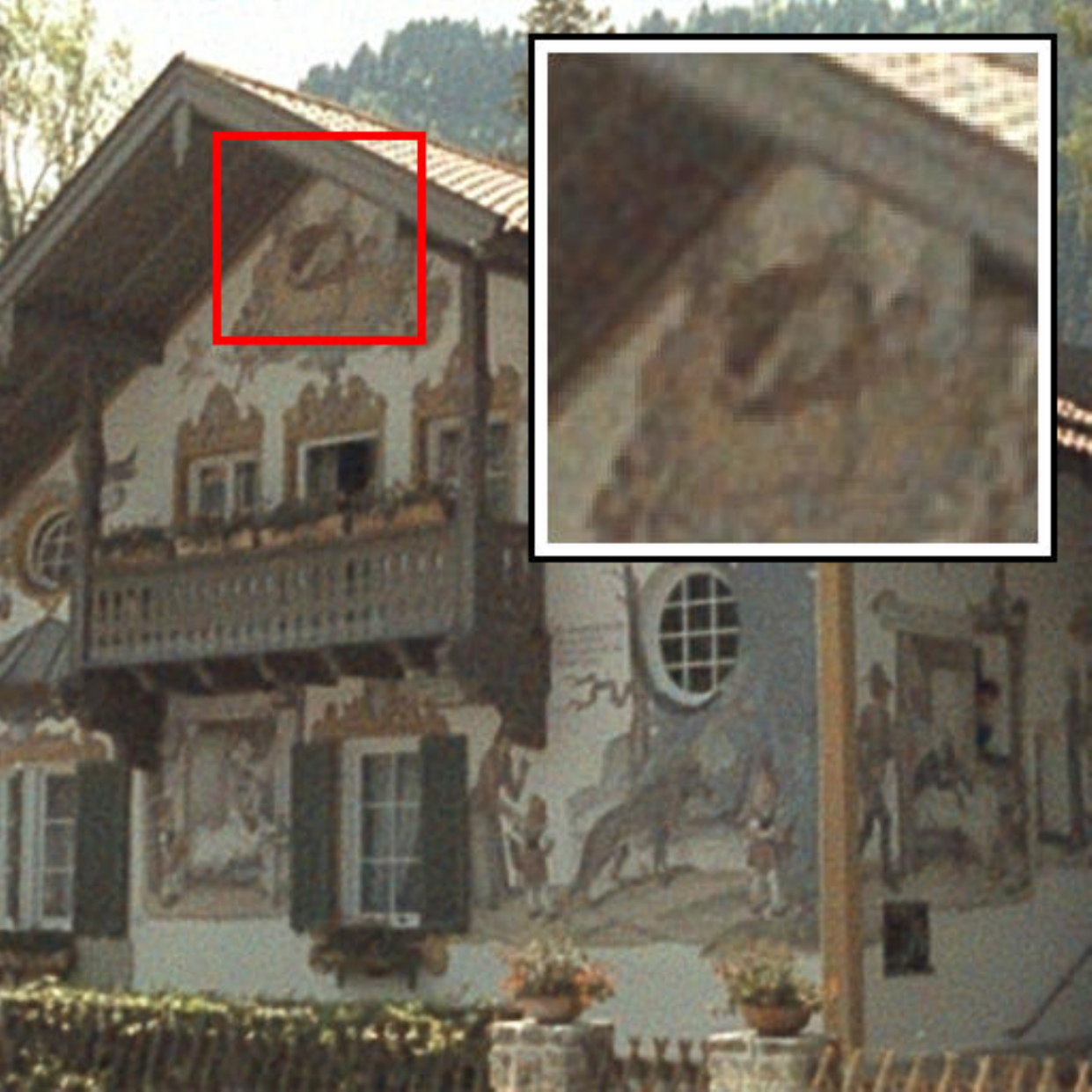} &
        \includegraphics[width=0.155\linewidth]{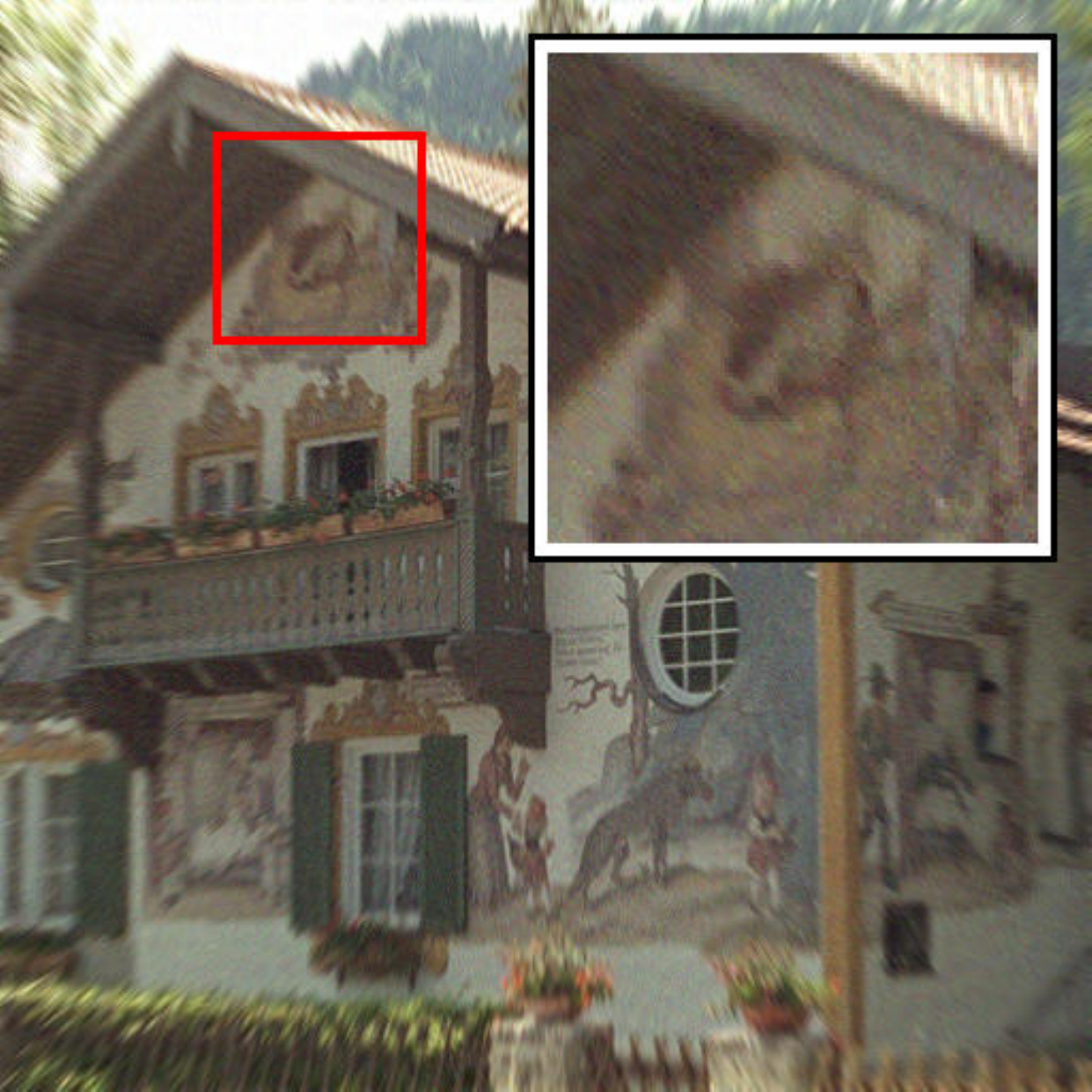} &
        \includegraphics[width=0.155\linewidth]{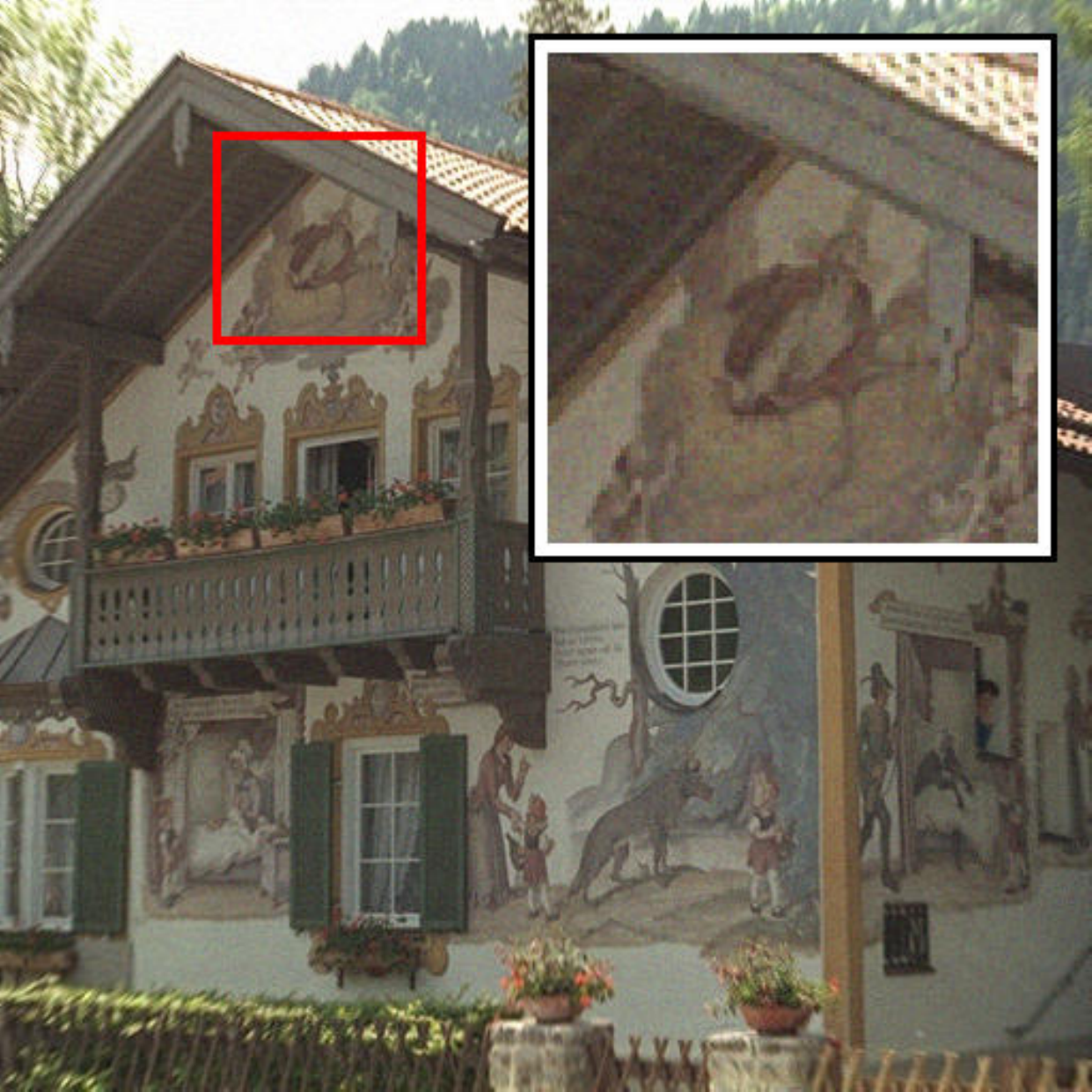} &
        \includegraphics[width=0.155\linewidth]{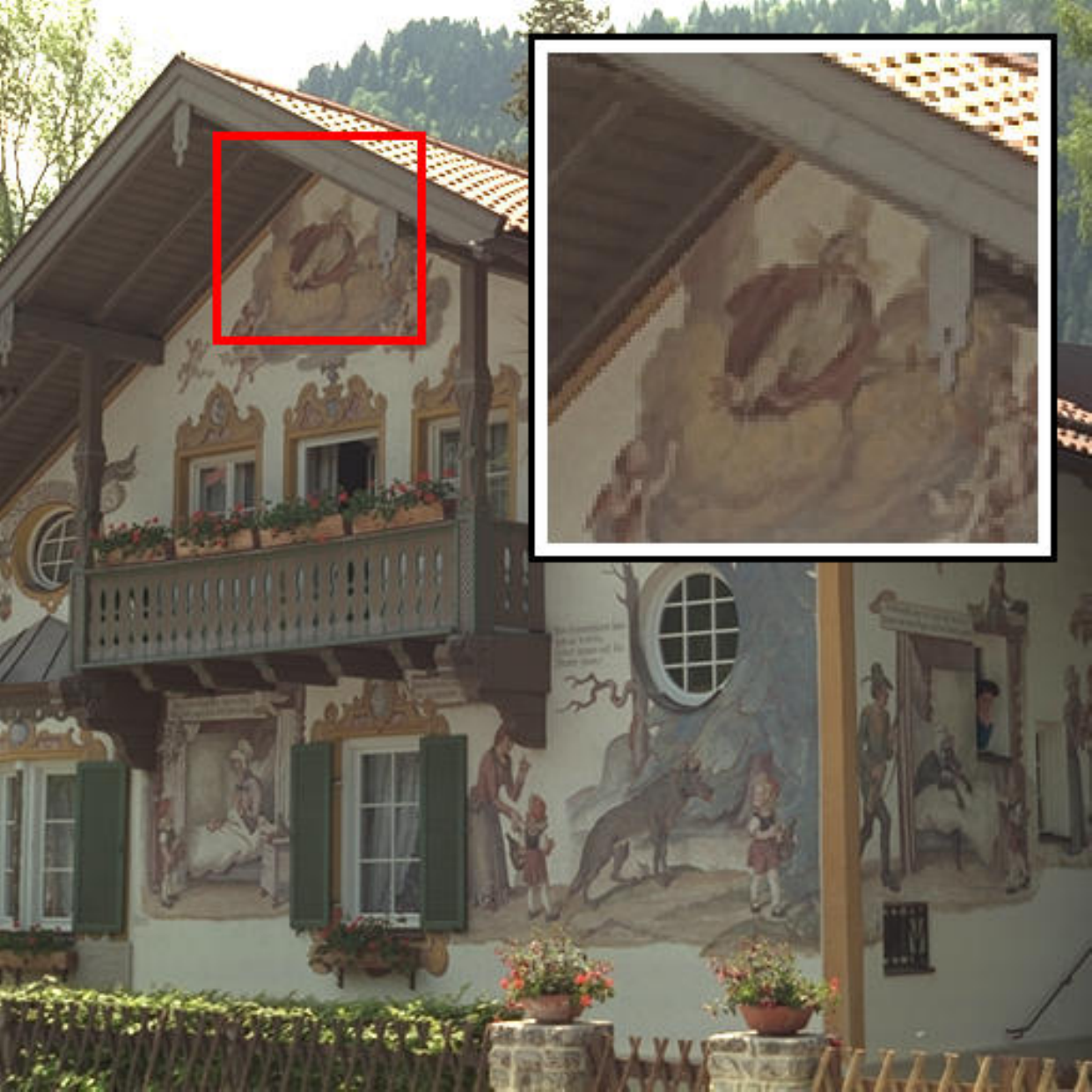} &
        \includegraphics[width=0.155\linewidth]{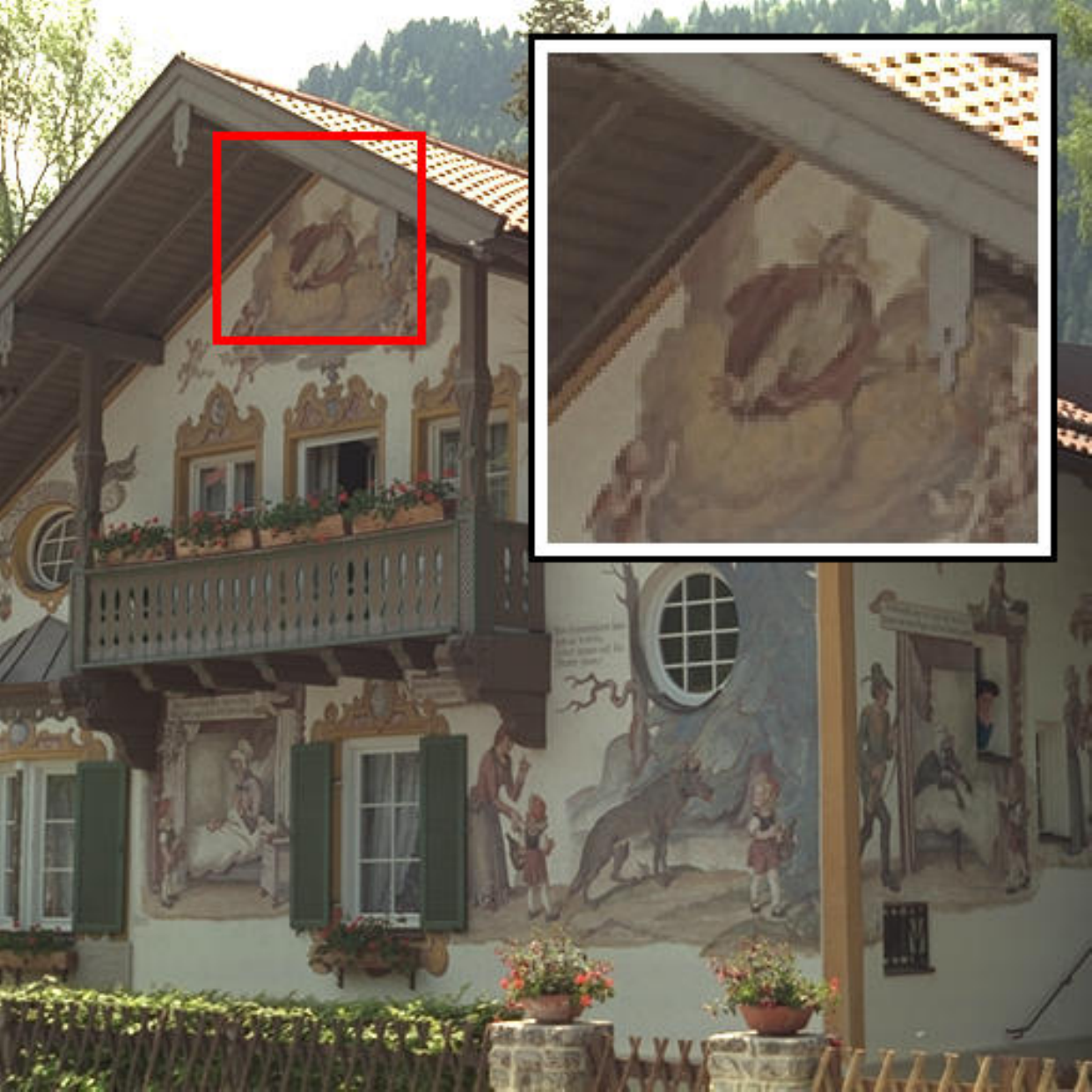}
    \end{tabular}

    \caption{
    Additional qualitative comparison on the \texttt{kodim24} image. ($512 \times 512$)
    }
    \label{fig:kodim24_appendix_qualitative}
\end{figure*}

\begin{figure*}[ht!]
    \centering
    \setlength{\tabcolsep}{2pt}
    \renewcommand{\arraystretch}{1.0}
    \begin{tabular}{ccc}
        \includegraphics[width=0.32\linewidth]{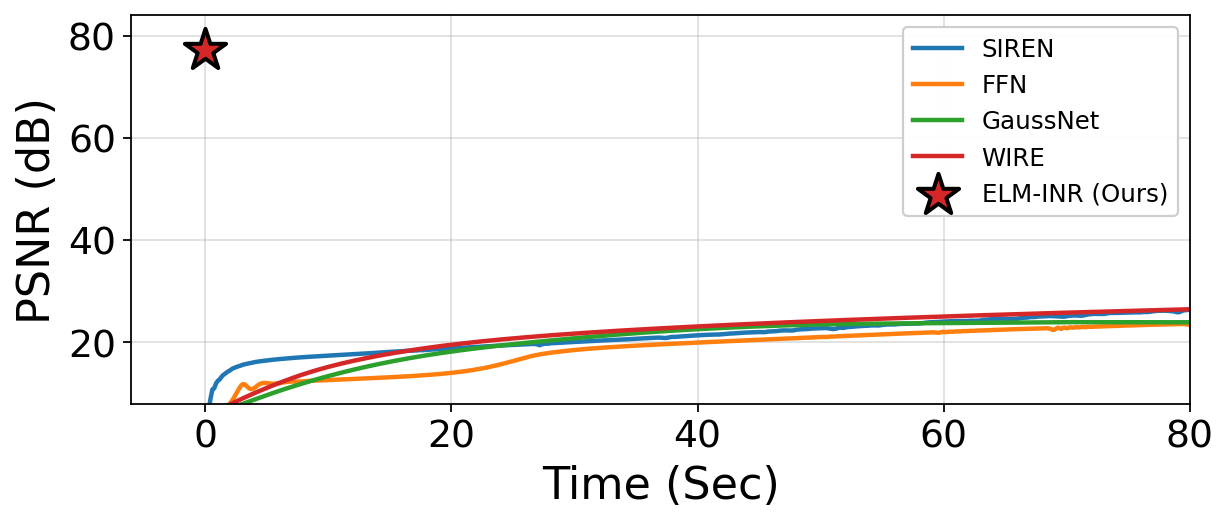} &
        \includegraphics[width=0.32\linewidth]{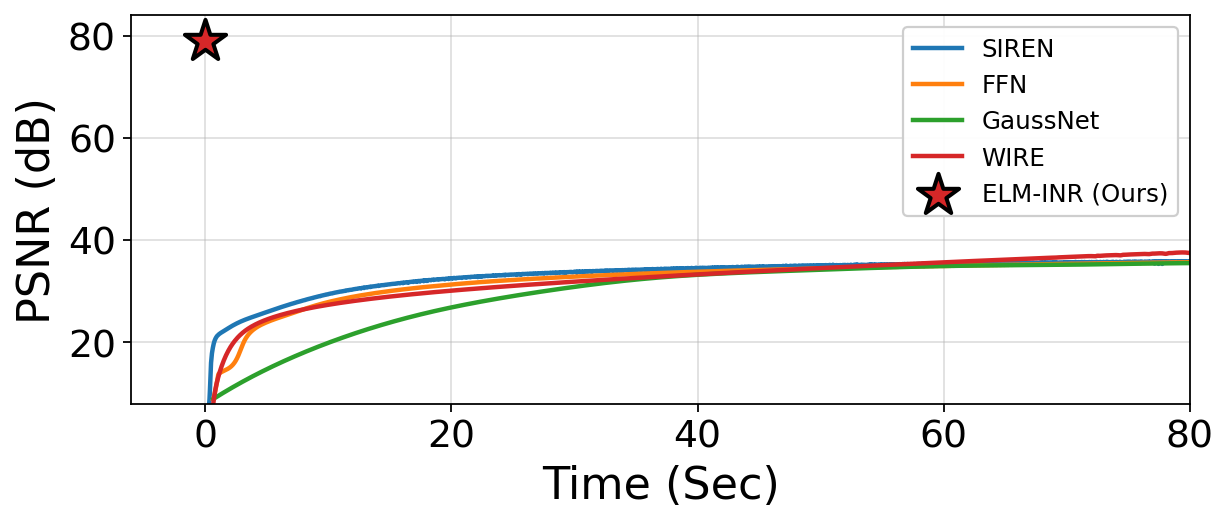} &
        \includegraphics[width=0.32\linewidth]{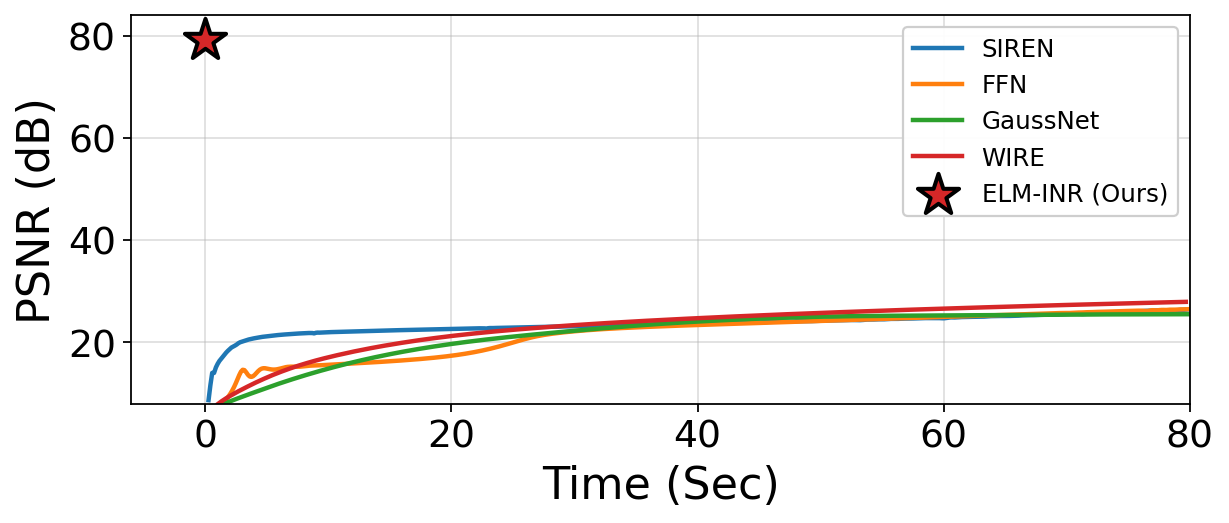} \\

        \scriptsize (a) \texttt{kodim08} ($512 \times 512$)&
        \scriptsize (b) \texttt{kodim20} ($256 \times 256$) &
        \scriptsize (c) \texttt{kodim24} ($512 \times 512$)
    \end{tabular}

    \caption{
    PSNR as a function of wall-clock time for different INR models on representative images.
    ELM-INR achieves substantially higher reconstruction quality within a short inference time compared to iterative INRs.
    }
    \label{fig:time_psnr}
\end{figure*}

We provide additional qualitative and efficiency comparisons on representative standard images from the Kodak dataset, including \texttt{kodim08}, \texttt{kodim20}, and \texttt{kodim24}. As shown in Figures~\ref{fig:kodim08_appendix_qualitative}--\ref{fig:kodim24_appendix_qualitative}, ELM-INR consistently achieves high-fidelity reconstructions across all datasets, accurately recovering fine-scale structures that are blurred or distorted by backpropagation-based INR baselines.

Figure~\ref{fig:time_psnr} further reports PSNR as a function of wall-clock time. While iterative INR methods exhibit gradual improvement due to repeated optimization steps, ELM-INR attains substantially higher PSNR within a short runtime, reflecting its closed-form, backpropagation-free training. These results demonstrate that the proposed method is both highly effective and computationally efficient across a range of standard computer vision benchmarks.


\subsection{Qualitative Comparison under a \texorpdfstring{$\mathrm{CO}_2$}{CO2} Emission Budget}\label{app:co2}
\begin{figure}[H]
    \centering
    \setlength{\tabcolsep}{2pt}
    \renewcommand{\arraystretch}{1.0}
    \begin{tabular}{cccccc}
        {\footnotesize SIREN}{\scriptsize (27.1 dB)} &
        {\footnotesize FFN}{\scriptsize (14.9 dB)} &
        {\footnotesize GaussNet}{\scriptsize (19.4 dB)} &
        {\footnotesize WIRE}{\scriptsize (13.4 dB)} &
        {\footnotesize \textbf{ELM-INR}}{\scriptsize \textbf{(77.8 dB)}} &
        {\footnotesize GT} \\
        \includegraphics[width=0.155\linewidth]{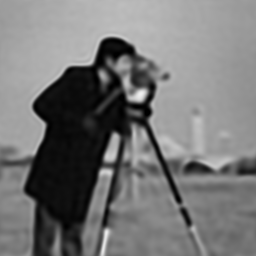} &
        \includegraphics[width=0.155\linewidth]{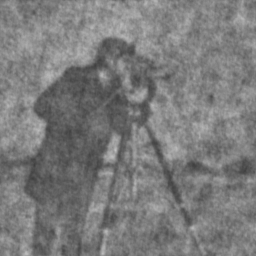} &
        \includegraphics[width=0.155\linewidth]{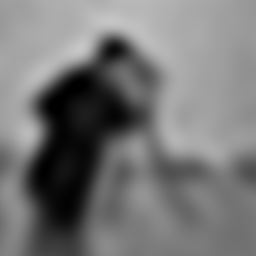} &
        \includegraphics[width=0.155\linewidth]{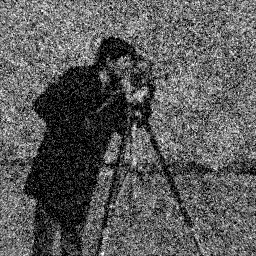} &
        \includegraphics[width=0.155\linewidth]{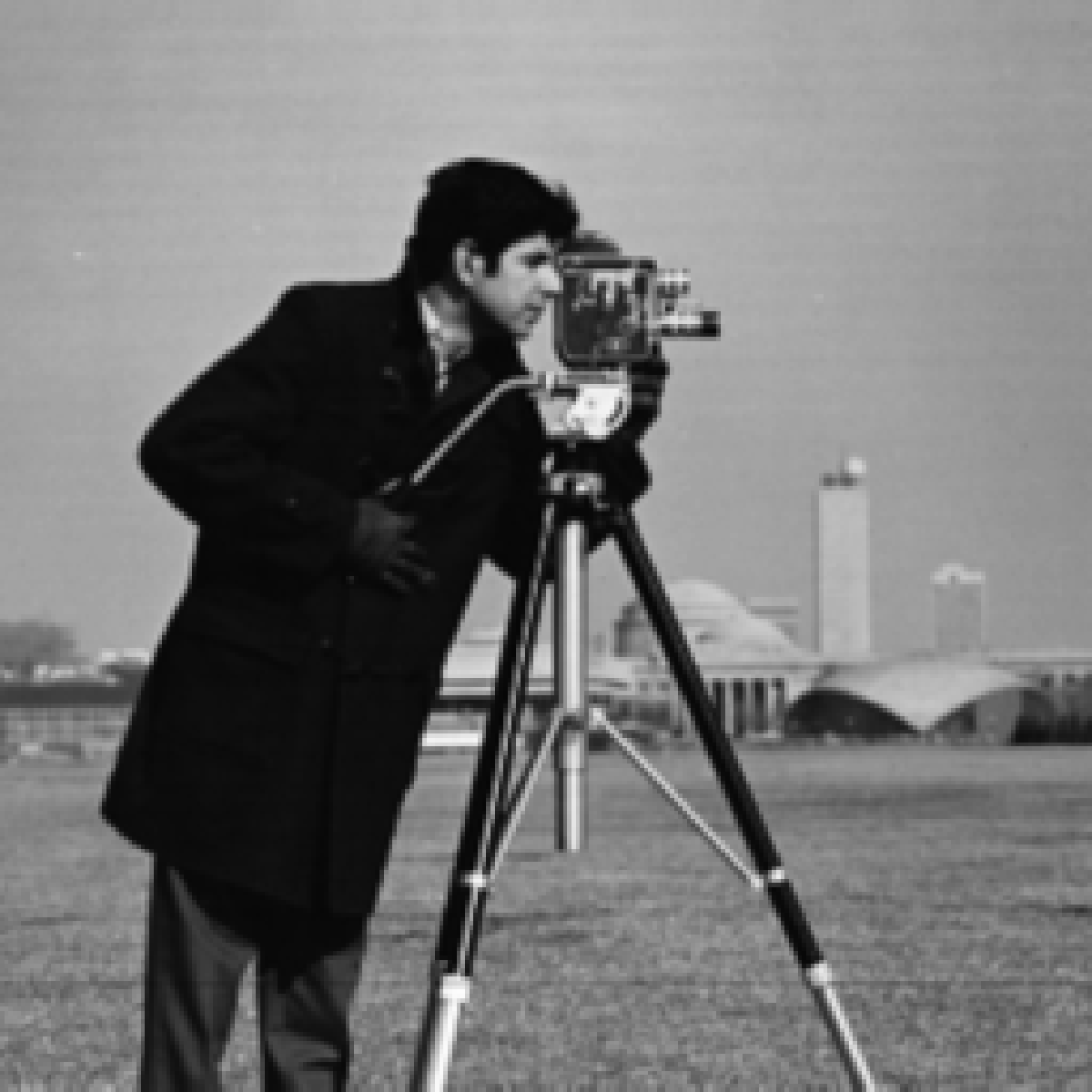} &
        \includegraphics[width=0.155\linewidth]{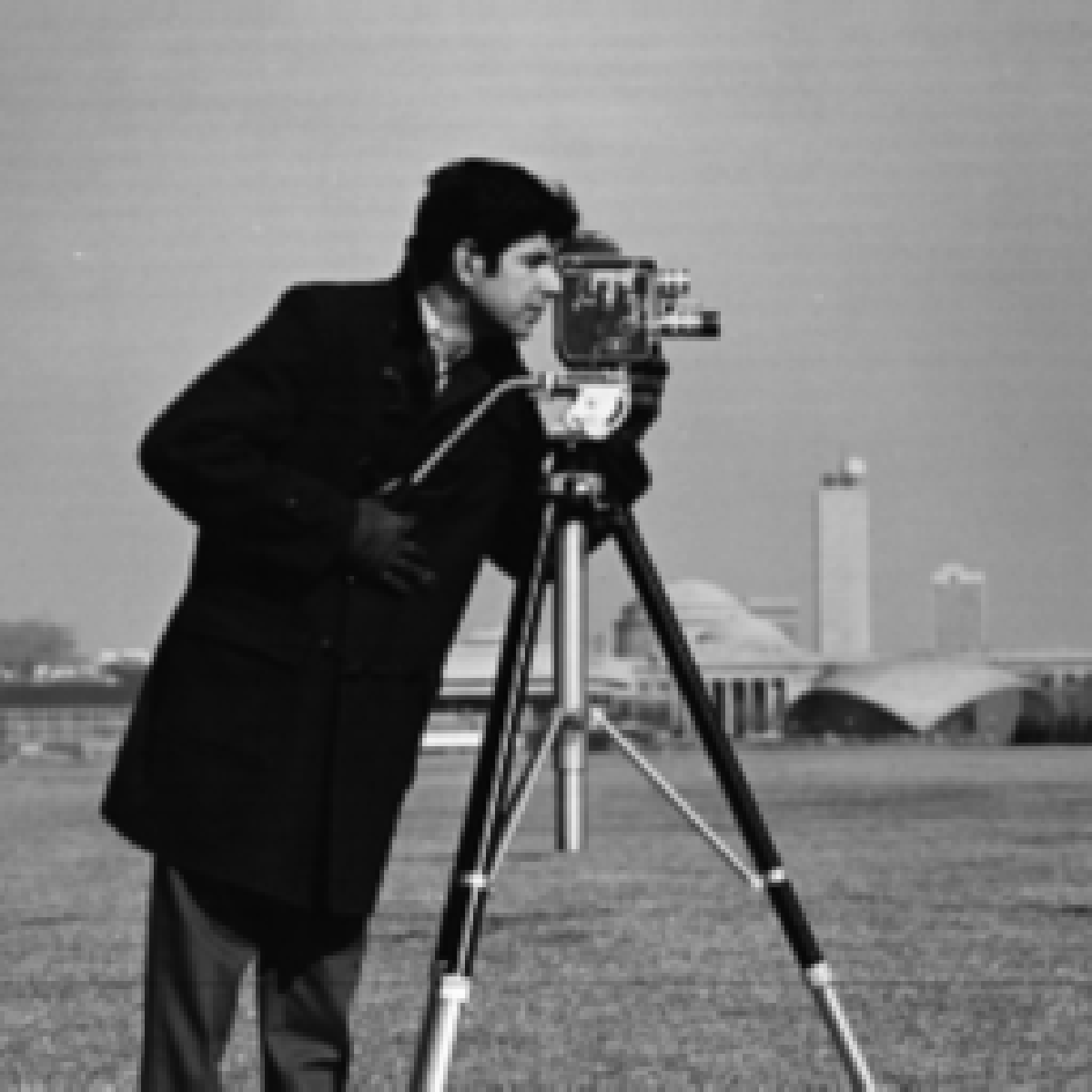}
    \end{tabular}
    \caption{
    Qualitative comparison on the \texttt{cameraman} grayscale image ($256 \times 256$) under CO$_2$-constrained training.
    ELM-INR is used as the reference, while baseline INR methods (SIREN, FFN, GaussNet, and WIRE) are trained with up to $10\times$ larger CO$_2$ emission budgets.
    PSNR values are reported next to each method, illustrating the reconstruction quality achieved under fixed environmental constraints.
    }
    \label{fig:cameraman_qualitative}
\end{figure}

\begin{table}[ht!]
\centering
\scriptsize
\setlength{\tabcolsep}{4pt}
\renewcommand{\arraystretch}{1.1}
\caption{
$\mathrm{CO}_2$-constrained training results on the \texttt{cameraman}.
Metrics are reported under increasing $\mathrm{CO}_2$ budgets relative to ELM-INR.
}
\label{tab:co2_cameraman}
\begin{tabular}{llccccc}
\specialrule{1pt}{2pt}{2pt}
Metric & CO$_2 \times$ & SIREN & FFN & GaussNet & WIRE & ELM-INR \\
\specialrule{1pt}{2pt}{2pt}

& $1\times$  & 20.2 & 11.5 & 11.7 & 5.8  & \textbf{77.8} \\
PSNR (dB)
& $5\times$  & 25.1 & 13.1 & 16.0 & 9.1  & -- \\
& $10\times$ & 27.1 & 14.9 & 19.4 & 13.4 & -- \\

\midrule
& $1\times$  & 34  & 20  & 11  & 5  & -- \\
Steps
& $5\times$  & 79  & 92  & 48  & 28 & -- \\
& $10\times$ & 135 & 178 & 93  & 54 & -- \\

\midrule
& $1\times$  & 1.01 & 0.36 & 0.38 & 0.37 & 0.011 \\
Time (sec)
& $5\times$  & 2.18 & 1.55 & 1.63 & 1.70 & -- \\
& $10\times$ & 3.62 & 2.99 & 3.15 & 3.20 & -- \\

\midrule
& $1\times$  & $1.21\!\times\!10^{-5}$ & $1.18\!\times\!10^{-5}$ & $1.15\!\times\!10^{-5}$ & $1.14\!\times\!10^{-5}$ & $1.13\!\times\!10^{-5}$ \\
$\mathrm{CO}_2$ (kg)
& $5\times$  & $5.70\!\times\!10^{-5}$ & $5.69\!\times\!10^{-5}$ & $5.80\!\times\!10^{-5}$ & $5.87\!\times\!10^{-5}$ & -- \\
& $10\times$ & $1.13\!\times\!10^{-4}$ & $1.13\!\times\!10^{-4}$ & $1.15\!\times\!10^{-4}$ & $1.15\!\times\!10^{-4}$ & -- \\

\specialrule{1pt}{2pt}{2pt}
\end{tabular}
\end{table}
This subsection evaluates the computational efficiency of ELM-INR under explicit $\mathrm{CO}_2$ emission constraints. All experiments are conducted on a single NVIDIA A6000 GPU, and $\mathrm{CO}_2$ emissions are estimated from measured wall-clock time using a consistent hardware-dependent conversion. Figure~\ref{fig:cameraman_qualitative} presents a qualitative comparison on the \texttt{cameraman} image under matched and increased $\mathrm{CO}_2$ budgets, while Table~\ref{tab:co2_cameraman} reports the corresponding reconstruction quality, runtime, and emission statistics.

As summarized in Table~\ref{tab:co2_cameraman}, backpropagation-based INR methods exhibit gradual improvements as the $\mathrm{CO}_2$ budget increases, but their gains remain limited even at significantly higher emission levels. In contrast, ELM-INR achieves substantially higher reconstruction quality with a single closed-form solve, highlighting a fundamentally different efficiency profile under constrained computational and environmental budgets.


\subsection{WorldView-3 8-band Visualizations}\label{app:wv3_bands}
\begin{figure}[H]
    \centering

    \begin{subfigure}{0.23\linewidth}
        \centering
        \includegraphics[width=\linewidth]{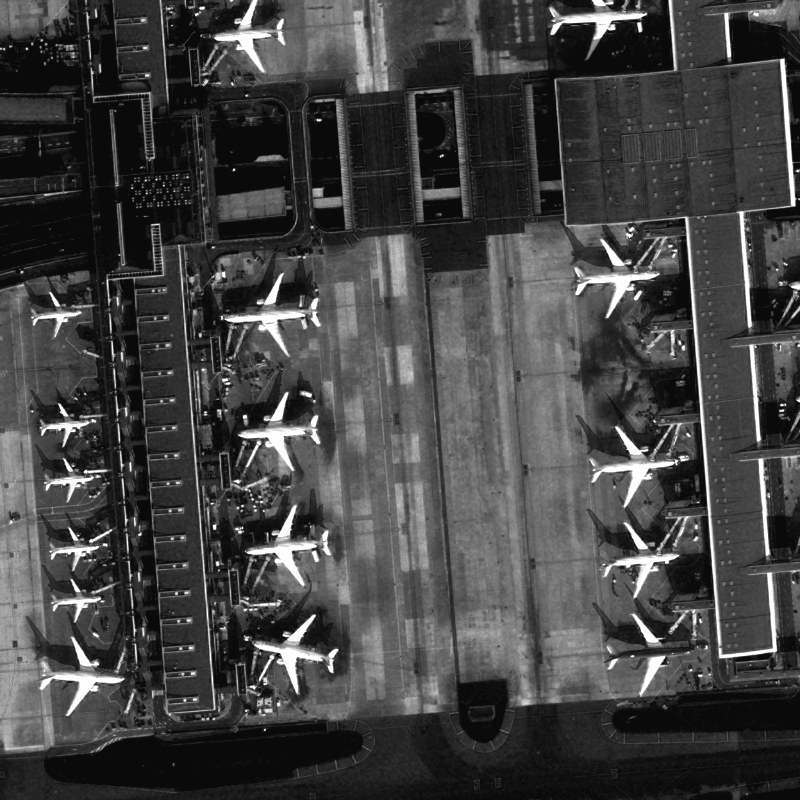}
        \caption{B1}
    \end{subfigure}
    \hfill
    \begin{subfigure}{0.23\linewidth}
        \centering
        \includegraphics[width=\linewidth]{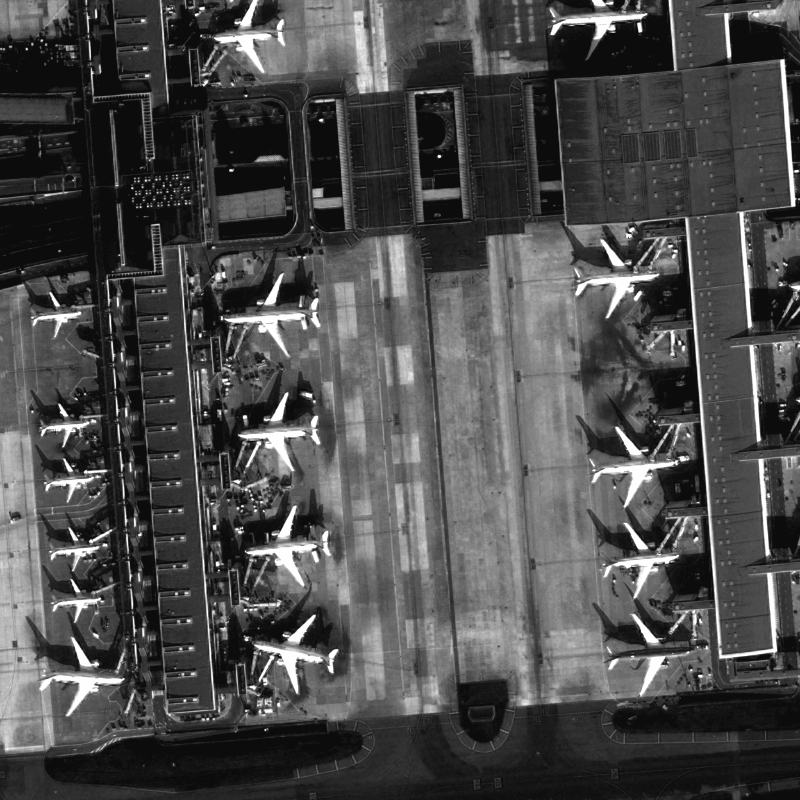}
        \caption{B2}
    \end{subfigure}
    \hfill
    \begin{subfigure}{0.23\linewidth}
        \centering
        \includegraphics[width=\linewidth]{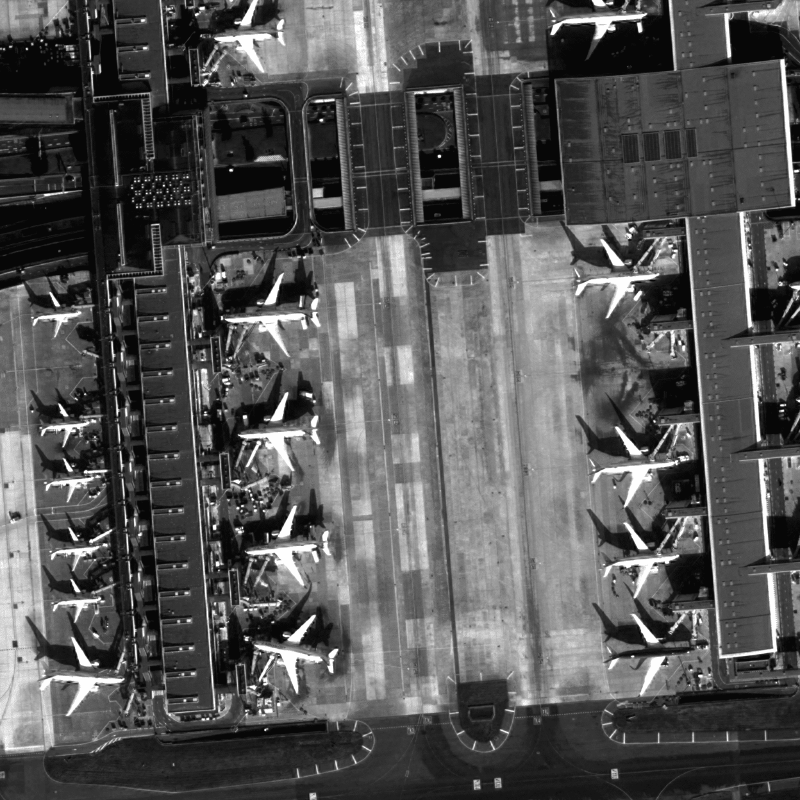}
        \caption{B3}
    \end{subfigure}
    \hfill
    \begin{subfigure}{0.23\linewidth}
        \centering
        \includegraphics[width=\linewidth]{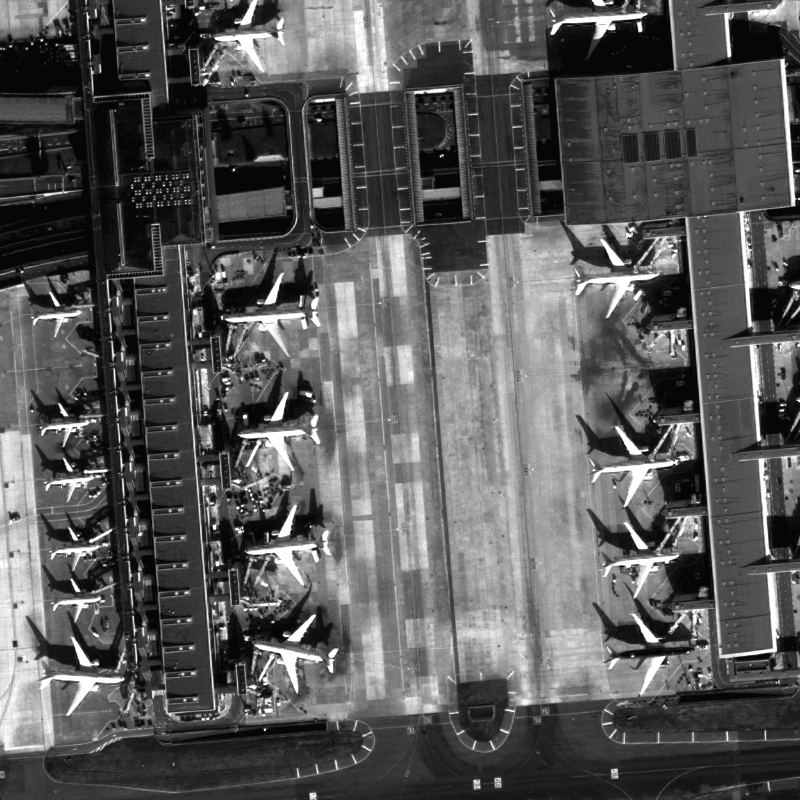}
        \caption{B4}
    \end{subfigure}

    \begin{subfigure}{0.23\linewidth}
        \centering
        \includegraphics[width=\linewidth]{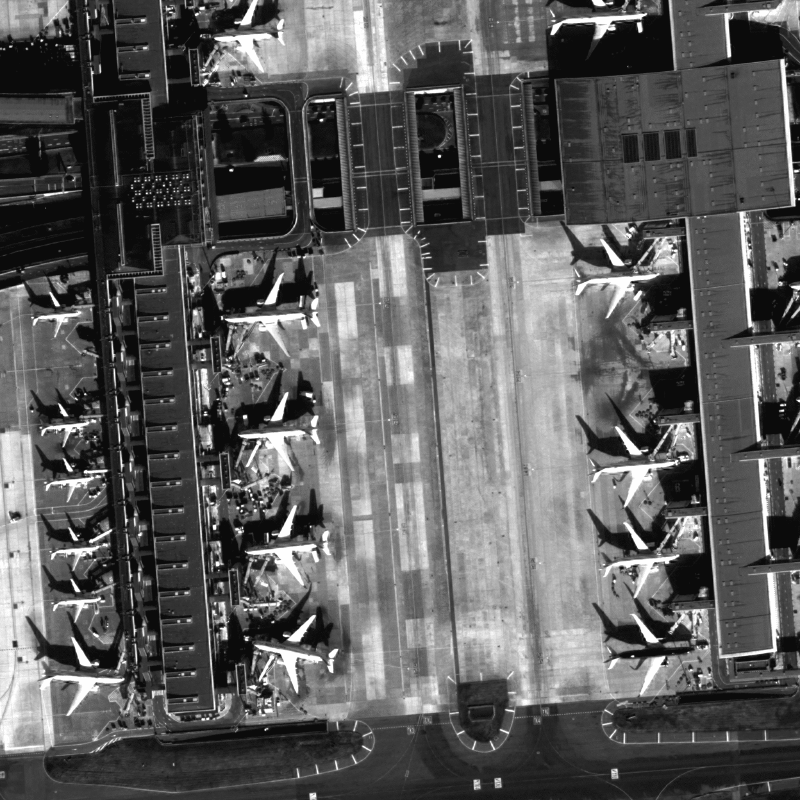}
        \caption{B5}
    \end{subfigure}
    \hfill
    \begin{subfigure}{0.23\linewidth}
        \centering
        \includegraphics[width=\linewidth]{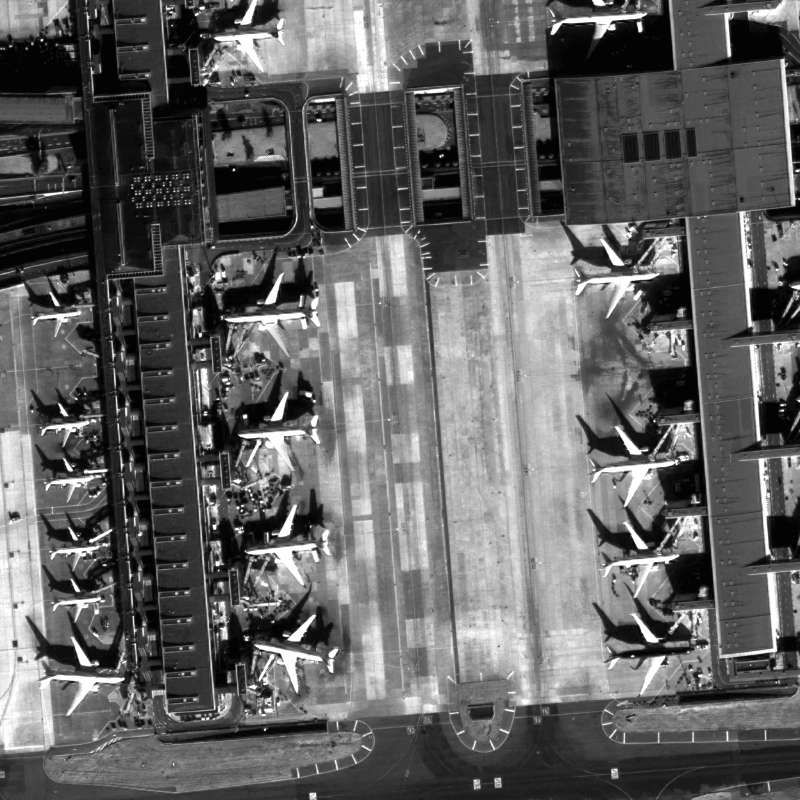}
        \caption{B6}
\end{subfigure}
    \hfill
    \begin{subfigure}{0.23\linewidth}
        \centering
        \includegraphics[width=\linewidth]{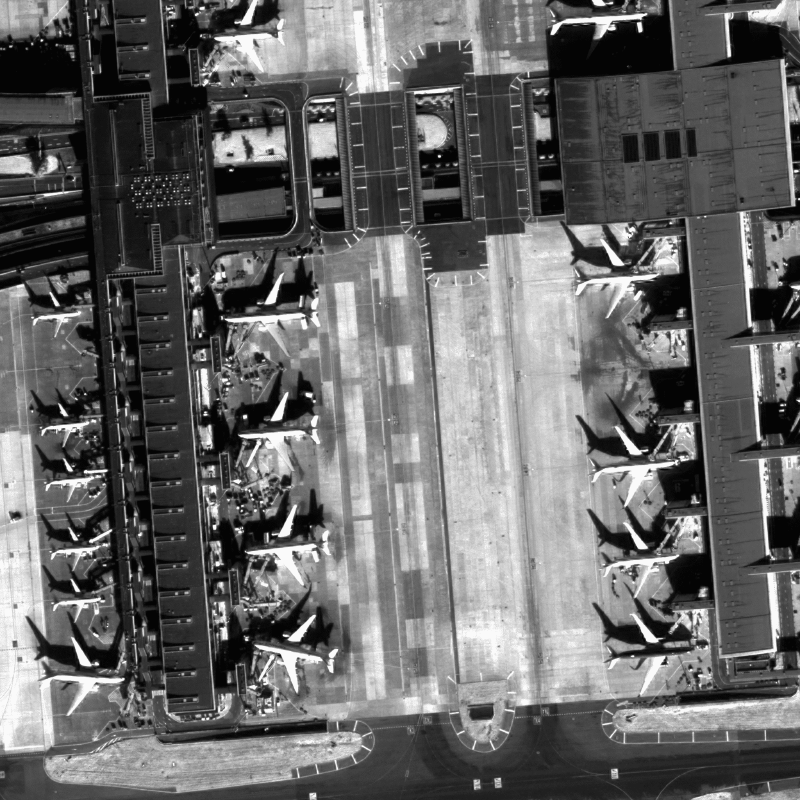}
        \caption{B7}
        \end{subfigure}
    \hfill
    \begin{subfigure}{0.23\linewidth}
        \centering
        \includegraphics[width=\linewidth]{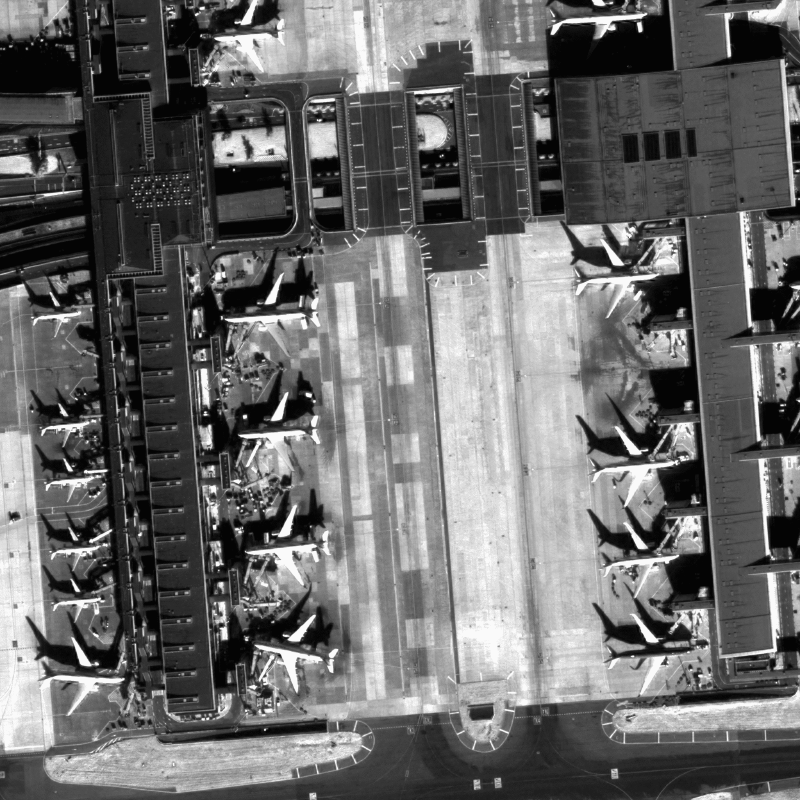}
        \caption{B8}
    \end{subfigure}
    \caption{Grayscale visualization of the eight spectral bands (B1--B8) from a WorldView-3 multispectral image capturing the Charles de Gaulle Airport (CDG).}
    \label{fig:wv3_scene2_bands}
\end{figure}

\begin{figure}[H]
    \centering
    \begin{subfigure}{0.23\linewidth}
        \centering
        \includegraphics[width=\linewidth]{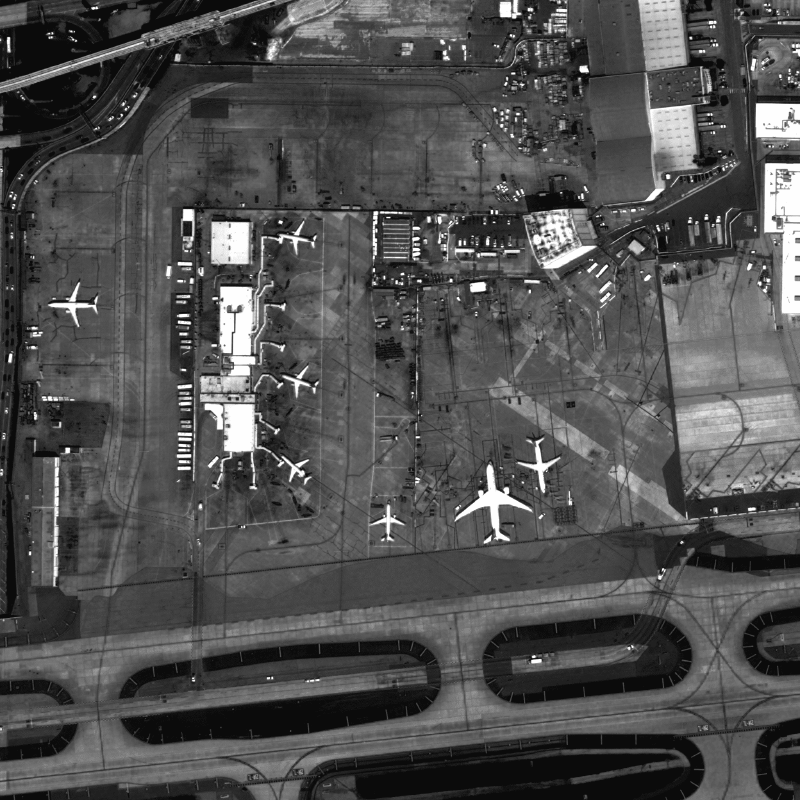}
        \caption{B1}
    \end{subfigure}
    \hfill
    \begin{subfigure}{0.23\linewidth}
        \centering
        \includegraphics[width=\linewidth]{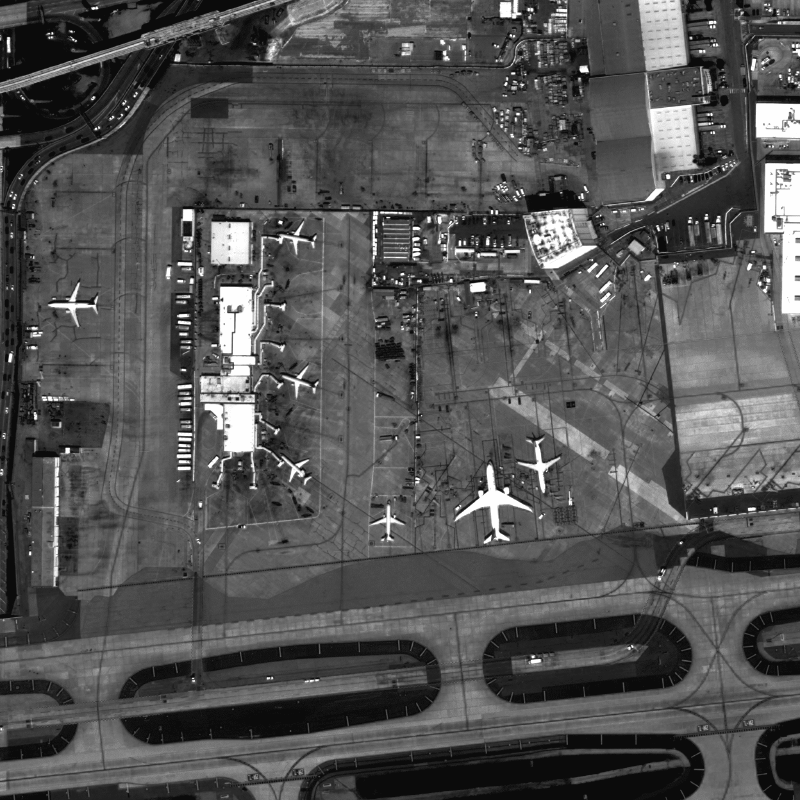}
        \caption{B2}
    \end{subfigure}
    \hfill
    \begin{subfigure}{0.23\linewidth}
        \centering
        \includegraphics[width=\linewidth]{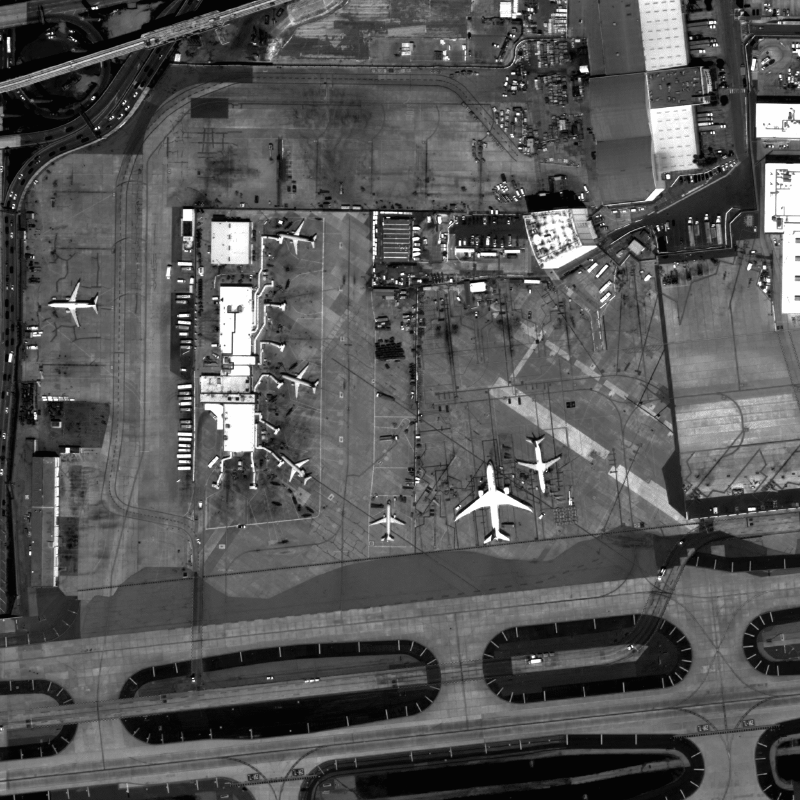}
        \caption{B3}
    \end{subfigure}
    \hfill
    \begin{subfigure}{0.23\linewidth}
        \centering
        \includegraphics[width=\linewidth]{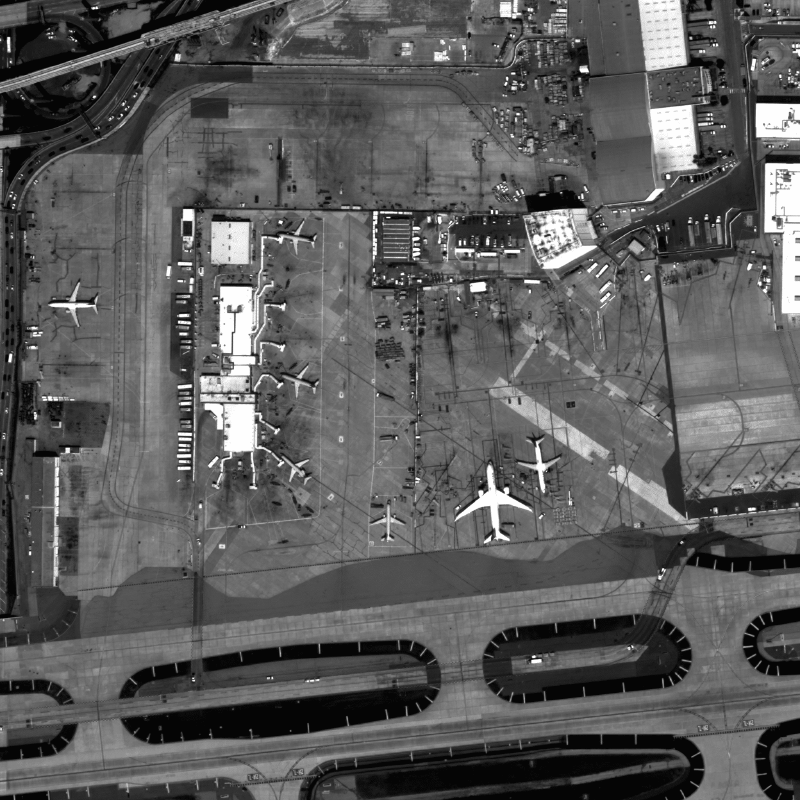}
        \caption{B4}
    \end{subfigure}

    \begin{subfigure}{0.23\linewidth}
        \centering
        \includegraphics[width=\linewidth]{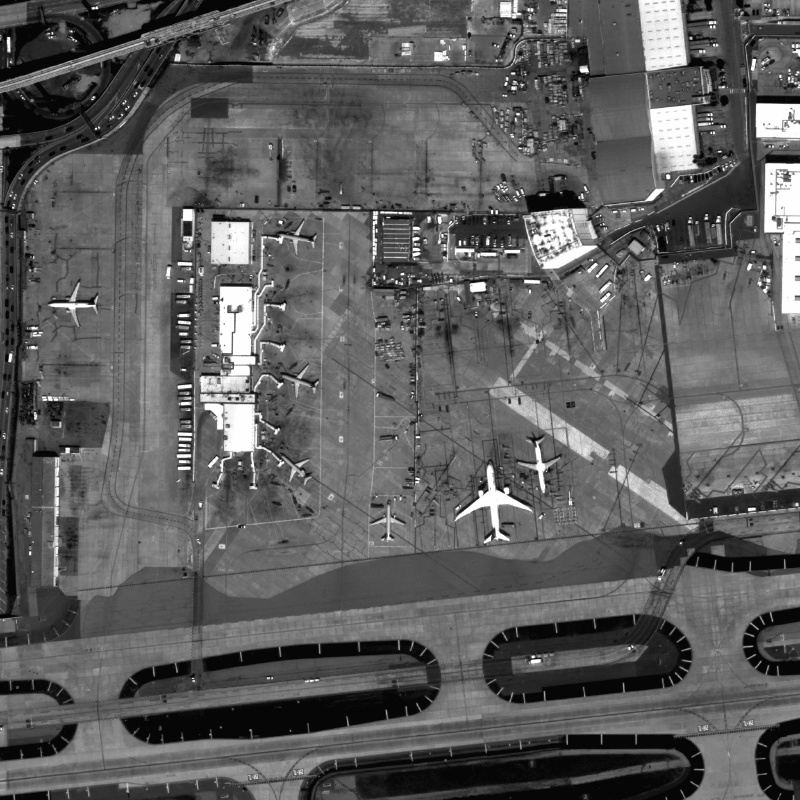}
        \caption{B5}
    \end{subfigure}
    \hfill
    \begin{subfigure}{0.23\linewidth}
        \centering
        \includegraphics[width=\linewidth]{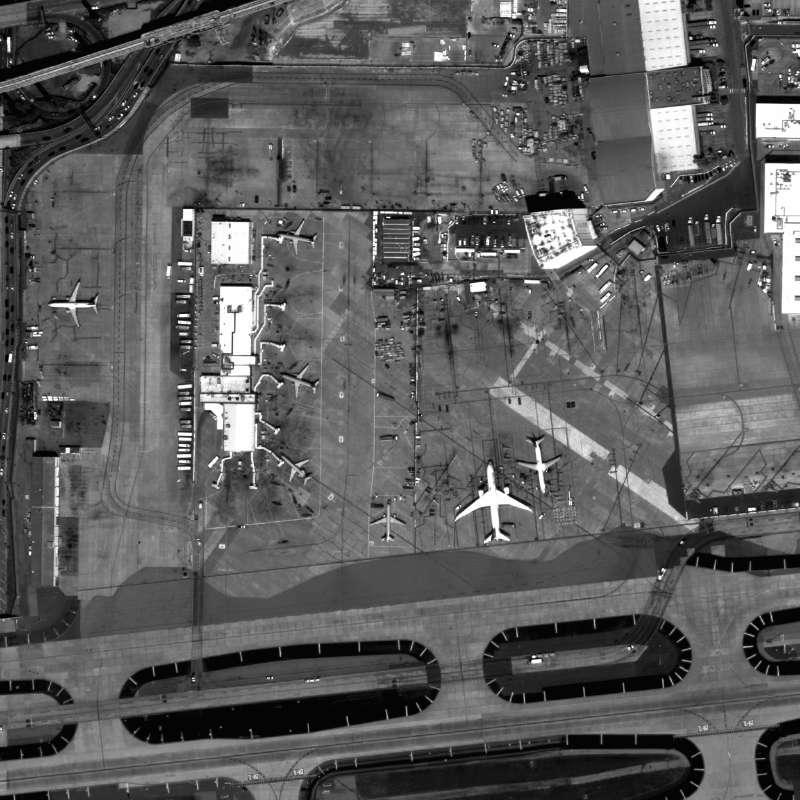}
        \caption{B6}
\end{subfigure}
    \hfill
    \begin{subfigure}{0.23\linewidth}
        \centering
        \includegraphics[width=\linewidth]{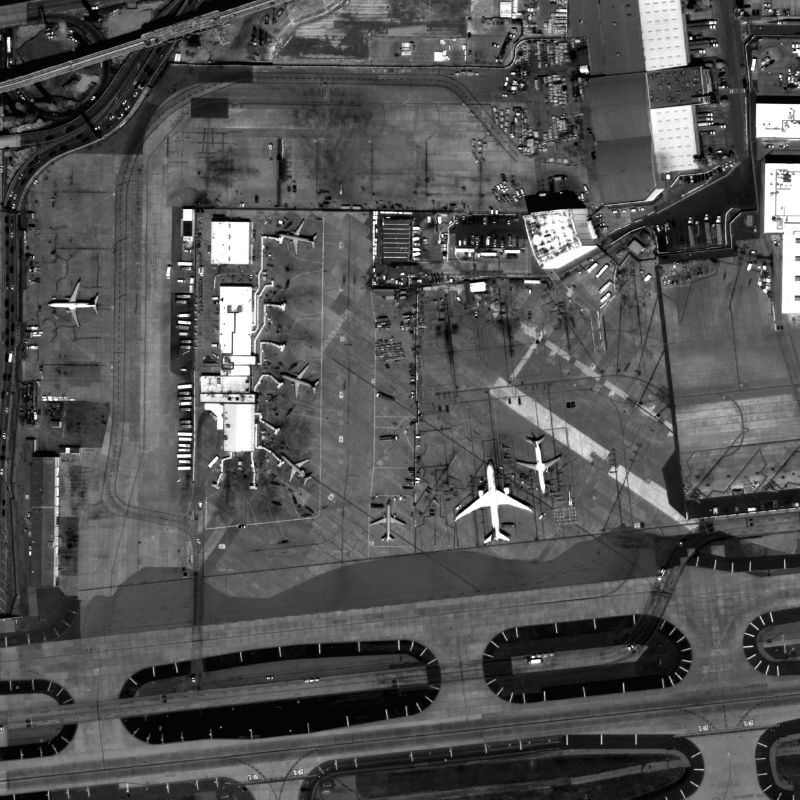}
        \caption{B7}
        \end{subfigure}
    \hfill
    \begin{subfigure}{0.23\linewidth}
        \centering
        \includegraphics[width=\linewidth]{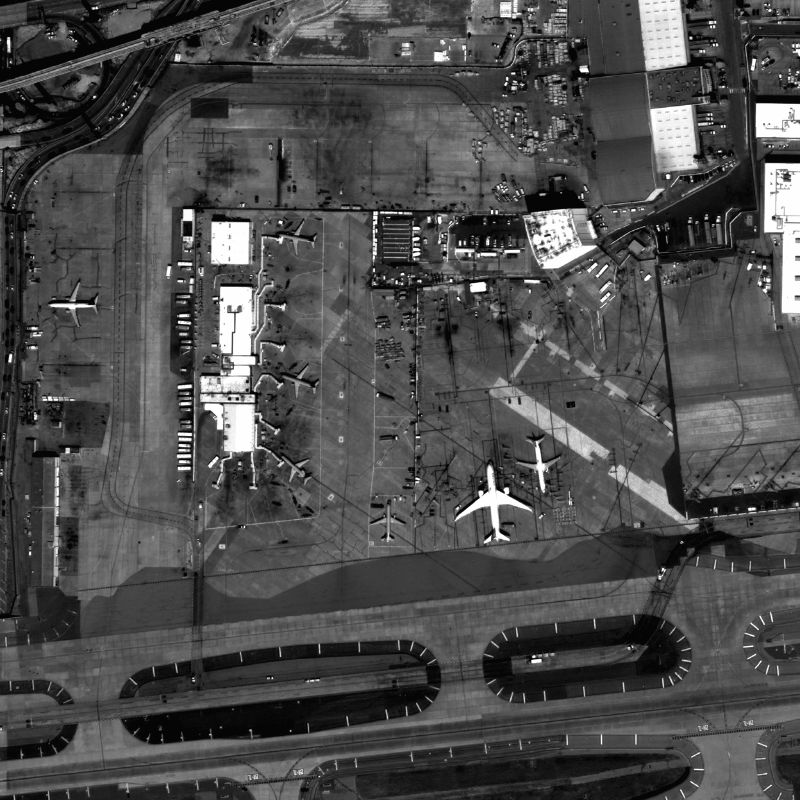}
        \caption{B8}
    \end{subfigure}
    \caption{Grayscale visualization of the eight spectral bands (B1--B8) from a WorldView-3 multispectral image capturing the Los Angeles International Airport (LAX).}
    \label{fig:wv3_scene1_bands}
\end{figure}

To improve reproducibility and provide visual context for our multispectral experiments, we include the two WorldView-3 scenes used in this study. For each scene, we visualize all eight multispectral bands individually. These visualizations highlight band-specific visual patterns and help interpret reconstruction behavior under multi-band supervision. Figures~\ref{fig:wv3_scene2_bands} and~\ref{fig:wv3_scene1_bands} show the eight-band decompositions for the two scenes.


\subsection{Experimental Results on Multispectral Satellite Imagery}
Beyond standard computer-vision benchmarks (i.e., 1-channel grayscale or 3-channel RGB images), we additionally evaluate ELM-INR on multispectral satellite imagery with a substantially larger number of channels. The input is a 2D spatial coordinate $\mathcal{X}=(x_1, x_2)$ on the image plane and the output is the per-pixel spectral vector $V(\mathcal{X}) \in \mathbb{R}^C$. For all satellite experiments, we train the INR by regressing the full spectral vector at each coordinate; for qualitative visualization, we only display the RGB composite. Since satellite scenes are typically higher-resolution and contain diverse, fine-grained structures within a single frame, we train all baseline INR models for 10{,}000 epochs to ensure fair and stable convergence. We additionally perform basic hyperparameter tuning for each baseline (e.g., hidden dimension, number of layers, and learning rate) under this longer training schedule.
\begin{figure*}[ht!]
    \centering
    \setlength{\tabcolsep}{2pt}
    \renewcommand{\arraystretch}{1.0}
    \begin{tabular}{cccccc}
        {\footnotesize SIREN}{\scriptsize (30.56 dB)} &
        {\footnotesize FFN}{\scriptsize (35.08 dB)} &
        {\footnotesize GaussNet}{\scriptsize (28.87 dB)} &
        {\footnotesize WIRE}{\scriptsize (29.20 dB)} &
        {\footnotesize \textbf{ELM-INR}}{\scriptsize \textbf{(78.49 dB)}} &
        {\footnotesize GT} \\

    \includegraphics[width=0.155\linewidth]{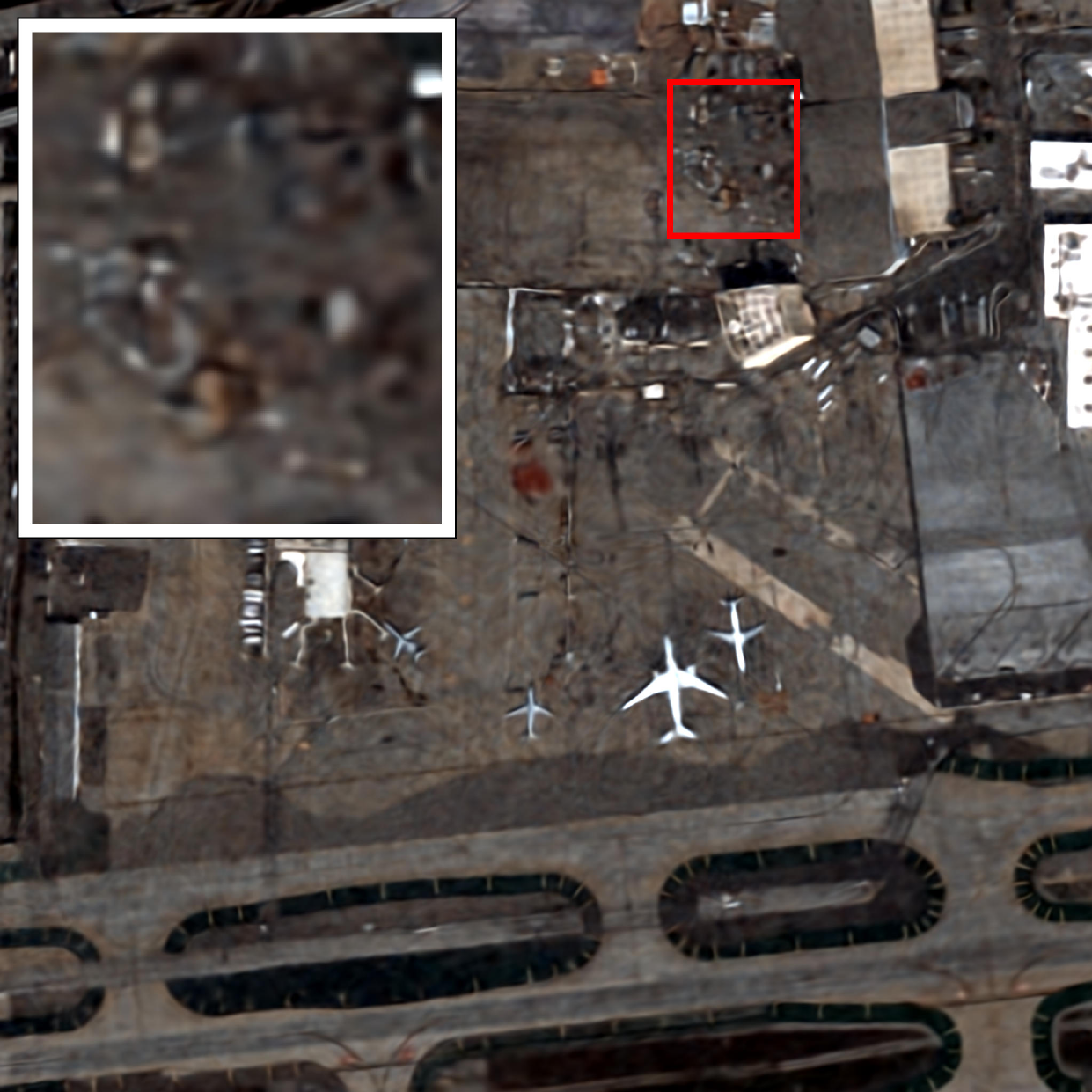} &
    \includegraphics[width=0.155\linewidth]{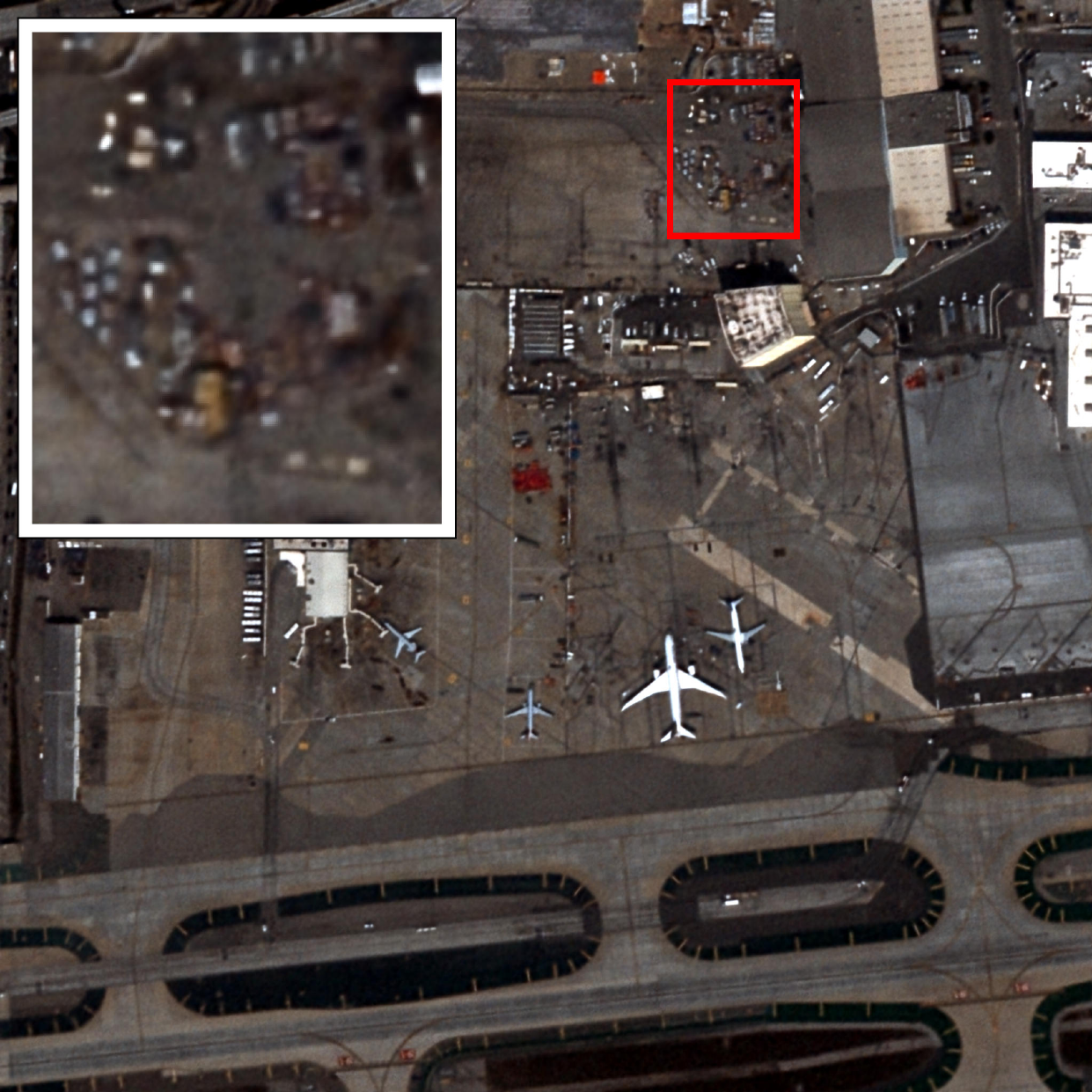} &
    \includegraphics[width=0.155\linewidth]{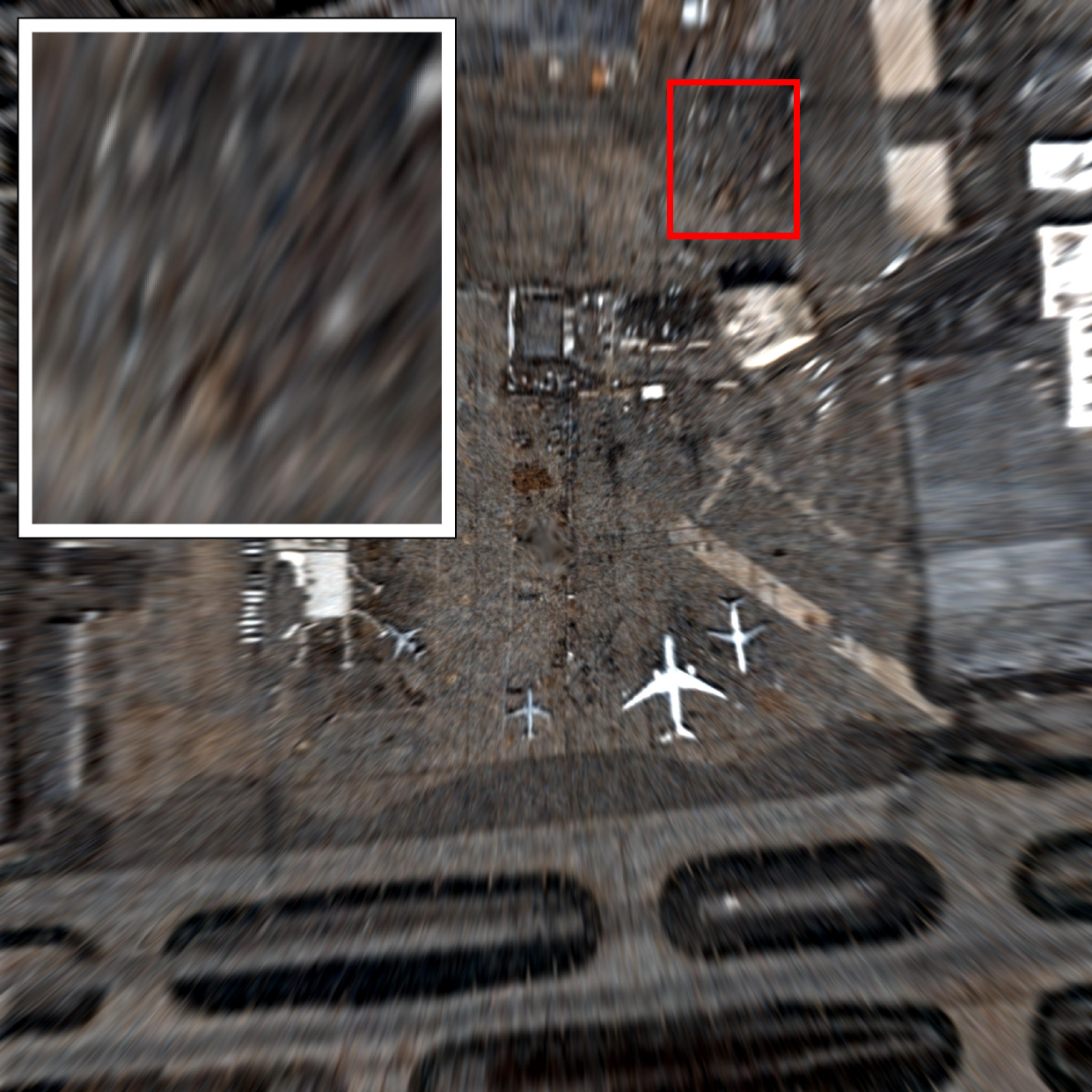} &
    \includegraphics[width=0.155\linewidth]{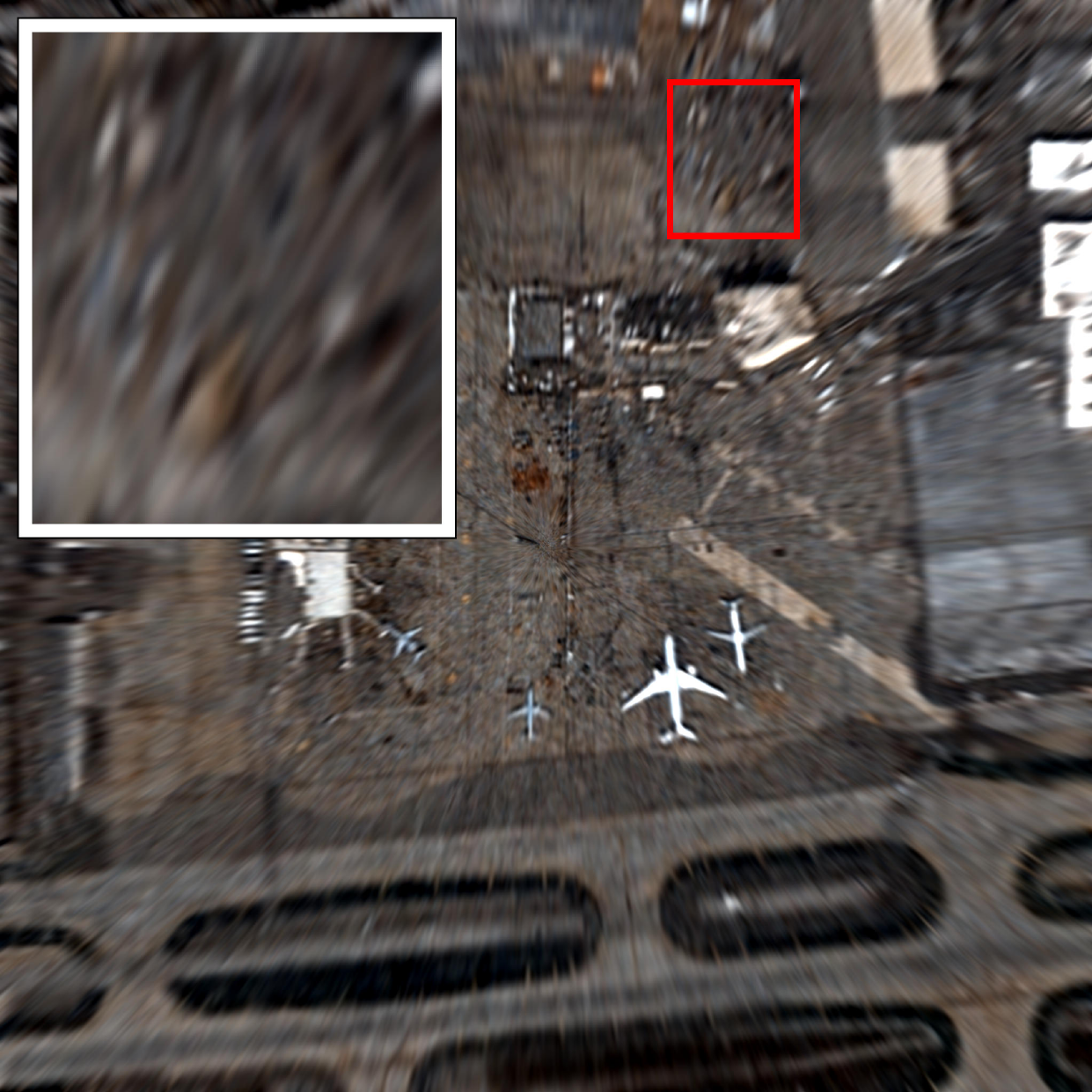} &
    \includegraphics[width=0.155\linewidth]{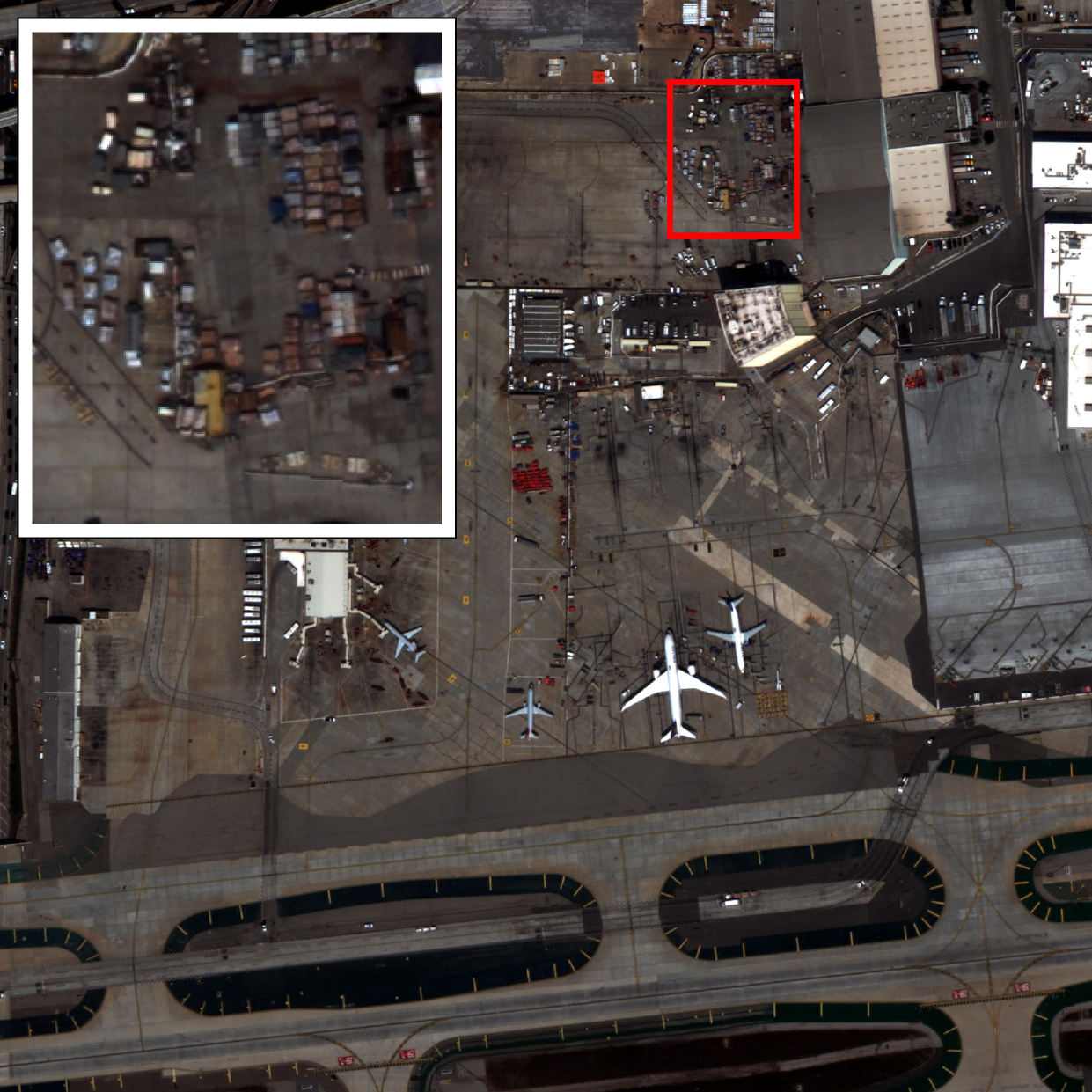} &
    \includegraphics[width=0.155\linewidth]{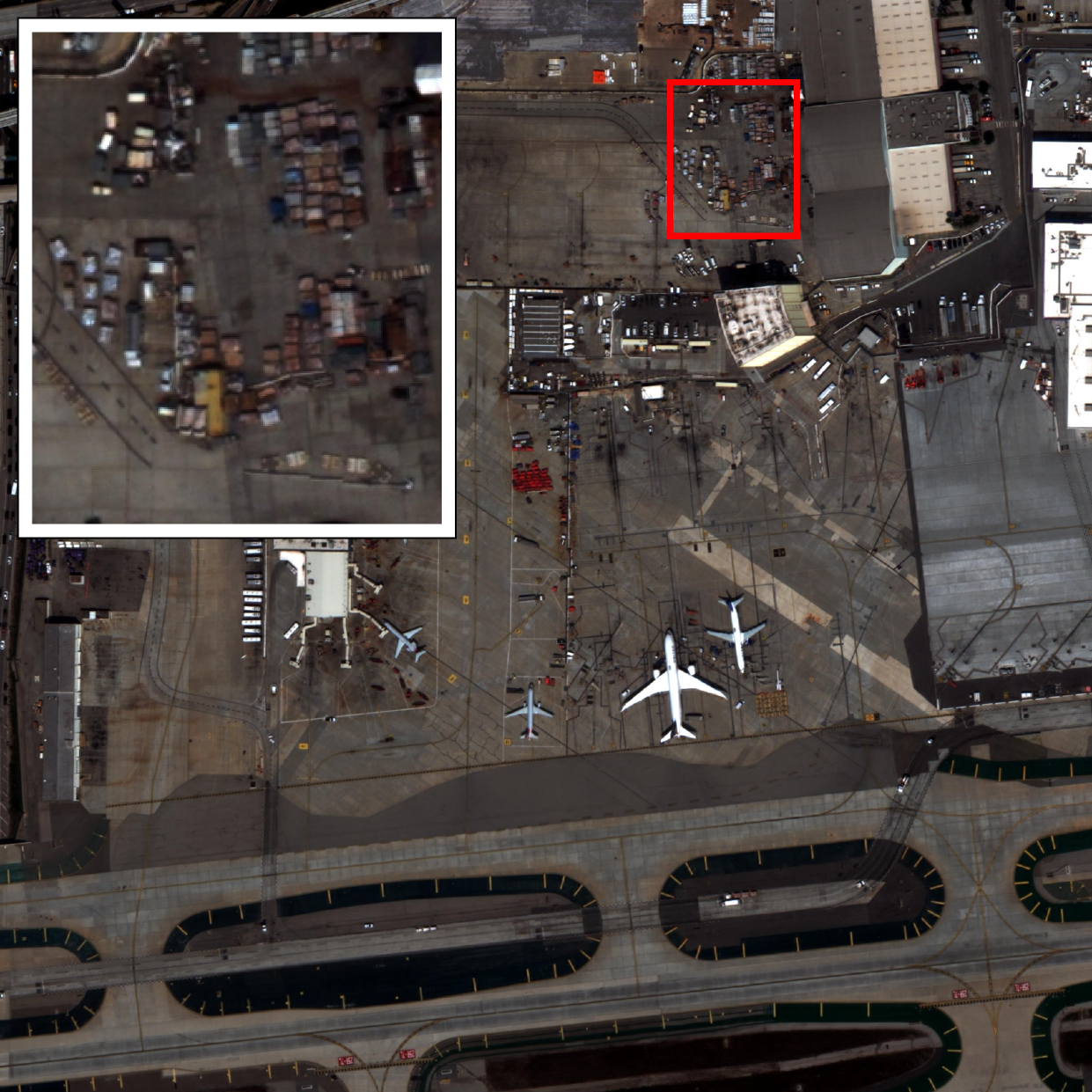}
    \end{tabular}
    \caption{
    Qualitative comparison on a high-frequency region from a WorldView-3 multispectral image.
    The image consists of 8 spectral bands with a spatial resolution of $2048 \times 2048$ pixels, while the visualization shows only the RGB channels for clarity.
    ELM-INR achieves substantially higher reconstruction fidelity, recovering fine-grained structures that are heavily blurred or distorted by baseline INR models.
    }
    \label{fig:wv3_la}
\end{figure*}

\begin{table}[ht!]
\centering
\small
\caption{WorldView-3 8-Band Specifications and Usage}
\label{tab:wv3_description}
\renewcommand{\arraystretch}{1.3}
\resizebox{\columnwidth}{!}{%
\begin{tabular}{llll}
\specialrule{1pt}{2pt}{2pt}
\textbf{Band} & \textbf{Description} & \textbf{Wave. Range (nm)} & \textbf{Primary Usage} \\
\specialrule{1pt}{2pt}{2pt}
B1 & Coastal & $400-450$ & Coastal monitoring and atmospheric correction \\ \cmidrule(lr){1-4}
B2 & Blue & $450-510$ & Aerosol/haze effects and water quality cues \\ \cmidrule(lr){1-4}
B3 & Green & $510-580$ & Vegetation vigor and true-color rendering \\ \cmidrule(lr){1-4}
B4 & Yellow & $585-625$ & Vegetation stress and chlorosis cues  \\ \cmidrule(lr){1-4}
B5 & Red & $630-690$ & Chlorophyll absorption and vegetation discrimination \\ \cmidrule(lr){1-4}
B6 & Red Edge & $705-745$ & Chlorophyll-related sensitivity and crop condition monitoring \\ \cmidrule(lr){1-4}
B7 & NIR-1 & $770-895$ & Biomass estimation, soil moisture, water body delineation \\ \cmidrule(lr){1-4}
B8 & NIR-2 & $860-1040$ & Vegetation and moisture-related sensitivity \\
\specialrule{1pt}{2pt}{2pt}
\end{tabular}}
\end{table}

\paragraph{WorldView-3.}
Figure~\ref{fig:wv3_la} illustrates a high-frequency crop from a WorldView-3 multispectral scene (a commercial satellite operated by Maxar Technologies, formerly DigitalGlobe) covering the Los Angeles International Airport (LAX).
The sample consists of $C=8$ spectral bands and is processed at a spatial resolution of $2048\times2048$ pixels.
As shown in Figure~\ref{fig:wv3_la}, the airport area contains complex man-made structures such as runways, taxiways, and buildings, which induce strong high-frequency components. This makes the region a particularly challenging testbed for evaluating spectral bias in implicit neural representations.

\begin{figure*}[ht!]
    \centering
    \setlength{\tabcolsep}{2pt}
    \renewcommand{\arraystretch}{1.0}
    \begin{tabular}{cccccc}
        {\footnotesize SIREN}{\scriptsize (44.20 dB)} &
        {\footnotesize FFN}{\scriptsize (42.54 dB)} &
        {\footnotesize GaussNet}{\scriptsize (38.41dB)} &
        {\footnotesize WIRE}{\scriptsize (37.40 dB)} &
        {\footnotesize \textbf{ELM-INR}}{\scriptsize \textbf{(79.51 dB)}} &
        {\footnotesize GT} \\
        \includegraphics[width=0.155\linewidth]{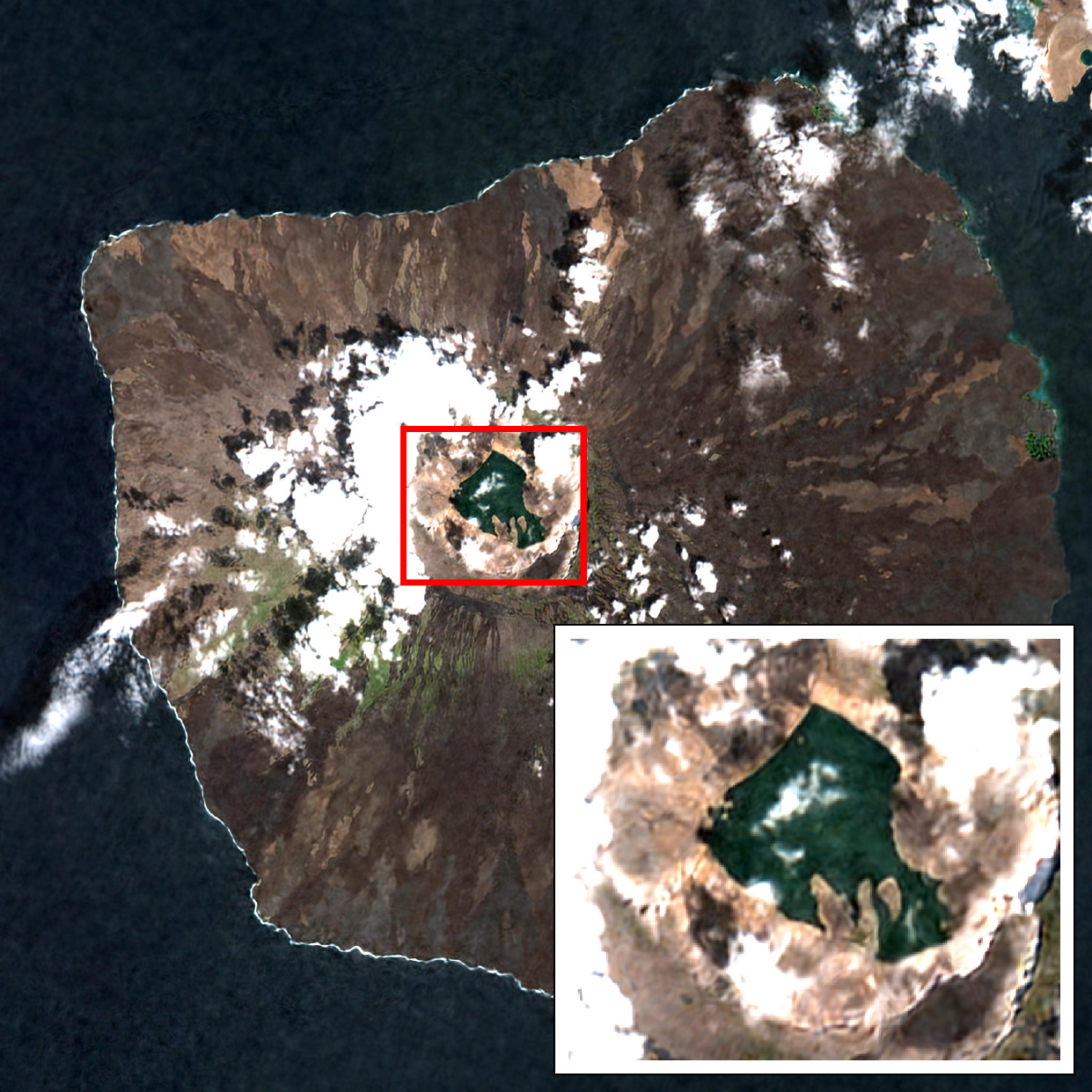} &
        \includegraphics[width=0.155\linewidth]{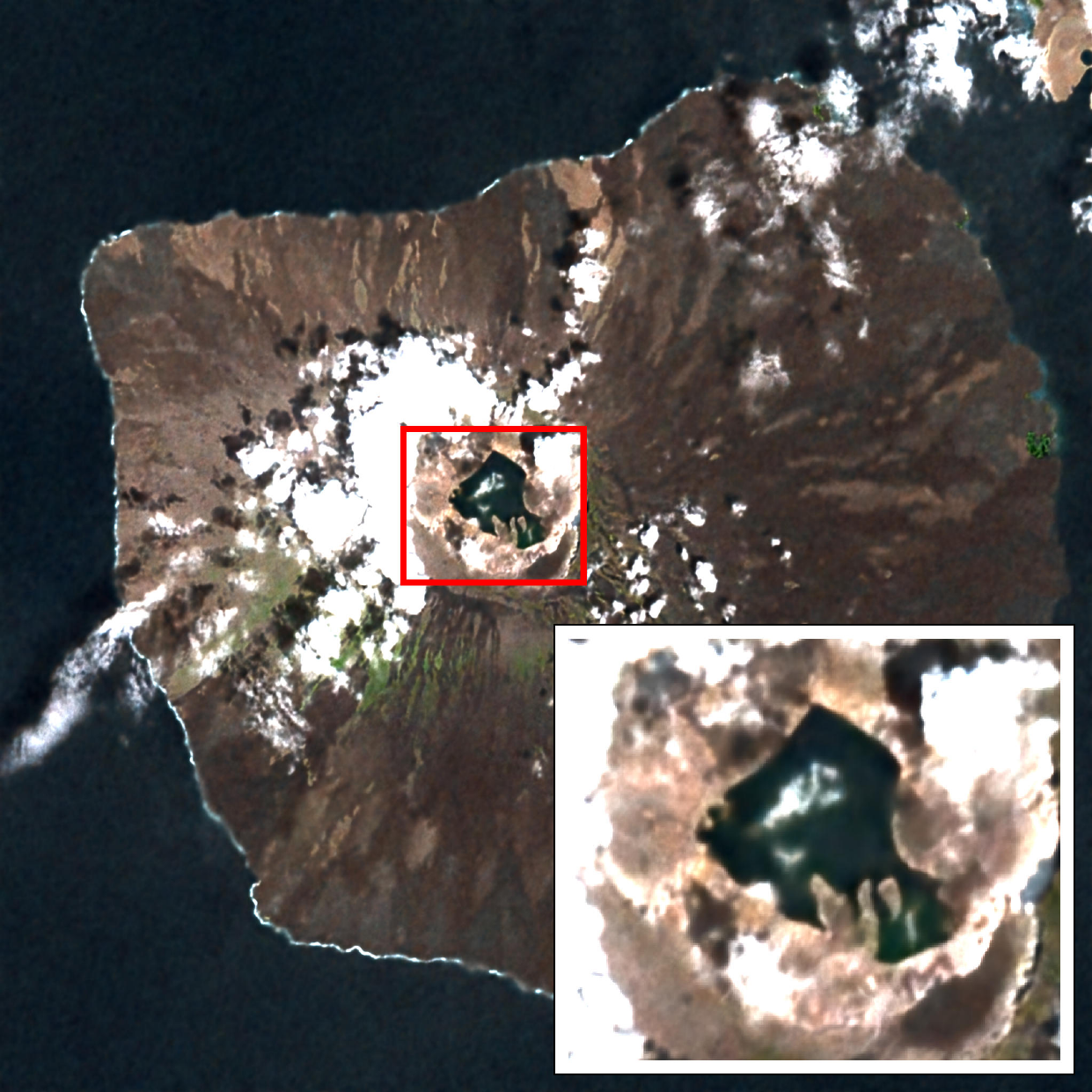} &
        \includegraphics[width=0.155\linewidth]{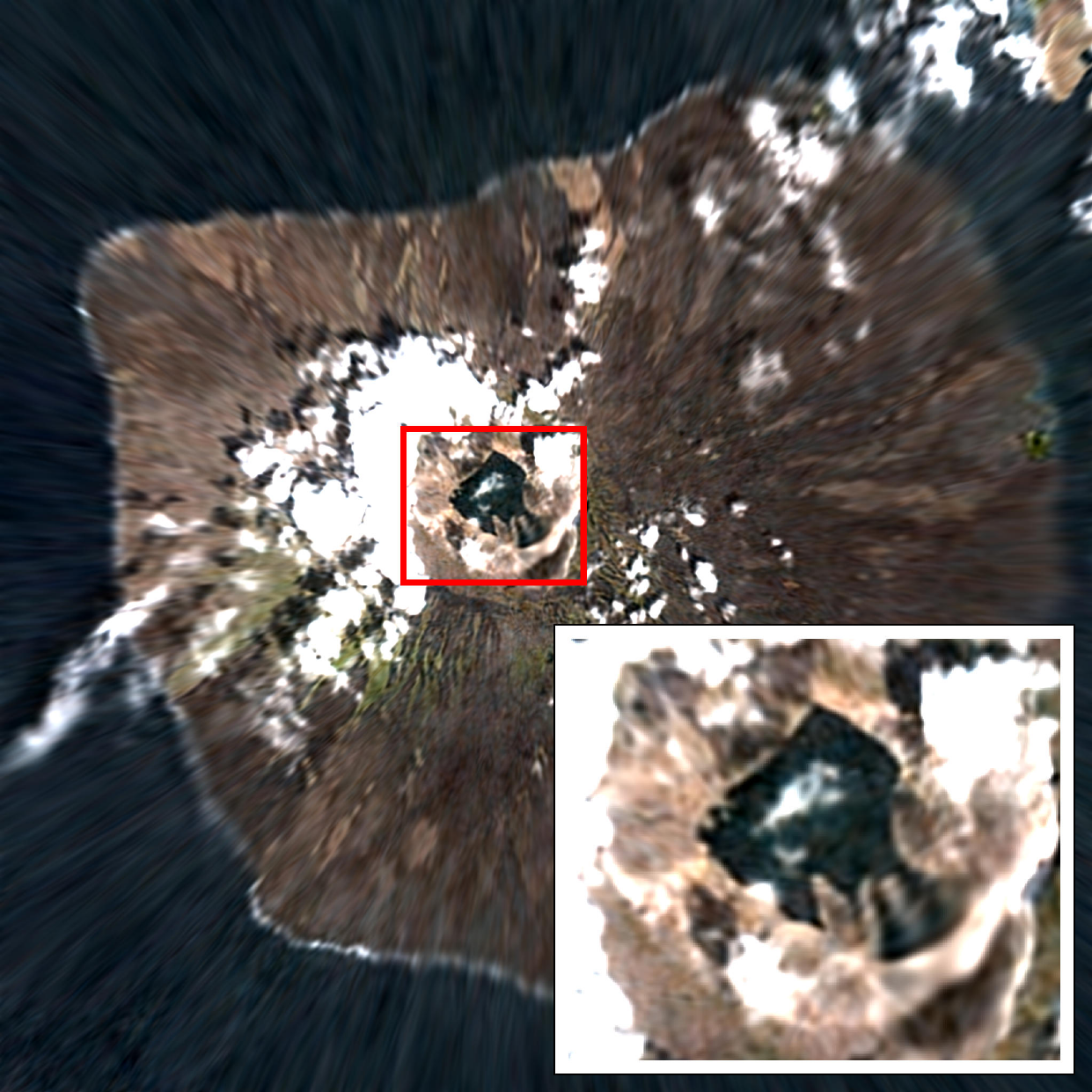} &
        \includegraphics[width=0.155\linewidth]{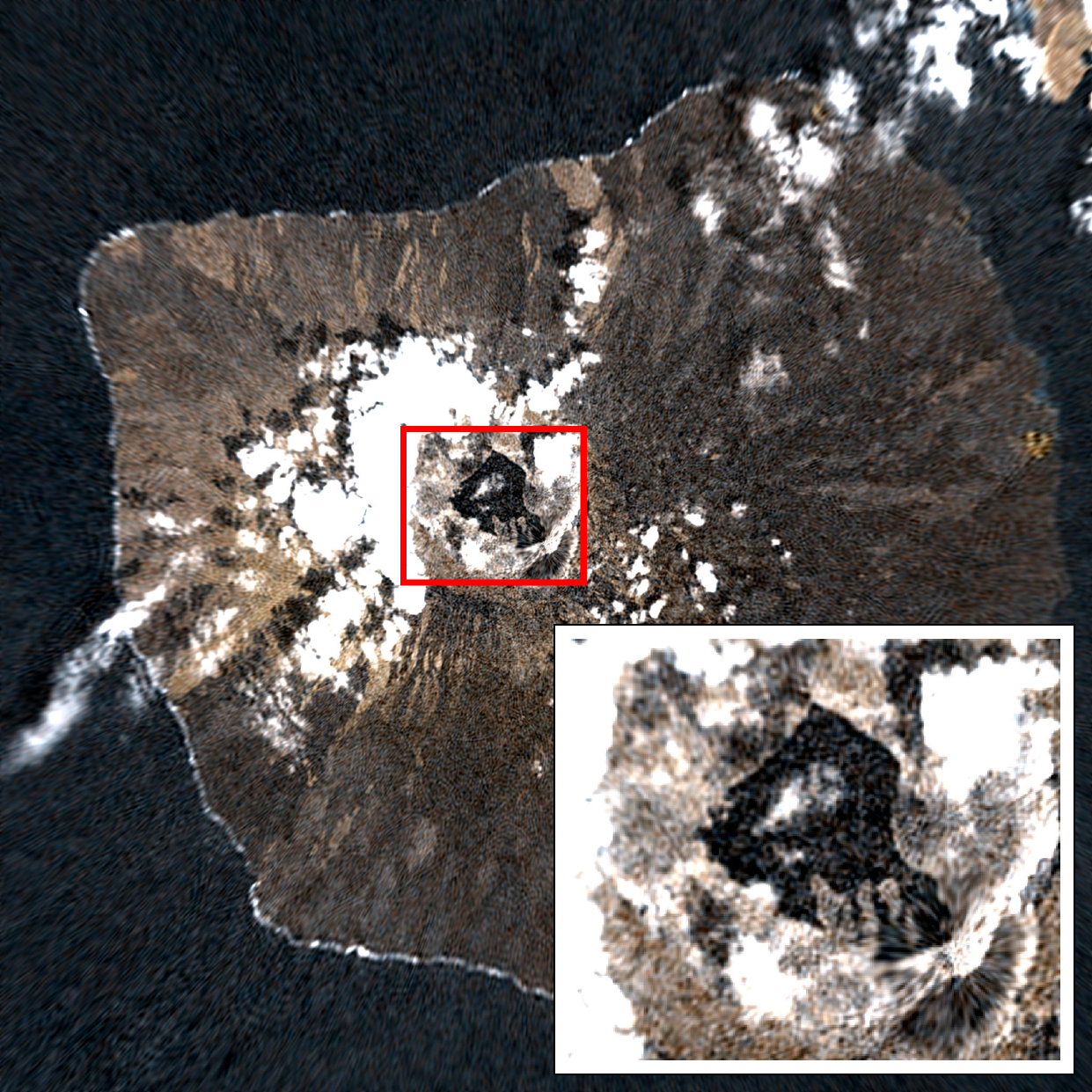} &
        \includegraphics[width=0.155\linewidth]{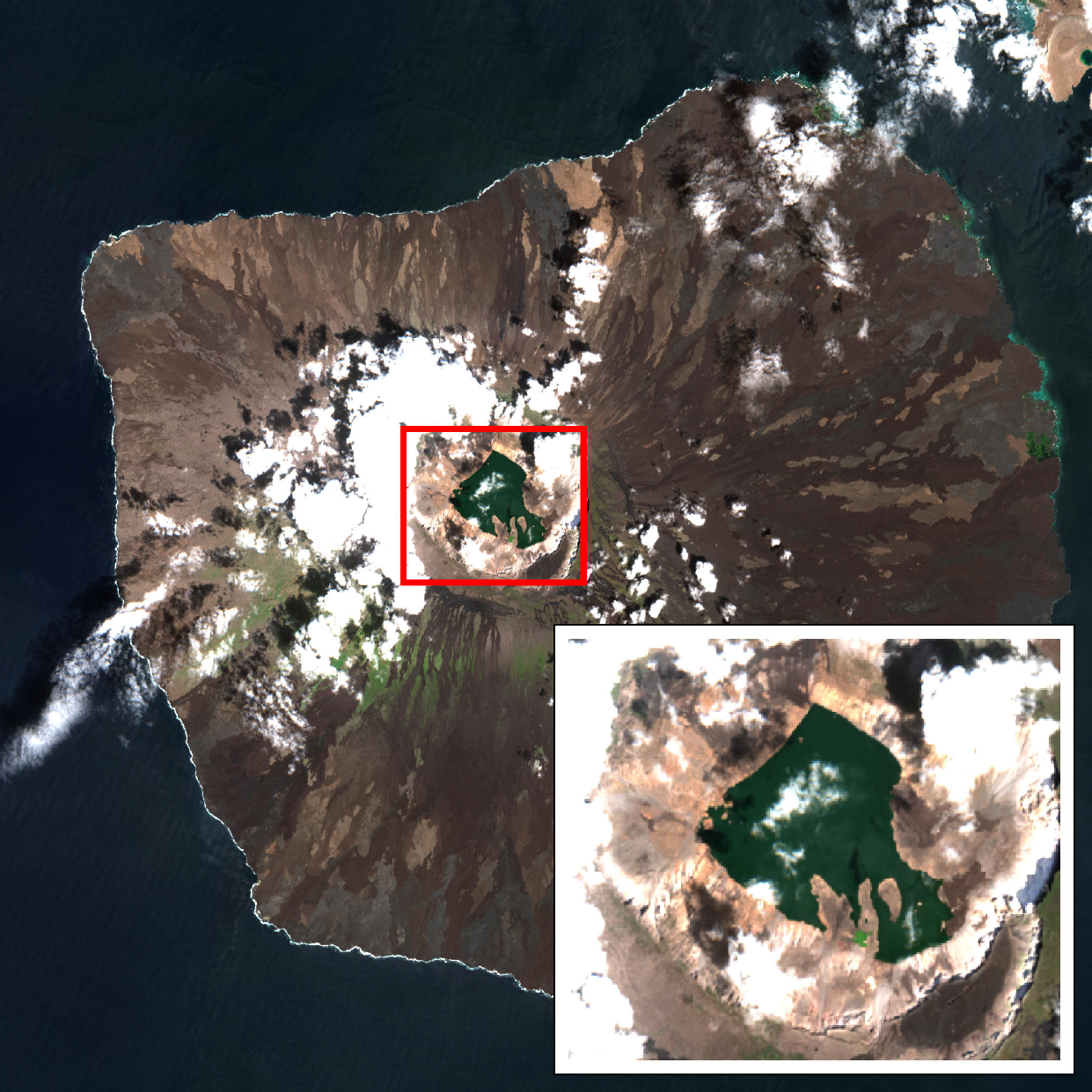} &
        \includegraphics[width=0.155\linewidth]{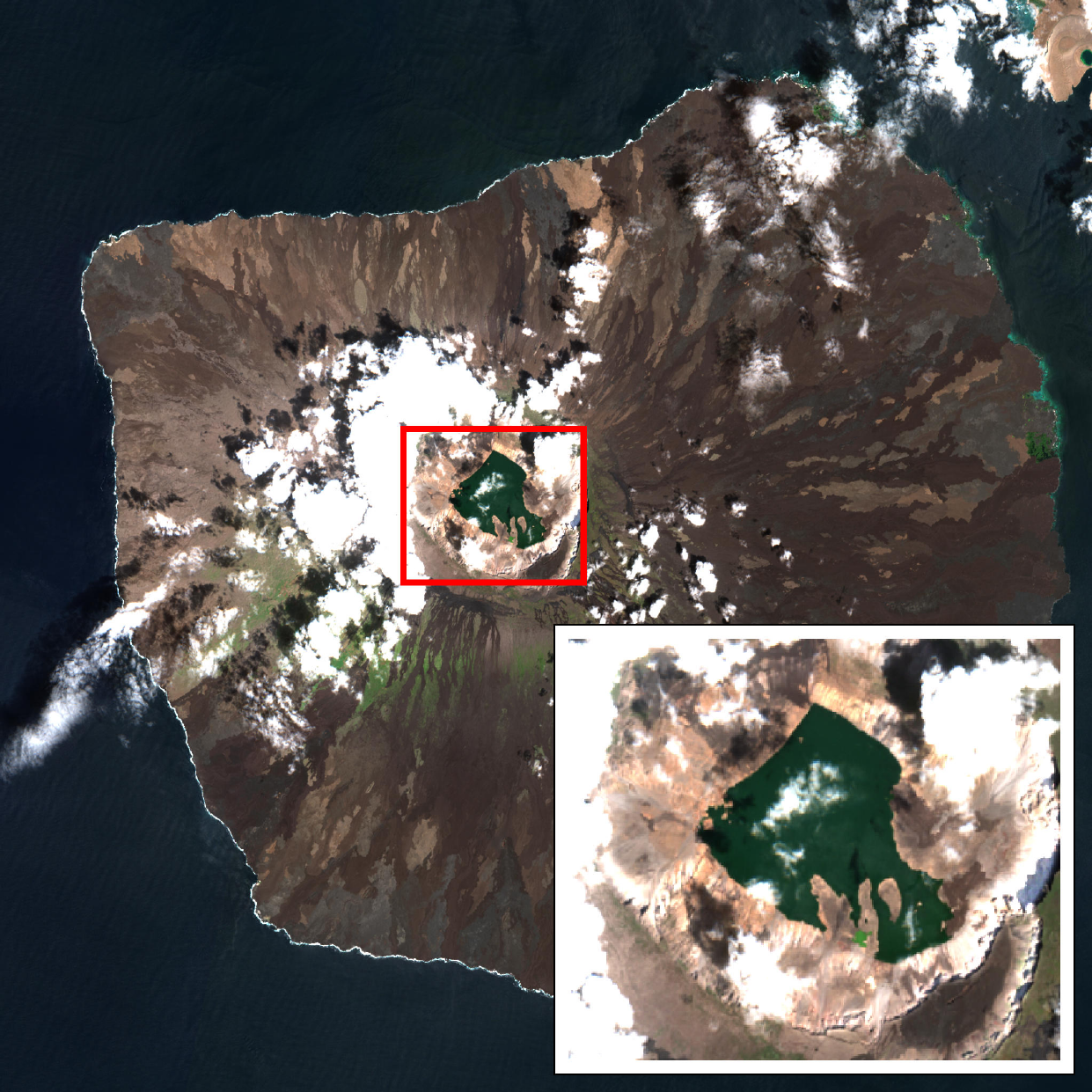}
    \end{tabular}
    \caption{
    Qualitative comparison on a high-frequency region from a Sentinel-2 L2A multispectral image~\cite{drusch2012sentinel}.
    The image consists of 12 spectral bands with a spatial resolution of $2048 \times 2048$ pixels, while the visualization shows only the RGB channels for clarity.
    ELM-INR achieves substantially higher reconstruction fidelity, recovering fine-grained structures that are heavily blurred or distorted by baseline INR models.
    }
    \label{fig:s2}
\end{figure*}

\begin{table}[h!]
\small
\centering
\caption{Multispectral Bands in Sentinel-2 Satellite Imagery}
\begin{tabular}{llcl}
\specialrule{1pt}{2pt}{2pt}
\textbf{Band} & \textbf{Description} & \textbf{Wave. Range (nm)} & \textbf{Primary Usage} \\ 
\specialrule{1pt}{2pt}{2pt}
B1   & Coastal aerosol & 433–453     & Atmospheric correction \\ \cmidrule(lr){1-4}
B2   & Blue            & 458–523     & Water body analysis    \\ \cmidrule(lr){1-4}
B3   & Green           & 543–578     & Vegetation monitoring  \\ \cmidrule(lr){1-4}
B4   & Red             & 650–680     & Vegetation and soil analysis \\ \cmidrule(lr){1-4}
B5   & Red-edge 1      & 698–713     & Vegetation monitoring  \\ \cmidrule(lr){1-4}
B6   & Red-edge 2      & 733–748     & Vegetation structure analysis \\ \cmidrule(lr){1-4}
B7   & Red-edge        & 773–793     & Vegetation chlorophyll assessment \\ \cmidrule(lr){1-4}
B8   & NIR             & 785–900     & Biomass and vegetation vigor monitoring \\ \cmidrule(lr){1-4}
B8A  & Narrow-NIR      & 855–875     & Vegetation monitoring  \\ \cmidrule(lr){1-4}
B9   & Water vapour    & 935–955     & cloud screening and water vapour analysis \\ \cmidrule(lr){1-4}
B11  & SWIR            & 1565–1655   & Cloud and snow monitoring \\ \cmidrule(lr){1-4}
B12  & SWIR            & 2100–2280   & Soil and water content analysis \\ 
\specialrule{1pt}{2pt}{2pt}

\end{tabular}\label{tbl:band information}
\end{table}

\paragraph{Sentinel-2.}
Figure~\ref{fig:s2} shows a Sentinel-2 Level-2A multispectral image capturing the Galápagos Islands.
The sample contains $C=12$ spectral bands and is processed at a spatial resolution of $2048 \times 2048$ pixels.
As illustrated in Figure~\ref{fig:s2}, sharp island–ocean boundaries and heterogeneous land-cover patterns introduce localized high-frequency structures, making this scene well-suited for evaluating reconstruction fidelity in a multi-band setting.
Following the WorldView-3 experiment, only the RGB channels are visualized for clarity, while the INR is trained to predict the full spectral vector at each spatial coordinate.

\subsection{Experimental Results on MRI data}
\begin{figure}[ht]
    \centering
    \setlength{\tabcolsep}{2pt}
    \renewcommand{\arraystretch}{1.0}
    \begin{tabular}{cccccc}
        {\footnotesize SIREN}{\scriptsize (33.7 dB)} &
        {\footnotesize FFN}{\scriptsize (33.4 dB)} &
        {\footnotesize GaussNet}{\scriptsize (29.2 dB)} &
        {\footnotesize WIRE}{\scriptsize (34.0 dB)} &
        {\footnotesize \textbf{ELM-INR}}{\scriptsize \textbf{(79.8 dB)}} &
        {\footnotesize GT} \\
        \includegraphics[width=0.155\linewidth]{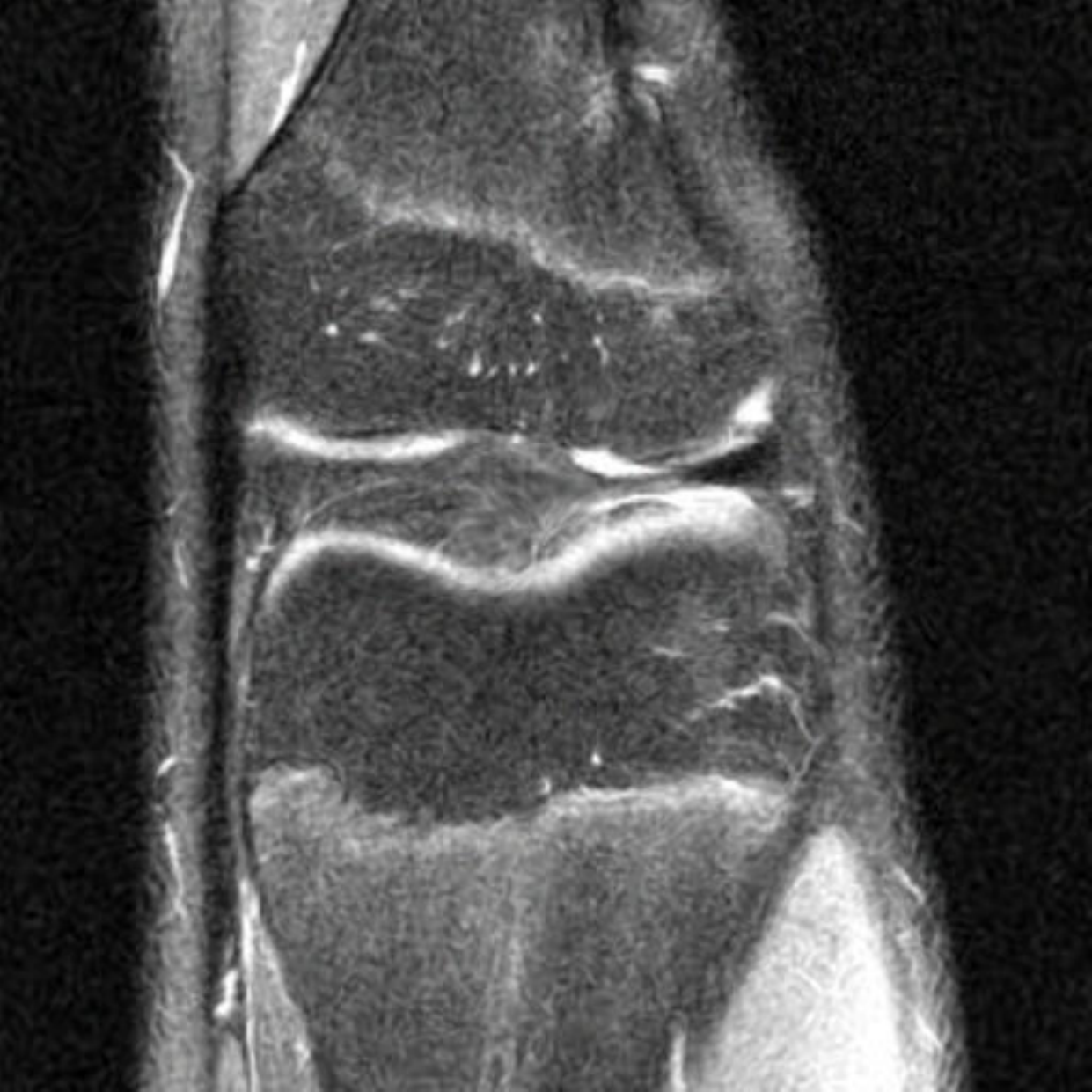} &
        \includegraphics[width=0.155\linewidth]{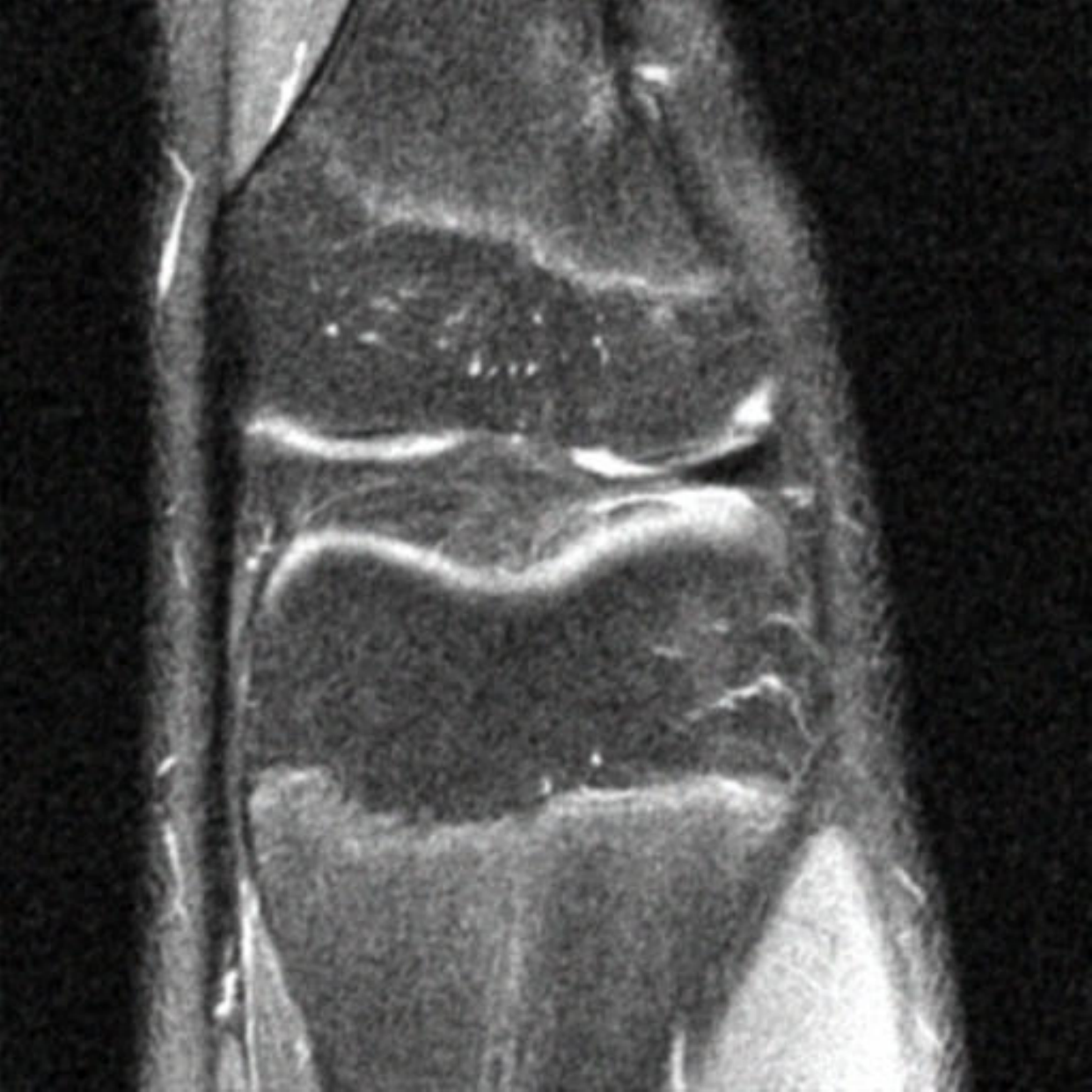} &
        \includegraphics[width=0.155\linewidth]{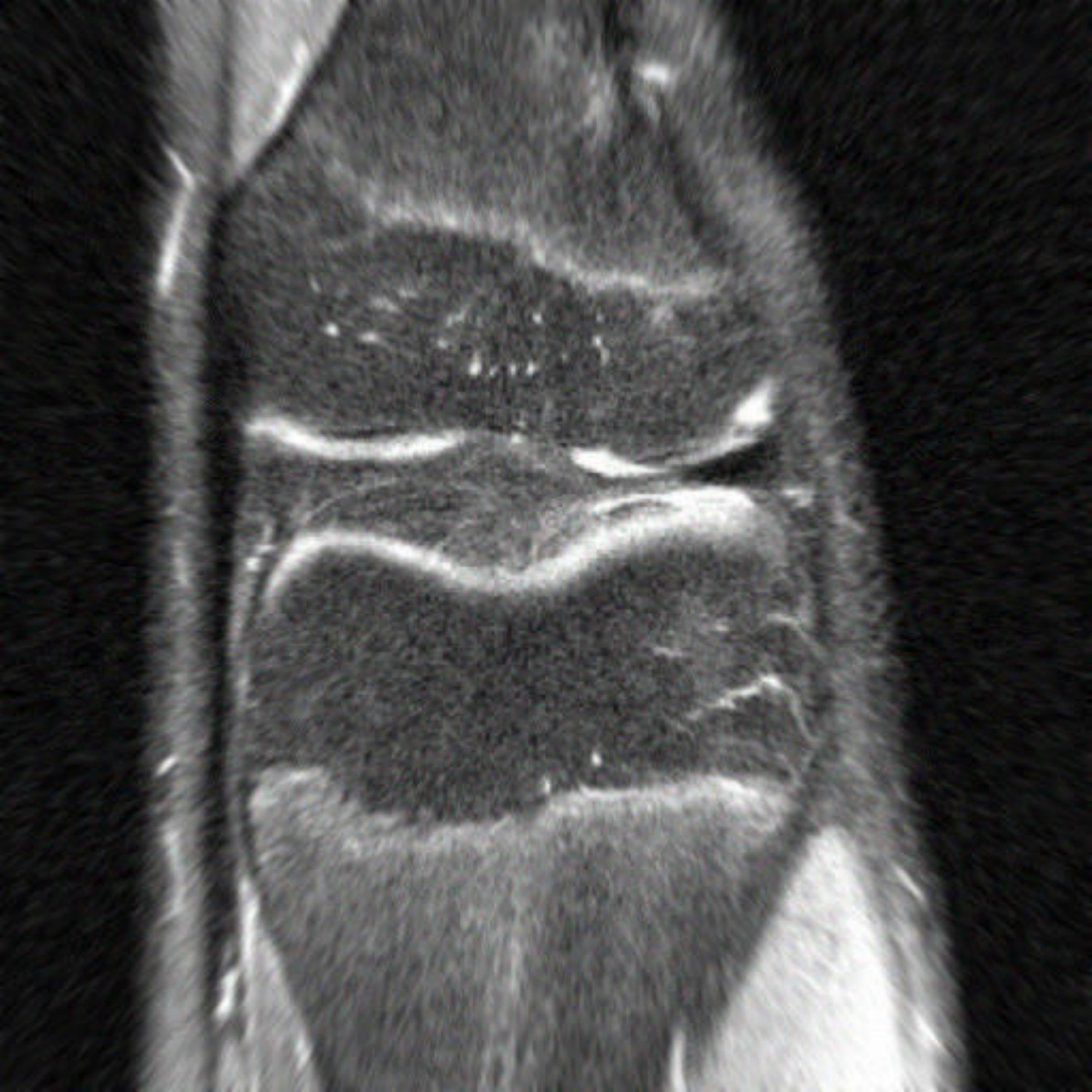} &
        \includegraphics[width=0.155\linewidth]{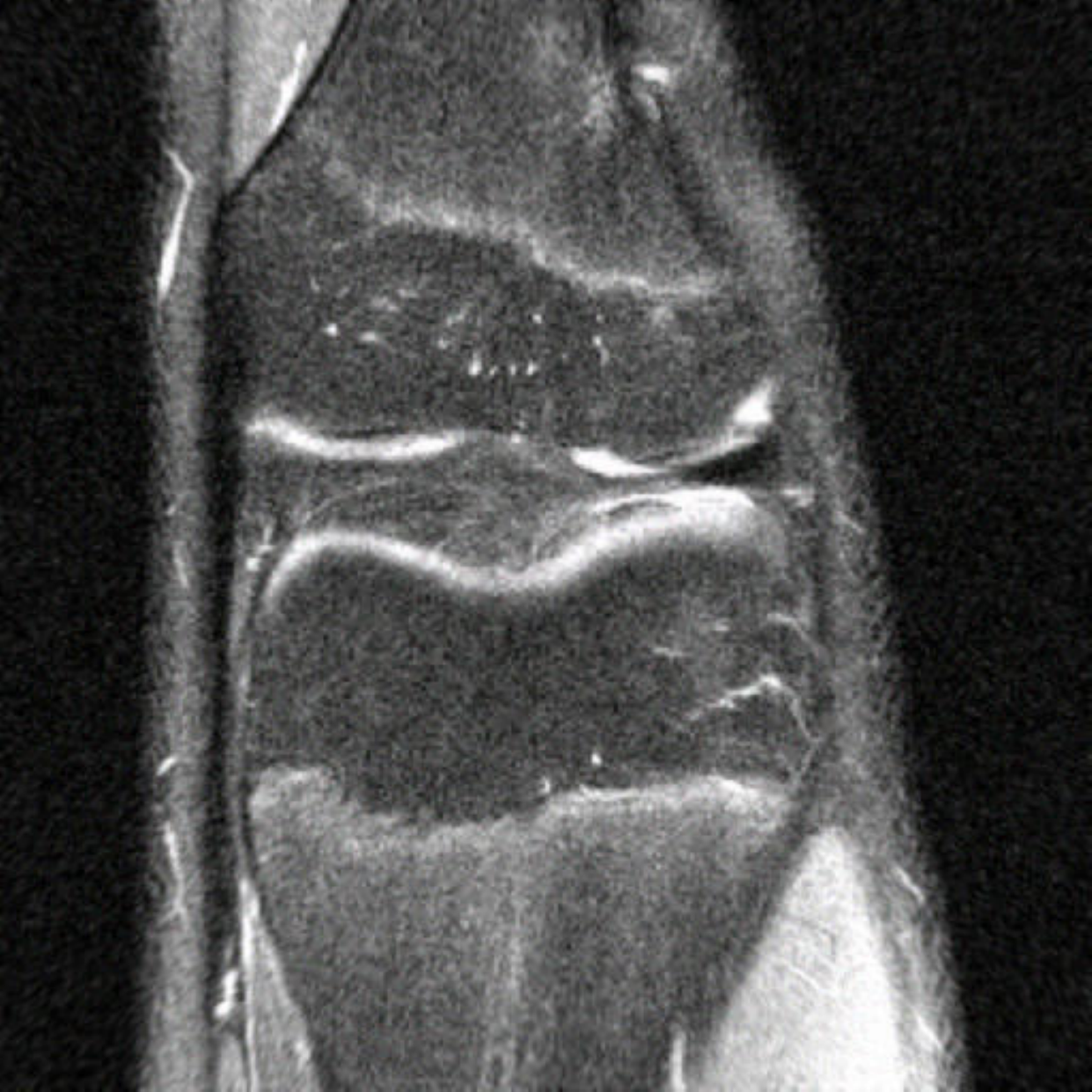} &
        \includegraphics[width=0.155\linewidth]{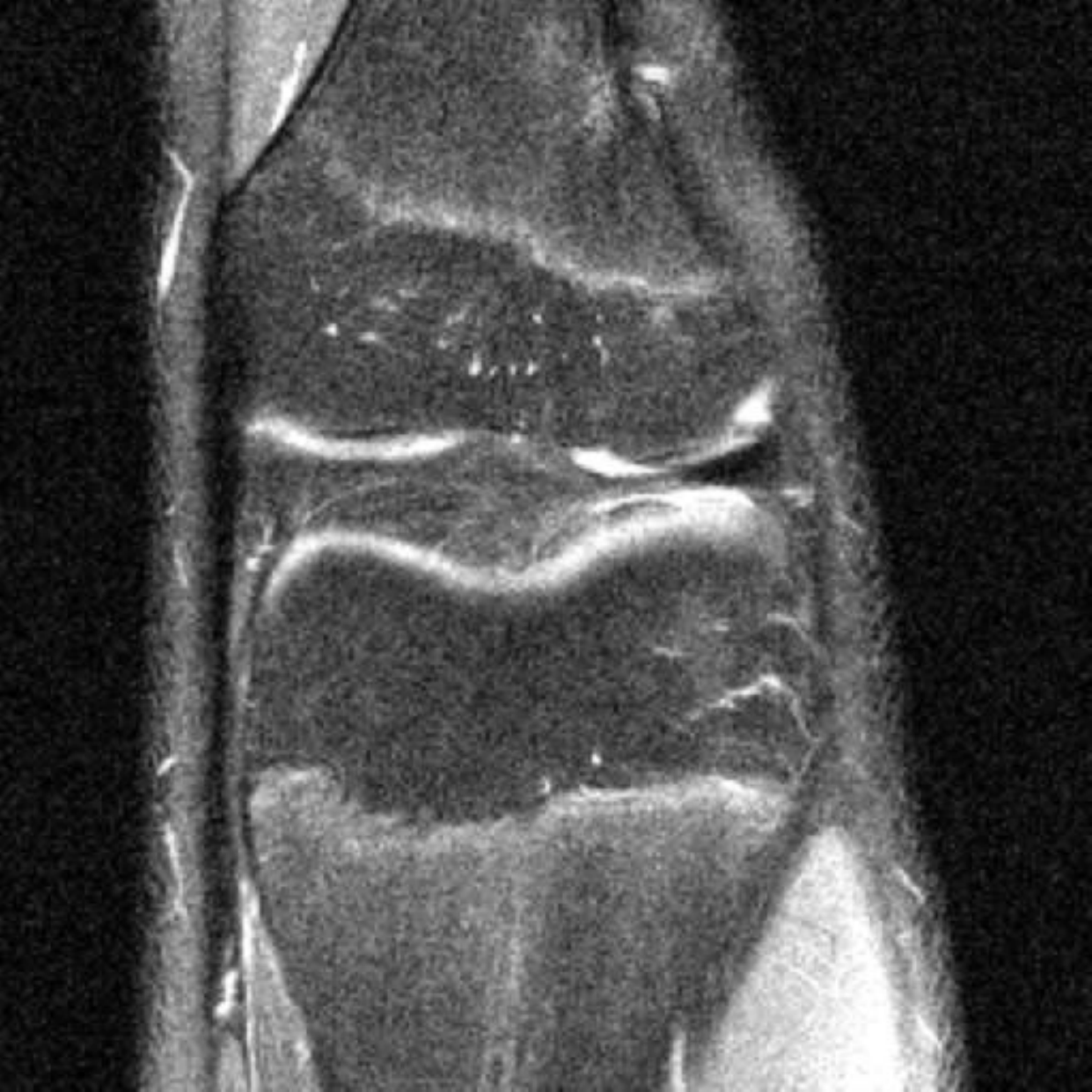} &
        \includegraphics[width=0.155\linewidth]{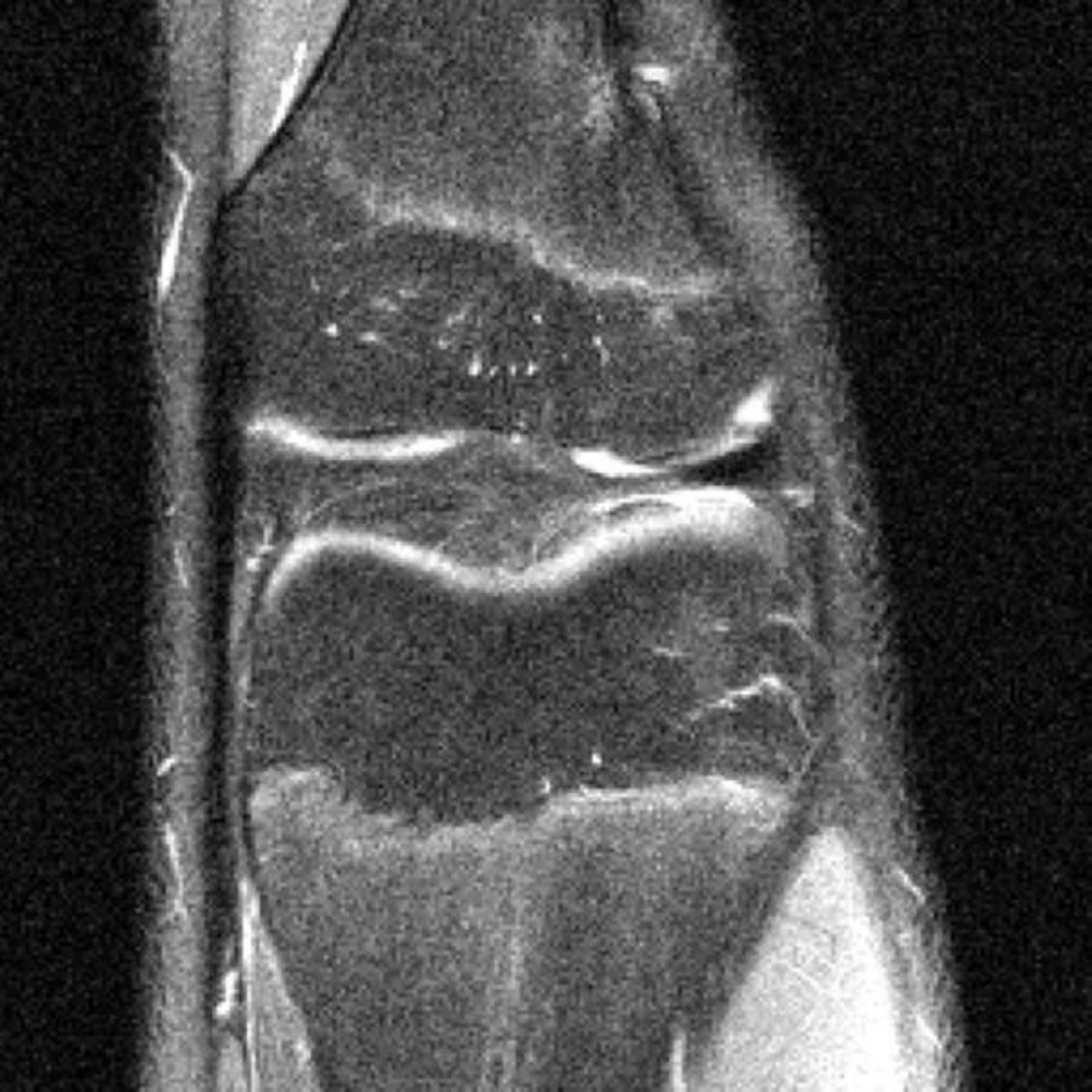}
    \end{tabular}
    \caption{
    Qualitative comparison on an MRI image with a spatial resolution of 512 $\times$ 512.}
    \label{fig:mri_appendix_qualitative}
\end{figure}

We further evaluate ELM-INR on MRI data using the dataset setup from InverseBench~\cite{zheng2025inversebench}, treating the $512\times512$ magnitude image as the target signal for coordinate-based regression. ELM-INR uses square subdomains with $S_i = 32^2$ samples and $m = 1024$ hidden units, mirroring our standard-image configuration. As shown in Figure~\ref{fig:mri_appendix_qualitative}, backpropagation-based baselines remain in the 29--34\,dB range and exhibit residual blurring along fine anatomical structures, whereas ELM-INR attains 79.8\,dB and faithfully recovers sharp tissue boundaries. This indicates that the closed-form, partition-of-unity formulation transfers naturally to medical imaging signals, where preserving high-frequency anatomical detail is essential for downstream interpretation.
\clearpage
\section{Additional Experiments: Audio Signal Reconstruction}
\label{app:audio}

We additionally evaluate ELM-INR on the audio signal-fitting task
introduced in SIREN~\cite{sitzmann2020implicit}, where the goal is to
learn an implicit representation of an audio waveform as a continuous
function of time. Following the original protocol, we use the two
audio clips released with the SIREN code, namely a music excerpt
(Bach) and a spoken-voice recording (Counting). Each audio sample
provides a pair consisting of a time index, linearly normalized to the unit interval, and the corresponding waveform amplitude, so that the model parameterizes a continuous mapping from a single temporal
coordinate to a scalar amplitude. All baselines use the per-method
architecture and learning rate found via the hyperparameter sweep
described in Appendix~\ref{app:detailed_experimental_setups}, while ELM-INR partitions the one-dimensional time domain into overlapping
subdomains and fits each local ELM in closed form.

\begin{table}[ht]
\centering
\footnotesize
\setlength{\tabcolsep}{4pt}
\caption{Audio signal reconstruction on the Bach and Counting clips
from SIREN~\cite{sitzmann2020implicit}. We report per-sample MSE
($\downarrow$), wall-clock training time ($\downarrow$), and parameter
count for each method. Best results in each column are shown in
\textbf{bold}.}
\label{tab:app_audio}
\begin{tabular}{lcccccc}
\specialrule{1pt}{2pt}{2pt}
 & \multicolumn{3}{c}{Bach} & \multicolumn{3}{c}{Counting} \\
\cmidrule(lr){2-4}\cmidrule(lr){5-7}
Method   & MSE & Time (s) & Params & MSE & Time (s) & Params \\
\midrule
SIREN    & $6.93\!\times\!10^{-3}$ & 373.7 & 0.8M
         & $2.00\!\times\!10^{-3}$ & 644.8 & 0.8M \\
FFN      & $6.91\!\times\!10^{-3}$ & 232.9 & 1.1M
         & $2.00\!\times\!10^{-3}$ & 534.4 & 1.1M \\
GaussNet & $6.93\!\times\!10^{-3}$ & 534.4 & 1.1M
         & $2.00\!\times\!10^{-3}$ & 896.1 & 1.1M \\
WIRE     & $6.67\!\times\!10^{-3}$ & 671.8 & 1.1M
         & $2.04\!\times\!10^{-3}$ & 653.1 & 1.1M \\
ELM-INR  & $\mathbf{1.90\!\times\!10^{-5}}$ & \textbf{0.015}  & \textbf{0.1M}
         & $\mathbf{9.86\!\times\!10^{-5}}$ & \textbf{0.020} & \textbf{0.1M} \\
\specialrule{1pt}{2pt}{2pt}
\end{tabular}
\end{table}

\paragraph{Results.}
Table~\ref{tab:app_audio} reports per-sample MSE, wall-clock training
time, and parameter count for both clips. Across both signals,
ELM-INR reduces the reconstruction error by roughly two orders of
magnitude relative to every backpropagation-based baseline, attaining
an MSE on the order of $10^{-5}$ on both Bach and Counting while the
strongest baseline remains in the $10^{-3}$ range. In this setting ELM-INR also uses fewer parameters than the baselines, and the gap in wall-clock training time is wider still: ELM-INR fits each clip within roughly 0.02 seconds, whereas the iterative baselines require several minutes to nearly fifteen minutes, corresponding to a $25000$--$75000\times$ speedup. These trends mirror the
findings on images and Navier--Stokes: closed-form local ELM solves
remain effective on dense one-dimensional temporal signals dominated
by high-frequency content, while entirely avoiding the iterative cost
of backpropagation-based INR training.

\section{Additional Experiments: Comparison with Structured and Plane-Based Representations}
\label{app:plane_based}

In addition to coordinate-based INRs that parameterize the entire
signal with a single MLP, a separate line of work decomposes the
signal across spatial sub-blocks, multiple scales, or factorized
feature grids. We compare ELM-INR against five representative methods
from this family: two structured INR variants
(Local-Global INR~\cite{ashkenazi2024towards}, MINER~\cite{saragadam2022miner}) and three
plane-based factorized representations
(TensoRF~\cite{chen2022tensorf}, K-Planes~\cite{fridovich2023k}, and
GA-Planes~\cite{sivgin2024geometric}).

\paragraph{Method descriptions.}
Local-Global INR~\cite{ashkenazi2024towards} decomposes the signal
into a shared global component and per-region local components,
enabling region-wise editing and cropping while sharing low-frequency
content across regions. MINER~\cite{saragadam2022miner} builds a
Laplacian-pyramid-style multiscale decomposition in which a small INR
is fit to the residual at each resolution level, allowing each
sub-network to focus on a narrow band of frequencies and to converge
quickly. TensoRF~\cite{chen2022tensorf} represents the signal through
a low-rank tensor decomposition (vector--matrix or CP components),
replacing dense MLP evaluations with explicit feature lookups. K-Planes~\cite{fridovich2023k} factorizes a $d$-dimensional
volumetric signal into pairwise 2D plane features that are combined
multiplicatively, yielding a compact explicit representation that
scales gracefully to higher dimensions. GA-Planes~\cite{sivgin2024geometric}
is a recent generalization of plane-based factorization that broadens
the K-Planes-style construction to a wider class of signals while
preserving the memory efficiency of factorized features.

\begin{figure}[ht]
\centering
\includegraphics[width=0.75\linewidth]{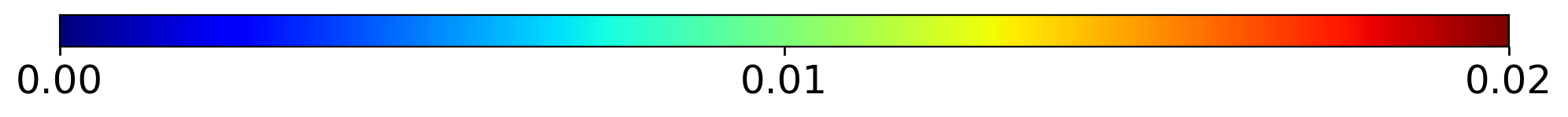}\\[4pt]

\begin{subfigure}{0.155\linewidth}
    \centering
    \includegraphics[width=\linewidth]{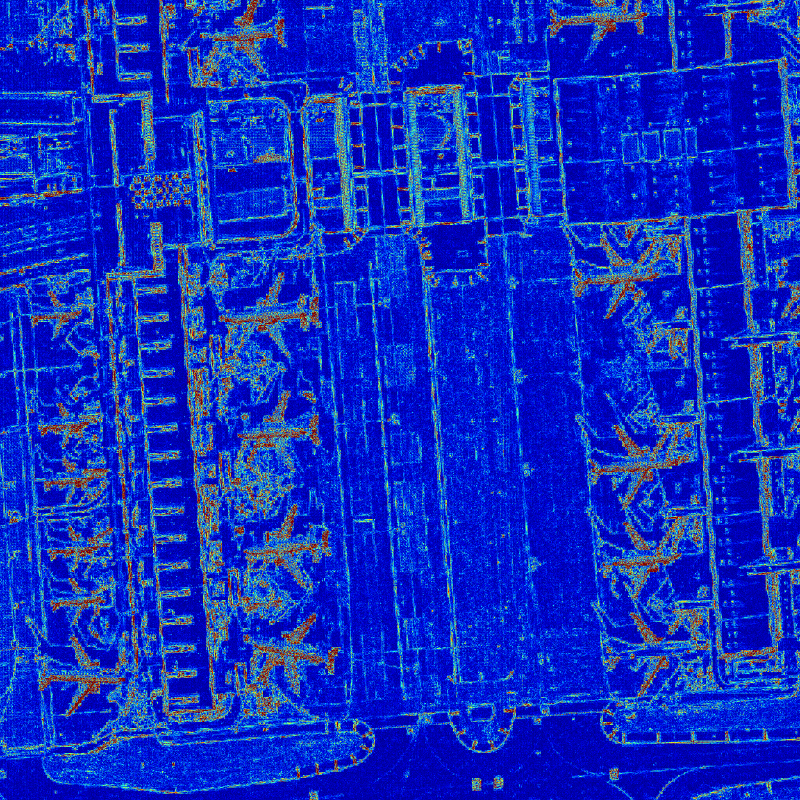}
    \caption*{Local-Global}
\end{subfigure}
\hfill
\begin{subfigure}{0.155\linewidth}
    \centering
    \includegraphics[width=\linewidth]{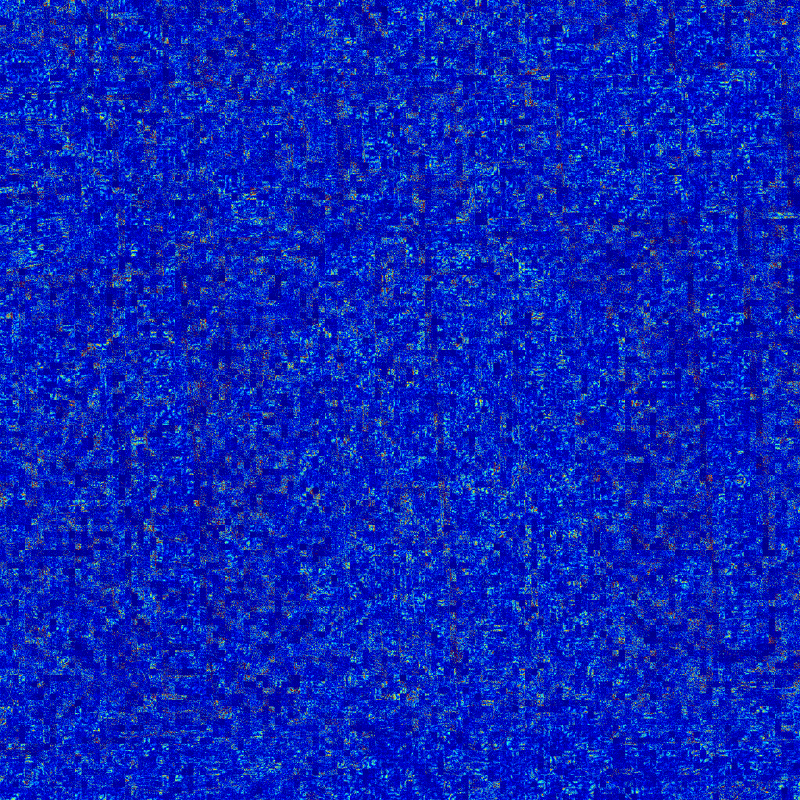}
    \caption*{MINER}
\end{subfigure}
\hfill
\begin{subfigure}{0.155\linewidth}
    \centering
    \includegraphics[width=\linewidth]{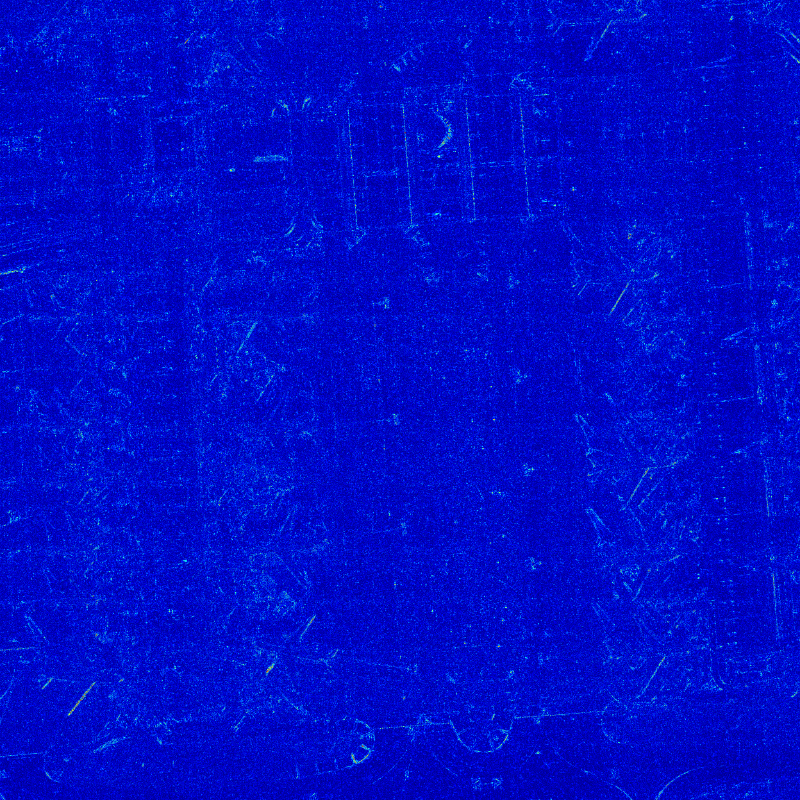}
    \caption*{TensoRF}
\end{subfigure}
\hfill
\begin{subfigure}{0.155\linewidth}
    \centering
    \includegraphics[width=\linewidth]{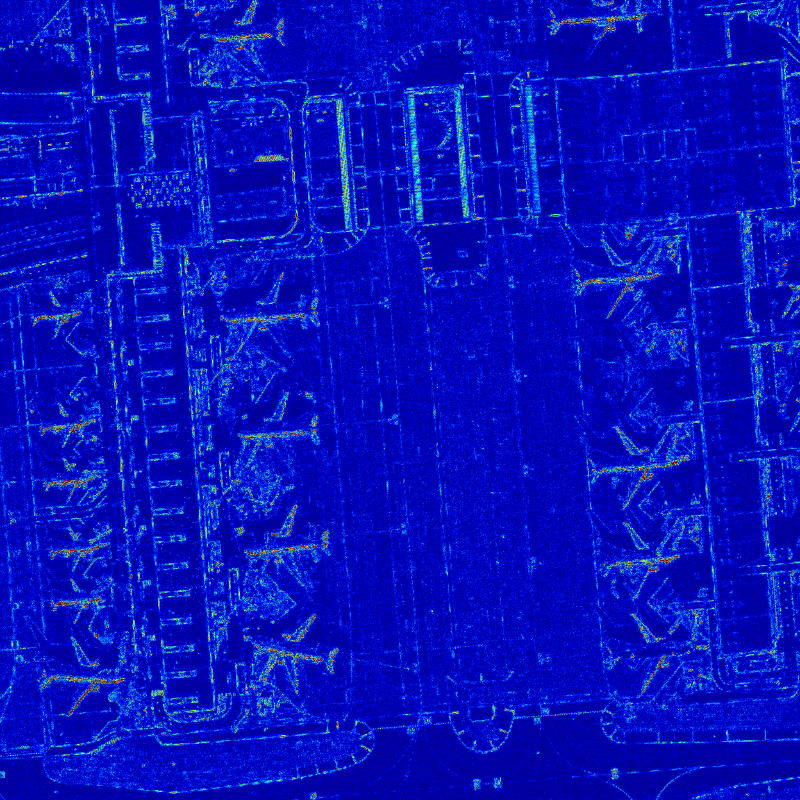}
    \caption*{K-Planes}
\end{subfigure}
\hfill
\begin{subfigure}{0.155\linewidth}
    \centering
    \includegraphics[width=\linewidth]{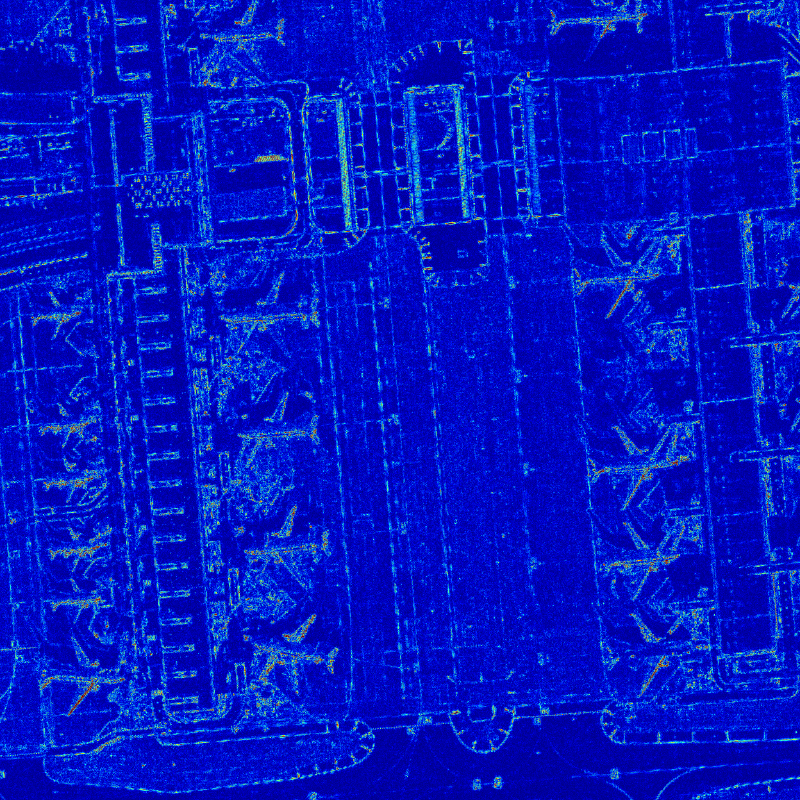}
    \caption*{GA-Planes}
\end{subfigure}
\hfill
\begin{subfigure}{0.155\linewidth}
    \centering
    \includegraphics[width=\linewidth]{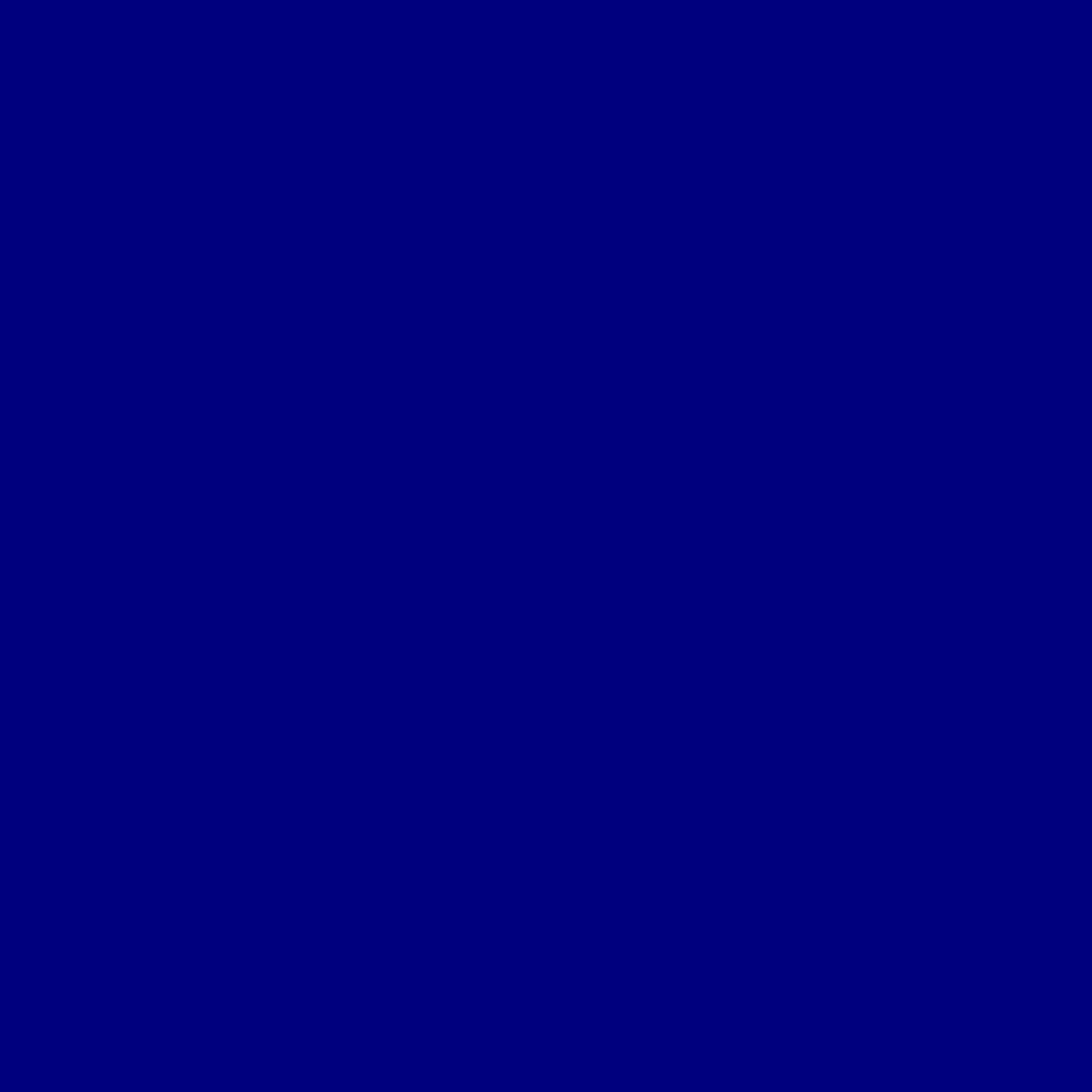}
    \caption*{ELM-INR}
\end{subfigure}\\[2pt]
{\small (a) WorldView-3 multispectral image (CDG)}\\[6pt]

\begin{subfigure}{0.155\linewidth}
    \centering
    \includegraphics[width=\linewidth]{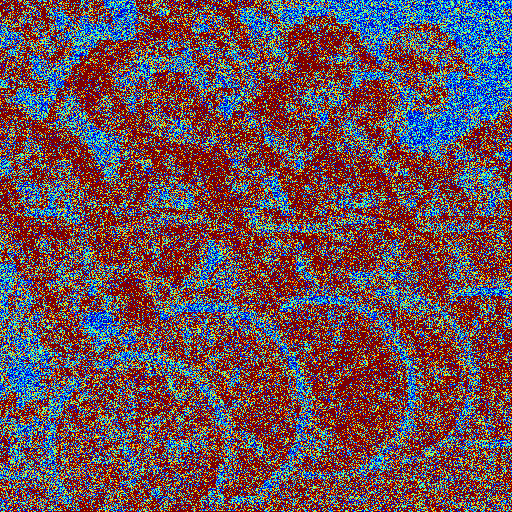}
    \caption*{Local-Global}
\end{subfigure}
\hfill
\begin{subfigure}{0.155\linewidth}
    \centering
    \includegraphics[width=\linewidth]{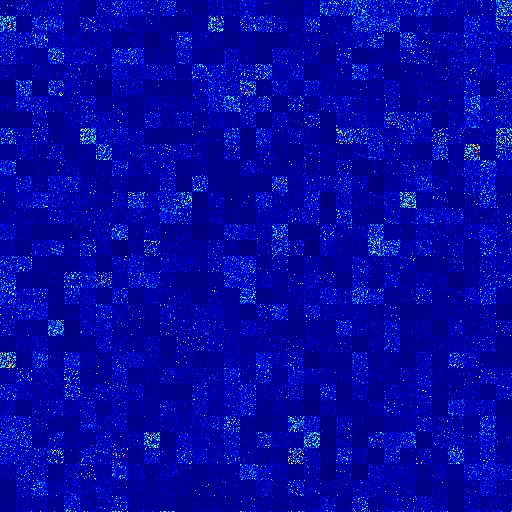}
    \caption*{MINER}
\end{subfigure}
\hfill
\begin{subfigure}{0.155\linewidth}
    \centering
    \includegraphics[width=\linewidth]{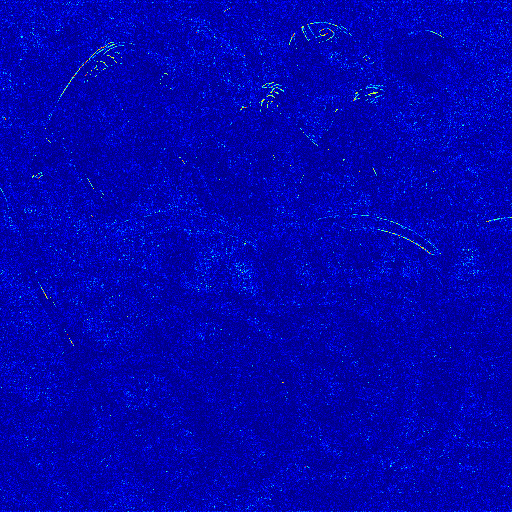}
    \caption*{TensoRF}
\end{subfigure}
\hfill
\begin{subfigure}{0.155\linewidth}
    \centering
    \includegraphics[width=\linewidth]{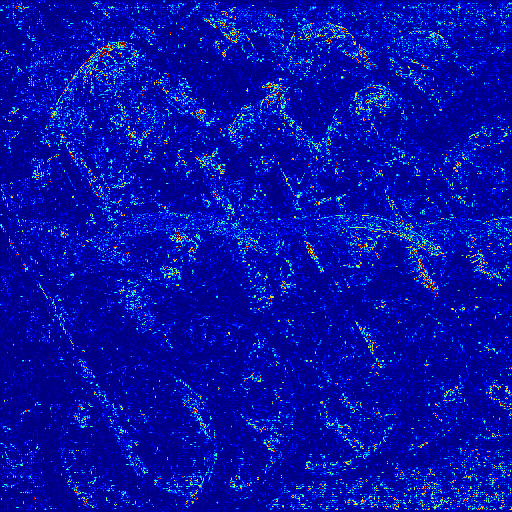}
    \caption*{K-Planes}
\end{subfigure}
\hfill
\begin{subfigure}{0.155\linewidth}
    \centering
    \includegraphics[width=\linewidth]{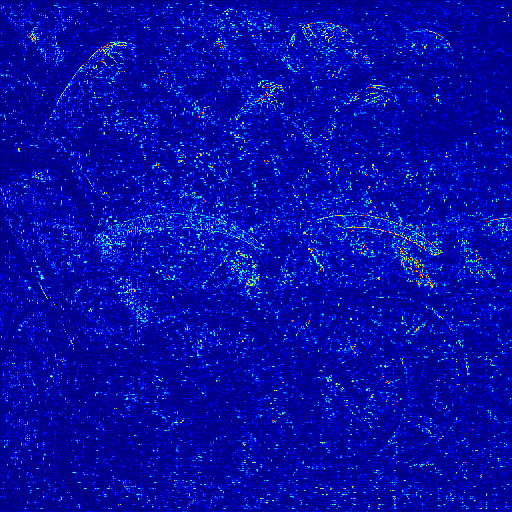}
    \caption*{GA-Planes}
\end{subfigure}
\hfill
\begin{subfigure}{0.16\linewidth}
    \centering
    \includegraphics[width=\linewidth]{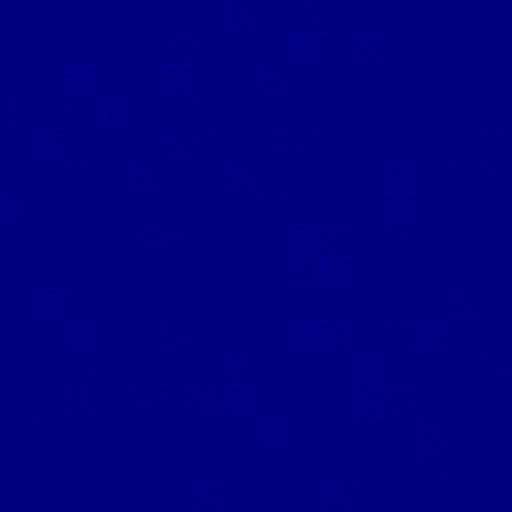}
    \caption*{ELM-INR}
\end{subfigure}\\[2pt]
{\small (b) kodim05}

\caption{Per-pixel absolute error maps on (a) the WorldView-3
multispectral image (CDG; errors averaged across all spectral
bands) and (b) the Kodak kodim05 image, visualized with a shared colorbar (top). Across both datasets, ELM-INR yields substantially lower reconstruction error than baseline methods, particularly around
high-frequency structures such as edges and fine textures.}
\label{fig:error_maps_combined}
\end{figure}

\paragraph{Results.}
Table~\ref{tab:app_planes} reports PSNR and SSIM (mean\,$\pm$\,std over
5 random seeds) on the kodim05 image and on a WorldView-3 multispectral
image. Plane-based methods clearly improve over the coordinate-based
baselines reported in Table~\ref{tab:app_psnr}, reaching the
mid-to-high $40\,\text{dB}$ range on both signals thanks to their
explicit factorized features, and MINER similarly benefits from its
multiscale decomposition. Despite this, ELM-INR consistently attains
the highest PSNR and SSIM on both signals, reaching above
$79\,\text{dB}$ in PSNR and saturating SSIM at $1.0000$, while
remaining backpropagation-free.
The corresponding ground-truth images and qualitative reconstructions
against coordinate-based baselines are shown in
Figures~\ref{fig:kodim05_final_qualitative} and~\ref{fig:wv3_fr}.
Extending this comparison to the plane-based and multiscale baselines
considered here, Figure~\ref{fig:error_maps_combined} visualizes
per-pixel error maps under a shared color scale, where ELM-INR exhibits
markedly lower residuals than all baselines, particularly around
high-frequency structures such as edges and fine textures.

\begin{table}[ht]
\centering
\small
\setlength{\tabcolsep}{4pt}
\caption{Reconstruction quality on the \texttt{kodim05} image and a WorldView-3, CDG multispectral image (PSNR averaged over all spectral bands for the
multispectral case). Results are reported as mean\,$\pm$\,std over 5
random seeds. Best results in each row are shown in \textbf{bold}.}
\label{tab:app_planes}
\resizebox{\columnwidth}{!}{%
\begin{tabular}{llcccccc}
\specialrule{1pt}{2pt}{2pt}
Dataset & Metric & Local-Global & MINER & TensoRF & K-Planes & GA-Planes & ELM-INR \\
\midrule
\multirow{2}{*}{kodim05}
 & PSNR
   & 28.34{\scriptsize\,$\pm$0.42} & 45.36{\scriptsize\,$\pm$0.18}
   & 44.44{\scriptsize\,$\pm$0.16} & 46.09{\scriptsize\,$\pm$0.01}
   & 46.36{\scriptsize\,$\pm$0.81}
   & \textbf{79.03}{\scriptsize\,$\pm$1.59} \\
 & SSIM
   & 0.9810{\scriptsize\,$\pm$0.0003} & 0.9996{\scriptsize\,$\pm$0.0000}
   & 0.9996{\scriptsize\,$\pm$0.0000} & 0.9997{\scriptsize\,$\pm$0.0000}
   & 0.9997{\scriptsize\,$\pm$0.0001}
   & \textbf{1.0000}{\scriptsize\,$\pm$0.0000} \\
\midrule
\multirow{2}{*}{WorldView-3}
 & PSNR
   & 41.34{\scriptsize\,$\pm$0.27} & 43.84{\scriptsize\,$\pm$0.15}
   & 48.38{\scriptsize\,$\pm$0.06} & 43.03{\scriptsize\,$\pm$0.52}
   & 48.79{\scriptsize\,$\pm$0.36}
   & \textbf{79.19}{\scriptsize\,$\pm$0.42} \\
 & SSIM
   & 0.9840{\scriptsize\,$\pm$0.0004} & 0.9910{\scriptsize\,$\pm$0.0002}
   & 0.9962{\scriptsize\,$\pm$0.0000} & 0.9950{\scriptsize\,$\pm$0.0006}
   & 0.9987{\scriptsize\,$\pm$0.0001}
   & \textbf{1.0000}{\scriptsize\,$\pm$0.0000} \\
\specialrule{1pt}{2pt}{2pt}
\end{tabular}}
\end{table}

\paragraph{Summary.}
Both structured INR variants (Local-Global INR, MINER) and plane-based
factorizations (TensoRF, K-Planes, GA-Planes) move beyond the
single-MLP coordinate baselines of
Appendix~\ref{app:comprehensive_comparison}, producing markedly higher
fidelity by exploiting structural priors over the signal domain.
ELM-INR nevertheless retains a substantial margin across all five
methods on both Kodak and WorldView-3 imagery, indicating that
closed-form local ELM solves combined with partition-of-unity coupling
exploit local structure at least as effectively as factorized feature
representations, while entirely avoiding iterative optimization.

\section{Additional Experiments: Comparison with a Closed-Form Linear Model}
\label{app:linear}
ELM-INR obtains its output coefficients through the closed-form least-squares solve of Eq.~\eqref{eq:ls} on local subdomains, but the local predictors themselves are the nonlinear random-feature models of Eq.~\eqref{eq:elm}. A natural question is whether the nonlinear random features are necessary at all, since a linear model also admits a closed-form solution, avoids iterative optimization, and introduces no randomness. To answer this, we implemented a linear model using the same local fitting pipeline as ELM-INR: the domain is decomposed into the same subdomains, each local predictor is fitted by the same closed-form least-squares solve, and the local predictions are blended through the same partition of unity as in Eq.~\eqref{eq:pou}. The only difference is that the random nonlinear hidden layer of Eq.~\eqref{eq:elm} is removed and each local predictor is a linear model in its inputs. We report PSNR and SSIM on four Kodak images as mean\,$\pm$\,std over five random seeds.
\begin{table}[H]
\centering
\small
\caption{Closed-form linear model vs.\ ELM-INR on Kodak images (PSNR in dB /
SSIM, mean\,$\pm$\,std over five random seeds). Both use the same subdomains,
partition-of-unity blending, and closed-form least-squares solve. Best results
in each row are shown in \textbf{bold}.}
\label{tab:linear}
\begin{tabular}{lcccc}
\specialrule{1pt}{2pt}{2pt}
& \multicolumn{2}{c}{Linear model} & \multicolumn{2}{c}{ELM-INR} \\
\cmidrule(lr){2-3}\cmidrule(lr){4-5}
Image & PSNR (dB) & SSIM & PSNR (dB) & SSIM \\
\midrule
\texttt{kodim05} & 16.81{\scriptsize\,$\pm$0.58} & 0.6667{\scriptsize\,$\pm$0.0614}
                 & \textbf{79.03}{\scriptsize\,$\pm$1.59} & \textbf{1.0000}{\scriptsize\,$\pm$0.0000} \\
\texttt{kodim08} & 15.46{\scriptsize\,$\pm$0.53} & 0.7256{\scriptsize\,$\pm$0.0424}
                 & \textbf{78.82}{\scriptsize\,$\pm$1.73} & \textbf{1.0000}{\scriptsize\,$\pm$0.0000} \\
\texttt{kodim20} & 18.89{\scriptsize\,$\pm$0.60} & 0.8165{\scriptsize\,$\pm$0.0301}
                 & \textbf{78.42}{\scriptsize\,$\pm$1.20} & \textbf{1.0000}{\scriptsize\,$\pm$0.0000} \\
\texttt{kodim24} & 19.76{\scriptsize\,$\pm$0.51} & 0.8174{\scriptsize\,$\pm$0.0258}
                 & \textbf{79.11}{\scriptsize\,$\pm$1.49} & \textbf{1.0000}{\scriptsize\,$\pm$0.0000} \\
\specialrule{1pt}{2pt}{2pt}
\end{tabular}
\end{table}
As shown in Table~\ref{tab:linear}, ELM-INR consistently and substantially
outperforms the linear model: the linear model reaches only 15--20\,dB PSNR
and 0.67--0.82 SSIM, whereas ELM-INR exceeds 78\,dB with SSIM saturated at
1.0000 under the identical decomposition and solver. Closed-form optimization
alone is therefore insufficient, and the nonlinear random features are
essential for high-fidelity reconstruction. Domain decomposition does not
replace this nonlinearity; rather, it mitigates the capacity limitation of
each shallow local model by letting it approximate a simpler local function
with a fixed finite basis, while the closed-form solve keeps the formulation
backpropagation-free. This is consistent with the ablation in
Figure~\ref{fig:ablation_hidden_nodes_subdomains}, where the reconstruction
quality of ELM-INR is governed by the number of random hidden units $m$, a
capacity knob that a linear model does not have.

\section{Additional Experiments: Choice of Activation Function}
\label{app:activation}
 
ELM-INR uses ReLU as the hidden-layer activation in all main experiments. To
examine whether the method depends on this particular nonlinearity, we
evaluated ReLU, sine, tanh, GELU, and Swish on two signals, the
\texttt{cameraman} image and \texttt{kodim05}, under the default
configuration. We additionally included the identity activation to test
whether the nonlinear mapping is necessary at all. All results are averaged
over five random initializations of the hidden parameters.
 
\begin{table}[ht!]
\centering
\footnotesize
\setlength{\tabcolsep}{3pt}
\caption{Activation-function ablation for ELM-INR on the \texttt{cameraman}
image and \texttt{kodim05}. PSNR is reported as mean\,$\pm$\,std over five
random initializations; SSIM and MSE are averaged over the same runs. Best results in each column are shown in \textbf{bold}.}
\label{tab:activation}
\begin{tabular}{lcccccc}
\specialrule{1pt}{2pt}{2pt}
& \multicolumn{3}{c}{\texttt{cameraman}} & \multicolumn{3}{c}{\texttt{kodim05}} \\
\cmidrule(lr){2-4}\cmidrule(lr){5-7}
Activation & PSNR (dB) & SSIM & MSE & PSNR (dB) & SSIM & MSE \\
\midrule
ReLU     & 79.98{\scriptsize\,$\pm$0.01} & 1.0000 & $3.5\!\times\!10^{-11}$
         & 79.88{\scriptsize\,$\pm$0.10} & 1.0000 & $2.7\!\times\!10^{-10}$ \\
Sine     & 79.97{\scriptsize\,$\pm$0.01} & 1.0000 & $6.3\!\times\!10^{-11}$
         & 79.98{\scriptsize\,$\pm$0.01} & 1.0000 & $4.1\!\times\!10^{-11}$ \\
Tanh     & \textbf{79.99}{\scriptsize\,$\pm$0.00} & 1.0000 & $\mathbf{4.7\!\times\!10^{-12}}$
         & \textbf{79.99}{\scriptsize\,$\pm$0.00} & 1.0000 & $\mathbf{1.5\!\times\!10^{-11}}$ \\
GELU     & 79.01{\scriptsize\,$\pm$1.31} & 1.0000 & $3.1\!\times\!10^{-9}$
         & 76.91{\scriptsize\,$\pm$3.90} & 1.0000 & $2.1\!\times\!10^{-8}$ \\
Swish    & 79.96{\scriptsize\,$\pm$0.04} & 1.0000 & $9.9\!\times\!10^{-11}$
         & 79.67{\scriptsize\,$\pm$0.31} & 1.0000 & $8.1\!\times\!10^{-10}$ \\
Identity & 21.48{\scriptsize\,$\pm$1.58} & 0.9306{\scriptsize\,$\pm$0.029} & $7.4\!\times\!10^{-3}$
         & 17.88{\scriptsize\,$\pm$0.92} & 0.6956{\scriptsize\,$\pm$0.090} & $1.7\!\times\!10^{-2}$ \\
\specialrule{1pt}{2pt}{2pt}
\end{tabular}
\end{table}
As shown in Table~\ref{tab:activation}, the nonlinear activations consistently achieve high reconstruction quality on both signals, indicating that ELM-INR is not strongly dependent on a specific nonlinear activation; GELU shows a slightly larger variance across initializations but remains far above the baselines of Table~\ref{tab:app_psnr}. In contrast, the identity activation degrades reconstruction quality substantially, confirming the importance of the nonlinear feature mapping and agreeing with the linear-model comparison in Appendix~\ref{app:linear}. We use ReLU in the main experiments as the simplest standard choice.

\section{Additional Experiments: Runtime Overhead of BEAM}
\label{app:beam_runtime}
Algorithm~\ref{alg:beam} is executed once per signal before the local ELMs are fitted. To quantify its overhead, we reran the \texttt{kodim20} experiment of Figure~\ref{fig:kodim20_beam} and separately measured the partition-construction time and the ELM-fitting time. 
\begin{table}[H]
\centering
\small
\caption{Runtime and reconstruction quality of the regular mesh and BEAM on
the \texttt{kodim20} experiment of Figure~\ref{fig:kodim20_beam}
($256\times 256$, $N = 64$). Times are in milliseconds; the total runtime is
the sum of partition construction and ELM fitting. Best quality for each
capacity is shown in \textbf{bold}.}
\label{tab:beam_runtime}
\begin{tabular}{llccccc}
\specialrule{1pt}{2pt}{2pt}
$m$ & Partition & Construction (ms) & Fitting (ms) & Total (ms) & PSNR (dB) & SSIM \\
\midrule
256 & Regular mesh & --   & 3.4  & 3.4  & 26.02 & 0.9686 \\
256 & BEAM         & 14.7 & 18.4 & 33.1 & \textbf{27.54} & \textbf{0.9781} \\
\midrule
512 & Regular mesh & --   & 7.0  & 7.0  & 29.60 & 0.9864 \\
512 & BEAM         & 14.7 & 45.5 & 60.2 & \textbf{32.29} & \textbf{0.9928} \\
\specialrule{1pt}{2pt}{2pt}
\end{tabular}
\end{table}
As reported in Table~\ref{tab:beam_runtime}, BEAM requires only a one-time partition-construction step before ELM fitting. The ELM-fitting time differs between the two partitions because BEAM produces irregularly shaped patches, resulting in local least-squares systems with different sample layouts from those of the regular partition. In return, BEAM improves PSNR by 1.52\,dB for $m = 256$ and by 2.69\,dB for $m = 512$, while keeping the total runtime below
0.1 seconds.

\section{Additional Experiments: BEAM in Capacity-Limited Regimes}
\label{app:beam_additional}
\begin{figure*}[ht!]
    \centering
    \captionsetup[subfigure]{font=tiny}
    \begin{subfigure}[t]{0.48\textwidth}
        \centering
        \begin{subfigure}{0.32\linewidth}
            \includegraphics[width=\linewidth]{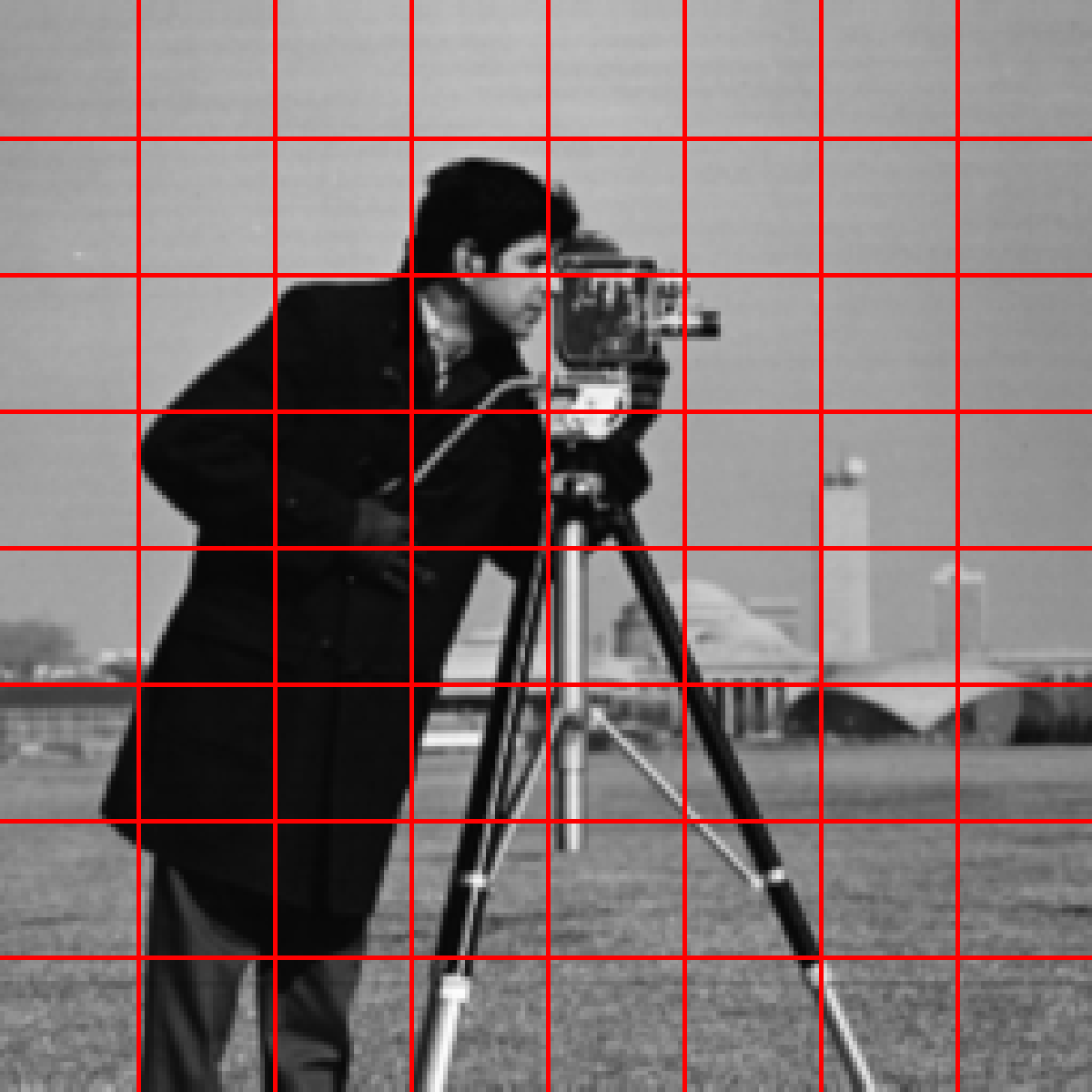}
            \caption{Regular Mesh}
        \end{subfigure}\hfill
        \begin{subfigure}{0.32\linewidth}
            \includegraphics[width=\linewidth]
            {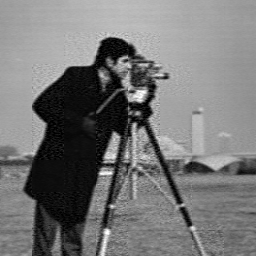}
            \caption{$m$=256 {\scriptsize(27.6 dB)}}
        \end{subfigure}\hfill
        \begin{subfigure}{0.32\linewidth}
            \includegraphics[width=\linewidth]
            {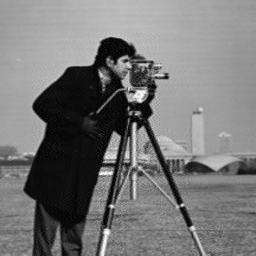}
            \caption{$m$=512 {\scriptsize(32.3 dB)}}
        \end{subfigure}

        \vspace{4pt}

        \begin{subfigure}{0.32\linewidth}
            \includegraphics[width=\linewidth]{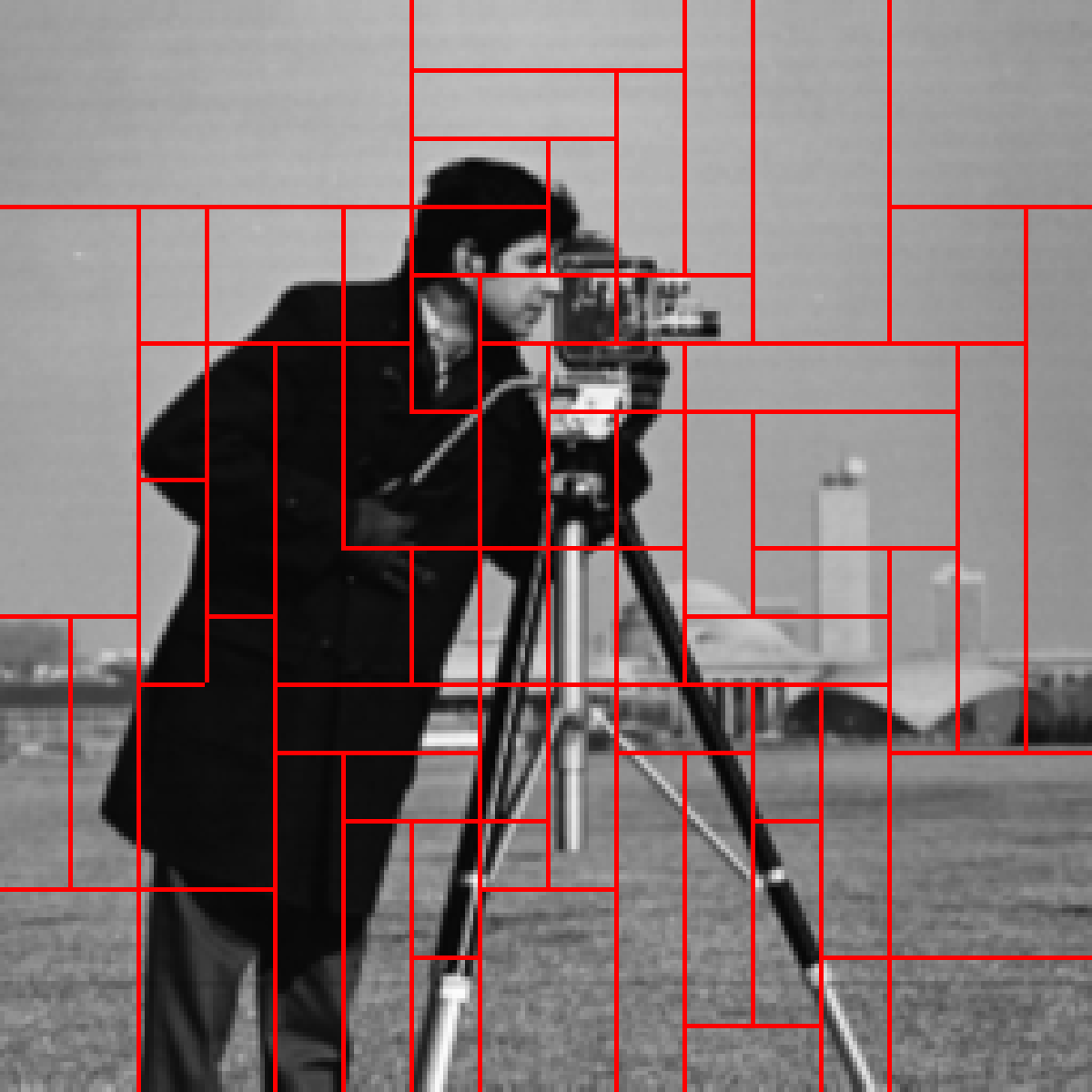}
            \caption{BEAM}
        \end{subfigure}\hfill
        \begin{subfigure}{0.32\linewidth}
            \includegraphics[width=\linewidth]
            {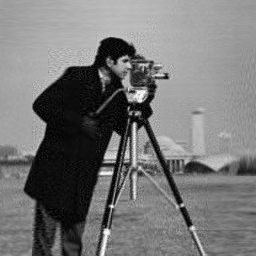}
            \caption{$m$=256 {\scriptsize(30.9 dB)}}
        \end{subfigure}\hfill
        \begin{subfigure}{0.32\linewidth}
            \includegraphics[width=\linewidth]
            {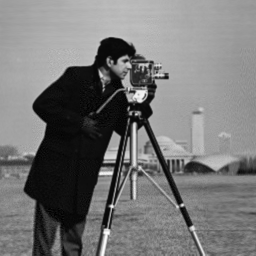}
            \caption{$m$=512 {\scriptsize(37.7 dB)}}
        \end{subfigure}

        \caption{\small
        \texttt{cameraman} (256$\times$256), $N$=64
        }
        \label{fig:cameraman_beam}
    \end{subfigure}
    \hfill
    \begin{subfigure}[t]{0.48\textwidth}
        \centering

        \begin{subfigure}{0.32\linewidth}
            \includegraphics[width=\linewidth]{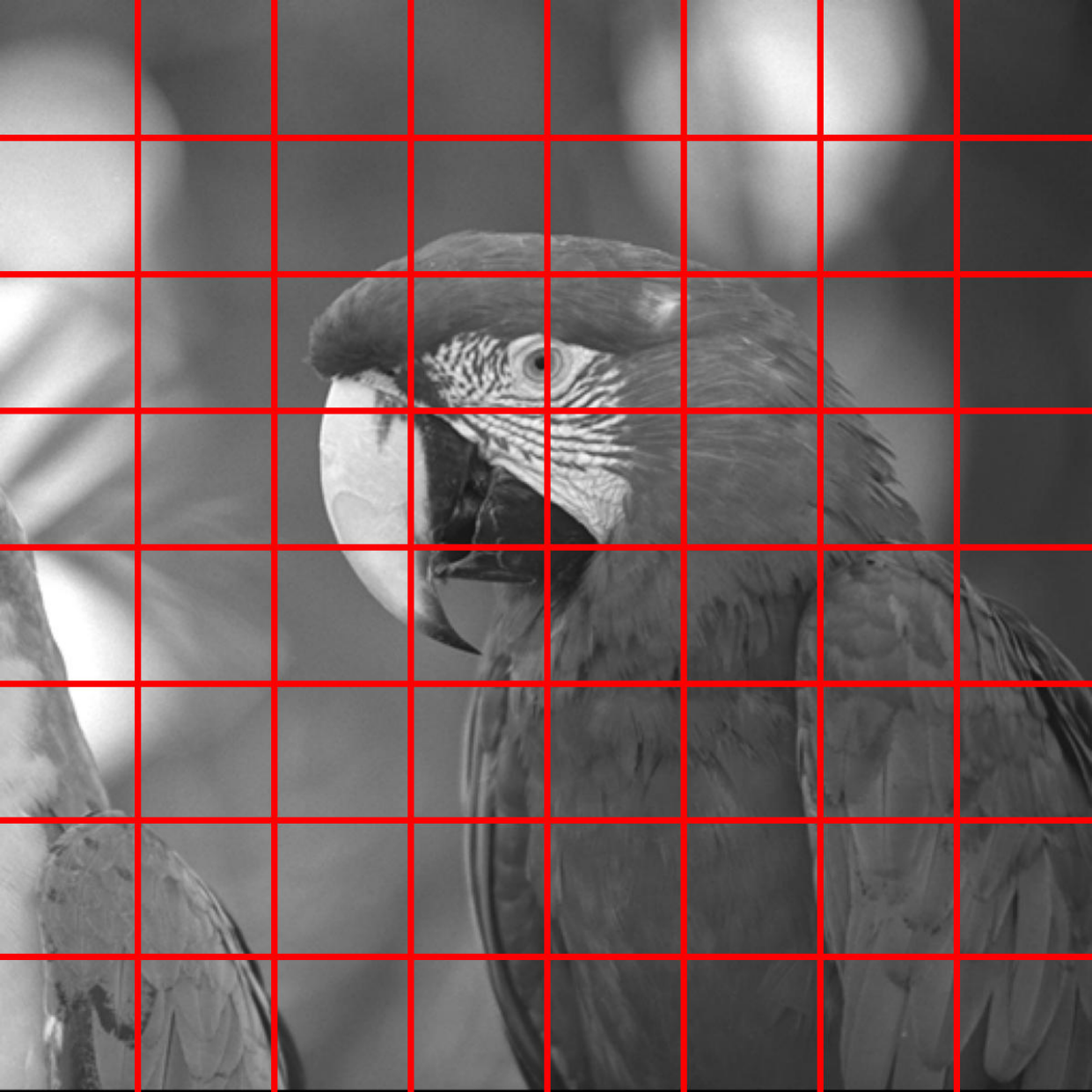}
            \caption{Regular Mesh}
        \end{subfigure}\hfill
        \begin{subfigure}{0.32\linewidth}
            \includegraphics[width=\linewidth]{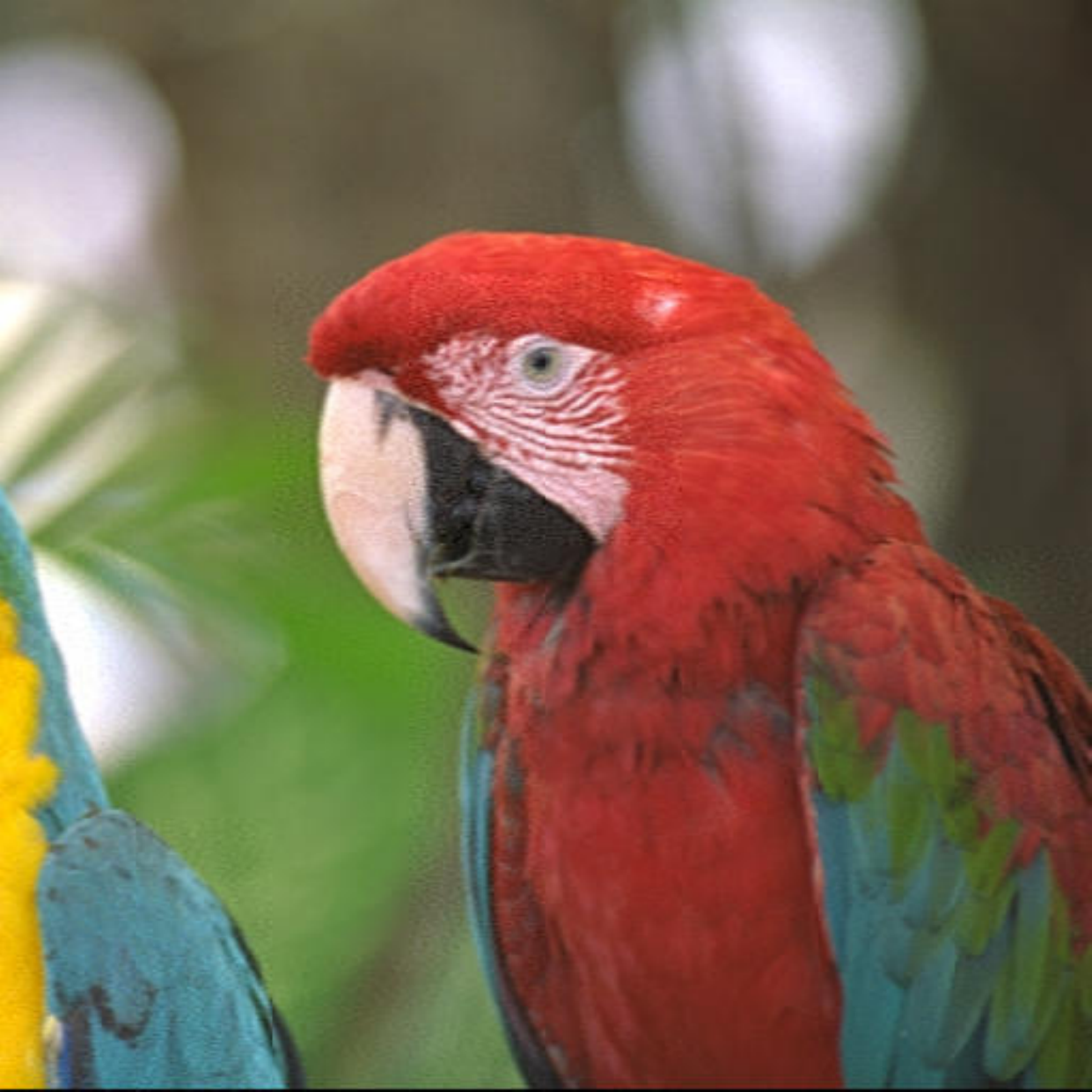}
            \caption{$m$=1024 {\scriptsize(34.1 dB)}}
        \end{subfigure}\hfill
        \begin{subfigure}{0.32\linewidth}
            \includegraphics[width=\linewidth]{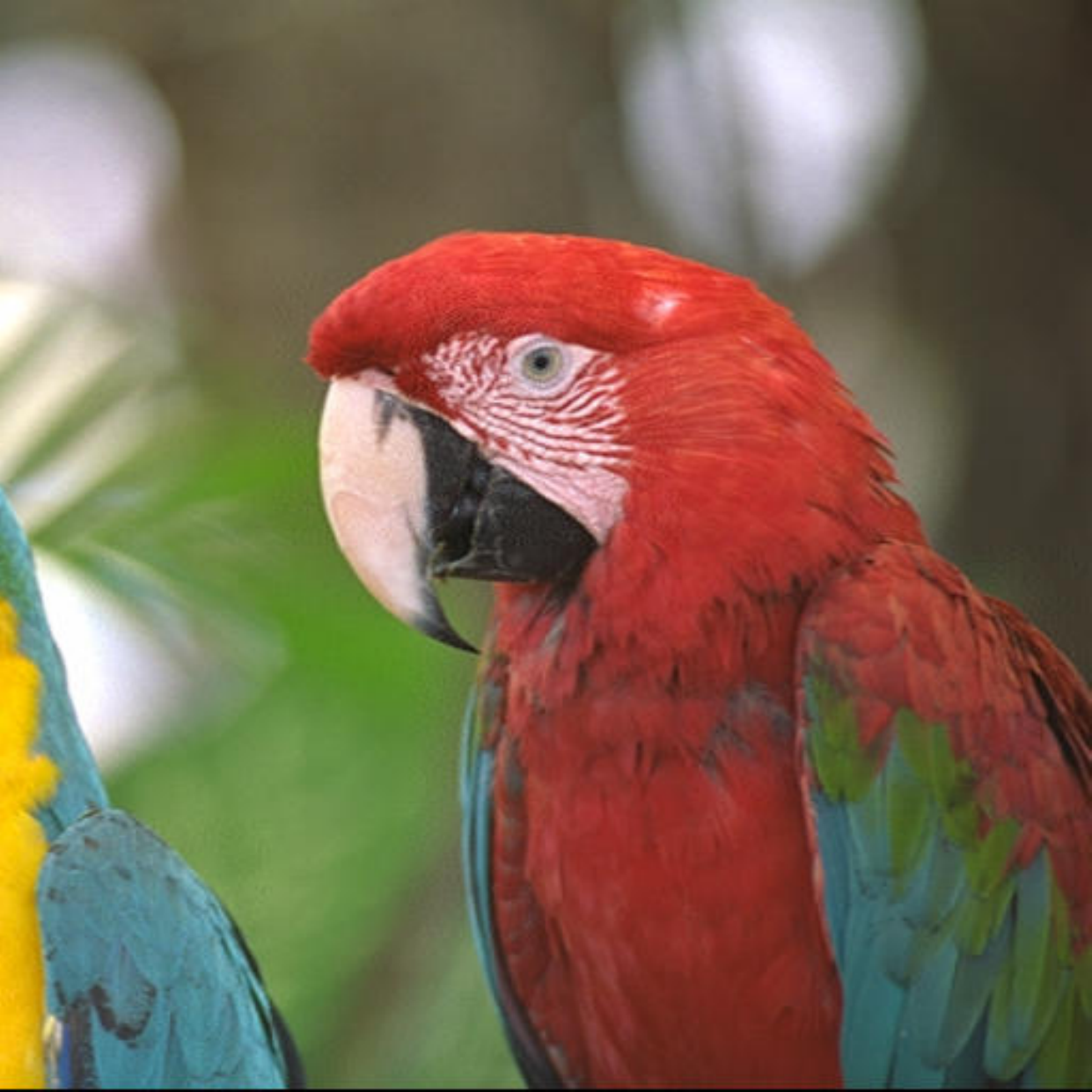}
            \caption{$m$=2048 {\scriptsize(38.3 dB)}}
        \end{subfigure}

        \vspace{4pt}

        \begin{subfigure}{0.32\linewidth}
            \includegraphics[width=\linewidth]{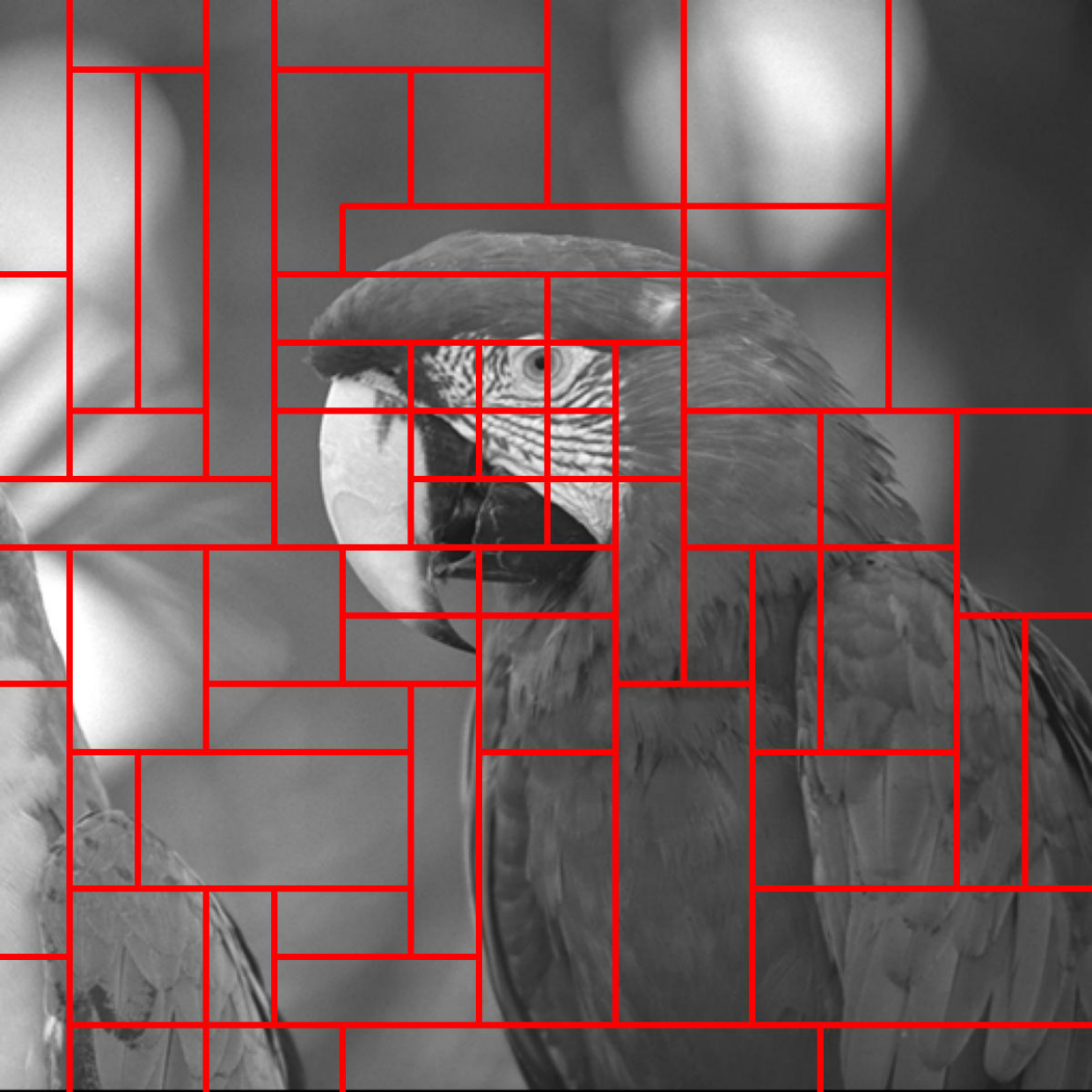}
            \caption{BEAM}
        \end{subfigure}\hfill
        \begin{subfigure}{0.32\linewidth}
            \includegraphics[width=\linewidth]{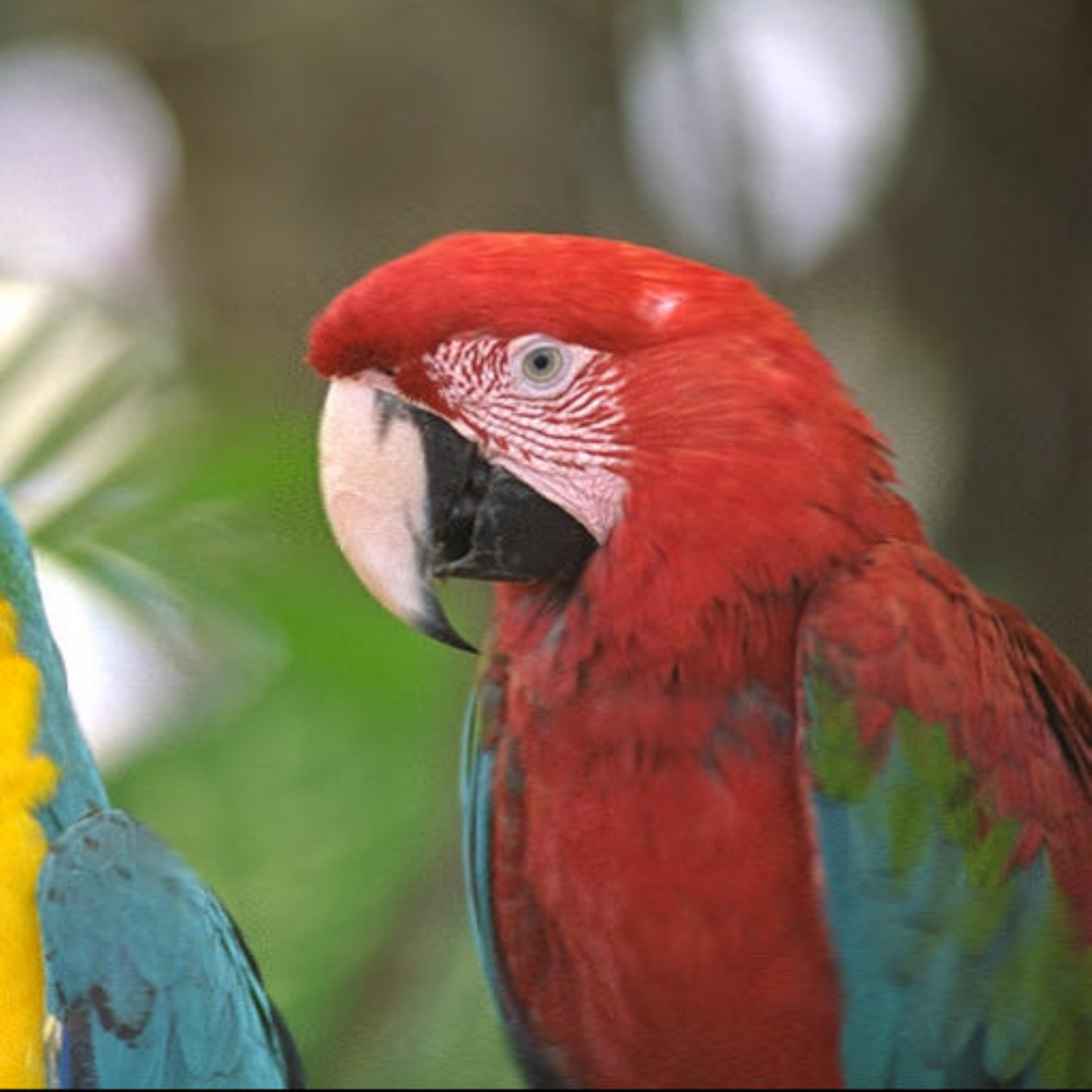}
            \caption{$m$=1024 {\scriptsize(36.9 dB)}}
        \end{subfigure}\hfill
        \begin{subfigure}{0.32\linewidth}
            \includegraphics[width=\linewidth]{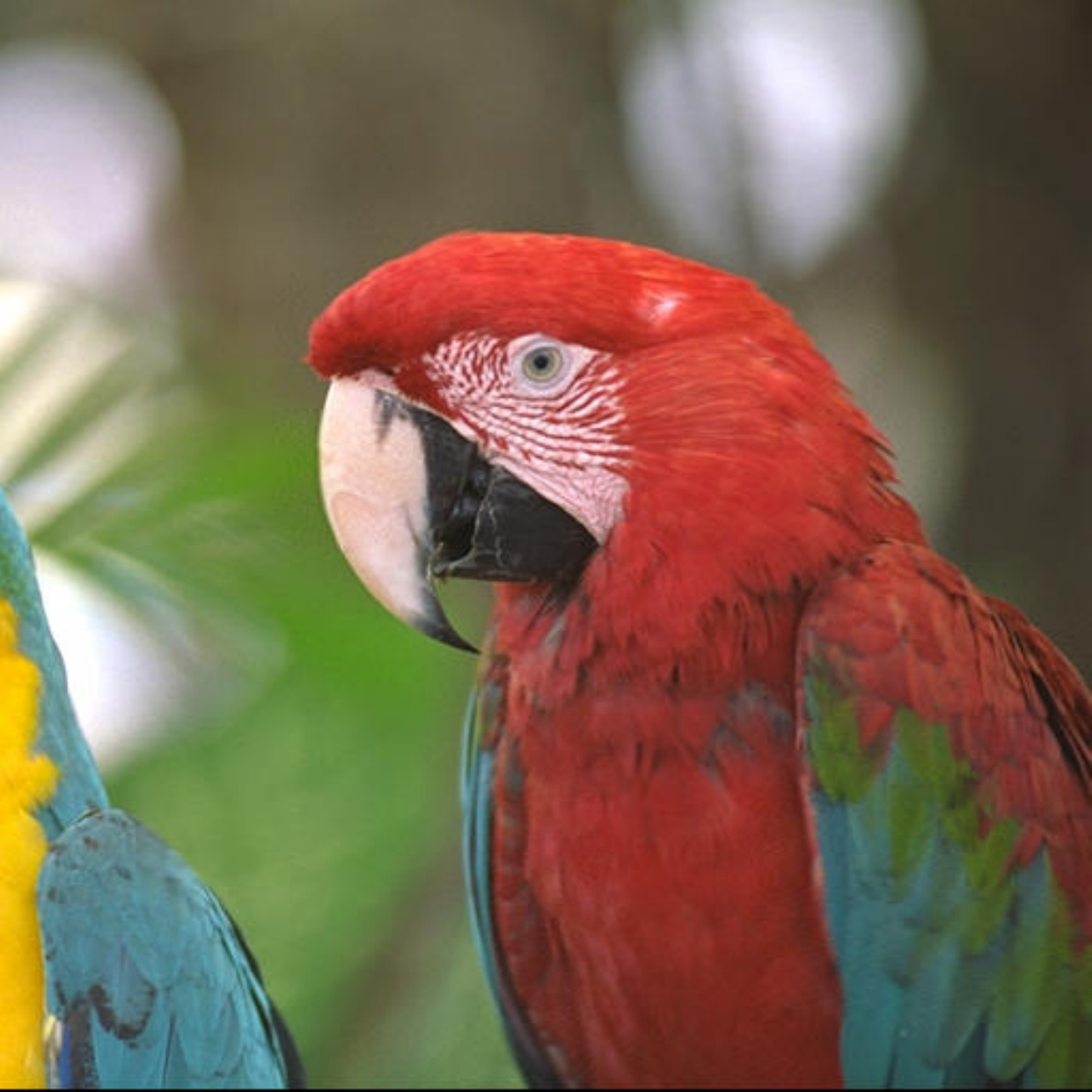}
            \caption{$m$=2048 {\scriptsize(41.0 dB)}}
        \end{subfigure}

        \caption{\small
        \texttt{kodim23} (512$\times$512), $N$=64
        }
        \label{fig:kodim23_beam}
    \end{subfigure}

    \caption{
    Side-by-side comparison of ELM-INR without BEAM (top rows) and with BEAM (bottom rows) on standard images.
    BEAM consistently improves reconstruction quality at identical model capacity $m$.
    }
    \label{fig:beam_side_by_side}
\end{figure*}

This section provides additional evidence of the effectiveness of BEAM in capacity-limited settings for ELM-INR. BEAM is designed to be most beneficial when the number of hidden nodes $m$ per local ELM is small, as reconstruction quality in this regime is dominated by the most spectrally complex regions (cf. Eq.~\eqref{eq:barron_bound}). By adaptively refining only difficult regions while allocating larger patches to smoother areas, BEAM enables a more effective use of a fixed budget of basis functions than uniform (regular) partitioning.

Figure~\ref{fig:beam_side_by_side} follows the same experimental protocol as in the main paper (e.g., Figures~\ref{fig:kodim20_beam}--\ref{fig:era5}), comparing ELM-INR with a regular mesh (top rows) against ELM-INR coupled with BEAM (bottom rows) at identical model capacity $m$. For the \texttt{cameraman} image ($256 \times 256$), we use spatial patches with $S_i = 32^2$ samples, resulting in $N = 64$ subdomains, and construct BEAM with an atomic size of $s=16$. Experiments are conducted for $m=256$ and $m=512$. For the \texttt{kodim23} image ($512 \times 512$), we use larger patches with $S_i = 64^2$, again fixing the number of subdomains to $N = 64$, and set the atomic size to $s=32$, with model capacities $m=1024$ and $m=2048$.

Across both datasets, BEAM consistently improves reconstruction quality, with especially pronounced gains at smaller $m$, where uniform partitioning tends to underfit localized high-frequency content such as edges and textured regions. In contrast, when $m$ is sufficiently large, reconstruction performance saturates, as also observed in the main experiments, leaving less room for improvement through adaptive partitioning. This behavior highlights that BEAM is particularly effective in regimes where model capacity is constrained.

\begin{figure*}[t]
    \centering
    \setlength{\tabcolsep}{1.5pt}
    \renewcommand{\arraystretch}{1.0}

    \begin{minipage}[t]{0.48\linewidth}
        \centering
        \textbf{Spring}
        \vspace{1pt}
        \begin{tabular}{ccc}
            \includegraphics[width=0.31\linewidth]{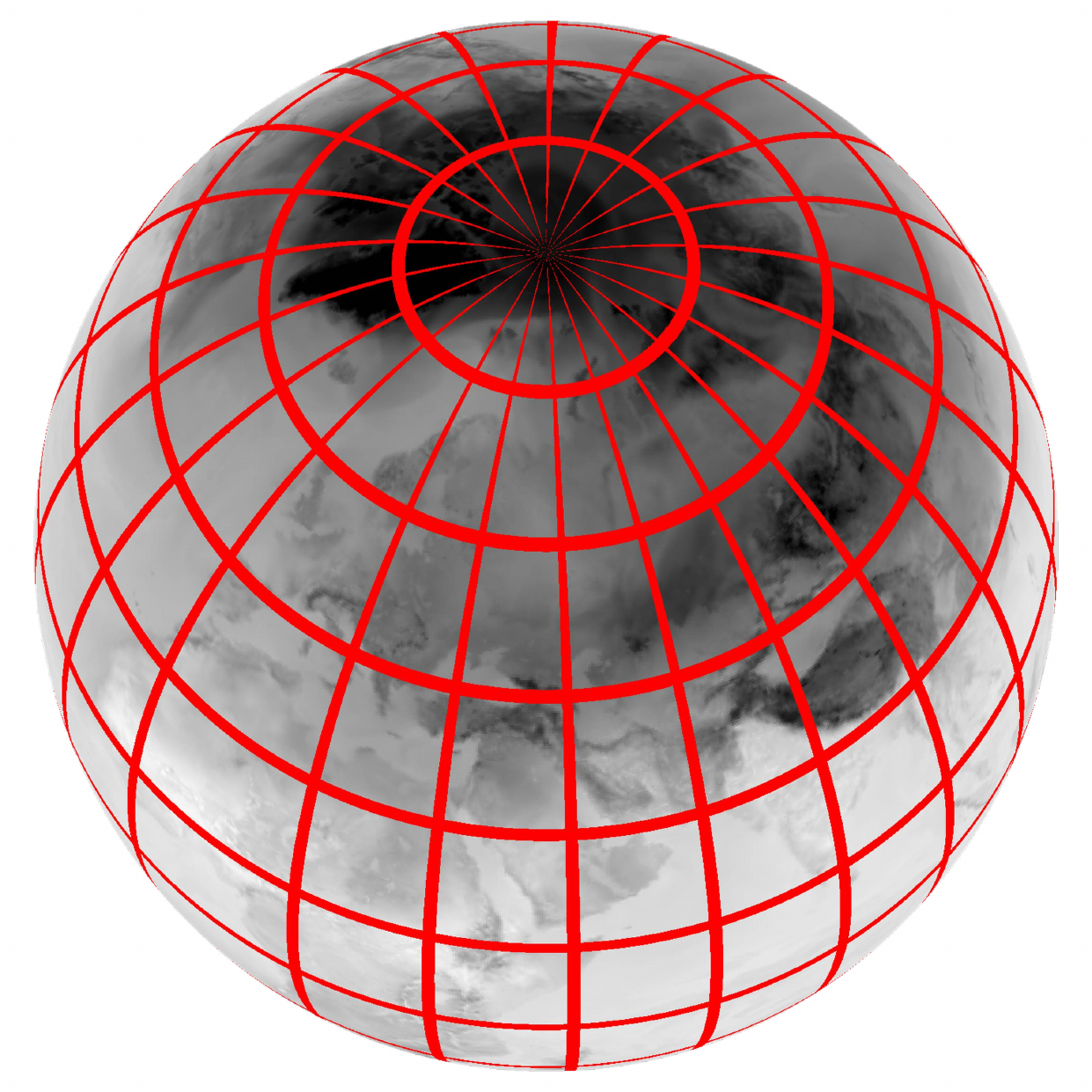} &
            \includegraphics[width=0.31\linewidth]{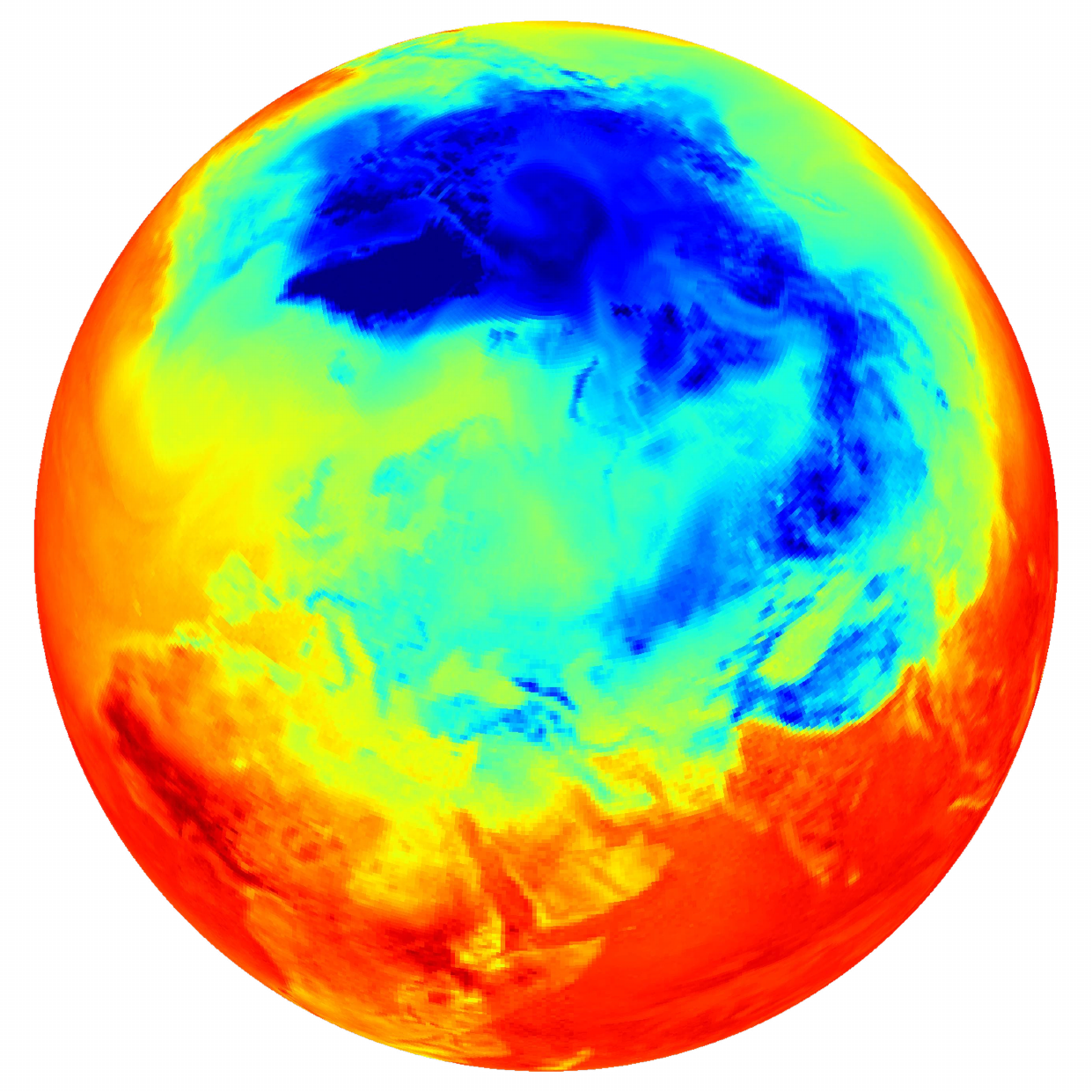} &
            \includegraphics[width=0.31\linewidth]{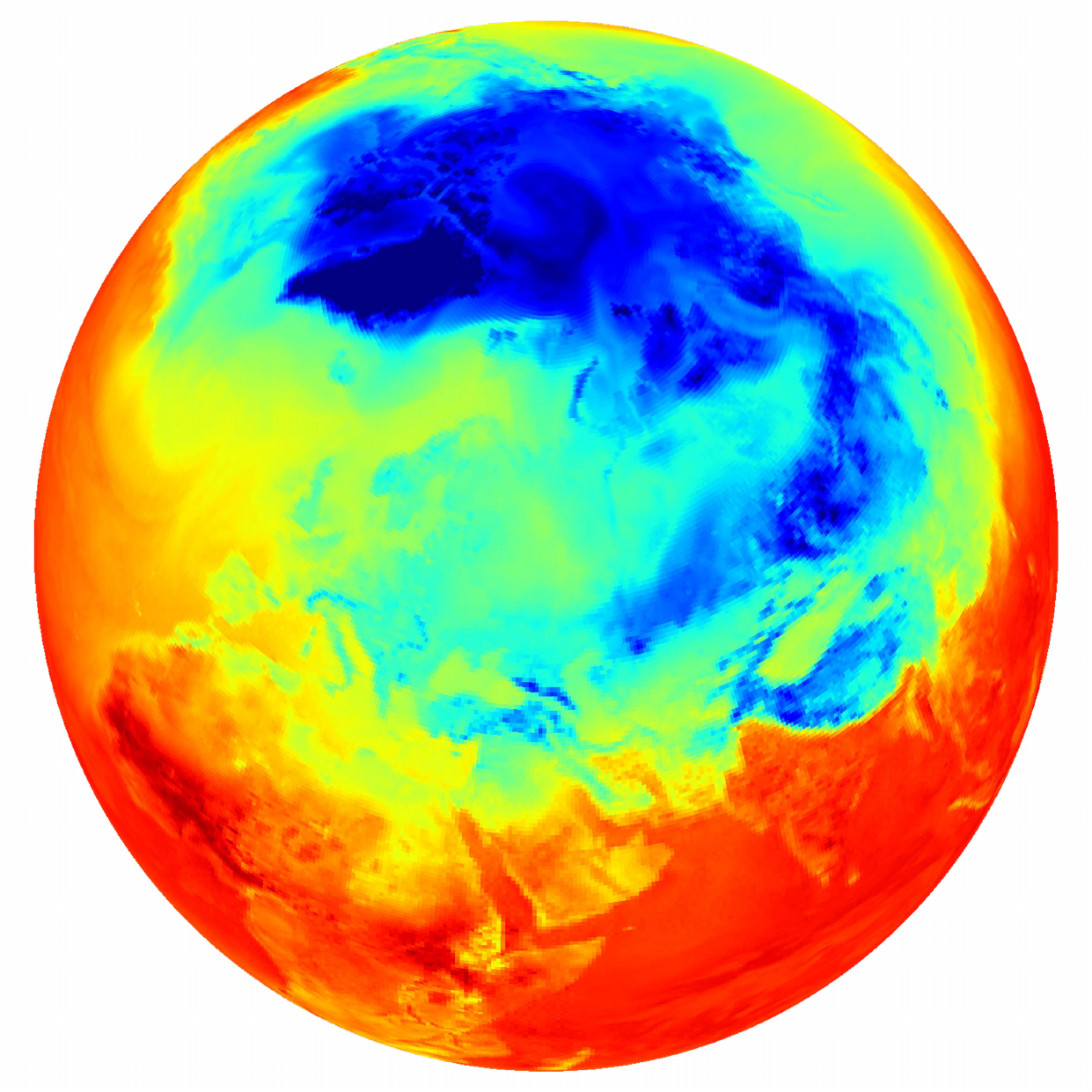} \\
            \scriptsize Regular Mesh &
            \scriptsize $m$=1024 (1.158) &
            \scriptsize $m$=2048 (1.075) \\[4pt]

            \includegraphics[width=0.31\linewidth]{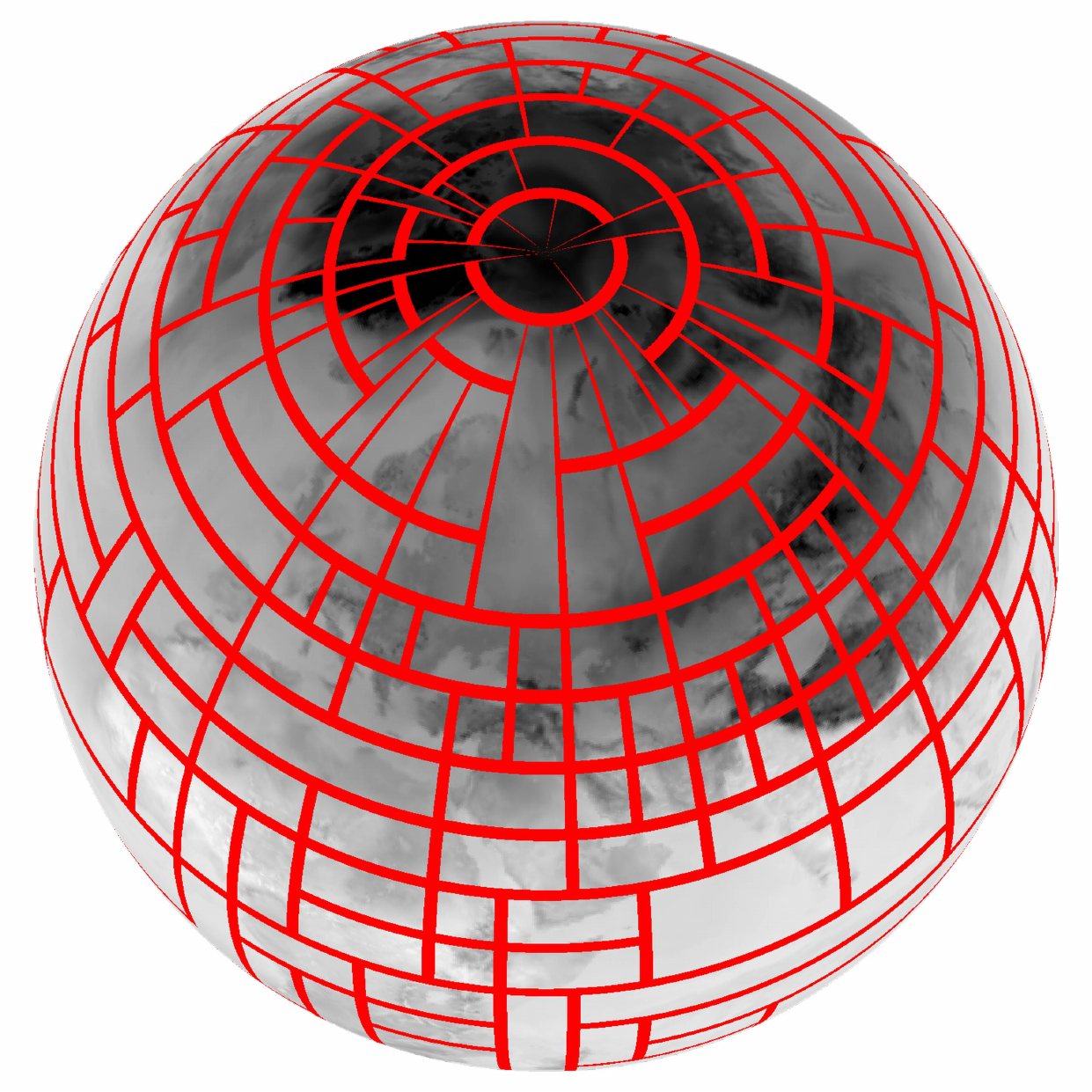} &
            \includegraphics[width=0.31\linewidth]{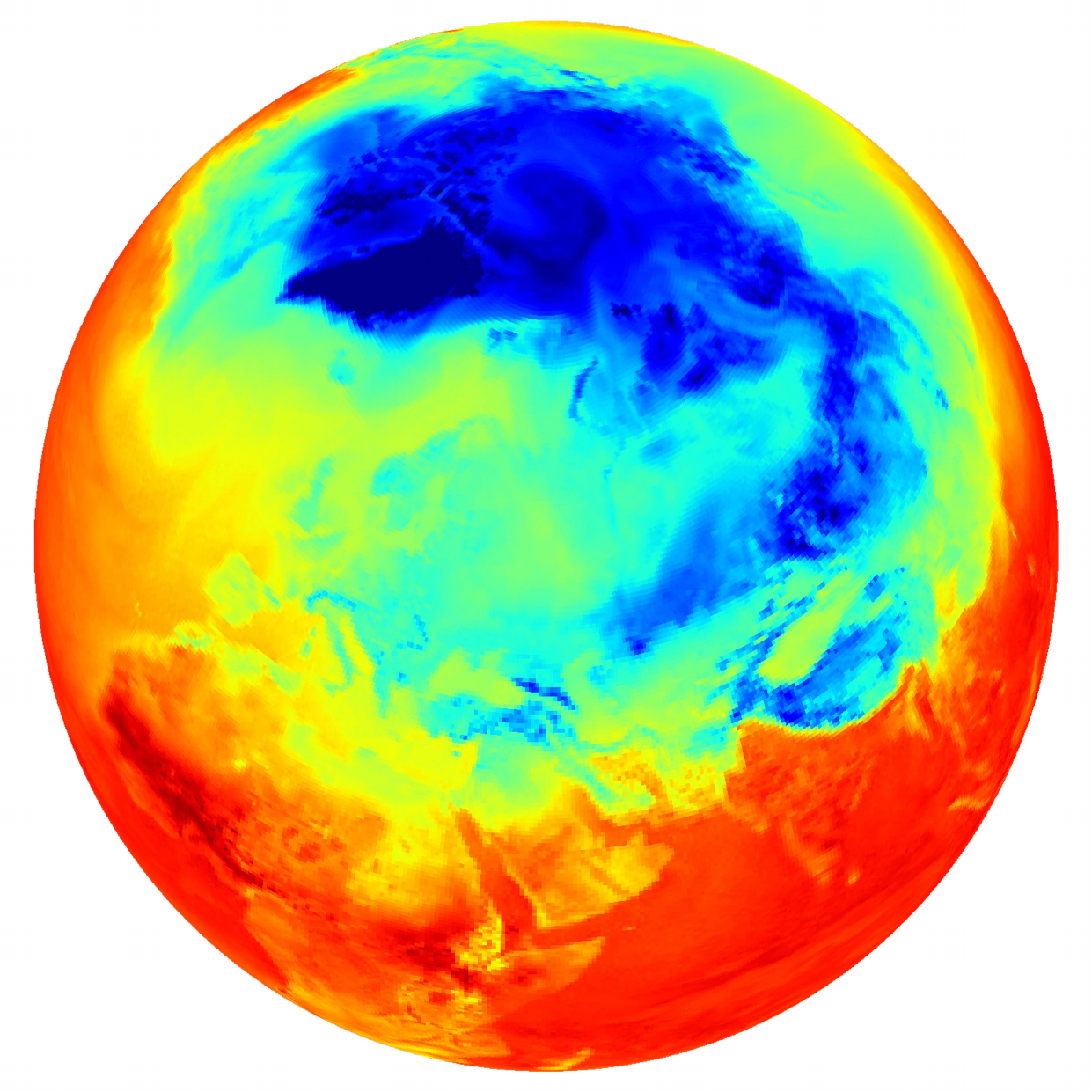} &
            \includegraphics[width=0.31\linewidth]{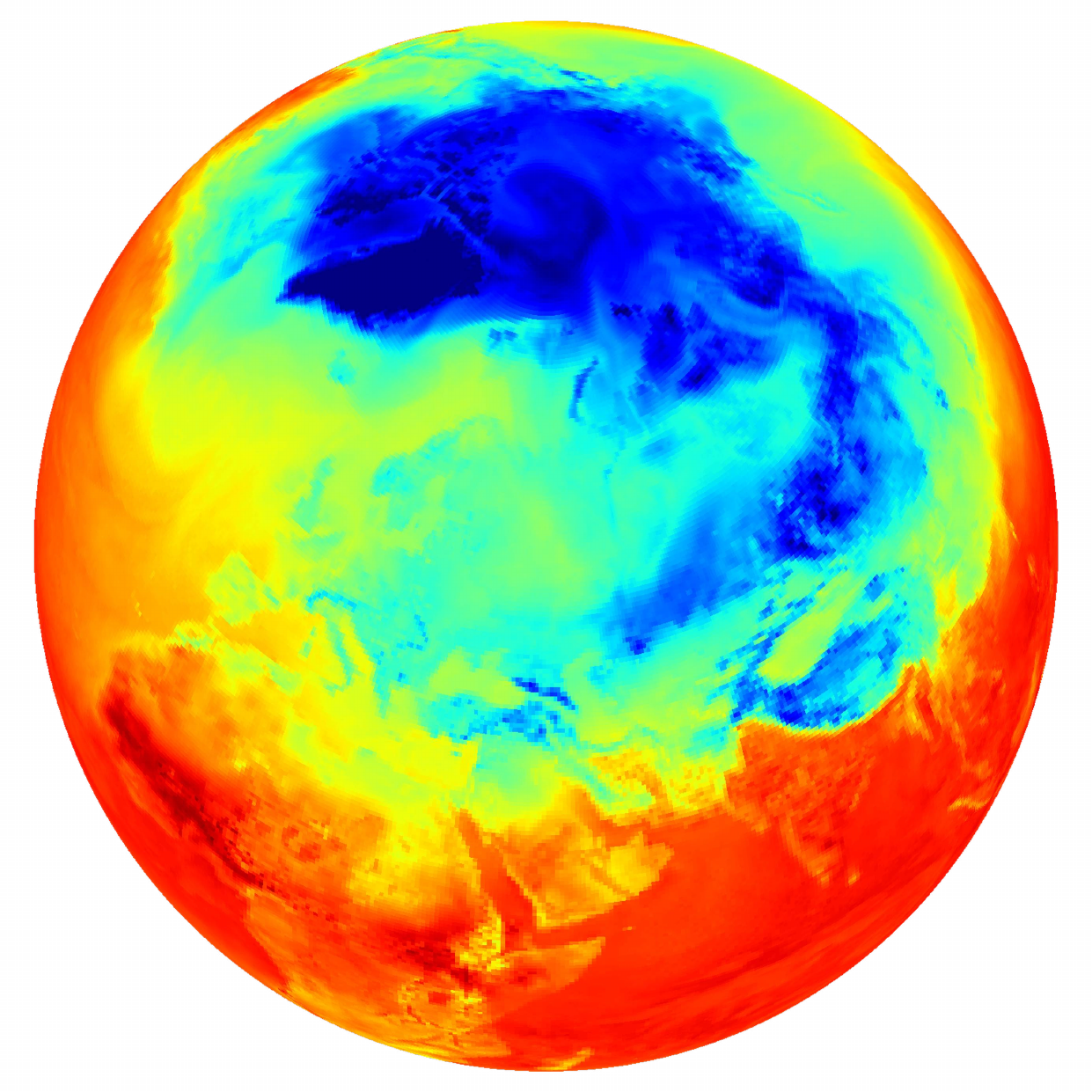} \\
            \scriptsize BEAM &
            \scriptsize $m$=1024 (1.118) &
            \scriptsize $m$=2048 (1.040)
        \end{tabular}
    \end{minipage}
    \hfill
    \begin{minipage}[t]{0.48\linewidth}
        \centering
        \textbf{Summer}
        \vspace{1pt}
        \begin{tabular}{ccc}
            \includegraphics[width=0.31\linewidth]{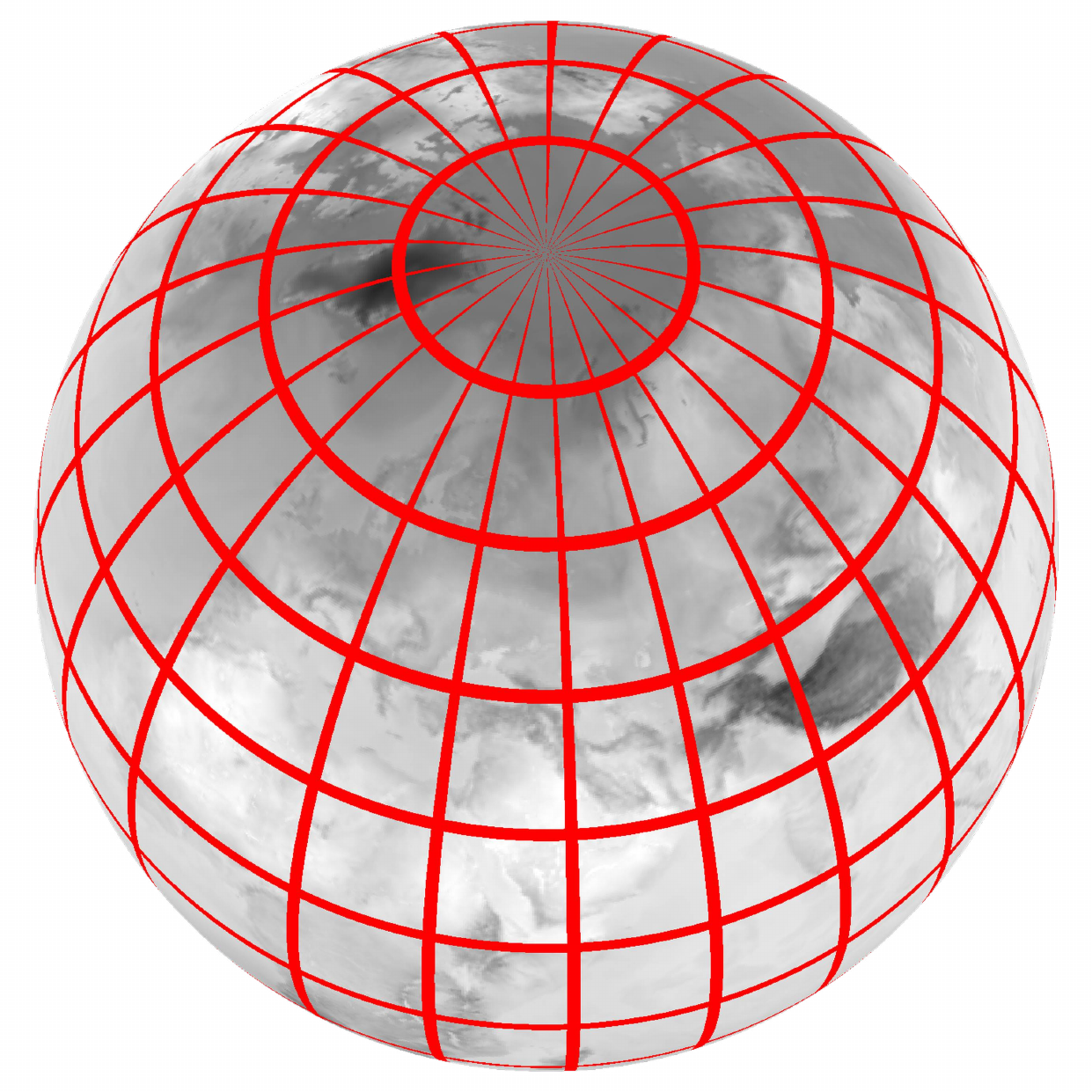} &
            \includegraphics[width=0.31\linewidth]{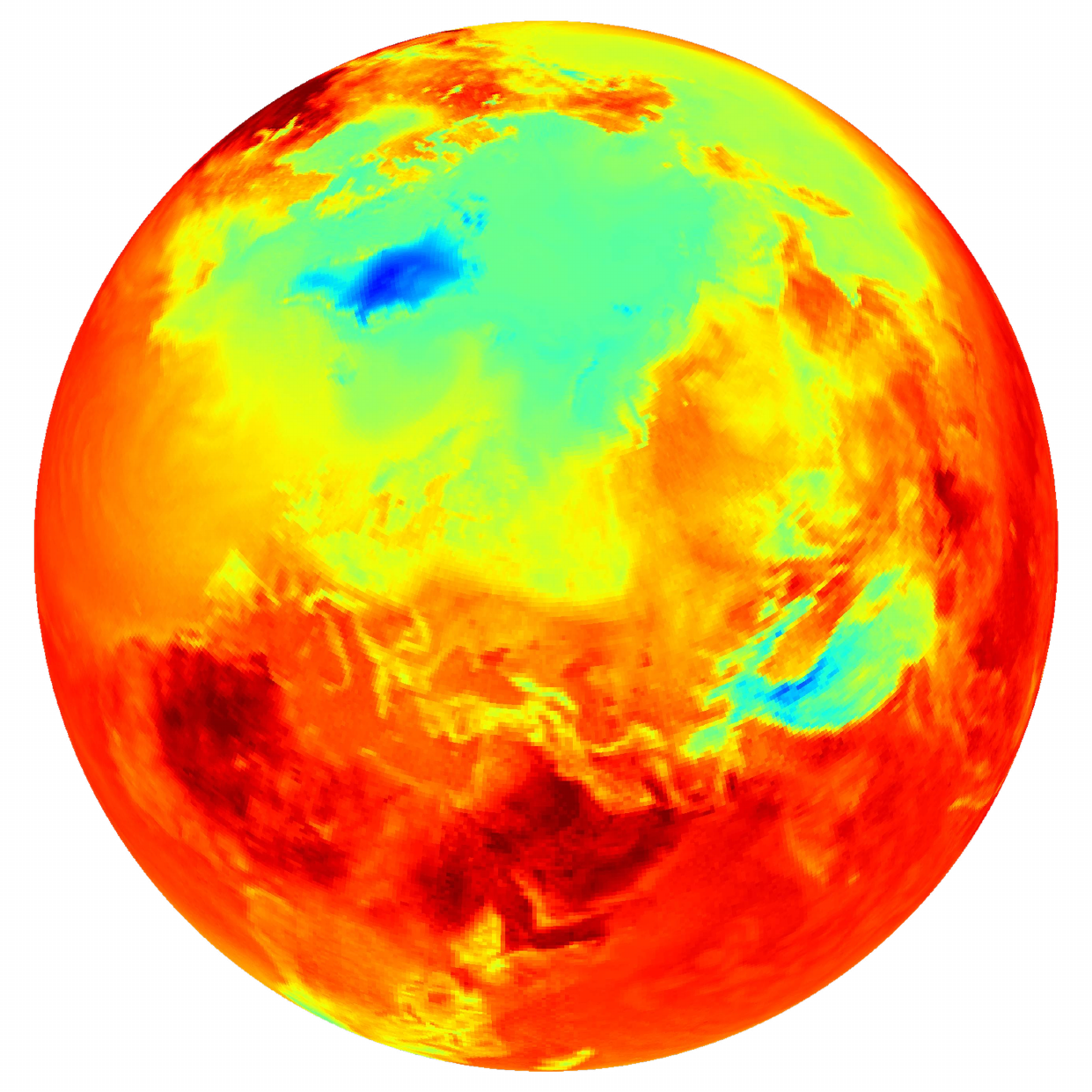} &
            \includegraphics[width=0.31\linewidth]{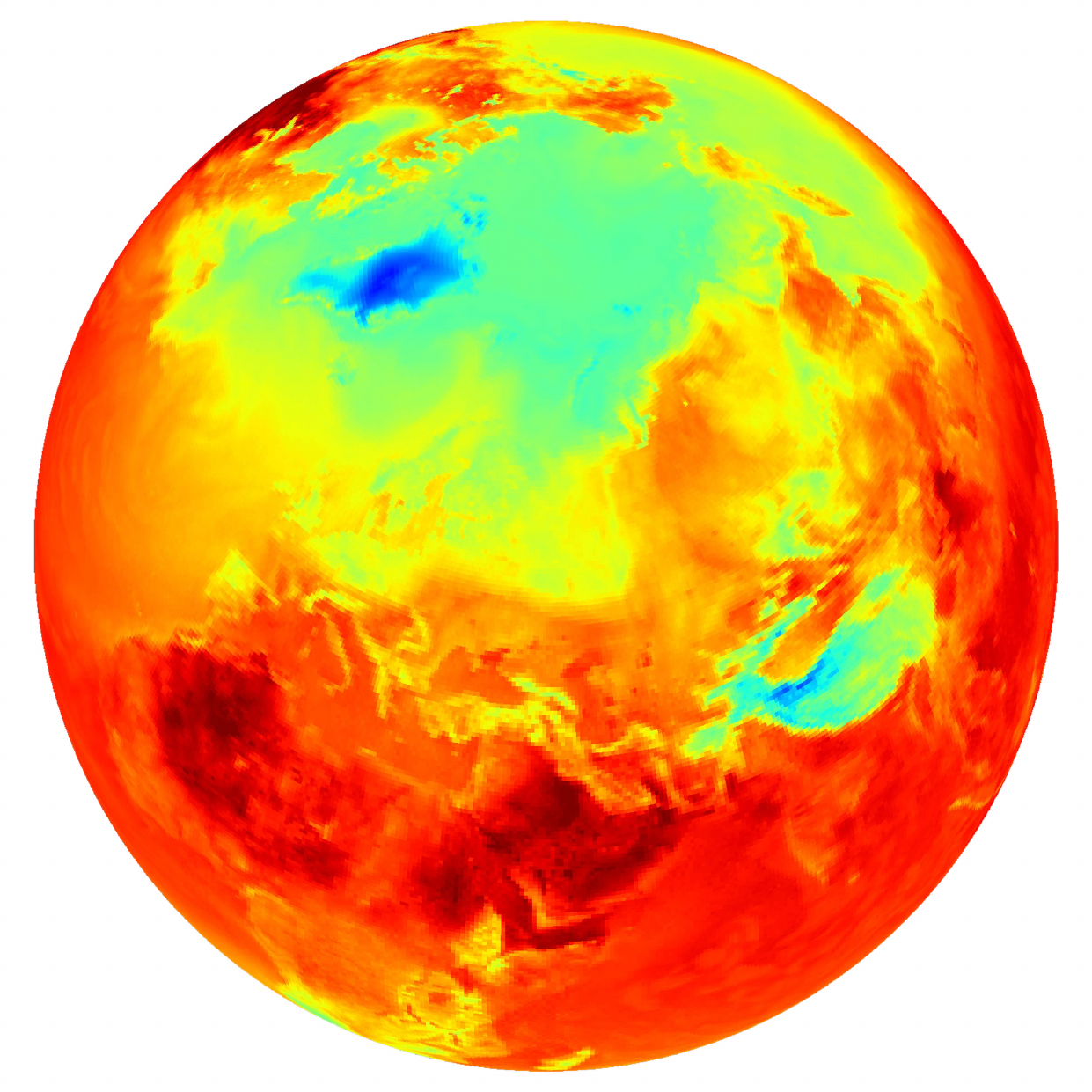} \\
            \scriptsize Regular Mesh &
            \scriptsize $m$=1024 (1.450) &
            \scriptsize $m$=2048 (1.367) \\[4pt]

            \includegraphics[width=0.31\linewidth]{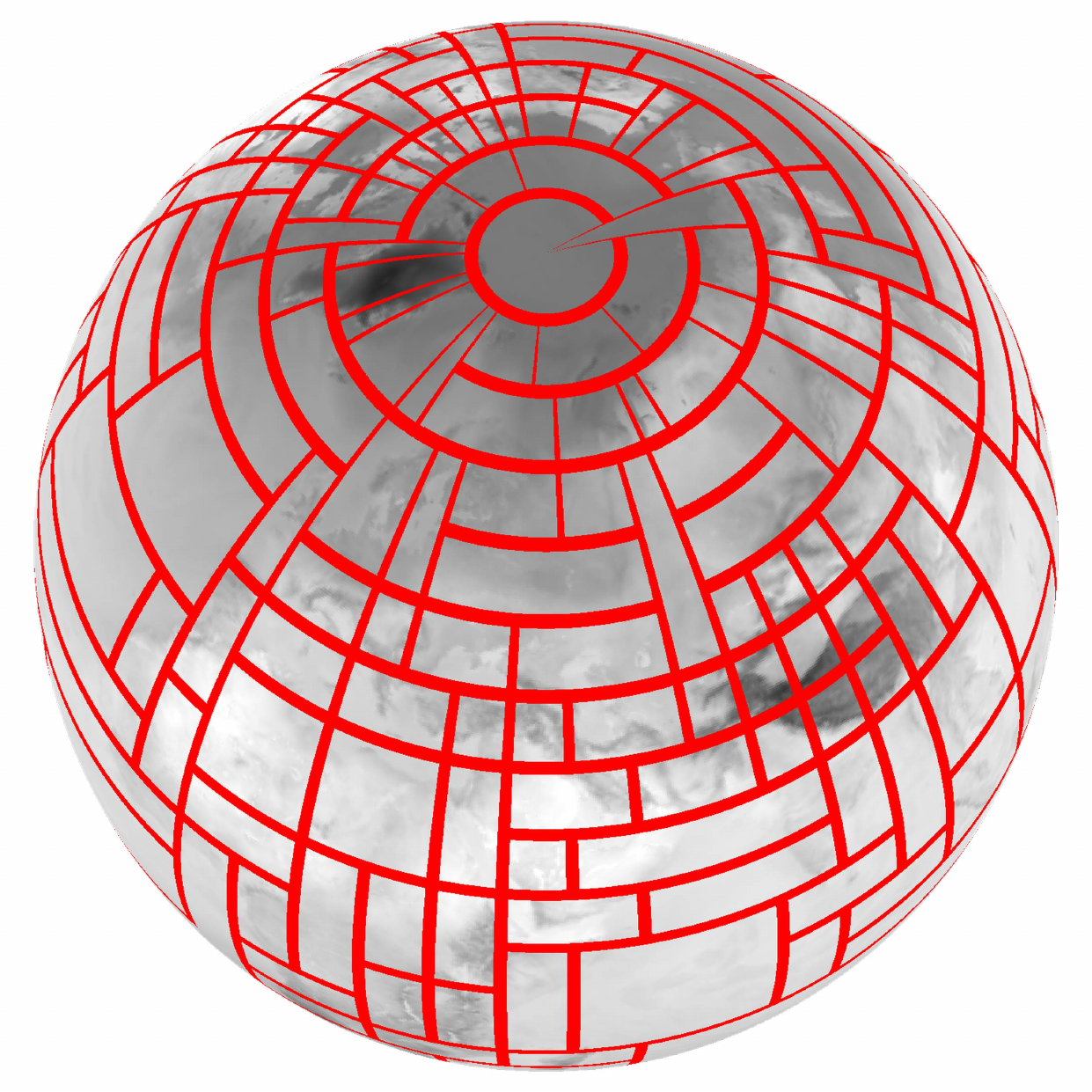} &
            \includegraphics[width=0.31\linewidth]{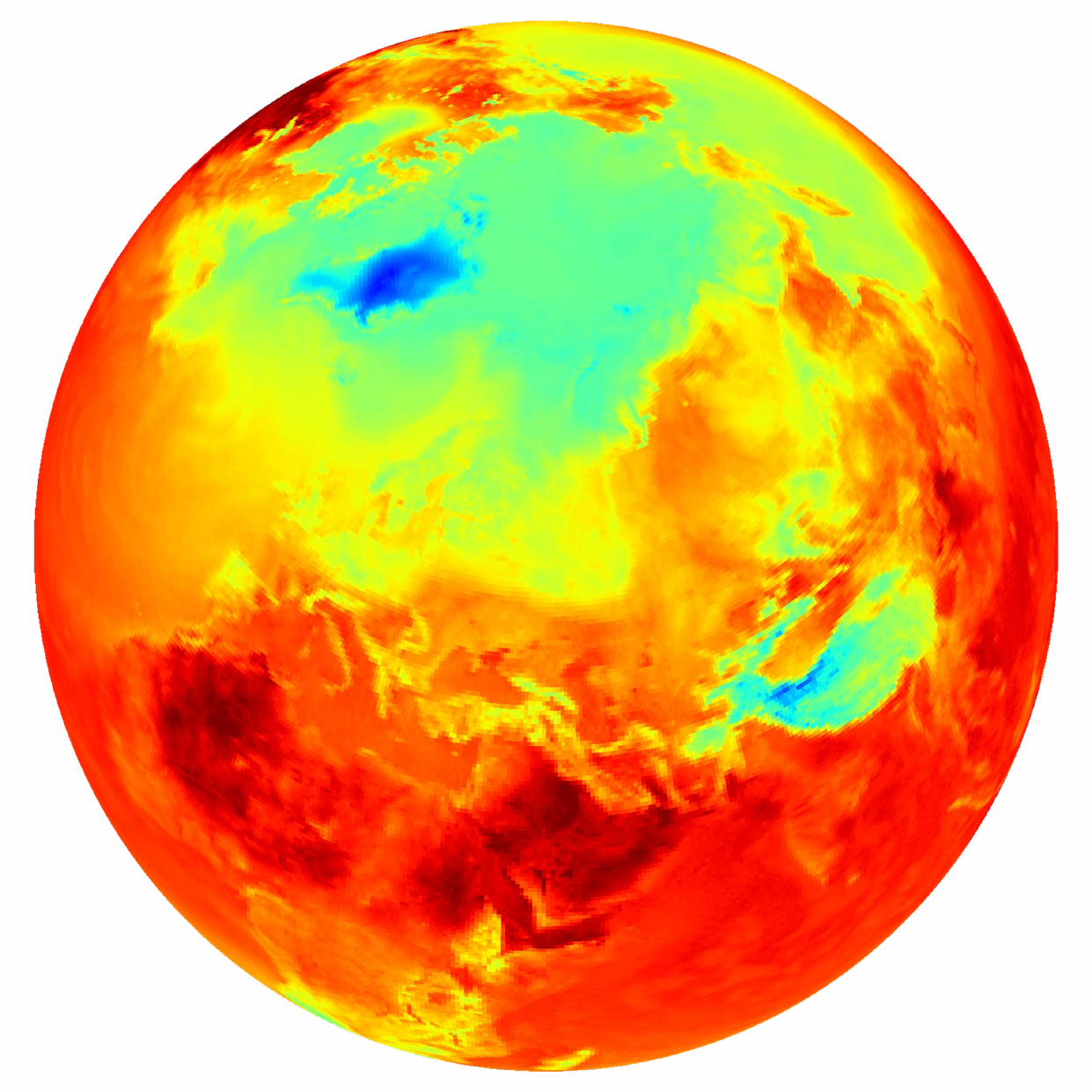} &
            \includegraphics[width=0.31\linewidth]{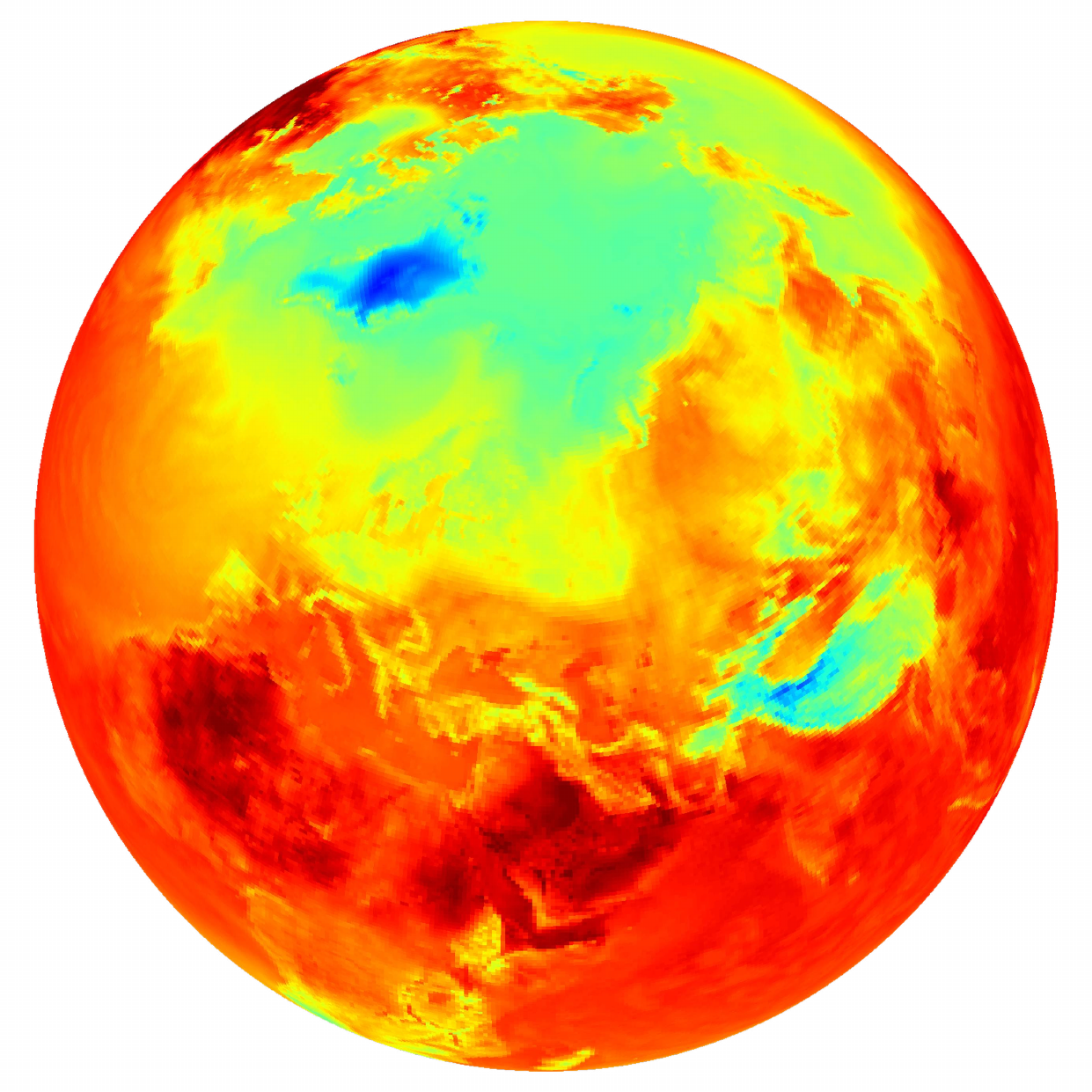} \\
            \scriptsize BEAM &
            \scriptsize $m$=1024 (1.408) &
            \scriptsize $m$=2048 (1.331)
        \end{tabular}
    \end{minipage}

    \vspace{6pt}

    \begin{minipage}[t]{0.48\linewidth}
        \centering
        \textbf{Autumn}
        \vspace{1pt}
        \begin{tabular}{ccc}
            \includegraphics[width=0.31\linewidth]{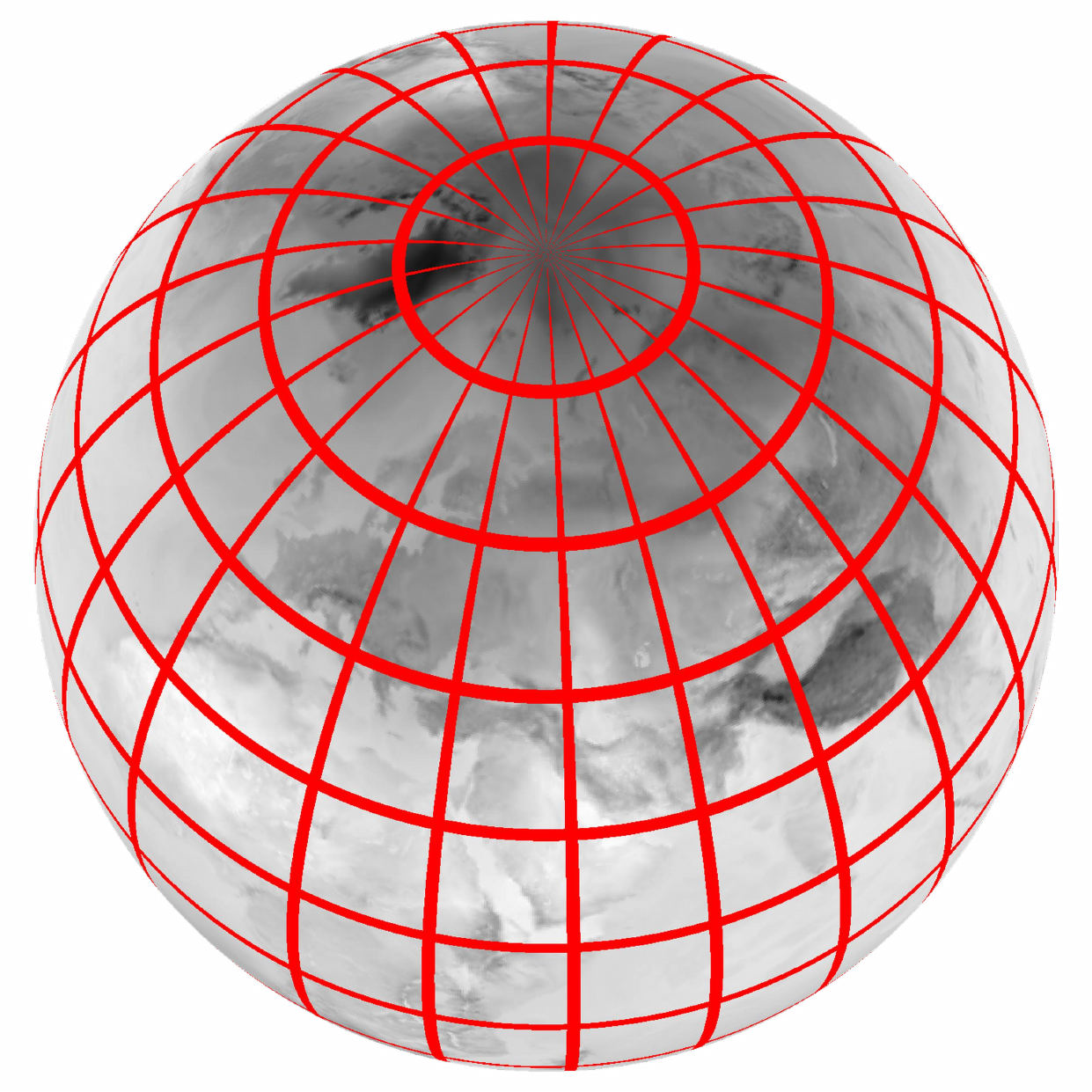} &
            \includegraphics[width=0.31\linewidth]{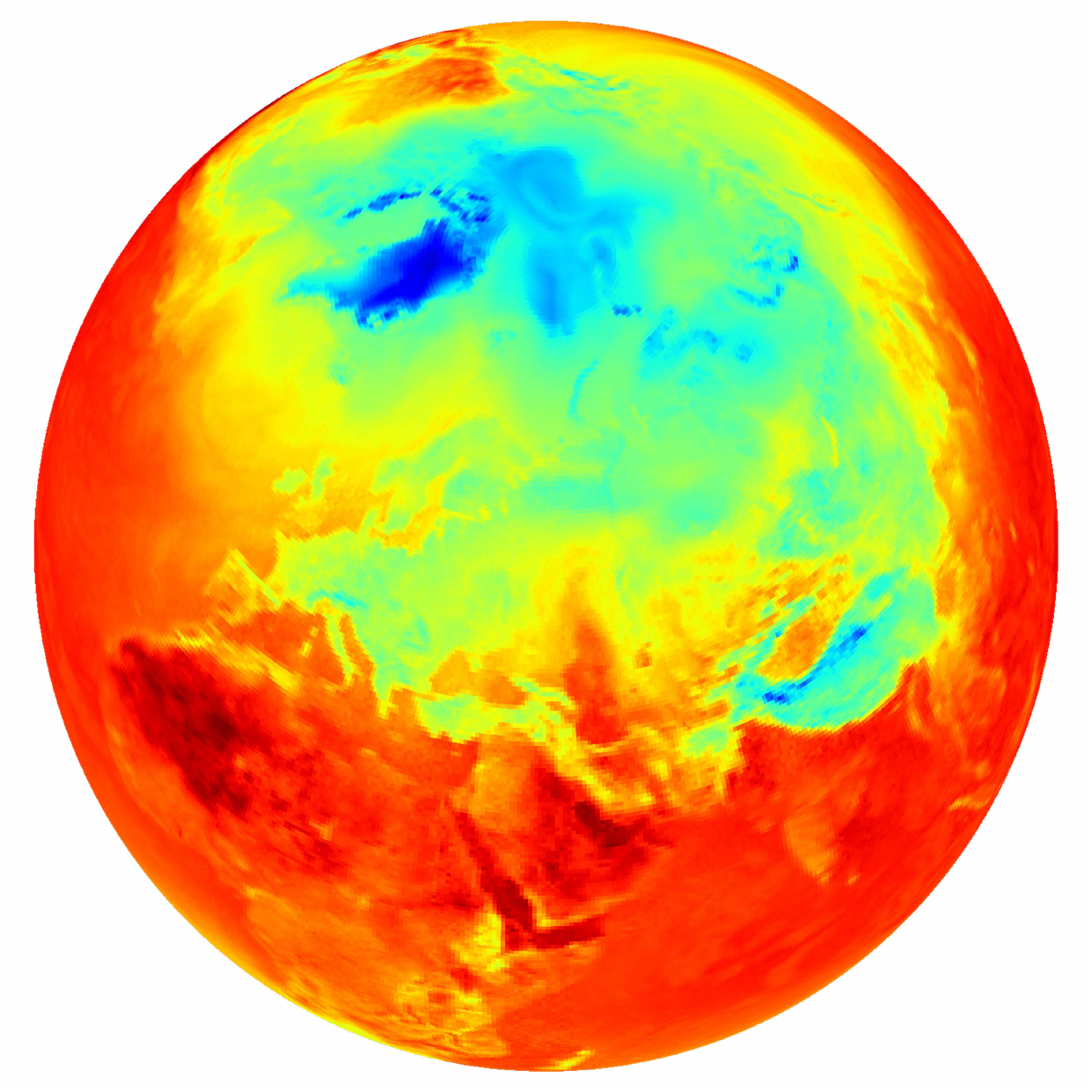} &
            \includegraphics[width=0.31\linewidth]{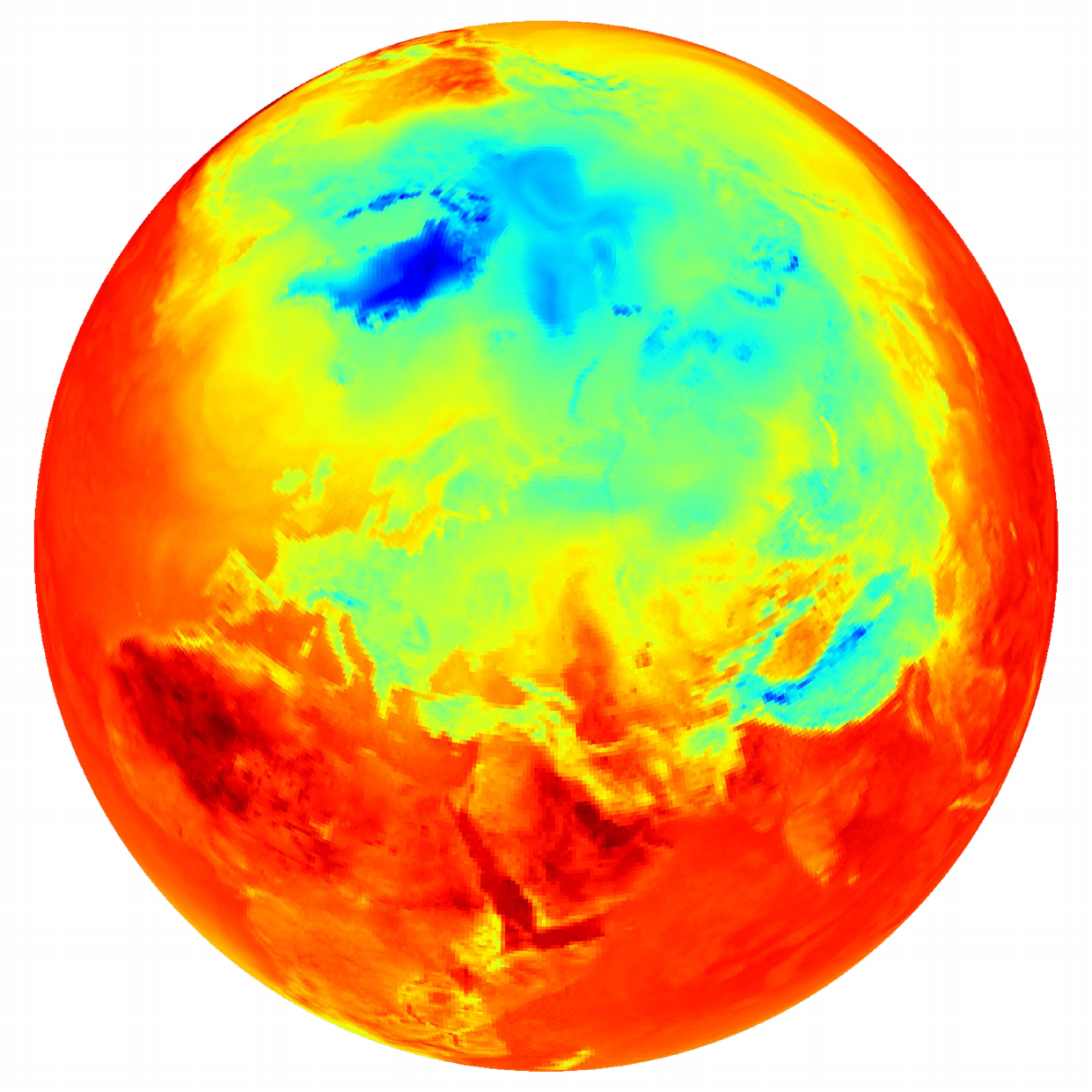} \\
            \scriptsize Regular Mesh &
            \scriptsize $m$=1024 (1.775) &
            \scriptsize $m$=2048 (1.698) \\[4pt]

            \includegraphics[width=0.31\linewidth]{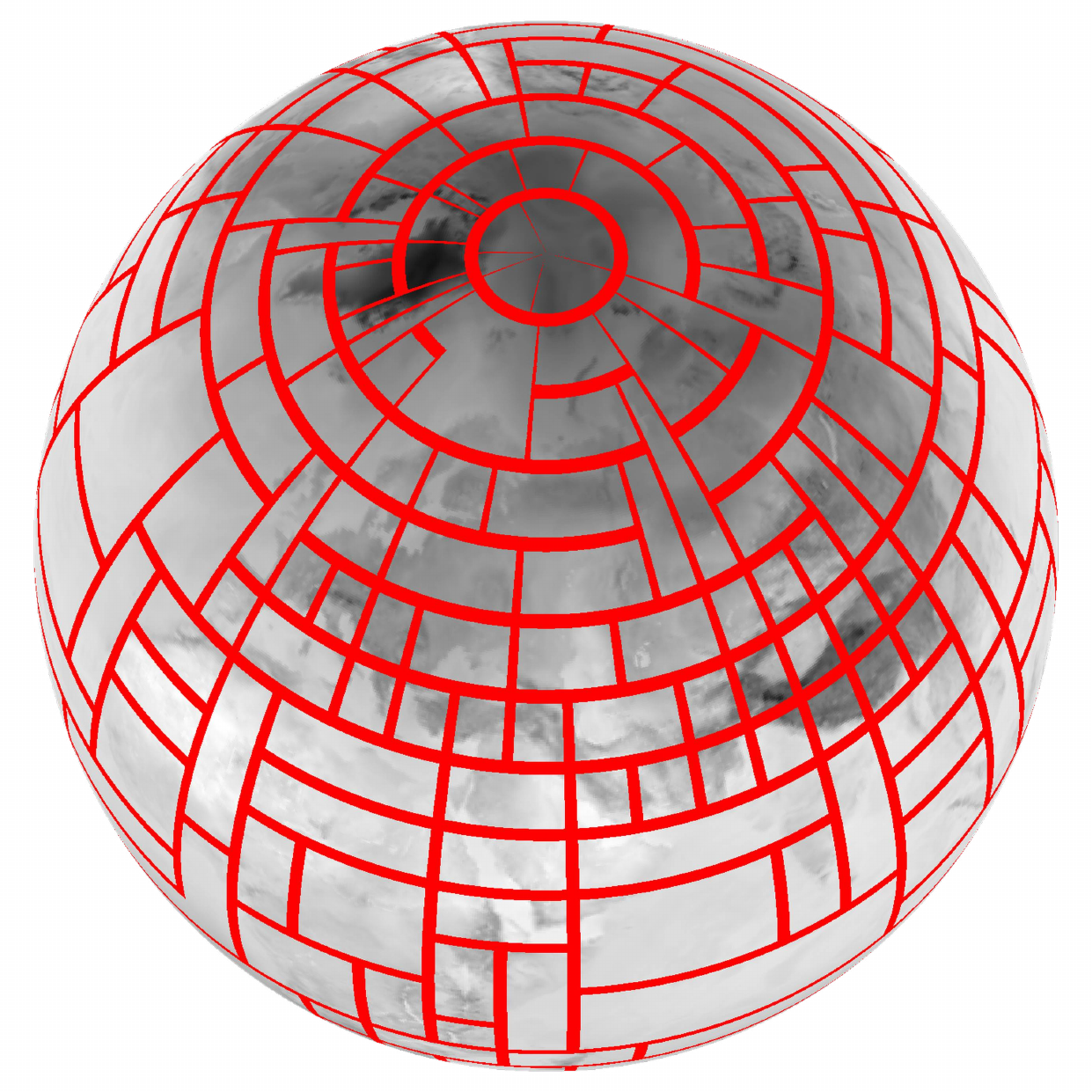} &
            \includegraphics[width=0.31\linewidth]{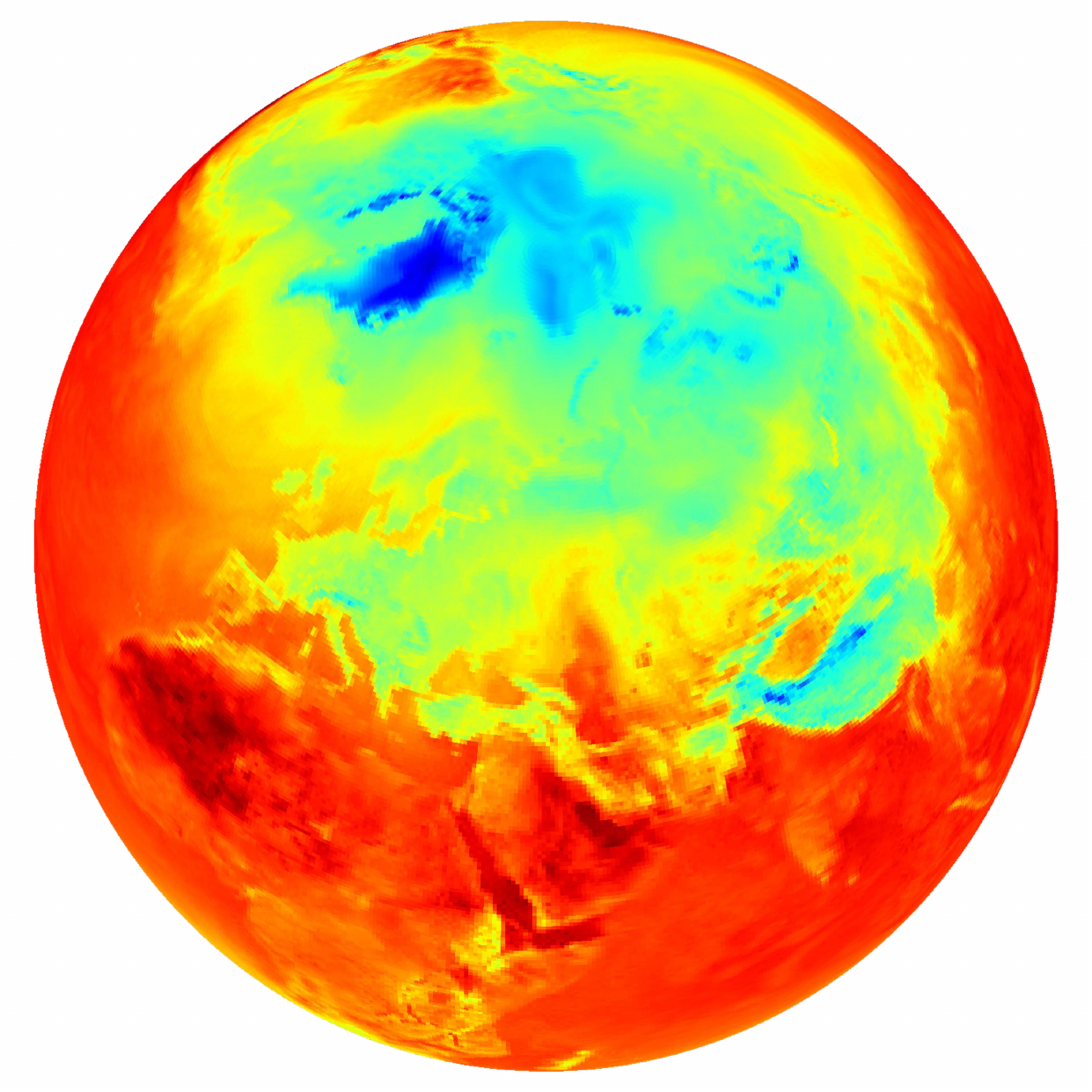} &
            \includegraphics[width=0.31\linewidth]{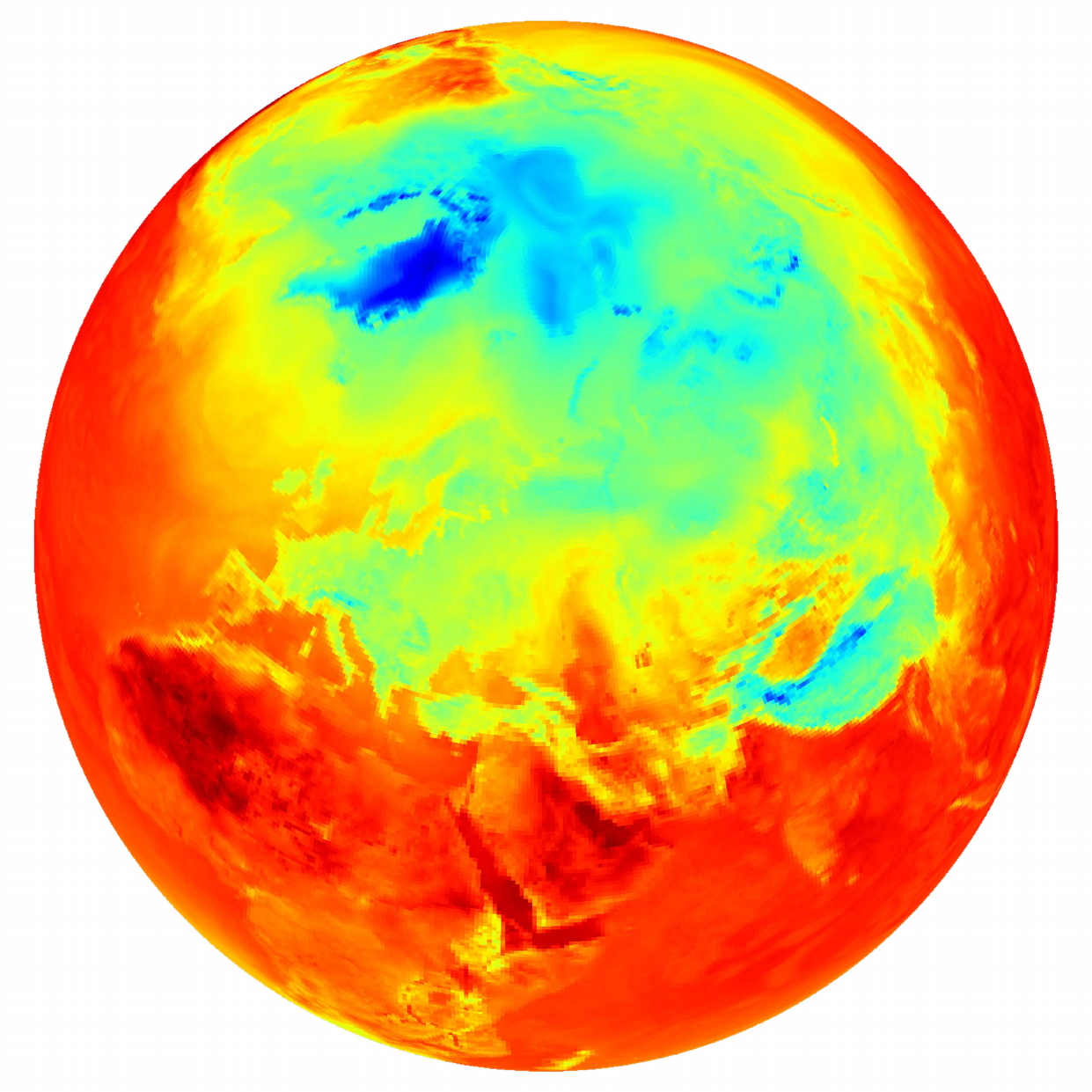} \\
            \scriptsize BEAM &
            \scriptsize $m$=1024 (1.737) &
            \scriptsize $m$=2048 (1.665)
        \end{tabular}
    \end{minipage}
    \hfill
    \begin{minipage}[t]{0.48\linewidth}
        \centering
        \textbf{Winter}
        \vspace{1pt}
        \begin{tabular}{ccc}
            \includegraphics[width=0.31\linewidth]{figures/era5/sphere_winter_patch_split_grid.pdf} &
            \includegraphics[width=0.31\linewidth]{figures/era5/sphere_winter_pred_mae0.398_from_npy_m1024.pdf} &
            \includegraphics[width=0.31\linewidth]{figures/era5/sphere_winter_pred_mae0.317_from_npy_m2048.pdf} \\
            \scriptsize Regular Mesh &
            \scriptsize $m$=1024 (0.398) &
            \scriptsize $m$=2048 (0.317) \\[4pt]

            \includegraphics[width=0.31\linewidth]{figures/era5/sphere_winter_patch_split_beam.pdf} &
            \includegraphics[width=0.31\linewidth]{figures/era5/sphere_winter_pred_mae0.363_from_npy_m1024.pdf} &
            \includegraphics[width=0.31\linewidth]{figures/era5/sphere_winter_pred_mae0.288_from_npy_m2048.pdf} \\
            \scriptsize BEAM &
            \scriptsize $m$=1024 (0.363) &
            \scriptsize $m$=2048 (0.288)
        \end{tabular}
    \end{minipage}

    \caption{
    Seasonal comparison of ELM-INR without BEAM (top row in each block) and with BEAM (bottom row) on full-resolution \texttt{ERA5} data.
    For identical model capacity ($m=1024$ or $m=2048$), BEAM consistently reduces MAE across all seasons by adaptively refining spectrally complex regions.
    }
    \label{fig:era5_seasons}
\end{figure*}

We further evaluate BEAM on full-resolution \texttt{ERA5} data. For these experiments, ELM-INR uses spatial patches with $S_i = 64^2$ samples and BEAM is constructed with an atomic size of $s=32$. Figure~\ref{fig:era5_seasons} reports seasonal reconstructions with and without BEAM for two capacities ($m=1024$ and $m=2048$). Across all seasons, BEAM consistently reduces the mean absolute error (MAE) at both capacities, with relatively larger improvements in the more capacity-constrained setting ($m=1024$). These results further corroborate that BEAM is most impactful when the underlying ELM-INR operates under limited representational budget.

\clearpage
\begin{figure}[t]
    \centering
    \includegraphics[width=0.9\columnwidth]{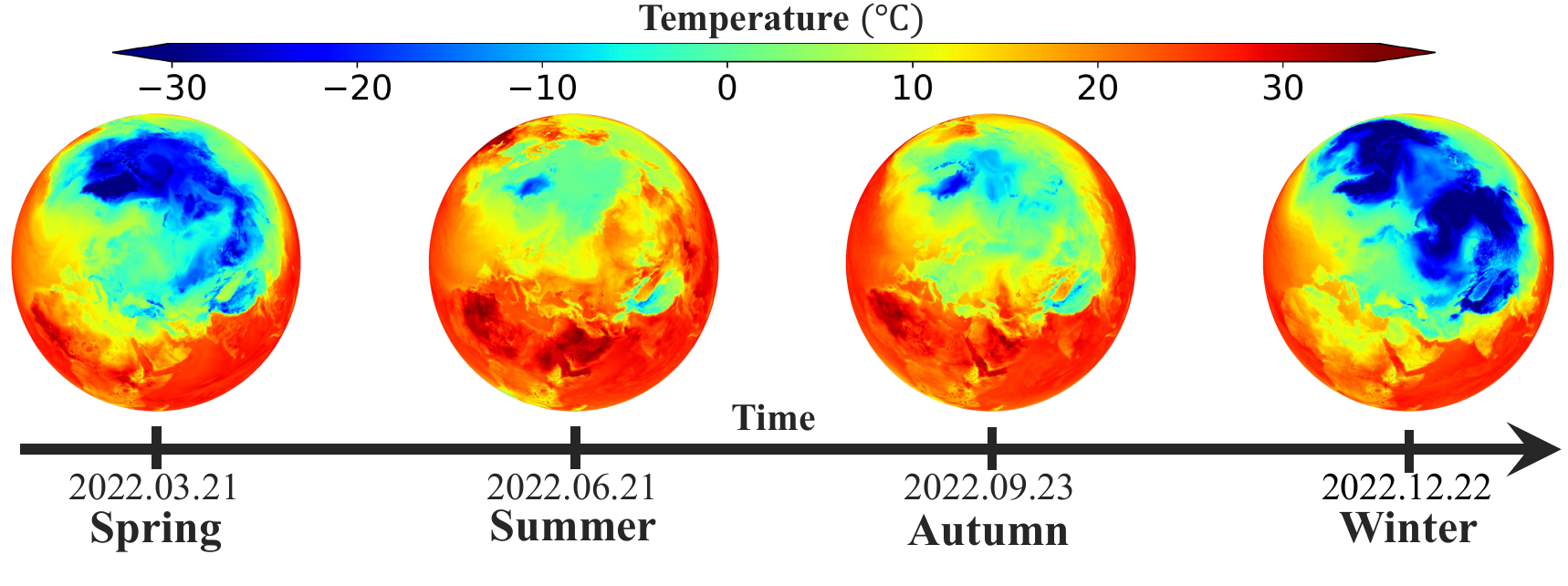}
    \caption{\textbf{ERA5 dataset samples.} Visualization of the ERA5 reanalysis data used in our experiments. We select four snapshots from 2022 corresponding to the vernal equinox, summer solstice, autumnal equinox, and winter solstice. Each panel shows the 2\,m air temperature field, highlighting seasonal variability and serving as the target signal for our reconstruction benchmarks.}
    \label{fig:time_psnr_era5}
\end{figure}

\clearpage
\section*{NeurIPS Paper Checklist}

\begin{enumerate}

\item {\bf Claims}
    \item[] Question: Do the main claims made in the abstract and introduction accurately reflect the paper's contributions and scope?
    \item[] Answer: \answerYes{} 
    \item[] Justification: The three contributions stated in Section~\ref{sec:intro} are supported by Sections~\ref{sec:elm_inr}–~\ref{sec:beam} and the experiments in Section~\ref{sec:experiments}.

\item {\bf Limitations}
    \item[] Question: Does the paper discuss the limitations of the work performed by the authors?
    \item[] Answer: \answerYes{} 
    \item[] Justification: Limitations are discussed in Section~\ref{app:limitation}.

\item {\bf Theory assumptions and proofs}
    \item[] Question: For each theoretical result, does the paper provide the full set of assumptions and a complete (and correct) proof?
    \item[] Answer: \answerYes{} 
    \item[] Justification: All theoretical results are stated with their assumptions in Section~\ref{sec:elm_inr}.

    \item {\bf Experimental result reproducibility}
    \item[] Question: Does the paper fully disclose all the information needed to reproduce the main experimental results of the paper to the extent that it affects the main claims and/or conclusions of the paper (regardless of whether the code and data are provided or not)?
    \item[] Answer: \answerYes{} 
    \item[] Justification: Architectures, hyperparameters, and training protocols for ELM-INR and all baselines are described in Appendix~\ref{app:detailed_experimental_setups}.

\item {\bf Open access to data and code}
    \item[] Question: Does the paper provide open access to the data and code, with sufficient instructions to faithfully reproduce the main experimental results, as described in supplemental material?
    \item[] Answer: \answerYes{} 
    \item[] Justification: All datasets used are publicly available, and the implementation is provided in the supplementary material.

\item {\bf Experimental setting/details}
    \item[] Question: Does the paper specify all the training and test details (e.g., data splits, hyperparameters, how they were chosen, type of optimizer) necessary to understand the results?
    \item[] Answer: \answerYes{} 
    \item[] Justification: Training details such as the optimizer and hyperparameters are provided in Appendix~\ref{app:detailed_experimental_setups}.

\item {\bf Experiment statistical significance}
    \item[] Question: Does the paper report error bars suitably and correctly defined or other appropriate information about the statistical significance of the experiments?
    \item[] Answer: \answerYes{} 
    \item[] Justification: We report the mean and standard deviation over five random seeds in Appendix~\ref{app:comprehensive_comparison}.

\item {\bf Experiments compute resources}
    \item[] Question: For each experiment, does the paper provide sufficient information on the computer resources (type of compute workers, memory, time of execution) needed to reproduce the experiments?
    \item[] Answer: \answerYes{} 
    \item[] Justification: Computational resources are described in Appendix~\ref{app:detailed_experimental_setups}.
    
\item {\bf Code of ethics}
    \item[] Question: Does the research conducted in the paper conform, in every respect, with the NeurIPS Code of Ethics \url{https://neurips.cc/public/EthicsGuidelines}?
    \item[] Answer: \answerYes{} 
    \item[] Justification:  The research conforms to the NeurIPS Code of Ethics.

\item {\bf Broader impacts}
    \item[] Question: Does the paper discuss both potential positive societal impacts and negative societal impacts of the work performed?
    \item[] Answer: \answerYes{} 
    \item[] Justification: Broader impacts are discussed in Section~\ref{app:limitation}.
    
\item {\bf Safeguards}
    \item[] Question: Does the paper describe safeguards that have been put in place for responsible release of data or models that have a high risk for misuse (e.g., pre-trained language models, image generators, or scraped datasets)?
    \item[] Answer: \answerNA{} 
    \item[] Justification:  The paper does not release any high-risk data or models.

\item {\bf Licenses for existing assets}
    \item[] Question: Are the creators or original owners of assets (e.g., code, data, models), used in the paper, properly credited and are the license and terms of use explicitly mentioned and properly respected?
    \item[] Answer: \answerYes{} 
    \item[] Justification:  All baselines and datasets used in this work are cited at first use.

\item {\bf New assets}
    \item[] Question: Are new assets introduced in the paper well documented and is the documentation provided alongside the assets?
    \item[] Answer: \answerYes{} 
    \item[] Justification: The proposed implementation, together with documentation, is included in the supplementary material.

\item {\bf Crowdsourcing and research with human subjects}
    \item[] Question: For crowdsourcing experiments and research with human subjects, does the paper include the full text of instructions given to participants and screenshots, if applicable, as well as details about compensation (if any)? 
    \item[] Answer: \answerNA{} 
    \item[] Justification: This work does not involve crowdsourcing or human subjects.

\item {\bf Institutional review board (IRB) approvals or equivalent for research with human subjects}
    \item[] Question: Does the paper describe potential risks incurred by study participants, whether such risks were disclosed to the subjects, and whether Institutional Review Board (IRB) approvals (or an equivalent approval/review based on the requirements of your country or institution) were obtained?
    \item[] Answer: \answerNA{} 
    \item[] Justification: This work does not involve human subjects.

\item {\bf Declaration of LLM usage}
    \item[] Question: Does the paper describe the usage of LLMs if it is an important, original, or non-standard component of the core methods in this research? Note that if the LLM is used only for writing, editing, or formatting purposes and does \emph{not} impact the core methodology, scientific rigor, or originality of the research, declaration is not required.
    \item[] Answer: \answerNA{}{} 
    \item[] Justification: LLMs are not used as a core component of the proposed method.

\end{enumerate}

\end{document}